# THE REALITY OF MULTI-LINGUAL MACHINE TRANSLATION

Tom Kocmi, Dominik Macháček, Ondřej Bojar

## STUDIES IN COMPUTATIONAL AND THEORETICAL LINGUISTICS

Tom Kocmi, Dominik Macháček, Ondřej Bojar

# THE REALITY OF MULTI-LINGUAL MACHINE TRANSLATION

This book has been printed with the support of the project 18-24210S of the Grant Agency of the Czech Republic and of the institutional funds of Charles University.
Printed by Reprostředisko MFF UK.

# Contents

# 1
# Introduction

## 1.1 Neural Networks Inspired by Humans

**The more languages you speak the better network you are?**

It definitely holds for humans: the more languages you speak, the easier it is for you to understand each of them as well as other languages, the better job you may aspire to,[1] the easier it is for you to learn a new language,[2] the more resilient you might be to some of mental aging diseases,[3] or enjoy other possible benefits.[4]

**Neural networks (NN)** were very initially inspired by the design of natural neuron cells and their interconnections (Rosenblatt, 1958). It took sixty years to come up with training methods and scale up the network size to allow these networks to carry out complex tasks across a very broad range of areas. **Deep learning (DL)**, that is the research field for designing and training of neural networks with a certain level of internal complexity, has invaded essentially all types of tasks where training data (sample inputs and expected outputs) can be gathered or created. For the non-expert community, deep learning has been equated with the term **artificial intelligence (AI)** and the AI hype is seen in all media.

This book focuses on a subfield of artificial intelligence, one particular area of **natural language processing (NLP)**, namely **machine translation (MT)**, and more specifically **NMT (neural machine translation)**, but the observations collected here will be generally applicable to many other situations, as soon as the characteristics of the processed data are similar.

As will be seen from the following chapters, machine translation belongs to one of very influential tasks for deep learning, thanks to several of its features:

- translation is a very complex cognitive task. In many situations, it is truly **AI-complete**, i.e. the system needs some “full understanding” of the problem in order to solve even a part of it. The AI-completeness of MT becomes apparent as soon as gaps in “world knowledge” occur.

[1] https://www.uei.edu/blog/can-speaking-two-languages-increase-your-job-prospects/

[2] https://www.studyfinds.org/more-languages-easier-for-brain/

[3] https://thereader.mitpress.mit.edu/can-learning-a-foreign-language-prevent-dementia/

[4] https://thecorporatemagazine.com/the-benefits-of-linguistic-diversity

- on the one hand, large datasets of inputs and expected outputs are relatively easy to obtain and their format (pairs of source text and its translation) is easy to work with,
- on the other hand, the intermediate representations, when humans are translating, are not available and will not become available any soon given the limited capabilities of brain scanning (Logothetis, 2008; Hansen-Schirra, 2017; Schleim and Roiser, 2009) and eye tracking (Hvelplund, 2014).

Deep learning is being designed with particularly such tasks in mind: where large training data are available but the task is too complex to be formally described. If a deep learning method succeeds in the task of machine translation, it confirms its ability to model very complex latent (hidden) processes.

Machine translation is also a very inspiring task because of the different linguistic properties of world's languages, see e.g. WALS[5] (Dryer and Haspelmath, 2013). While some languages are very close to each other and essentially only very simple "word substitution" is needed to translate the text, such as Czech and Slovak (Homola et al., 2009), some languages are very distant and translation actually amounts to "understanding the message and regenerating it in the target language". We expect that e.g. Chinese-Czech translation would need such an approach. For deep learning, this challenge is very attractive because ideally *the very same* learning method would (learn to) fit well across the wide range of task complexities.

Finally, there are two more and mutually related reasons why machine translation is a good playground for deep learning. The first one is the central point of this book: Machine translation can involve more than just two languages, which we call **multi-lingual** MT. In such a setting, the neural network can process or be expected to produce several closely related and yet different signals, the inputs or outputs in multiple languages. The second one is beyond the scope of this book but it is certainly the next critical component that MT research will absorb: **multi-modal** translation (Sulubacak et al., 2020), i.e. translation with additional, non-textual information such as sound, images or video. The relatedness of the multi-lingual and multi-modal extensions of machine translation may not be apparent at the first sight, but NNs are actually a great technical device for processing different modalities as we discuss in Chapter 3 in closer detail.

## 1.2 Multi-Lingual Machine Translation

Machine translation always works with at least two languages: the source and the target. Practical reasons can lead authors of MT systems to use one or more intermediate languages, called **pivot languages**. A system which translates the source first to the pivot and then from the pivot to the target is still not a multi-lingual system as we define it.

[5] `https://wals.info/`

A multi-lingual MT system is designed to work with more than two languages in a non-decomposable way; it is impossible to extract a component from the system that would translate only between a subset of these language.

Inspired by humans, MT researchers are hopeful that training MT systems on more than just two languages will be helpful for the translation quality. In the rest of this book, we examine this hope and desire very carefully from multiple angles. We show the reliable and promising pathways and warn against common pitfalls which can, for instance, inflate the benefits of an approach and create unreachable expectations. To summarize our main message very shortly:

**More than two languages in machine translation can help,**
**if you know what you are doing**
**and you work hard not to fool yourself about the real source of the gains.**

## 1.3 Aims of This Book

When writing this book, we had several rather general sources of motivation, stretching well beyond the particular task of translating with more languages at once:

**Advertise MT as an interesting domain.** As discussed above, we are convinced that MT is a fascinating example of tasks for deep learning. While we are not providing a tutorial to MT as such, and instead refer the reader e.g. to Koehn (2020), we are discussing the *learning problems* NNs are facing when trying to translate. These, in turn, are common to all application areas of deep learning.

**Highlight the discrepancy** between practical performance of NLP/MT systems and the assumed level of "understanding". One would think that in order to translate a text between natural languages, one has to understand it. The amazing translation quality achieved by recent systems (Hassan et al., 2018; Popel et al., 2020; Graham et al., 2020a) chips away some of this confidence: outputs hard or impossible to distinguish from manually translated texts are produced even by systems which obviously can not possess any reasoning capacity, any understanding of the world described.

Throughout this book, we are going to return to the question of understanding several times. By revisiting it, we want to highlight the sore point of NLP research: the rather common lack of understanding of understanding. This problem is very pressing (Bender and Koller, 2020), because it is likely to mislead the public about the expected performance and applicability of NLP or AI systems. Overestimation and the inevitably following disillusion from failing to meet the (unrealistic) expectations puts our whole field in high risk of yet another "AI winter", i.e. a period of decreased trust and interest, and severely limited funding.

**Focus again the question of generalization capability of NNs.** The lack of machine understanding is closely related to the generalization capability of trained systems. As argued by many, deep learning systems can excel at *interpolating*, i.e.

generalizing “within the area” documented by training examples, but they tend to fail “outside” of that area. The concept of “the area” in NLP, i.e. the spaces of possible inputs and outputs, is not as easy to grasp as the concept of a highly-dimensional vector space (as we discuss in Section 3.1.2) but in our experience, DL methods exhibit here the same lack of generalization, even in questions as simple as the length of the sentence (see Section 4.5).

Despite the rapid development in the area of machine translation, multilingual representations etc., which is hard to fully follow even by researchers active here every day, we believe that writing our observations in the form of a monograph makes sense. If we tried to propose any new method, models or algorithms in this book, it would become outdated before it is even printed.

This book thus aims not to chase the newest developments, but instead to provide a rather stable common ground for reasoning and research. Perhaps even more importantly, we want to *warn about common, frequent and recurring fallacies*, and provide and explain examples of these methodological and argumentation errors in the particular area of machine translation. We are, of course, not alone in this quest, see e.g. Marcus (2020) for a number of such warnings and a plan to re-inject knowledge into AI systems.

Throughout our journey, we will highlight various observations and express them often in fairly general ways. It is important to keep in mind that all these observations are based on some particular experimental setting, set of languages, specific training data and other conditions. Inevitably, there will be exceptions and perhaps completely different behaviours observed if the settings change. The list of all observations is at the end of the book in *List of Observations* on page 185.

**Observation 1:** *Observations in this book should be taken as general rules of thumb and ideally should be verified in your specific settings.*

## 1.4 Intended Audience for This Book

This book should be accessible and interesting for you if you belong to any of these groups:

**Researcher or student in machine translation** You are already knowledgeable about what one *should* do to successfully train MT systems. We hope you will find the set of observations we make nevertheless inspiring and extending your knowledge. We also think that reiterating what one *should not do or assume* is important for your progress in the research.

**Researcher or student in general NLP** We believe that you may benefit in two areas: (1) by learning more about the general deep learning techniques and common fallacies, (2) by learning more about multilinguality, which is known to help and to be needed in many areas of NLP.

**General DL/AI researcher** We would like you to find here inspiration and caveats regardless what your intended target application is, or if you are trying to improve deep learning techniques in general. We would also like to invite you to extend your domains of interest and test your methods and hypotheses in NLP or MT specifically.

**Curious user of NLP/MT** Regardless if you are an experienced practitioner from the translation and localization industry or if you are just an end user of online MT systems, you should benefit from reading this book by understanding better the inherent and still unsurpassed limitations of the systems. Through all the caveats we mention, we would like to "inoculate" you at least a little against exaggerated claims of MT or AI achievements. As researchers willing to foster our field in a long-term run, we need to manage your expectations. Yes, we are passionate about the advances we are achieving and seeing in the works of our colleagues, but we need to always question and double check the real reasons behind the observed gains, and then moderate any extrapolations into future developments accordingly.

## 1.5 Book Structure

Machine translation is not really moving as a pleasant Sunday afternoon walk. Having worked in the field for quite some time and esp. over the transition to deep learning methods, we see it more as a roller coaster ride.

We are going to mimic this and take you on board, running up and down over a few hills. These highs and lows correspond to phases of optimism and high expectations vs. phases of a little sobering up and perhaps disillusion when we reveal that the methods are not doing what we expected them to do. Overall, these highs and lows should cancel out in time, and you should arrive smoothly at stable and credible opinion.

In Part I, we start by building up expectations on how neural networks could break the language barrier in Chapter 2. Given their versatility discussed in Chapter 3, neural networks could indeed be the technical device to achieve that. Our ride starts to decline as we reveal the pitfalls of NN learning in Chapter 4 and we hit a bottom when we realize that often, we can be easily getting good results for very wrong reasons (Chapter 5).

Part II then delves fully into multi-lingual MT. We start with an overview, defining various types of multi-lingual solutions (Chapter 6). We then provide an in-depth analysis of the perhaps simplest type of multi-linguality, namely transfer learning in Chapter 7. We then jump over to the most recent advances, which however often require massive models, see Chapter 8. We strike the balance between what is amazing and what is practical in Chapter 9.

The overall conclusion (Chapter 10) discusses the persevering challenges and desired future developments, incl. one important aspect: the energy efficiency.

# I Background

# 2
# Reverting the Babel Curse

According to the myth, the people of Babel agreed to build a tall tower that could reach the heaven and God. However, God saw that they are united, have a single language, and that they have no restraints. Therefore, God confounded their language and scattered them along the earth so they were not longer able to build the tower and reach God. The word "babel" means "perplexity" or "confusion" in Hebrew, the language of the Bible where the story is recorded. It is also the Hebrew variant of the name Babylon, the ancient city in today Iraq where it was located.

Since the ancient times of the Babel tower, people live in nations speaking different languages. On the one hand, the distinct languages make nations unique and coherent. For many nations, the language is an important asset for their national self-determination. National languages also induce protective boundaries between the nations. On the other hand, the language barrier is an obstacle in international cooperation. Studies confirm that language barriers slow down decision processes and make them more expensive in international companies (Harzing et al., 2011). In the United States, lack of knowledge of other languages than English limits the business (ACTFL Survey, 2019).

Today, around 5 000 years after the Babel tower, we would like to revert the Babel curse in a way so that we keep language diversity for its benefits and at the same time help people to understand others, without the necessity to learn a foreign language at an advanced level. We use technology for that, namely machine translation.

## 2.1 The Benefits of Language Diversity

What are the benefits of language diversity? Why does the European Commission spend 349 millions Euro per year for translating 2.3 million pages between all 24 EU official languages?[1] Why there is parallel simultaneous interpreting of the plenary sessions of the European Parliament into 24 EU official languages, instead of choosing one that would be adopted as official?

Between the EU official languages, there are mutually intelligible pairs: Czech and Slovak, Slovenian and Croatian, and Spanish and Portuguese. Nearly all the speakers of one of the languages understand the other one. It would be possible to spare one interpreter in each pair, and the people in both nations would still understand. Why is it not practiced?

[1] Numbers from the year 2021. Source: `https://op.europa.eu/s/uWck`

The first reason is that the EU countries would not agree on a single official language. Unofficially, English is the most widespread lingua franca in the current Europe. It is the mostly used language for communication between people who do not have the same native language but share English as a foreign language. However, relying on English as the only official EU language is not adopted. After Brexit, English is ironically the official language of an EU country only in Ireland with the population of 5 million and Malta with the population of 0.5 million. English is thus the official language for less than 1.5% of the EU population. Van der Worp et al. (2016) observe other reasons why English is not adopted as lingua franca in a survey on Basque companies: while efficient, people appreciate to be dealt with in their mother tongue, and furthermore, some people see the spread of English as a threat to linguistic diversity.

The second reason is that there are many people who have difficulties to use any foreign language. Despite that, the EU wants to treat all its citizens as equal partners in communication, to make EU institutions more accessible and transparent. Those who would be forced to use intelligible, but non-native language, could feel unequal (Welle, 2020).

There are psycholinguistic studies about the effect of native language on visual consciousness (Maier and Rahman, 2018), and generally to world view and opinions. This concept is called **linguistic relativity**. The diversity of cultures and native languages might be beneficial especially in work teams. Multilinguality might thus support innovativeness.

Finally, we refer the reader also to Gorter et al. (2015) who discuss the possibilities and complications when trying to formalize the benefits of multilinguality for the society in economic terms.

## 2.2 Why More than Two Languages in MT?

Machine translation inherently handles two languages, the source and the target one. But would the system benefit, similarly as humans tend to do, if it was created to process more than just two languages? The reasons may be efficiency, flexibility, and quality. We overview and explain these three sources of motivation. In practice, it is usually a certain mix of these three sources that drives the design of translation systems.

### 2.2.1 Efficiency

If you need to translate from one language into 40 languages in a short time, you need either 40 bilingual models, or one multi-lingual. In current standards, you need around 800 megabytes of disk space for each bilingual model. When starting to translate, you need to load the model to GPU memory. This can take around 30 seconds or several minutes, depending on the framework. You can not parallelize the loading or translation processes of several different models on one GPU because it is not faster than running them in sequence. Therefore, when translating into 40 languages with bilingual models, you need either 40-times more time than into one language, or 40 GPUs for parallelization. Or one multi-lingual model.

With one multi-lingual model, you need only one GPU, and its loading and translation time is roughly the same as for one bilingual model. Batch processing in translation can be used, so e.g. 64 sentences can be translated at once. In a multi-lingual model, the same input sentence can be presented to the system in a batch of 64 copies, each to be translated to one of the desired target languages. This is especially useful in simultaneous speech translation to many targets, where one input sentence arrives every now and then and all the targets are needed as quickly as possible.

We discuss practical projects addressing this aspect in Chapter 9.

### 2.2.2 Flexibility

Multi-lingual NMT can be flexible. Depending on the type of the model, it can cover a set of languages that are accepted on the source side, and another set as supported targets. The user can then very flexibly provide all available sources and ask for various targets. While switching, there is no need to wait for loading another model in the background. The model could be trained for even more flexibility: It can detect the source language on its own, without marking it by the user.

The next level of flexibility is adaptability to "**code switching**" or code mixing which means alternating languages within one utterance, within one sentence. A multi-lingual model can be designed to translate code-switched inputs into one common target language.[2]

A multi-lingual model can also flexibly handle translation of language pairs for which there are only extremely few mutually parallel training data or no data at all. This case is called **zero-shot** translation. For example, there may be training data for English-Russian and for Russian-Kazakh. As Kazakh is culturally and geographically far from English, direct training data may be too small. When translating English-Kazakh, an old traditional approach is to cascade two separate bilingual models, English-Russian and Russian-Kazakh. Russian would be then the pivot language and the approach called **pivoting**. Pivoting is, however, prone to error accumulation which a multilingual model may bridge over. Some multi-lingual models can translate the unseen language pairs directly, with a quality comparable to or even higher than pivoting. Furthermore, a multi-lingual model can effectively exploit a small direct dataset along with the two indirect ones, while pivoting can not benefit from this extra resource.

[2] Alternatively, NMT systems can be designed to inject code-switching, e.g. when specific terminology is better understood in English (Bafna et al., 2021).

### 2.2.3 Quality

The next reason for multi-linguality in MT is the expected higher quality when neural network is trained for multiple languages at once than when trained separately for two languages. Two separate sources of gain are possible here: (1) an overall better model thanks to cross-lingual generalization and data reuse, and (2) benefits from more inputs.

For (1), the hope builds upon the idea of multi-tasking, i.e. the ability of the network to use knowledge from one task to improve a related task when the network is trained to both of them at the same time, see Section 3.5 below.

Specifically, it is expected that the neural network learns and benefits from generalization across languages. For example, Upper and Lower Sorbian are two Slavic languages similar to Czech and Polish. They are similar to each other and mutually intelligible. They contain many loan words from German because they are languages of two small Sorbian national minorities living in Germany. Due to the small number of speakers, the two languages are low-resourced, so there are only small parallel data. It is therefore possible that a neural network learns more generalizations for Upper and Lower Sorbian when learning them together with Czech, Polish and German, compared to learning them separately.

For (2), consider multi-national organizations such as the European Union or United Nations. They often need to translate documents from one source language into many targets in a short time and ensuring high quality. Professional translators are thus an inherent part of the process. In this situation, the machine translation for e.g. English-to-Greek may benefit from the fact that e.g. the German parallel version has already been processed and revised. It might be beneficial to translate from English and German parallel sources, because the additional source can help the disambiguation. For example, the German word "Schloss" can mean both "castle" or "lock" and conversely, the English word "chair" can mean both the piece of furniture ("Stuhl" in German) or the president ("der/die Vorsitzende"). Having access to the other language resolves these ambiguities when choosing the Greek work.

# 3
# The Versatility of Neural Networks

This chapter introduces the necessary technical background and also aims to maximize the optimistic view: use the Transformer architecture (see Section 3.3.2), throw the data in, have it trained and obtain a great performing system for any kind of task you like because neural networks are very versatile.

We refer the reader to any of the abundant introductions to neural networks and deep learning, with the book Deep Learning (Goodfellow et al., 2016) being the bible of the field.

In this chapter, we are going to assume that you are familiar with neural network structure at the finest level (individual neurons with their weights, biases and activation functions to introduce non-linearity), with more complex compositions (basic feed forward layers, convolutional neural networks), basic training methods (stochastic gradient descent, backpropagation) as well some of the basic tricks and knowledge needed for stable training (regularization methods such as dropout, learning curves and overfitting).

We will cover the mathematical background needed for processing and producing text with neural networks, to the extent needed for a clear and unambiguous presentation of our findings and conjectures. We first characterize the task of machine translation (Section 3.1), then discuss how the realistically big (i.e. huge) vocabularies of natural languages need to be presented to the network (Section 3.2). In Section 3.3, we explain the techniques used to feed neural networks with sequences of words and how to obtain output sequences of words from them. Section 3.4 discusses the flexibility that we are gaining when the network processes sequences of tokens into sequences of tokens: how we can trivially provide the network with additional information or ask it to report more than just the translation. Finally, Section 3.5 explains multi-tasking, i.e. the idea of learning to handle more tasks at once.

## 3.1 Characteristics of Machine Translation Task

When translating, human translators and more consciously interpreters consider all available information, including linguistic, paralinguistic and full world context. This is different from machine translation.

### 3.1.1 Sentence-Level Translation

Since very early attempts, see the Georgetown experiment in 1954 (Hutchins, 2004), and later followed by the statistical approaches (Berger et al., 1994), machine translation has been formalized as the task of translating *a single sentence in the source language to a single sentence in the target language.* This simplification generally persists even today; it is built deeply into the interfaces of machine translation services and computer-assisted tools. Only slowly, the field is moving towards considering larger units of text.[1]

It is assumed, that "one sentence expresses one thought", and that sentences are thus reasonable units for translation. Short sentences taken out of context ("Yes, I do."), are going to be very ambiguous and risky to translate, but they do not show up too often in the examined domains – mainly news – so the ambiguity resolution problem they pose is not striking enough for the research community. Moderately long sentences already bear enough information to disambiguate most words. While there are definitely very important translation choices that can be correctly made only when knowing the surrounding context in both the source and target language, document-level coherence remains so far too broad problem with benefits not so visible in evaluation.

### 3.1.2 Space of Possible Outputs for Sequence-to-Sequence Task

Machine translation is formalized as a **sequence-to-sequence** task: Given a sequence of input symbols (be it words, subword units, tokens, characters or bytes), produce a corresponding sequence of symbols in the target language.

First, let us assume that the basic unit is a single character. For simplicity, we consider just 26 English letters, space, full stop and question mark (an alphabet of 29 characters), and some maximum sentence length, e.g. 50 characters. Consider an input sentence, e.g. the Arabic:

أم انها نتيجة جهد مشترك؟

This setting offers $29^{50}$ = 13,179,529,148,667,419,313,752,621,434,193,363,370,806,933, 128,742,294,644,974,969,657,446,901,001 $\approx 13 \cdot 10^{72}$ (13 trevigintillion, if you wondered if this order of magnitude still had a name) admissible output strings. For a human being, most of them are outright wrong, e.g.:

$*$ lsdkjflkoieromcimerocimldklskdjksadmcolsikmr fsijf

[1] Consider for example the evolution of the news translation task at WMT: up until 2018 (Bojar et al., 2018), the whole shared task on translation was run and evaluated by considering independent sentences, although the test sets were generally compiled using full documents. Since 2019 (Bojar et al., 2019), document-aware evaluation is slowly being adopted but even today, in 2021 (Akhbardeh et al., 2021), the vast majority of MT systems translates individual sentences without considering the surrounding ones and the evaluation method for more than half of the language pairs uses only a very simple approach where the evaluator can not go back to previous sentences in the document.

∗ ?????.?????????????????.??????????????????.???????
∗ eeeeeeeeeeeeeeeeeeeeeeeeeeeeeeeeeeeeeeeeeeee

Some of the permissible outputs are surely good translations, e.g. the four reference translations provided as part of the NIST 2008-2012 Open Machine Translation (OpenMT) Progress Test Sets:[2]

- or is it a result of a combined effort?
- or are they the result of a joint effort?
- or are they the result of joint efforts?
- or is it the result of a joint effort?

As Dreyer and Marcu (2012) and Bojar et al. (2013) documented, the set of good translations for a given single sentence is actually *huge*. It easily contains hundreds of thousands of widely differing strings. The sample Arabic sentence with four reference translations can be also translated as:

- or rather is it the result of a team work?
- or should these be regarded as the outcome of attempts of both sides?
- or it is the result of combined endeavor?
- or does this come out of a unified effort?
- or perhaps is this arising because of a joint effort?
- or it is the consequence of joint aspiration?
- or could it be a product of an amalgamated push?
- or could it be a outcome of a unified push?
- or are these considered as their joint accomplishments?

At the same time, a very small change in the string can render the translations wrong (error underlined, correct word in parentheses):

∗ or is it a result of a combined error? (effort)
∗ or are hey the result of a joint effort? (they)
∗ or are you the result of joint efforts? (they)

We can thus summarize the characteristics of sentence-level MT as the transduction of the string in the source language to the string in the target language with:

- extremely many possible input values,
- extremely many technically permitted output values,
- extremely many output values which are clearly wrong,
- very many output values which are good (some levels of “goodness” are possible), where
- many of the good outputs differ from one another much more than they differ from a bad output value.

Our presentation using clearly garbage strings of letters as conceivable outputs may seem exaggerated but it reflects realistically the situation faced by MT systems if they process letters.[3] At the beginning of the training, the systems do not have any prior knowledge, any concept of words, any idea of sentence structure or most importantly any world experience which limits the set of sensible meanings.

[2] https://catalog.ldc.upenn.edu/LDC2013T07

This idea of very large space of possible strings, and hugely smaller but still extremely large space of good ones, is critical for understanding that almost *any* constraining of the search space is going to be useful. So there is quite a big chance for overestimating the utility of this constraining, see Chapter 5.

## 3.2 Processing Words

Neural networks work with inputs, outputs and intermediate results represented in continuous spaces. When used for NLP tasks, we need to bridge the gap between the world of discrete units of words and the continuous, differentiable world of neural networks.

Originally, the first step was to use a finite vocabulary, where each word had a different index, which was then represented as a one-hot vector of the size equal to the number of items in vocabulary. Sizes of 10–90k words were used in NLP and separate handling of unknown words was necessary. The one-hot representation, a vector containing only zeros except at a single position with one, moves us from the realm of words to the realm of numbers but still it is not continuous and differentiable. Furthermore, it does not reflect a key observation that words are more or less similar to other words. Similar words should be closer to each other within the representation.

One way of compressing the one-hot vectors is to assign each word a specific dense vector through an NN layer which can be called *lookup table* or *word embedding*.

Embeddings (Bengio et al., 2003) are dense vector representations of words commonly of 100-1000 dimensions. They are trained jointly with the whole network and learn word-specific features and cluster the words in the space so that similar words have vectors that are close to each other.

Mikolov et al. (2013) found that word embeddings in a language model neural network, i.e. a network predicting a word given some of its neighbours, contain semantic and syntactic information without being trained to do so. An example of embeddings clustering the space of words is in Figure 3.1.

Word embeddings can exhibit an interesting correspondence between lexical relations and arithmetic operations in the vector space (Mikolov et al., 2013). The most famous example is the following:

$$v(king) - v(man) + v(woman) \approx v(queen)$$

In other words, adding the vectors associated with the words *'king'* and *'woman'* while subtracting *'man'* should be close to the vector associated with the word *'queen'*. We can also say that the difference vectors $v(king) - v(queen)$ and $v(man) - v(woman)$ are almost identical and describe the gender relationship.

[3] Letter-based MT systems do exist and work reasonably well, they just need more training time. Commonly, the basic units are larger as we discuss in the following.

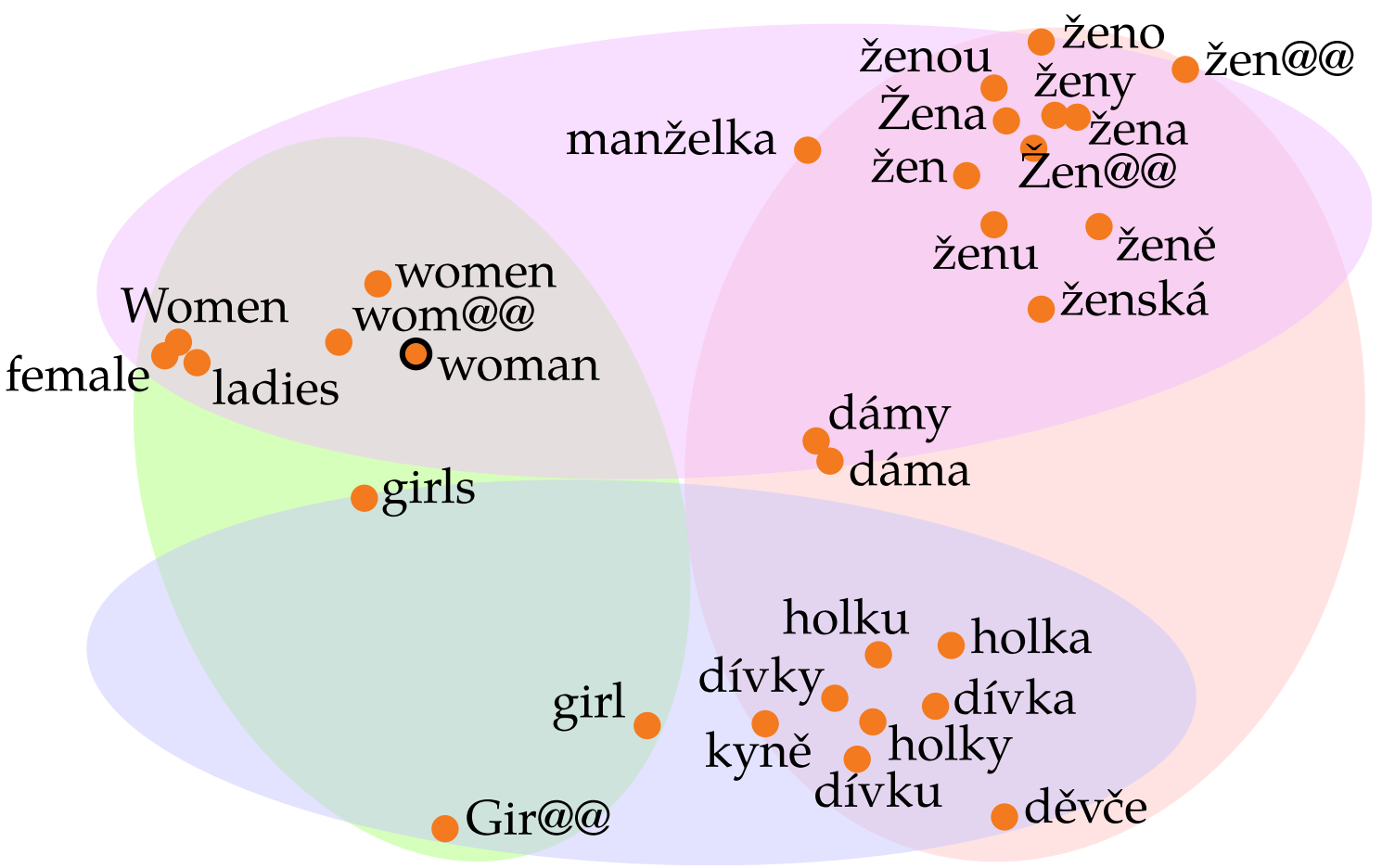

**Figure 3.1:** Thirty nearest neighbors in cosine similarity for the word "woman" visualized in 2D by principal component analysis. The large cluster bubbles were added manually for better presentation. The representation is from the encoder BPE subword embeddings (see Section 3.2.2) of our Czech→English model. This figure shows that the 30 nearest neighbors are variants of the word "woman". Interestingly there are two separate clusters for Czech and English words (left and right), which suggests that NN understands equivalence across languages. Furthermore, there are clusters dividing words for adult women and young women (top and bottom). Worth of mentioning is the subword "kyně", which is a Czech ending indicating the feminine variant of several classes of nouns, e.g. professions (pěvkyně – singer, plavkyně – swimmer, soudkyně – judge, etc.). It appears in the "young women" cluster probably because of the common word "přítelkyně" (girlfriend).

Mikolov et al. (2013) noticed that such relations emerge without specific training criteria naturally from training the language model with unannotated monolingual data.

### 3.2.1 Word Embeddings Aware of Word Structure

The Skip-gram model (Mikolov et al., 2013) uses one-hot representation of a word in the vocabulary as the input vector $x$. The embedding of a word then corresponds to the multiplication of the one-hot vector with the trained weight matrix (the lookup table). In other words, the $i$-th word in the vocabulary is represented with an embedding vector which is stored as the $i$-th row in the embedding matrix. Therefore weights $w_i$ of the input word $i$ can be directly used as word embeddings $E$:

$$E_j = \sum_{i=1}^{|x|} x_i \cdot w_{ij} = w_j \tag{3.1}$$

In Kocmi and Bojar (2016), we proposed a substring-oriented extension of Skip-gram model that induces vector embeddings from the character-level structure of individual words. Our approach gives the NN more information about the examined word reducing the issue of data sparsity and introducing the morphological information about the word to NN.

Our approach provides the neural network with a "multi-hot" vector representing the substrings contained in the word instead of the one-hot vector representing the whole word.

We use a vocabulary of substrings, instead of words, created in the following fashion: first, we take all character bigrams, trigrams, tetragrams, and so on up to the length of the word. This way, even the word itself is represented as one of the substrings. As an indication of the beginning and the end of words, we appended the characters ^ and $ to each word. Here we provide an example of the segmentation:

'cat' = {'^c', 'ca', 'at', 't$', '^ca', 'cat', 'at$', '^cat', 'cat$', '^cat$'}

Using all possible substrings would increase the size of vocabulary beyond a reasonable size. Thus, we select only the most frequent substrings based on the frequency in the training data.

In order to generate the vector of substrings, we segment the word and create a multi-hot vector, where "ones" indicate word's substrings indices in the vocabulary. In other words, each word is represented as a multi-hot vector indicating which substrings appear in the word.

The word embedding is created in the same fashion as in the one-hot representation: by multiplication of the input vector with the weight matrix. We have to keep in mind that each word has a different number of substrings. Thus the embeddings need to be normalized either by sigmoid function or by averaging over the number of substrings. We decided to use the mean value because it is computationally simpler than sigmoid:

$$E_j = \frac{\sum_{i=1}^{|x|} x_i \cdot w_{ij}}{\sum_{i=1}^{|x|} x_i} \tag{3.2}$$

where $x$ is the multi-hot vector, and the summation in denominator represents the number of found substrings of the word.

Our model, compared to Skip-gram, can encode even unseen words, and has a comparable or better performance on syntactic tasks (see Kocmi and Bojar, 2016). It could be useful for NLP tasks that do not produce any textual output, for example, sentiment analysis, language identification, or POS tagging. However, the approach is not reversible, and there is no simple way to transform embeddings back to word forms, which would be needed in word generation, such as the target side of machine translation.

While our paper Kocmi and Bojar (2016) was in the review process, Bojanowski et al. (2017) released *fasttext*,[4] an open-source implementation of a very similar idea. Thanks to the released tool and perhaps more importantly models pre-trained on very large text collections, *fasttext* vectors became very popular.

### 3.2.2 Subword Representation

Traditionally, NMT systems relied on a vocabulary to store all words used in the translation. The capacity of this vocabulary was typically 10–90k words. However, this is not enough to cover all words in a language. That is why the first NMT systems used a special OOV symbol as a replacement for the remaining rare (as well as not so rare) words.

The out-of-vocabulary words are a substantial problem especially for languages with inflection, agglutination or compounding, where many variants of a frequent word become rare. Increasing the size of the vocabulary in order to reduce the number of OOV words proportionally increases the training complexity as well as decoding complexity. Moreover, the NMT systems will not be able to learn good encoding for uncommon words or word-forms as they are seen only a few times within the training corpus or not at all. Huck et al. (2017) present an example of a Czech word which is observed in the first 50K sentences of a corpus but all its morphological variants are not seen even in 50 *million* sentences. Without a generalization capacity on word formation, some necessary word forms may never be accessible to the system.

To overcome the large vocabulary problem and avoid the OOV problem, translation models need mechanisms that go below the word level. There are two possible solutions: Either including more information into the representation of words, or splitting uncommon words and translating at the level of subword units.

The former approach makes word representations richer with information on linguistic classes or word structure. For instance, Tamchyna et al. (2017) removed the inflection by morphologically annotating training sentences and let the NMT translate only the lemmas and associated morphological tags, a joint multi-tasking as we will discuss in Section 3.5. Luong and Manning (2016) proposed to use a hybrid approach where they first get the word embeddings from characters followed by standard NMT on the computed embeddings. In Kocmi and Bojar (2016), we proposed to include the substring structure of a word into the word embedding as discussed above in Section 3.2.1.

The latter approach breaks uncommon words into subword units that are handled by the NN as standalone tokens. The trivial approach is to break the sentence into individual characters, but it needs much longer training times as the number of tokens per training example is several times higher than the number of words, and it creates a problem with long-range dependencies making the character-level translation suboptimal (Tiedemann, 2009; Ling et al., 2015). Thus, we need to split the words into some sensible subwords but avoid bloating the size of the subword vocabulary. For compounding, such an approach actually makes the segmented representation more informative than one vector for the whole word, consider e.g. the German word 'Abwasser|behandlungs|anlage' (sewage water treatment plant) or Czech 'velko|výroba' (mass-production).

[4] `https://fasttext.cc/`

In recent years, several segmentation algorithms have been proposed, however, only the byte pair encoding (Sennrich et al., 2016b) , wordpieces (Wu et al., 2016) and later SentencePiece (Kudo and Richardson, 2018) became widely used. We describe them in the next paragraphs.

It should be noted that linguistically more adequate approaches to word segmentation in NMT have been repeatedly explored, see e.g. Macháček et al. (2018). We disregard them in this book because they do not seem to perform any better than language-agnostic approaches, and primarily because they would be difficult to generalize across more languages.

**Byte-Pair Encoding** Using a word-based vocabulary in NMT leads to problems with OOV. Sennrich et al. (2016b) tackled this problem by segmenting the words into more frequent subword tokens with the use of byte-pair encoding (Gage, 1994).

**Byte-pair encoding** (BPE) is a simple data compression algorithm, which iteratively merges the most frequent pairs of consecutive characters or character sequences. A table of the merges, together with the vocabulary, is then required to segment a given input text.

The table of merges is generated in the following way. First, all characters from the training data are added into the vocabulary plus a special symbol for the word ending '$\langle /w \rangle$', which is used to restore original segmentation after the translation. Then we add the ending symbol to all words in the training set and separate them to individual characters. We iteratively find the most frequent symbol pairs and replace them with a new single symbol representing their concatenation. Each merge thus produces a new symbol that represents a character n-gram. We continue and stop when we have the same number of initial characters plus merges as is our desired size of the vocabulary. By this process, frequent words and word components become directly included in the vocabulary. A toy example is in Figure 3.2.

The merges are applied in advance on the training corpus by merging characters based on learned merges. The BPE segments the words into subword tokens, which can be used by NMT without any need for architecture modification. In other words, the NMT model handles subwords as regular words.

- Given a dictionary of token types and frequences.
  1. Replace the most frequent pair of characters with a new unit. (Record this merge operation.)
  2. Repeat until the desired number of merge operations is reached.

| Current vocabulary | The new merge |
|---|---|
| lo**we**r lo**we**st ne**we**r widest | we → we |
| lo**we** **r** lo we st ne **we** **r** widest | we r → we r |
| lo we r lo we **st** ne we r wide**st** | st → st |

- New input: Apply the *recorded sequence* of merges:
  newest → ne we st → ne we st ⇒ n@@ e@@ we@@ st

**Figure 3.2:** An example of BPE encoding construction and application.

In practice, the symbol for the end of the word is not produced during segmentation. Instead, a '@@' is added to all subword tokens that end in the middle of a word. For example, the word 'newest' is segmented into 'n@@ e@@ we@@ st', see Figure 3.2.

Sennrich et al. (2016b) also showed that using joint merges, generated from concatenated training sets for both the source and the target language, is beneficial for the overall performance of NMT. This improved consistency between the source and target segmentation is especially useful for the encoding of named entities, which helps NMT in learning the mapping between subword units.

BPE implementation has several disadvantages. It can not address well languages that do not use a space as a separator between words, for example, Chinese. It fails when encoding characters that are not contained in the vocabulary, for example, foreign words written in a different alphabet. Lastly, BPE algorithm relies on a tokenizer. Without its use, the punctuation attached directly to words would have different word segmentation than when separated. The wordpiece method (Wu et al., 2016) solves all these problems. We describe it in the next section.

**Wordpieces** Wordpiece is another word segmentation algorithm. It is similar to BPE and is based on an algorithm developed by Schuster and Nakajima (2012). Wu et al. (2016) adopted the algorithm for NMT purposes. We describe the algorithm in comparison to BPE.

The wordpiece segmentation differs mainly by using language model likelihood instead of highest frequency pair during the selection of candidates for new vocabulary units. Secondly, it does not employ any tokenization leaving it for the wordpiece algorithm to learn.

The algorithm works by starting with the vocabulary containing individual characters and building a language model on the character-segmented training data. Then it adds a combination of two units from the current vocabulary by selecting the pair of units that increases the likelihood on the training data the most, continuing until the vocabulary contains the predefined number of subword units.

The iterative process would be computationally expensive if done by brute-force. Therefore the algorithm uses several improvements, e.g. adding several new units at once per step or testing only pairs that have a high chance to be good candidates.

The segmentation works in a greedy way when applying. It finds the longest unit in the vocabulary from the beginning of the sentence, separates it and continues with the rest of the sentence. This way, it does not need to remember the ordering of merges; it remembers just the vocabulary. This makes it simpler than BPE.

The Tensor2tensor (Vaswani et al., 2018) framework slightly improves the wordpiece algorithm by byte-encoding OOV characters, which makes any Unicode character encodable. It uses an underscore instead of '$\langle/w\rangle$' as an indication of the word endings.

Furthermore, the implementation by Vaswani et al. (2018) optimizes the generation by counting frequencies for only a small part of the corpus. In the experiments described below, we prefer a more stable segmentation, so we created vocabularies from the first twenty million sentences. Additionally, Vaswani et al. (2018) introduce a 1%[5] tolerance for the final size of the vocabulary. Therefore instead of having 32000 subwords,[6] the vocabulary has between 31680 and 32320 items. For details see the source code.[7]

**SentencePiece** SentencePiece (Kudo and Richardson, 2018) is an implementation of language independent end-to-end subword tokenizer. It implements BPE (Sennrich et al., 2016b) and subword regularization (Kudo, 2018). It has useful features especially for multilingual NMT, e.g. guaranteed coverage of all Unicode characters and construction of frequent sentence pieces even across word boundaries, so e.g. "Mr. President" can become a single token.

## 3.3 Processing Sentences

As discussed in Section 3.1, machine translation has been always geared towards processing sentences, i.e. arbitrarily (but reasonably) long sequences of tokens. Arguably, sequences of varying length are not the best suited type of input for neural networks with their fixed-size vectors and matrices.

Standard books on neural networks will cover **convolutional neural networks (CNN)** which "condense" an arbitrarily long sequence into a shorter one using an aggregation function such as summation of maximization and **recurrent neural networks (RNN)** which learn to digest an arbitrarily long input sequence to a fixed-size vector one symbol at a time. Neither of these techniques is ideal for processing sentences, so we skip CNNs altogether, gloss over RNNs and describe the "self-attentive" Transformer model instead.

[5] The implementation in T2T tries to create vocabulary several times, and if it fails to create a vocabulary within this tolerance, it uses the generated vocabulary with the closest size.

[6] We use exactly 32000 as the desired vocabulary size instead of $2^{15} = 32768$.

[7] `https://github.com/tensorflow/tensor2tensor/blob/v1.8.0/tensor2tensor/data_generators/text_encoder.py#L723`

### 3.3.1 Sequence-to-Sequence Models

One of the first end-to-end NMT systems were Sutskever et al. (2014) and Cho et al. (2014b). The authors of both works used a recurrent neural network with LSTM cells that processes one word at a time until it reads the whole input sentence. Then a special symbol "<start>" is provided and the network produces the first word based on its inner state and the previous word. This generated word is then fed into the network, and the second word is generated. The process continues until the model generates the "<end>" symbol.

The main disadvantage of the works of Sutskever et al. (2014) and Cho et al. (2014b) is that the network has to fit the whole sentence into a single vector of 300–1000 elements, essentially a sentence embedding, before it starts generating the output. Therefore Bahdanau et al. (2014) proposed the so-called attention mechanism. The attention mechanism gives the network the ability to reconsider all input words at any stage and use this information when generating a new word.

Gehring et al. (2017) redesigned the previous architecture with a convolutional neural network, which handles all input words at the same time, therefore making the training and inference process faster.

In the same year as Gehring et al. (2017), another complete model redesign was proposed by Vaswani et al. (2017). The so-called Transformer architecture avoids both RNNs and CNN and instead uses feed-forward layers. Transformer fully dominated the field of NMT and we describe this architecture in detail in the next section.

### 3.3.2 Transformer Model

The **Transformer** architecture (Vaswani et al., 2017) consists of an **encoder** and **decoder**, similarly to the previous approaches. The encoder takes the input sentence and maps it into a high-dimensional state space. Its output is then fed into the decoder, which produces the output sentence. However, instead of going one word at a time from the left to the right of a sentence, the encoder sees the entire input sequence at once. This makes it faster in terms of training and inference speed in comparison to previous neural architectures because it allows better usage of parallelism. The decoder remains "autoregressive", i.e. always producing the output symbol with the knowledge of the previously produced output symbol. Non-autoregressive models (Libovický and Helcl, 2018) are still an open research question, although some progress is apparent (Agrawal et al., 2021).

Note that since 2017, a multitude of larger or bigger modifications of the original Transformer architecture as discussed here were proposed. See Lin et al. (2021) for a recent survey.

### Transformer Attention

Bahdanau et al. (2014) introduced the idea of **attention** mechanism to address an important problem of RNNs: words more distant in the preceding processing steps were "forgotten" and not influencing the translation enough. The attention allowed the decoder to look back at any position, so that no piece of information had to fade away. Technically, the attention of Bahdanau et al. (2014) was nothing more than a weighted combination of encoder states that the decoder was consulting at its every step. The decoder was also dynamically changing these weights to "focus" on different parts of the input.

The novel idea of self-attention in Transformer is to use the attention also within the encoder and decoder themselves. In other words, the attention allows the model to interpret the word it is currently processing in the context of relevant words from its surrounding.

For example, when processing the sentence "The kitten crawled over the room because it was hungry.", NMT needs to know the antecedent of the word "it". The self-attention mechanism solves this problem by incorporating the information into the representation of the word "it" at deeper layers of the encoder.

In general form, the Transformer attention function uses three vectors: queries (Q), keys (K) and values (V). The output is a weighted sum of values, where weights are computed from queries and keys. The attention is defined as follows:

$$\text{Attention}(Q, K, V) = \text{softmax}\left(\frac{QK^T}{\sqrt{d_k}}\right)V \tag{3.3}$$

where $d_k$ is the square root of the dimension of the key vectors, which is normalization necessary to stabilize gradients.

The intuition behind the attention is that we get a distribution over the whole sequence using the dot product of queries (which can be understood as hidden states representing all positions in the sequence) and keys followed by softmax. This distribution is then used to weight the values (another automatically established derivative hidden representation of every position, similarly like queries). The weighted combination is a vector, where features of relevant words are stressed. Which words are deemed relevant is defined by the match of keys and queries and what features should be used in the subsequent computation is defined by the values.

The attention is used separately in each of encoder and decoder as a "**self-attention**" where all queries, keys, and values come from the previous layer. It is also used in encoder-decoder attention, dubbed **cross attention**, where queries and keys come from encoder and values from the decoder.

### Multi-Head Attention

Having only one attention, Transformer would focus solely on some positions in the previous layer, leaving other relevant words ignored or conflating mutually irrelevant aspects into one overused attention. Transformer model solves this by using several heads within each layer, each with its own keys, queries and values. This allows concurrent observation of different aspects of the input.

Formally, the **multi-head attention** is defined as follows:

$$\text{MultiHead}(Q, K, V) = \text{Concatenate}(\text{head}_1, ..., \text{head}_h)W^O \tag{3.4}$$

$$\text{where head}_i = \text{Attention}(QW_i^Q, KW_i^K, VW_i^V) \tag{3.5}$$

where the projection matrices $W^{Q/K/V}$ are trainable and different for each attention head, and $h$ is the number of heads. The "Transformer-big" configuration has 16 heads. The concatenation in multi-head attention is then linearly projected by a matrix $W^O$.

Because the heads are trained unsupervised, their functions can vary a lot. For instance, Garg et al. (2019) use one of the cross attention heads to force the network to learn and produce the alignment between source and target tokens. We discuss the use on an encoder head to learn and predict syntactic structure of the source in Section 5.4.1. See the book by Mareček et al. (2020) for further examples of analyses of the attention.

### Positional Encoding

With the move from the recurrent structure to self-attention, an important piece of information is lost from the model: the information about the position of each word. This is solved by adding a special positional encoding to all input words, which allows Transformer to explicitly handle word order.

The absolute position encoding (PE) for position $w$ is a vector with each element $i$ defined as:

$$\text{PE}_{(w,2i)} = \sin(w/10000^{2i/d_{\text{model}}}) \tag{3.6}$$

$$\text{PE}_{(w,2i+1)} = \cos(w/10000^{2i/d_{\text{model}}}) \tag{3.7}$$

where $d_{\text{model}}$ is the dimension of the vector. In other words, each dimension of $\text{PE}$ corresponds to a sigmoid with a different frequency.

In Section 3.2, we explained how discrete words are mapped to embedding vectors. The positional encoding is added to the word embeddings and used as the input to the first layer of Transformer.

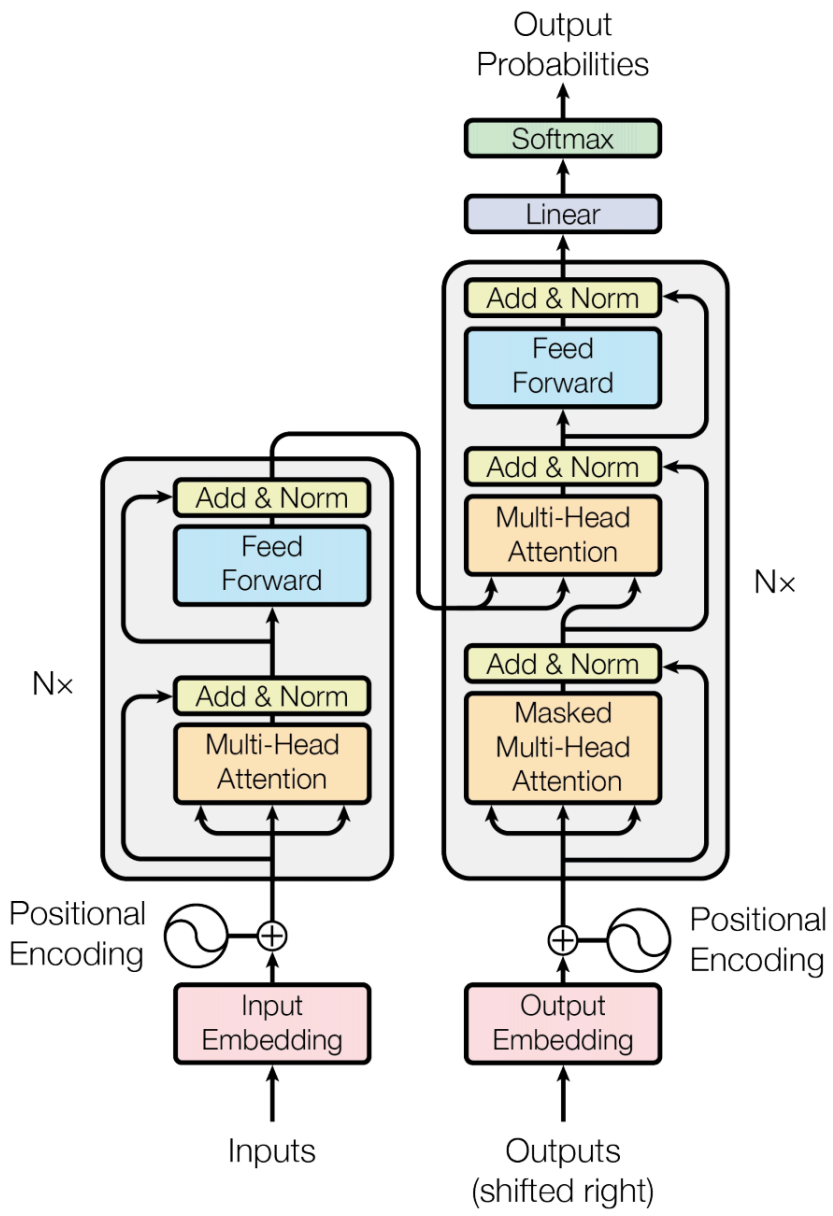

**Figure 3.3:** Transformer architecture. Reproduced from Vaswani et al. (2017).

While the formulas for PE seem to suggest that the model can understand various distances in the sentence by considering different elements (i.e. sigmoid frequencies), trainable random positional encodings works as well (Wang et al., 2020a).

### Summary of Transformer Architecture

The complete Transformer architecture is illustrated in Figure 3.3. Except parts mentioned above, there is a residual connection after each multi-head attention, which sums input of multi-head attention with its output followed by layer normalization (Ba et al., 2016; labeled as "Add & Norm" in Figure 3.3). The model stacks N layers of multi-head attention on top of each other, with position-wise feed-forward layers after the attentions. In the original model, six layers are used.

The output of the decoder is finally modified by linear transformation followed by the softmax function that produces probabilities of words over the model vocabulary.

For further reading, see the original paper (Vaswani et al., 2017) or various blog posts describing the model.[8]

[8] `http://jalammar.github.io/illustrated-transformer/`

## 3.4 Input and Output Versatility

Neural networks have been used in a wide range of tasks with inputs and outputs of diverse types, including:

- Scalars – one number as a score of a language model, confidence or quality estimation score, binary classification, or regression, e.g. sentiment analysis.
- Vectors – e.g. an autoencoder transforms text sequences, documents or images into vectors for compression or further processing, e.g. classification, clustering, etc.
- Classes – a result of any classification task, e.g. language identification, stance detection, etc.
- Text sequences – machine translation, simplification, style transfer (e.g. formal tone to informal), question answering, and any tasks that can be represented as a text sequence, e.g. lemmatization, tagging, parsing, alignment, etc.
- Audio – speech recognition and synthesis, text-to-speech, speech-to-text and speech-to-speech translation, etc.
- Images – e.g. optical character recognition and any other image processing tasks
- Graphs or other structures – e.g. a syntactic parse tree, word alignment, a lattice of top hypotheses from speech recognition, etc.

Additionally, N-best lists of possible outputs can be useful in various applications, e.g. NMT can produce list of N best translations instead of the single best one. Such a list can be then exploited, e.g. by a CAT tool, or reordered by another neural network.

The input and output types can be repeated, combined or mixed in one neural network. For example, the quality estimation task (Specia et al., 2020) is processing two text sequences (the source and its machine translation) into a scalar that represents the MT quality score. Multi-modal MT (Sulubacak et al., 2020) processes e.g. an image and its caption, and produces the caption translation in the context of the image.

The versatility actually goes one step further in that the network is often capable of learning to magically *convert* among the types as needed. These conversions have, of course, some limitations and a cost, so e.g. converting between unary and decimal number representations can be too difficult for a given network structure and prevent it from learning the main task. With a reasonable representation, we see many papers successfully relying on this conversion in the sequence-to-sequence tasks. One interesting example is a sequence-to-sequence model learning to carry out symbolic mathematical operations on algebraic expressions such as derivation or integration (Lample and Charton, 2020), outperforming common tools like Matlab or Mathematica.

In the area of NLP, many "supplementary" pieces of information can be directly put into the sequences of input tokens or sought for in the output sequences. For instance some sentence class tokens can be prepended or appended to the source or target sequence, words can be interleaved with tokens expressing various aspects of formal linguistic description or other token-level information. Instead of designing the network to process two sequences of inputs, one can simply concatenate these two sequences (with or without an explicit delimiter) and the network will easily identify them as observed already by Niehues et al. (2016), see Figure 3.4.

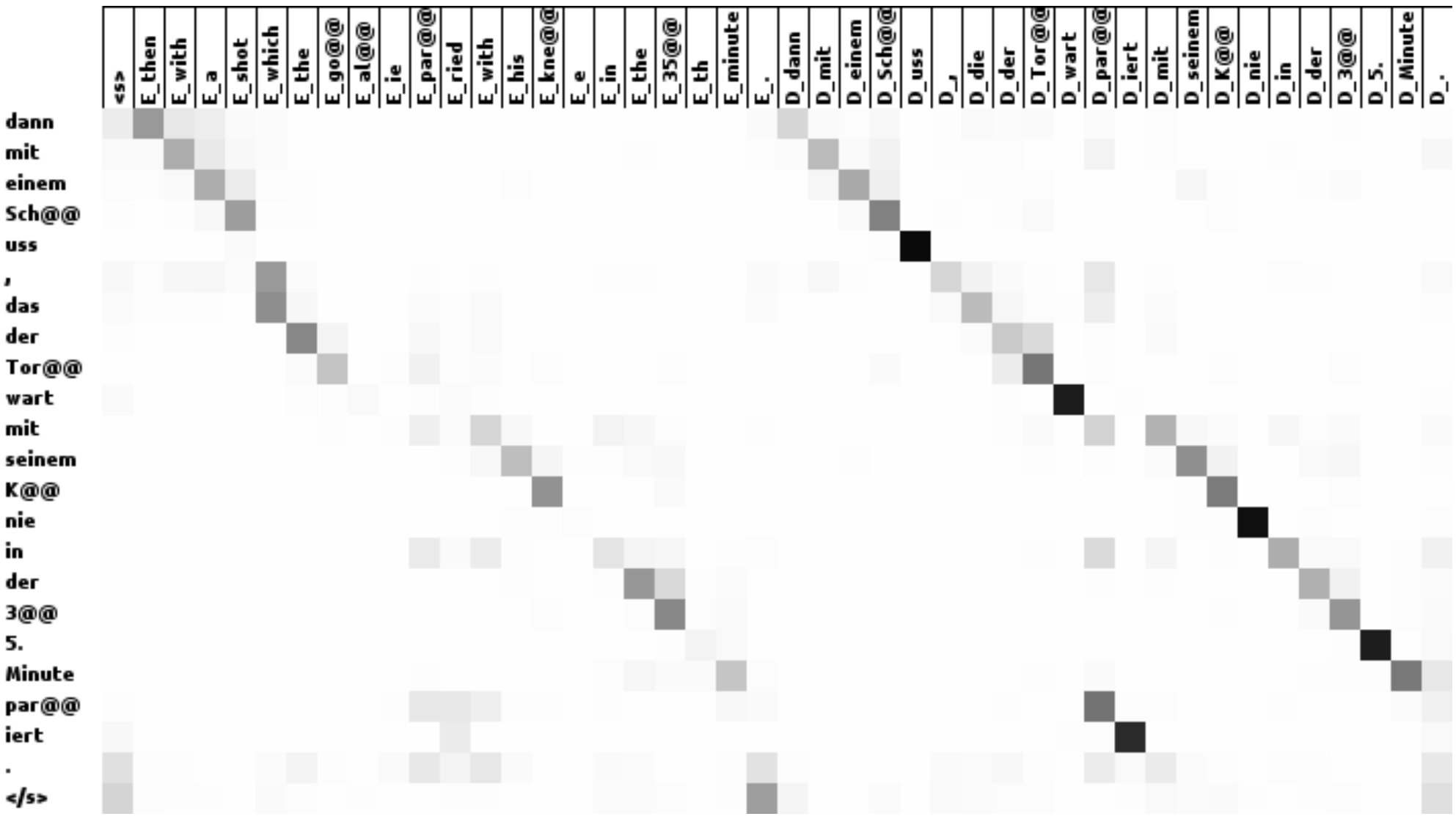

**Figure 3.4:** Visualization of the attention in an RNN-based NMT model where the input contains the concatenation of the source sentence and its pre-translation by a PBMT system. The NMT learned to follow the source and pre-translation sequences in parallel, sometimes paying more attention to the source token, sometimes to the target token. The example also illustrates that source tokens (top row) are annotated with the language they come from ("E" for English, "D" for German) so that the vocabularies are kept distinct. Reproduced from Niehues et al. (2016).

## 3.5 Multi-Tasking

The idea of transferring knowledge between different algorithms is as old as machine learning itself, and researchers were trying to reuse previously trained features on different tasks. Pratt et al. (1991) published a paper properly formulating the transfer learning problem by pretraining a neural network on a different task.

**Multi-tasking** or multi-task learning refers to the ability of a neural network to process more than one task. For example, when an NMT produces text sequence as a translation and a confidence score as a scalar, it can be considered multi-tasking. For a comprehensive survey on multi-tasking in NLP in 2020's, we refer to Worsham and Kalita (2020). Multi-tasking has been studied even before the deep learning era, in classical machine learning (Caruana, 1997).

The motivation for multi-tasking is knowledge transfer and generalization across the tasks. It is supposed to be more beneficial than training NN for the tasks separately. The tasks in multi-tasking should be therefore mutually related and somehow similar to each other, such as MT and linguistic annotation tasks (lemmatization, POS tagging, syntactic and semantic parsing, named entity recognition; Eriguchi et al., 2017; Nădejde et al., 2017; Niehues and Cho, 2017; Zaremoodi and Haffari, 2018), or MT from or into multiple languages (Johnson et al., 2017; Wang et al., 2020b).

Luong et al. (2016) proposed one of the first multi-task neural approaches in NLP, where they combined tasks of machine translation, part-of-speech tagging, autoencoding[9] and image captioning. Different tasks may require different encoders or decoders but whenever possible, these components are shared across more tasks. For instance, the image captioning processes images with its custom encoder but produces English text and thus the decoder is shared with machine translation into English. The network is still the very first sequence-to-sequence model using RNN without attention. Another confounding factor, from current point of view, is that they still do not use any subword units and instead limit the vocabulary to the 50k most common word forms. The behaviour of BLEU in this setting is not very well known.

Luong et al. (2016) observe a substantial improvement in English→German translation accuracy by adding a secondary decoder for predicting the parse of the source English. The tasks are trained in an alternating way. Interestingly, they gain most when the parsing is included in 1 out of only 101 training batches. With more room for parsing, the gain decreases. In Section 5.4, we observe that regularizing attention of the Transformer model via multi-tasking with parsing has a similar effect and the true syntax is actually not the reason for the improvement. It is unclear where the gain comes from in Luong et al. (2016); aside from the hoped-for better understanding of source syntax, it could be e.g. due to some global properties like more explicit modelling of sentence length distribution in the English encoder.

For German→English translation, Luong et al. (2016) observe gains by an additional task of image captioning, sharing the decoder. The captioning task brings another 0.5M of English sentences, increasing the target language data by 11%. This extended dataset can improve language modelling of English.

### 3.5.1 Multi-Tasking Benchmarks

Specific examples of multi-tasking in NLP are benchmarks such as decaNLP, (McCann et al., 2018), GLUE (General Language Understanding Evaluation Wang et al., 2018) or SuperGLUE (Wang et al., 2019a). DecaNLP represents 10 tasks as question answering problems. Their list and evaluation metrics are in Table 3.1.

[9] **Autoencoding** is encoding source into an intermediate representation and than back into the source. The intermediate vector or matrix representation may be useful in some applications.

| Task | Dataset | Evaluation Metric |
|---|---|---|
| Question Answering | SQuAD | Normalized F1 |
| Machine Translation | IWSLT | BLEU |
| Summarization | CNN/DM | ROUGE |
| Natural Language Inference | MNLI | Exact Match |
| Sentiment Analysis | SST | Exact Match |
| Semantic Role Labeling | QA-SRL | Normalized F1 |
| Zero-Shot Relation Extraction | QA-ZRE | Corpus-Level F1 |
| Goal-Oriented Dialogue | WOZ | Dialogue State Exact Match |
| Semantic Parsing | WikiSQL | Logical Form Exact Match |
| Pronoun Resolution | MWSC | Exact Match |

**Table 3.1:** Tasks and their evaluation metrics in decaNLP. Reproduced from McCann et al. (2018).

Recently, multi-tasking benchmark datasets started targeting also cross-lingual and language-universal processing, see XTREME (Hu et al., 2020)[10] or XTREME-R (Ruder et al., 2021). These benchmarks cover a relatively wide range of typically classification tasks such as prediction whether a pair of sentences is a pair of paraphrases, or natural language inference, i.e. prediction whether one of the sentences implies the other. Some of the tasks have more complicated outputs, such as sequence labelling tasks (e.g. named entity detection or part-of-speech tagging) or question answering formulated as the identification of the span in a text which represents the answer (SQuAD, Rajpurkar et al., 2016). Multilingual versions of these benchmarks have these tasks covered in many languages spanning multiple language families. Translations itself is typically not among the examined tasks, although e.g. XTREME-R includes two tasks that expect the model to identify translations of sentences in a large collection of candidates.

### 3.5.2 Task Related Topics

**Task Relatedness** Intuitively, the more related tasks, the higher gains from multi-tasking could be expected. However, the task relatedness can not be assessed easily in NLP. Tan et al. (2019) compare two clustering methods for joint learning of multi-lingual NMT. The automatic clustering method based on embeddings from a universal multi-lingual NMT model that requires costly training the model first, outperforms the prior expert knowledge of language relatedness. The optimal selection of NLP tasks for multi-tasking requires experimenting and empirical evaluation.

[10] `https://sites.research.google/xtreme`

**Task Importance** From the user's point of view, the tasks can be either of the same importance, for example MT into language 1 as the first task, and into language 2 as the second task, or there can be a hierarchy of primary and auxiliary tasks, e.g. MT primary and POS tagging auxiliary. The users usually do not care about the results of the auxiliary tasks, they use them only to enhance the quality of the primary ones.

**Scheduling** From NN's point of view, the importance of the tasks is deduced from the loss function. The overall loss function can be a sum of task specific losses. Then, multiplication factors can balance the particular task losses. Other method for balancing is oversampling or undersampling the training instances of particular tasks.

Kiperwasser and Ballesteros (2018) propose a scheduler that continuously changes the focus between tasks during training, from auxiliary to primary task.

With the current behaviour of NNs, as we will discuss in Chapter 4 such as catastrophic forgetting, scheduling is one of the critical problems of multi-lingual models addressing many languages, see Chapter 8.

### 3.5.3 Multi-Tasking Architectures

Multi-tasking NN model can be either technically identical to the single tasking model (so called **universal model**; Johnson et al., 2017), or have task-specific components, such as multiple encoders and decoders, in addition to the components that are shared for all tasks. See sharing options in e.g. Niehues and Cho (2017) and illustration in Figure 3.5.

With the task specific components, the user has a control for distributing the network capacity to the tasks, while with the universal model, the network controls it fully on its own. Both methods have their challenges. In the universal model, the easier tasks can outweigh more difficult ones, and degrade their performance. With the non-universal model, the challenge is to find an appropriate component size for each task. Too small size may lead to outweighing by the easier task, while too large leads to longer training and suboptimal generalization across the tasks.

**Task Identification** With the universal multi-tasking model, there is a question how the model recognizes the task to process. A very simple but effective method by Johnson et al. (2017) is representation in data. They insert a task identification token into the source sequence. In case of multi-lingual translation by Johnson et al. (2017), the token has a form e.g. `<2en>`, and meaning "translate from German (de) to English (en)".

The next option is to always process all the tasks at once, by interleaving two sequences into one, for example CCG supertags and MT. See e.g. Nădejde et al. (2017) and Figure 3.6.

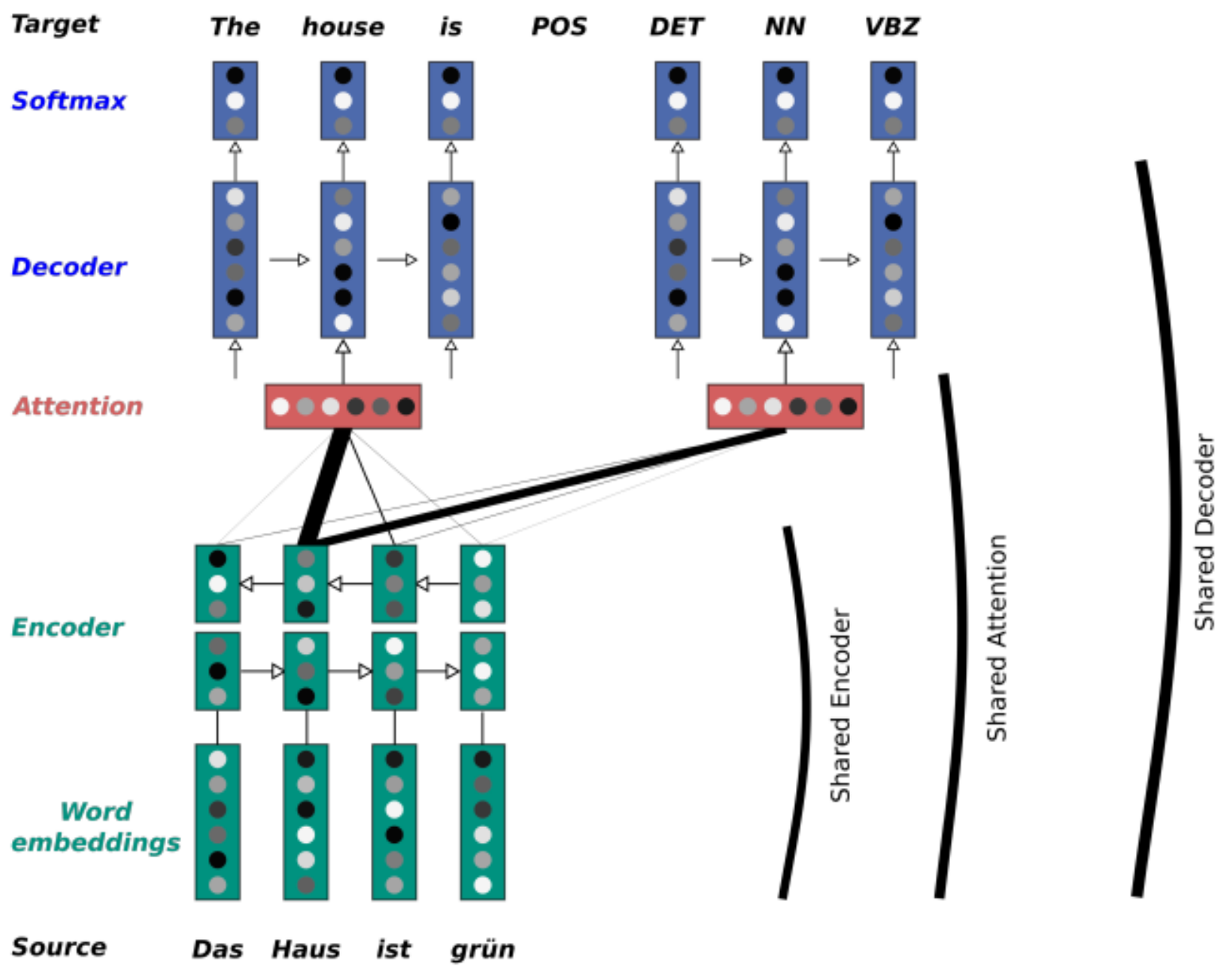

**Figure 3.5:** Several options of sharing components for two-task NN. The first task is German →English MT, and the second task is POS tagging of the English target translation. Figure reprinted from Niehues and Cho (2017).

**Alternating vs Joint Multi-Tasking** The model architecture allows either alternating between the tasks, so that only one task is handled in one decoding step, or joint processing, where multiple tasks are processed within a single decoding process. For the latter, all inputs have to be present and all outputs are concurrently produced.

Multi-way processing discussed in Chapter 6 is a particular example of alternating multi-tasking and it is a common technique in multilingual MT, see e.g. Johnson et al. (2017). Multi-source and multi-target models, on the other hand, are examples of joint multi-tasking.

Pham et al. (2019) show examples of both alternating and joint multi-tasking. An illustration of the latter one is in Figure 3.7.

## 3.6 Back-Translation

Parallel data are usually small and expensive, in contrast to monolingual, untranslated, texts. **Back-translation** is a method to exploit monolingual data for high quality MT. You only need an MT system to synthesize the “missing side of the parallel cor-pus", i.e. to make the authentic monolingual data "pseudo"-parallel. Back-translation refers to the fact that we trust more in the quality of the authentic texts and therefore put them on the more important target side of the synthetic parallel corpus.

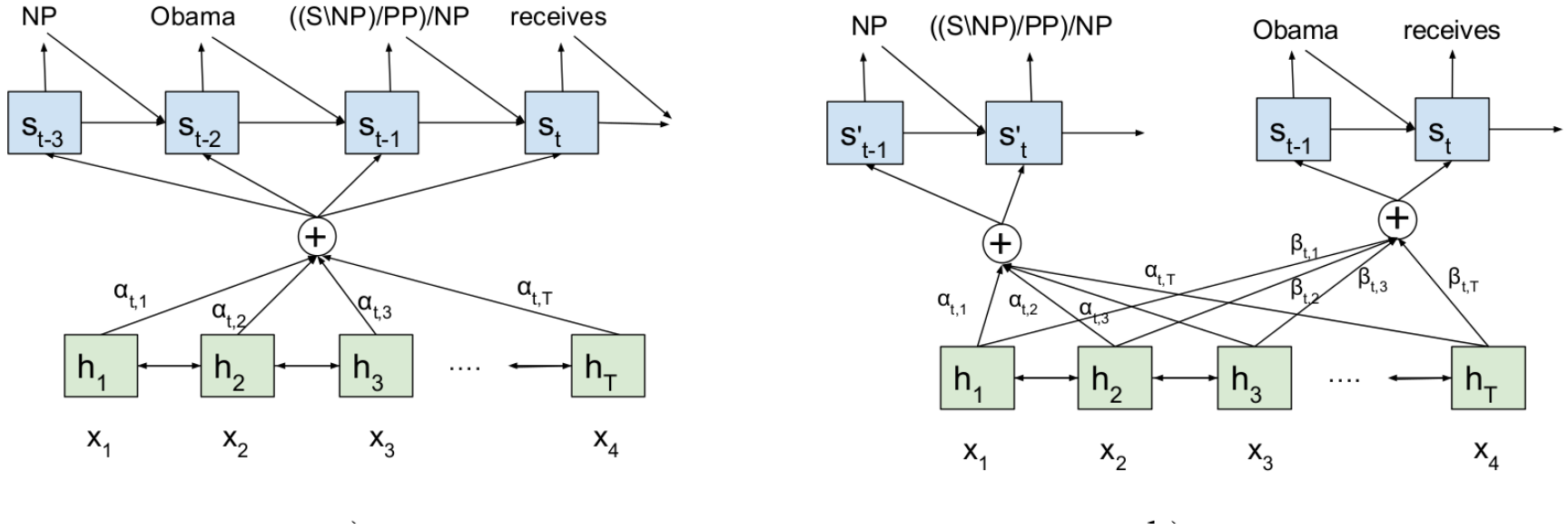

**Figure 3.6:** Left: Interleaving two tasks in a universal model with one encoder and one decoder. The tasks are CCG supertags and MT. Right: The same two tasks with non-universal model that has one shared encoder, and two decoders, one for each task. Figure reprinted from Nădejde et al. (2017).

The target side is then an authentic sample of human language, text created by humans. The source side is synthetic, created by MT that might be flawed, ungrammatical into some extent, but hopefully close to human language, so that when a machine is trained on it, it understands the authentic source language in production.

So, to create an MT system for a language pair A→B by back-translation, you need monolingual data of B language, and B→A MT. If you don't have it, or the MT is of inferior quality, but you happen to have at least A monolingual data, use back-translation first to get B→A MT to get synthetic data to get your desired A→B MT. In other words, back-translation process can be iterated. As a fallback MT at the back of back-translations use MT trained on the small parallel data.[11] See the illustration in Figure 3.8.

So far in this section, we have used a generic term MT, and not specifically NMT. Back-translation is applicable both for PBMT, as e.g. Bojar and Tamchyna (2011) showed in 2011. Back-translation introduced by (Sennrich et al., 2016a) into NMT was one of the major moments of NMT, because it finally allowed to use the available large monolingual data. In fact, back-translation is just one particular data augmentation technique. Data augmentation is widely used for many tasks in machine and deep learning. See e.g. survey of data augmentation for NLP in Feng et al. (2021).

[11] Or unsupervised MT (Lample et al., 2018).

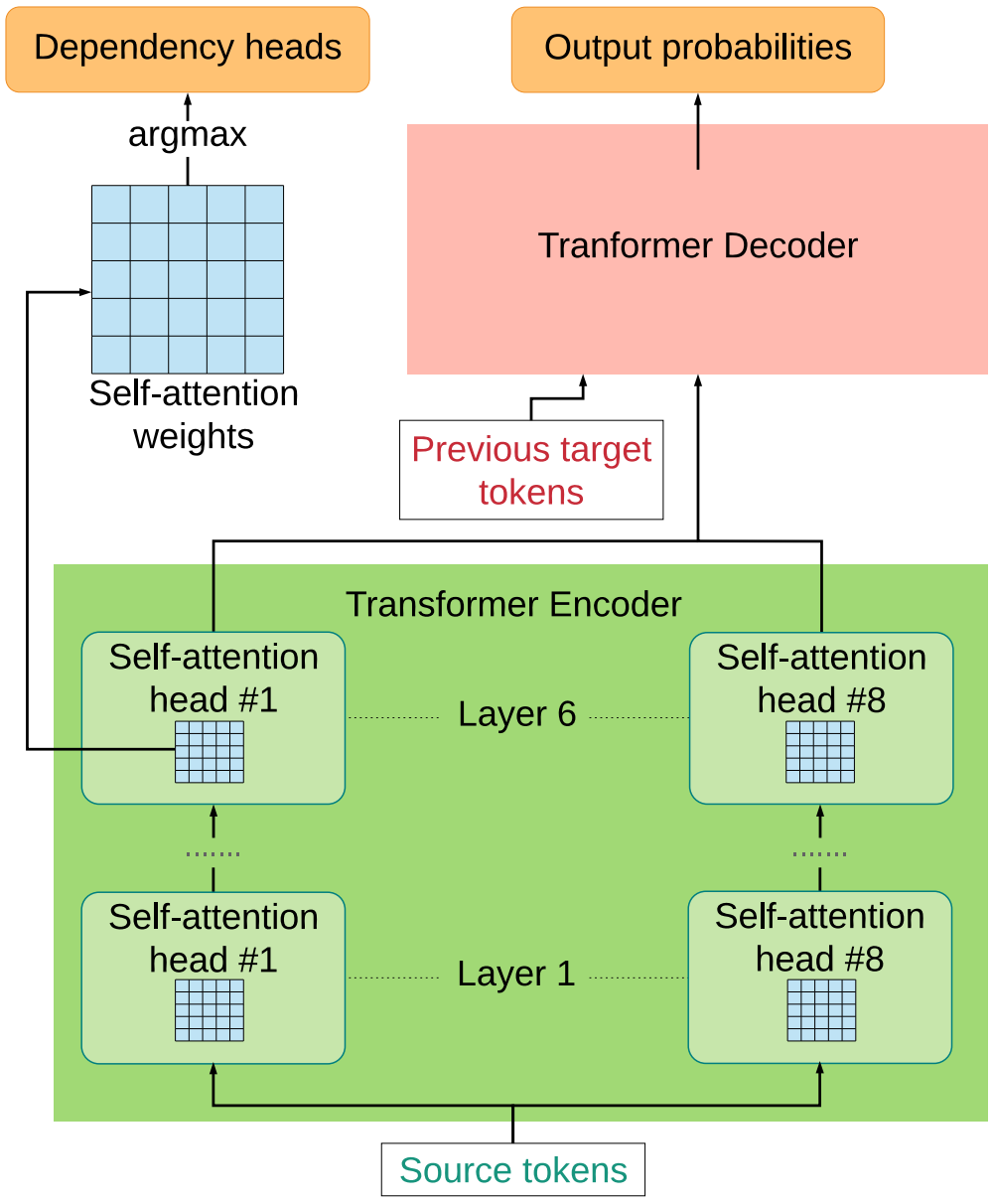

**Figure 3.7:** Example of joint multi-tasking in Transformer NMT. The first task is MT, and it is provided by a standard Transformer decoder. The second task is dependency parsing of the source. It is retrieved as a matrix of probabilities being a head in the parsing tree, from the 1st attention head in the 6th encoder layer. Figure reprinted from Pham et al. (2019).

Although back-translation on its own is not a technique that leverages more than two languages, it is still worth mentioning in this book. Besides others, also from the following reasons:

**Strong Baselines and Applicability of Research** A realistic baseline in your research should take into account the publicly available MT systems[12] that you, or the one who considers applying your researched method in practice, can use for back-translation.

Disregarding back-translation in research is, of course, simple, cheap and fast. On the other hand, you may spend your time on a complicated solution of a problem of your weak baseline that a strong back-translated baseline does not face. The applicability of your research may be limited.

Furthermore, if you see an improvement on a weak baseline, your method may hopefully work even on the strong baseline, but still, it is not certain. E.g., Denkowski and Neubig (2017) compared three methods with a weak and strong baseline, and only one of them was beneficial with the stronger one.

[12] Consider e.g. the MT models at `https://huggingface.co/models`.

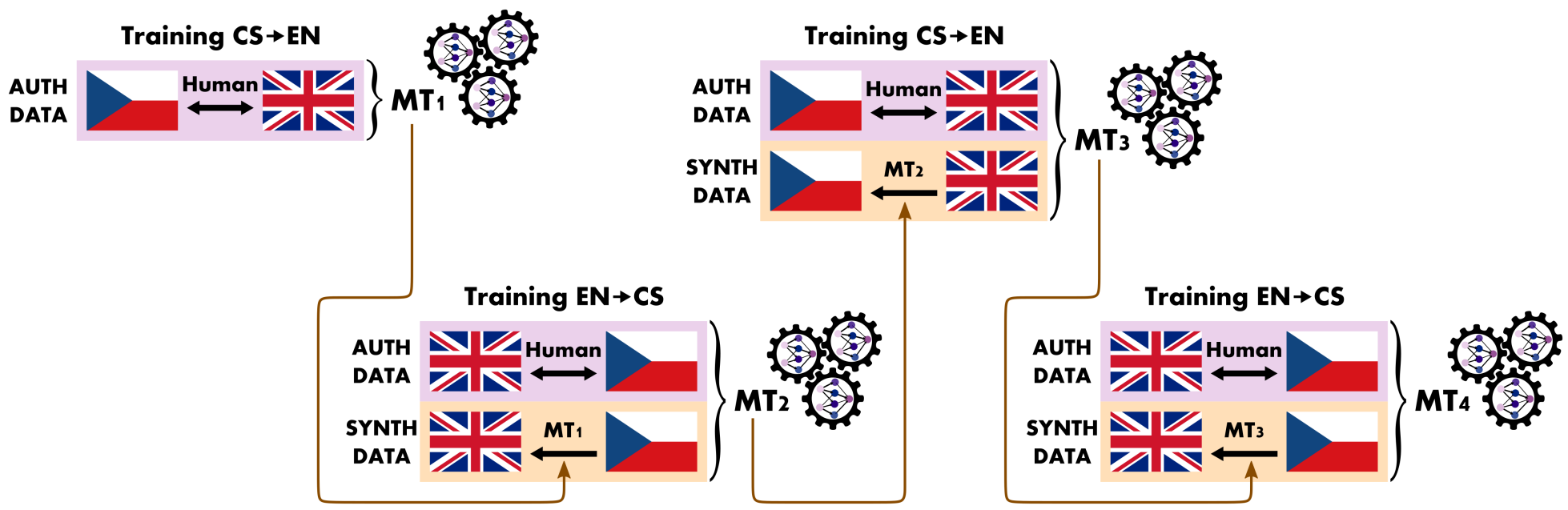

**Figure 3.8:** Diagram of iterated back-translation: the system MT1 trained only on authentic parallel data is used to translate monolingual Czech data into English, which are used to train system MT2; this step can be iterated one or more times to obtain MT3, MT4, etc. Figure and caption reprinted from Popel et al. (2020).

An easy solution for the this issue is to ask research questions that are realistic even for the current state-of-the-art baselines. The annual shared tasks (WMT: Akhbardeh et al., 2021, IWSLT: Anastasopoulos et al., 2021, etc.) are supposed to be good sources of examples.

Another solution is to propose language-independent methods. Then you can hope that between the 7 000 human languages and countless domains and special cases someone finds your method useful and applicable. Or, you can at least inspire someone else who proposes an applicable method.

**Combination of Techniques** Back-translation is applicable together with multilingual and many other techniques. An example of such a combination in non-multilingual NMT is in Figure 3.9: Popel et al. (2020) examine the combination of averaging eight last checkpoints[13] with two types of back-translated data usage: alternating blocks of synthetic and authentic data in training (block-BT) compared to the standard approach of mixing them (mix-BT).

We want to explicitly point out that when focusing for the highest quality MT, multilinguality may be not always the most advisable option. Simple techniques such as data filtering, back-translation, checkpoint averaging, ensembling, tuning for translationese (see Popel, 2018), etc., can be very effective.

[13] **Checkpoint averaging** is a technique to prevent overfitting to mini-batch. The trainable parameters of e.g. last 8 checkpoints saved each 10 minutes of training are averaged and saved as the model checkpoint for inference. It was used in Transformer NMT (Vaswani et al., 2017), and it is explained and investigated in Liu et al. (2018).

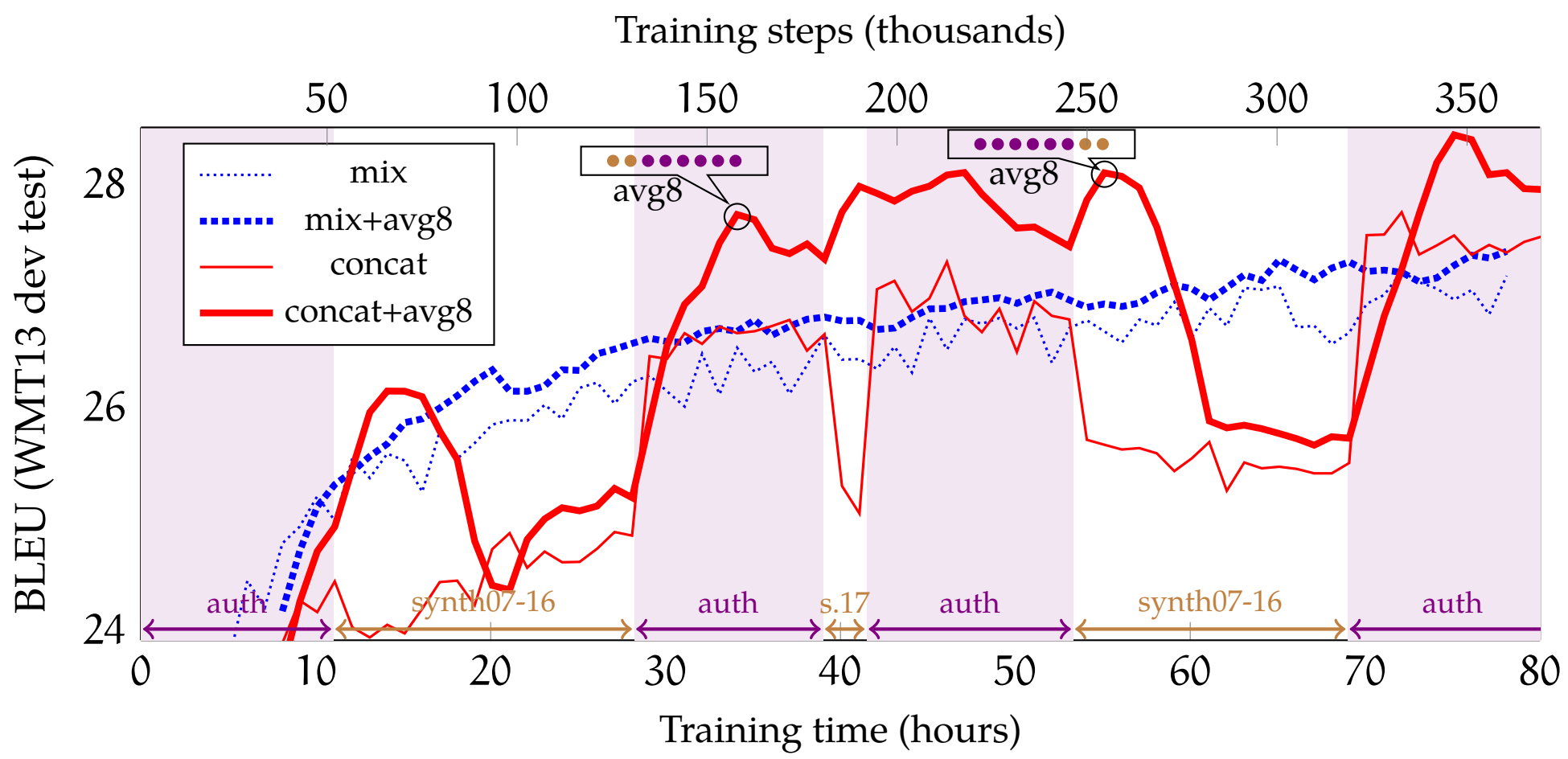

**Figure 3.9:** The effect of averaging eight last checkpoints with block-BT and mix-BT on the translation quality as measured by BLEU on the development set WMT13 newstest. The callouts (pointing to the initial and final peaks of the block-BT + avg8 curve) illustrate the 8 averaged checkpoints (synth-trained ones as brown circles, auth-trained ones as violet circles). Reproduced from the defense slides of Martin Popel, also appears in Popel et al. (2020).

On the other hand, back-translation is useful also in combination with multilingual MT. For example, Nishimura et al. (2018) propose to synthesize missing translations for multi-source NMT. Kocmi and Bojar (2019) combines back-translation with transfer learning. See more on the latter in Section 7.10.

# 4
# Learning Skills and Pitfalls of Neural Networks

In the previous chapter, we have seen the flexibility of neural networks. As if we were saying: create some structure of calculations, run a few epochs of training and the network parameters will settle to perform the task well. For sequence-to-sequence, we suggested: represent any potentially useful information as additional tokens, throw them into input or train the system to produce them in the output and the performance will grow thanks to the multi-tasking.

This chapter will highlight the *many* problems that you can run into when applying the above procedure. We start with a little caveat on differences among implementations (Section 4.1), then discuss the issue of data acquisition (Section 4.2) and then move to the equally important area of training (Section 4.3).

There is some black art to successful training of complex neural networks. According to Kuynghuyn Cho (p.c.), the first end-to-end neural success by Cho et al. (2014b) or Bahdanau et al. (2014) was possible not only because of the careful engineering but also thanks to some luck and unprecedented patience.

These days, the need for patience persists, and there is also a certain skill in deciding when to stop (Section 4.3.1). Furthermore, Transformer model tends to suffer from training instability (Section 4.4). The next problem we are going to expose is the surprising lack of generalization (Section 4.5), and then the catastrophic forgetting (Section 4.6).

Given our goal to work with more languages at once, we also provide a study on the "cost" of multi-task training in Section 4.7.

## 4.1 Devil in the Detail

The devil is often in the detail and while many differences in the mathematical formulation or the implementation of the discussed methods may be unimportant and magically resolved in the training, equally many subtle differences can and do play a critical role in replicability of the results and in achieving the expected performance.

NMT saw a quick development of machine translation toolkits. This is different from phrase-based machine translation where Moses (Koehn et al., 2007) was essentially the one and only widespread toolkit that was either used as a black box or to which everyone was contributing. A standard, "canonical" baseline was thus undisputable and knowledge and techniques were slowly piling up in Moses' repository.

NMT is much broader, with many larger teams building their own toolkits and with several people reporting to have implemented the attentional sequence-to-sequence over the weekend. The toolkits thus heavily vary not only in advanced features (for which one would pick one toolkit over another) but also in the tiny details of the very baselines.

The situation is not made any easier by the common habit to simplify the formulas in papers describing the methods, e.g. by omitting bias terms for presentation purposes. Bias terms however introduce further training parameters and can thus easily affect the final performance of the implementation. Replicating the score achieved by one toolkit using another one may be simply impossible due to the differences in the computation of the two toolkits.

We discussed different styles of subword representations in Section 3.2.2. While all the details on tokenization or word splitting may seem negligible to be mentioned in a paper, Macháček et al. (2018) observed that for an inflected language like Czech, the performance considerably differed when each word was appended by an underscore before the application of BPE, which allowed BPE to discover a more adequate vocabulary. The base form has one entry in this dictionary and the various suffixes, including the "empty suffix" one, are in the subsequent token. At the same time, the experiment was confounded by the fact that the gain was observed only when this underscore was not added to the final full stop.

## 4.2 Language Resources

Finding and collecting training data is a crucial step when creating MT systems. In general, we can classify textual data based on their domain, their size, and their quality. Another criterion is whether they are monolingual or parallel.

The availability of parallel texts, i.e. texts in two languages that correspond to each other, is crucial for NMT. When processed into an aligned form where we know for almost every sentence its translation, we talk about a **parallel corpus**.

It is expensive and hard to obtain large numbers of parallel sentences. On the other hand, it is easier to obtain monolingual texts by crawling the web or accessing various online sources.[1] The amount of available monolingual data in the target language typically far exceeds the number of parallel sentences. A whole new area of **unsupervised machine translation** has emerged that attempts to train translation systems using monolingual data only (Ravi and Knight, 2011; Lample et al., 2018), but we do not cover this area in the present book.

A comparably important criterion to the size of the parallel corpus is its domain and quality. It is well-known that domain-specific training data are better for the final performance than some general data. The same holds for the quality where a large, noisy training set can notably hinder the performance of an NMT system, see e.g. the survey by Chu and Wang, 2018.

[1] Printed documents are too difficult to access at a larger scale, although the Google Books effort (`https://books.google.com/`) should not be forgotten, including the simple corpus browser: `https://books.google.com/ngrams/`.

Whenever we talk about a language pair, we use a dash between the languages, e.g. Czech–English. On the other hand, whenever we talk about actual translation direction, we use an arrow to specify a direction from which language to which we translate, e.g. English→Czech. The translation direction is, of course, critical, even though both rely on the same parallel data because each direction exhibits its specific ambiguities and reordering phenomena. In the case of a parallel corpus, we often use "sentences" as a shortcut for "sentence pairs".

In this section, we introduce a definition of an MT domain, followed by an examination of the quality of parallel corpus and approaches for dataset cleaning.

### 4.2.1 What is Domain in Machine Translation?

The definition of a domain varies among papers. In general, **domain** is considered any set of instances from a dataset possessing a common feature. In most of the papers concerning domain adaptation, the authors define the domain as the source, an origin, of the dataset. The domain is also closely related to the topic or genre of the documents (Hildebrand et al., 2005; Chu et al., 2017; Servan et al., 2016). Examples of such domains are subtitles, literature, news, medical reports, patents, IT, and many more. All of them vary in the used vocabulary, style of writing, and content.

Another feature linked with the domain is the formality or informality of the documents, which is closely related to honorifics in languages like Czech, German, or Japanese (Sennrich et al., 2016c). It is a way of encoding the relative social status of speakers to the readers, and for many styles, like official documents. Correctly rendering honorifics and preserving formality of the language it is important for the quality of the translation.

Further, we can distinguish documents based on sentiment. The sentiment tone of a text can change with machine translation (Glorot et al., 2011; Mohammad et al., 2016) mainly because of language differences and ambiguity. Another issue with sentiment is that the same information can be written from a positive, neutral, or negative stance. We can go even further and distinguish documents based on the writing style of the author or expected style preference of the reader, as of formality of a speech, specialized vocabulary, or dialects (Jeblee et al., 2014).

It is important to realize that the domain is not a feature of individual sentences. There are many sentences which are fully neutral with respect to the domain, and there are many sentences or expressions that clearly indicate a domain.

While boundaries of domains are very soft in general, it is seen as a clear flaw of the translation if the domain consistency is not preserved throughout the document. It should be noted that while the current research is clearly moving towards document-level evaluation, the domain consistency is not explicitly evaluated in any way yet.

### 4.2.2 Definition of Low-Resource Languages

Recently, rapid development of NMT systems led to the claims that the human parity (performing on the same level as a human translator) has been reached on high-resource language pairs like Chinese–English (Hassan et al., 2018) or Czech–English (Bojar et al., 2018). However, NMT systems tend to be very data-hungry. Koehn and Knowles (2017) have observed that in the low-resource scenario, off-the-shelf setup of NMT lags behind the previous PBMT approaches. This problem led to the rise of interest in low-resource NMT.

A precise determination which language pairs should be counted as low-resource is a research question itself. One must consider all aspects of available language resources as well as the properties of the languages themselves.

One of the aspects is the domain of the parallel corpus. Having a large amount of domain-specific parallel sentences can be considered high-resource in the given domain, but low-resource in the general domain, where the performance can be terrible. For example, one common source of parallel sentences for low-resource languages is the Bible, which is translated into hundreds of languages (Christodouloupoulos and Steedman, 2015). However, it is a highly domain and style specific text.

Highly-inflected languages further complicate the definition of low-resource by presenting a notable sparsity problem with the various forms of inflected words and therefore requiring more parallel sentences to reach comparable performance as the translation of less inflected languages. This has been the case for statistical machine translation (Bojar, 2015) as well as for NMT (Denkowski and Neubig, 2017).

Gu et al. (2018) define the *extremely low-resource scenario* by the minimal amount of data needed for NMT to obtain a reasonable translation quality. They showed that extremely low-resource scenario could be considered up to the 13-28k parallel sentences for English→Romanian translation.

In recent years, researchers have organized several machine translation shared tasks aimed specifically at the low-resource scenario. Niehues et al. (2018) introduced a task on Basque→English low-resource translation with an available in-domain corpus of 6k sentence pairs and 940k out-of-domain sentence pairs. The low-resource translation tasks in WMT 2018 (Bojar et al., 2018) have been Estonian–English with 880k and Turkish–English with 208k parallel sentences. In WMT19 (Bojar et al., 2019), the low-resource language was Gujarati–English with 170k parallel sentences and Kazakh–English with 220k parallel sentences. In 2020, the task of low-resource was merged with the task on unsupervised MT (Fraser, 2020) with Upper Serbian↔German as the translation directions (about 60k sentence pairs). The same year, the news task (Barrault et al., 2020) also included low-resource languages, paired with English: Inuktitut (1.5M pairs), Khmer (300k in a curated corpus and 4M in a noisy corpus from a previous corpus cleaning task Koehn et al., 2018), and Pashto (125k pairs curated and 1M noisy pairs). Similar interest was apparent in 2021.

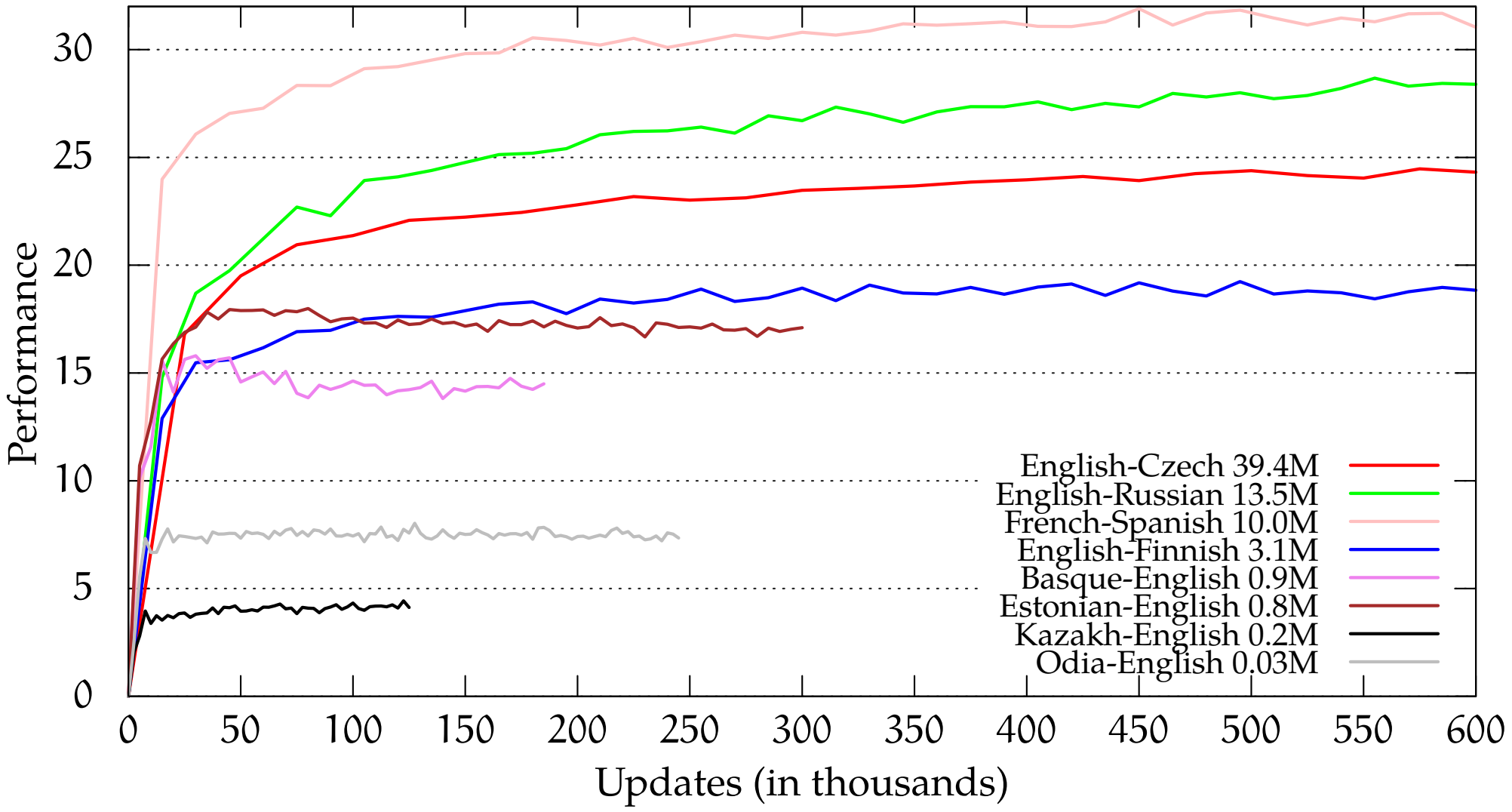

**Figure 4.1:** Learning curves for various language pairs with various sizes of parallel corpus. We can see that language pairs with less than 1M data quickly flatten out or even start overfitting as in the case of Basque→English.

Notably, Koehn and Knowles (2017) found out that NMT outperforms SMT when more than 24.1M words are available, i.e., approximately 1M Spanish–English parallel sentences. However, we need to add that Sennrich and Zhang (2019) recently revisited the training condition for low-resource and showed that current systems outperform SMT even in the low-resource scenario if careful training is performed.

Summarizing across the diverse conditions, we take the stance that language pairs with less than a million training pairs are deemed low-resource.

Another point of view which can be used to delineate low-resources is behavior during training. Section 4.2.2 shows so-called learning curves, i.e., the performance of a given system on a development set (also called held-out set) throughout the training. When the curves bend down (e.g. the performance starts decreasing), it is an indication that the model is overfitted, usually by memorizing training sentences.

During our experiments, we noticed that low-resource languages often overfit or flatten out within first 50-100k of weight updates, which in our setting is roughly half a day of training on a single GPU. In contrast, the higher-resource language pairs usually do not overfit at all. Instead, their learning curves only slowly flatten out after several hundred thousand steps, see Section 4.2.2 for an illustration.

We should also note that a language pair considered low-resource today might not be considered low-resource in the future: this can happen thanks to newly available data or also thanks to the improvements in NMT training techniques.

### 4.2.3 Resource Quality

Language resources for machine translation come in varying degrees of quality. On the one extreme, there are carefully revised translations made by professional translators, which result in parallel corpora with more aligned translations and a generally very good sentence-to-sentence alignment. On the other extreme, parallel corpora can be sentences automatically extracted from noisy crawled web pages, which results in not-quite-parallel sentences with mismatching content, non-word parts, and even sentences in a wrong language.

Another issue with parallel corpora is the phenomenon called **translationese** (Gellerstam, 1986). A text translated from one language to the second one has different linguistic properties than text written in the second language originally. According to Baker et al. (1993), translated texts are often:

- Simplified – when translators subconsciously simplify the message.
- Normalized – to conform the typical features of target languages up until a point of a slight change in meaning.
- Explicitated – the notion that structures of text are explained in more detail due to the rarity of the phenomenon in the target language, for example, explaining abbreviations. It is, to some extent, an inverse to the simplification.

Stymne (2017) evaluated the effect of translationese on the MT systems and found out that the translation direction indeed influences the final quality. However, as the authors mention, their study would need to be evaluated over a larger sample of language pairs and MT systems to be more reliable. For sure, translationese has a pronounced effect on MT evaluation (Graham et al., 2020b).

### 4.2.4 Corpus Cleaning

Collecting training data is the first step needed in order to start training machine translation models. For low-resource languages, crawled data are often one of the largest sources of parallel sentences. Unfortunately, the resources based on crawled data are usually noisy. In order to use the crawled data, we need to clean them first.

There is a whole field for filtering parallel corpora. Koehn et al. (2018) organized a shared task intending to study various techniques of filtering parallel corpus for NMT to improve its performance by removing noisy sentence pairs. Filtering usually consists of pre-filtering rules, like removing non-word tokens, characters with malformed encoding or various tags. Very often, sentences with a too big difference in length are also removed. More advanced techniques rely on various scoring functions.

Furthermore, we can remove short segments of up to a few tokens. These sentences are often relics of misaligned pairs. A standard is to remove sentences with less than five tokens (Koehn et al., 2018).

In contrast, we can also think about removing long segments or breaking them to shorter ones. For practical reasons of batch training on GPU, sentences within one batch are padded to a fixed length according to the longest sentence in the batch. Thus, in order to increase the batch size, we can remove very long sentences from the corpus, because they result into large padding of other sentences in the same batch. However, removing long sentences should be done only in the case if there are very few of them compared to the size of the whole corpus so that their removal will not affect the overall number of parallel sentences. Note however that skewing the distribution of sentence length is rather risky for the model performance, see Section 4.5. Popel and Bojar (2018) recommends to set the threshold to 100 or 150 tokens.

Whenever we are dealing with a corpus that was automatically collected, we may want to check all sentences by an automatic language identification tool in order to remove sentences that are in a different language. Although this step can remove correct sentences due to language identification errors, it is usually beneficial because it removes a part of noisy sentences. More advanced approach of corpus filtering it so use dual conditional cross-entropy (Junczys-Dowmunt, 2018) which relies on training MT model that is used to reassess training sentences.

## 4.3 Measuring Training Progress

The progress of NN training can be measured in various ways. We can measure, for example, the wall-clock time passed, the number of processed training examples or the financial cost of the training. Each method is more relevant for various applications.

The most common approach is to report the number of processed training steps. It can be reported either by the number of individual seen examples, number of training epochs or as a number of batches. The individual examples are preferred in tasks, where each example has the same informative value. In MT, one training example is usually a sentence pair, which can be of varying length. Reporting the number of epochs does not rely on the ordering of sentences in corpus because it is reported after each complete pass over the training corpus. Lastly, we can report the number of batches, also called training steps, which specifies the number of updates (error backpropagation) through the network.

The batch size can be either fixed to a given number of sentence pairs (e.g. in Neural Monkey Helcl et al., 2018) or to a number of subword tokens in all sentences (e.g. Vaswani et al., 2018). The former definition correlates with measuring the number of training examples. However, the latter is optimized for training as it can better use the available GPU memory.

Another option is to measure the time passed to achieve a given result. Popel and Bojar (2018) use this reporting of wall-clock time and justify it by stating that the training speed computed as steps per second fluctuates at most by 2% during the training. Wall-clock time can be more informative than reporting a number of seen examples since the training step can contain a variable number of sentences or tokens. However, the training time is heavily influenced by the hardware and the other load on a given machine.[2]

Lastly, practitioners are primarily concerned with the error rate and the cost they need to pay to achieve that error level. It is referred to as a hardware cost (Shallue et al., 2018). This cost can be measured by multiplying the number of training steps by the average price per one step. It heavily depends on the respective hardware, but the number of training steps is hardware-agnostic and can be computed for any hardware given the average cost per step.

### 4.3.1 Convergence and Stopping Criterion

The standard recommendation of neural network training is to train until **overfitting**, i.e. the moment where the performance on the training set is improving, but the score on the development set is worsening.

Training of NMT models is complicated by the fact that the learning curves, showing the performance of the model over the learning period on a fixed development set, usually never fully flatten or start overfitting on reasonably large datasets. Signs of overfitting are noticeable in low-resource settings only, as discussed in Section 4.2.2 and also exemplified on an interesting highly-multilingual example in Section 8.4. Especially with the recent models trained on a large parallel corpus, we can get some improvements, usually around tenths of a BLEU point, even after several weeks of training and the learning curve will not fully flatten out.

The common practice in machine learning is to use a stopping criterion. One option is to set the maximum number of training steps or epochs. The second option is to evaluate the model every X steps (or minutes) on the development set and stop the training whenever the last N updates do not improve the performance by at least some small delta.

The former approach relies on an intuition of how long approximately is enough to train the model. Usually, more complex models need more time to reach the maximal score, and an incorrectly set number of steps could stop the training prematurely. The latter approach is sensitive to the number of steps (and their duration) between individual evaluations: if the evaluations are too close to each other, the training can stop too early due to the training fluctuations, and when they are too far apart the training would not stop in a reasonable time on big datasets.

[2] The NN research is usually carried out on clusters, where several people run various processes. Even when the GPU is allocated only for a given job, other shared resources, like CPU or network disks, can slow down the training process.

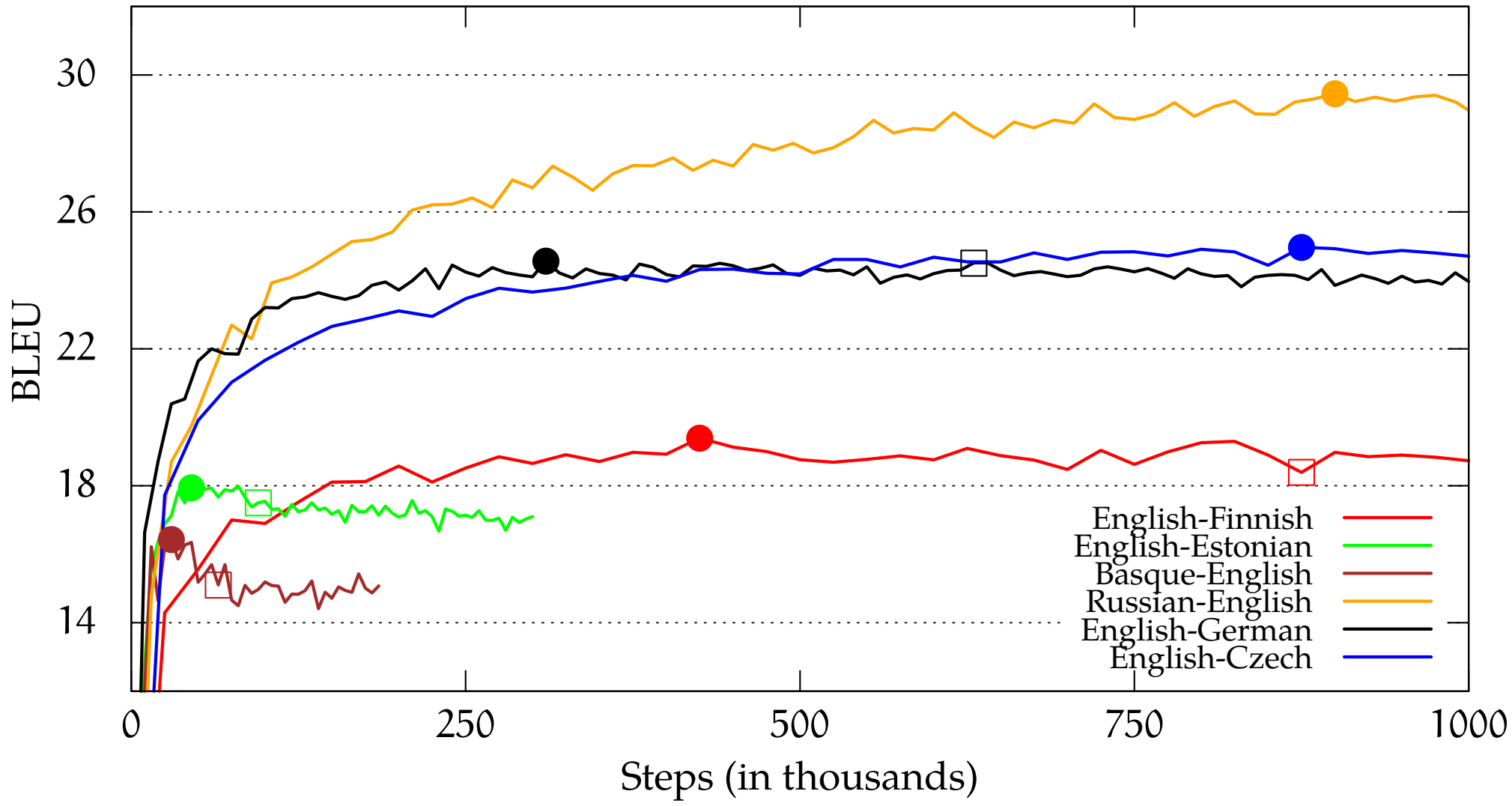

**Figure 4.2:** Examples of real learning curves. Full dots represent the best performance, squares represent the stopping criteria.

Many papers do not specify the stopping criteria or only mention an approximate time or the number of steps for how long the model was trained (Bahdanau et al., 2014; Vaswani et al., 2017). Presumably, the models are trained until no apparent improvement is visible on the development set. However, this stopping criterion is not perfect since the models could be stopped at various stages of training, and the comparison could be unfair.

As shown in Figure 4.2, convergence times differs significantly depending on a language. For example, low-resource language pairs can converge within 50k steps and high-resource pairs improve even after 1000k steps. Kocmi (2019) defined a more general convergence criterion: to stop the training whenever there was no improvement bigger than 0.5% of maximal reached BLEU within the past 50% of evaluations. This criterion is comparable to stopping after X batch updates without any improvement, and it is less sensitive to the number of steps between evaluations as the low-resource languages are evaluated up to ten times more often.

This stopping criterion is especially useful for low and middle resource languages. Figure 4.2 presents with square a point where training stops. Without any doubt, the model already passed its best performance, and thus the stopping is valid. We can notice that for high-resource languages, the stopping criterion does not trigger within 1000k steps.

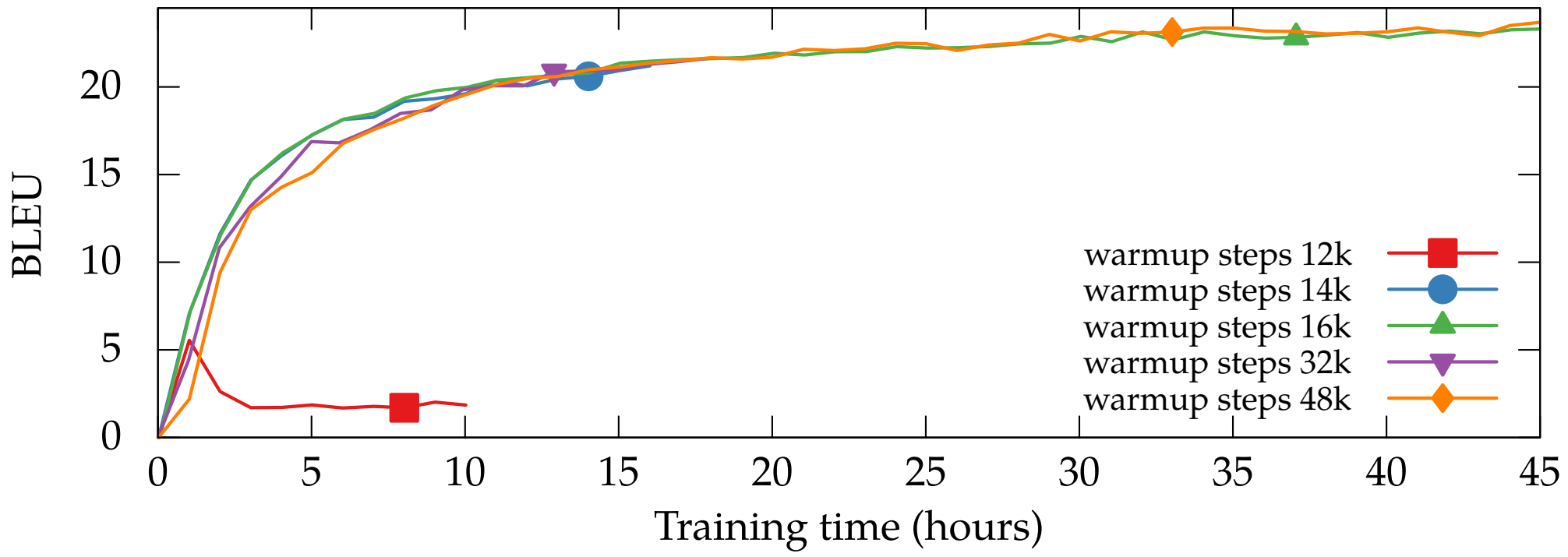

**Figure 4.3:** Transformer training instability if warm-up steps are too low. Transformer Big model trained on English-to-Czech translation with the default batch size (1500) and learning rate (0.20) on a single GPU. Reproduced from Popel and Bojar (2018).

With the stopping criteria in mind, once the training stops, it is common practice to take the best performing model on the development set instead of the last step where the training stopped.

## 4.4 Training Instability

We discussed the core elements of the Transformer model in Section 3.3.2 but did not mention all the hyperparameters of the model that affect the success of the training and the reachable performance given a fixed amount of training data. The hyperparameters are intertwined and finding a working configuration may be difficult if one does not start from an established setting. We illustrate the aspects we found most important for the overall convergence, i.e. those that may prevent the model from learning altogether. We are not discussing the *optimization* of these parameters to reach the best possible performance, because such an optimization is too much dependent on the particular dataset and other conditions. See Popel and Bojar (2018) for an example of such a hyperparameter optimization.

**The Importance to Start Slow** We assumed you are familiar with the basics of NN training: the weights of the model are gradually updated as new training data are coming in. The amount by which the weights are modified depends on how much they contribute to the current prediction error and globally on the learning rate, a hyperparameter. With too small learning rate, the training would take too long, with too high learning rate, the weights would be jumping too much there and back, overshooting the local optima.

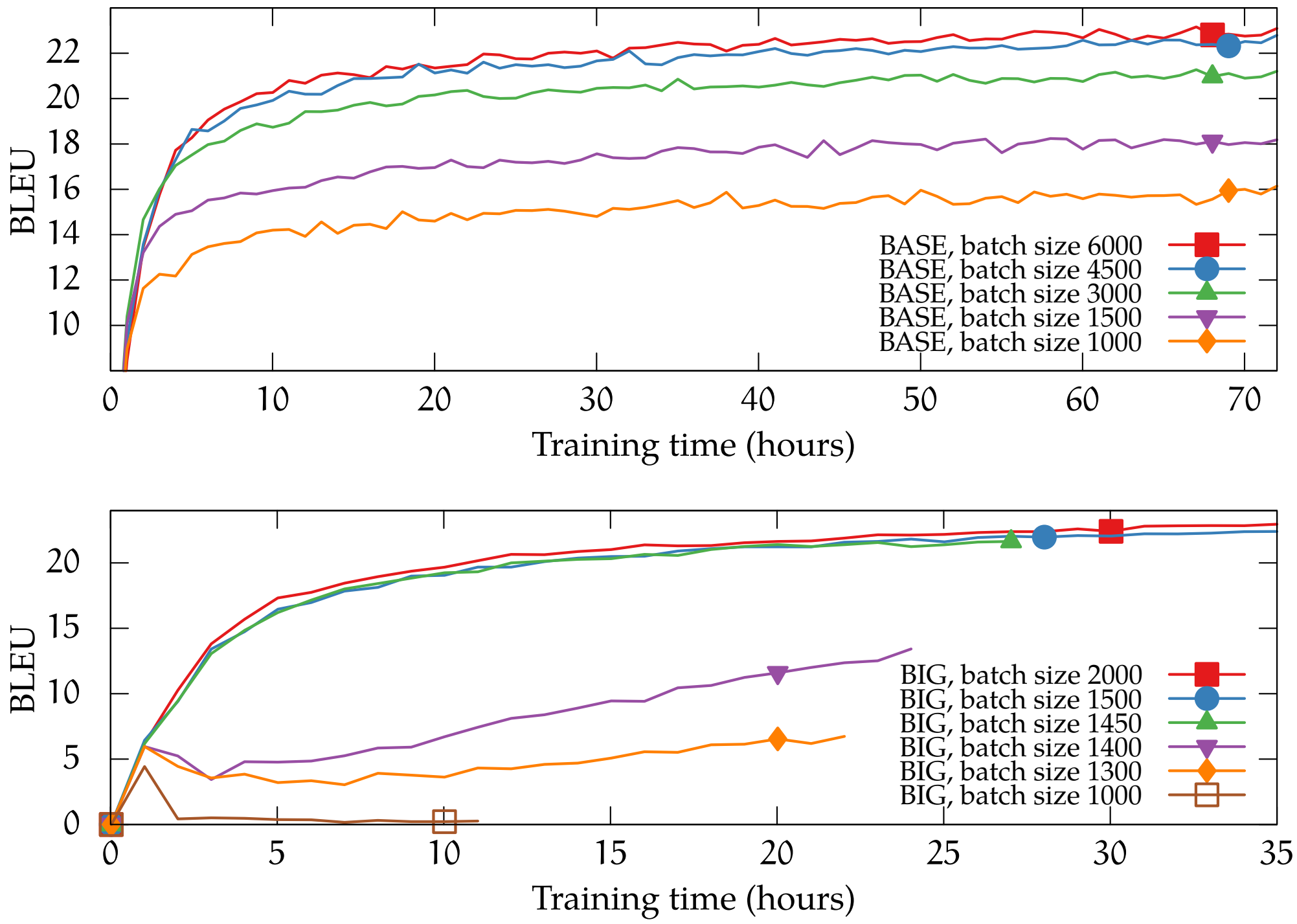

**Figure 4.4:** Effect of the batch size with the Transformer model in the Base setting (top) and Big setting (bottom) with the same training data. Reproduced from Popel and Bojar (2018).

The complexity of the Transformer model demands yet a cleverer approach. Instead of a fixed learning rate, standard implementations follow the so-called "**Noam** scheme". The learning rate is first linearly increased up to its maximum desired value and then exponentially decayed. The linear "warm-up" phase is probably needed so that the model moves to a reasonable area of the parameter space and the decay the helps to settle on a particular local optimum.

Figure 4.3 illustrates what happens if the warm-up phase is too short, i.e. if the learning rate reaches too quickly its maximum value: The model starts learning but soon then diverges, losing all its translation ability.

**Batch Size** The batch size limits how many sentence pairs are used in one training step, i.e. for the estimation of the training loss and thus for the calculation of the nec-essary weights update. Ideally, and in very early days of NN training, the *full* training dataset was used at each training step.

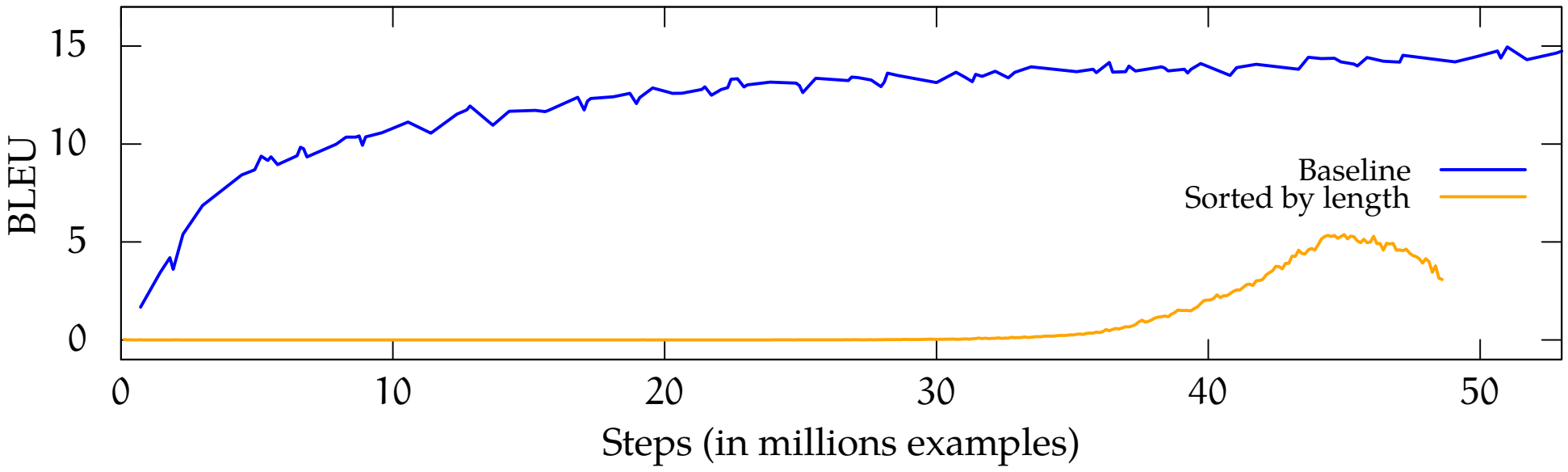

**Figure 4.5:** Failure to learn from a sorted corpus.

As we see from Figure 4.4, the size of the batch (measured in the number of subword units) can have a very pronounced effect on both the overall learnability as well as on the final performance. For the smaller configuration of the Transformer model ("Base"), larger batch sizes lead to higher overall scores. For the bigger configuration ("Big"), smaller batch sizes prevent the model from learning while larger ones seem to lead to identical levels of performance.

Given the limited resources, here specifically the memory capacity of the GPU, one has to carefully balance the amount of memory used for the weights of the model itself and the data of the current batch. Due to the size of the "Big" model, Popel and Bojar (2018) could not afford more than 2000 subwords in the batch.

**The Importance of Shuffling** Figure 4.5 reproduced from Kocmi and Bojar (2017a) documents the importance of *representative batches*. The best way of obtaining a representative sample from a data collection is to take a random sample. When batches are created from shuffled data, which is the default in almost all MT toolkits, each of the batches is representing the distributions in the dataset reasonably well.

If the shuffling is forgotten and the dataset exhibits some regular evolution of a parameter across the training, the training can easily fail. Figure 4.5 exemplifies this with the length of the sentences. The training starts with batches that contain only very short, typically one-word, sentences. Because of the deduplication strategy of the underlying parallel corpus (Bojar et al., 2016) which avoid duplicated *documents* but does not generally remove duplicated sentences, a large portion of these training examples is actually identical. This happens because many documents contain the same short and simple sentences.

As a result, none of the sorted batches is representative of the overall translation task. The model probably memorizes the current batch and a few preceding but it never learns to translate. Only at the very end of the training, when the parameters are already rather stabilized in a very wrong way, long sentences start to illustrate enough of translation and the model attempts to re-learn this.

**Observation 2:** *Transformer (as all deep neural networks) requires representative batches. Never forget to shuffle your dataset.*

## 4.5 Lack of Generalization

Given the good performance of NMT on high-resource language pairs, one would assume that the model indeed learns to use the language, that "the quantity brought the quality" and the model will correctly translate all new sentences, as long as the domain is preserved. Unfortunately, the generalization capability of Transformer is substantially worse.

As e.g. Variš and Bojar (2021) document, Transformer is failing to generalize already in such a trivial feature as the sentence length. In the early days of NMT, before the attention mechanism, it was known that recurrent neural networks handle longer sentences worse, because the fixed-size vector can not well represent the whole sentence (Cho et al., 2014a). Transformer is fully built on attention, so it has constant-time access to any position in the sentence and has no reason to suffer from such and information loss. The exact technique of the model for predicting the target sentence length is however unclear and only empirical observations reveal the limitations.

It is not overly surprising that Transformer is reluctant to produce sentences longer that the original training examples were.[3] The key interesting observation of Variš and Bojar (2021) is that the default Transformer can not generalize to *shorter* sentences, too.

Section 4.5 summarizes the experiment: The test set is divided into buckets depending on the target sentence length (source sentence bucketing leads to identical observations). The baseline model is trained for English-to-Czech translation on the whole corpus CzEng and it performs close to 20 BLEU points across all the test buckets. Eight contrastive models are trained on partitions of the training data ("Train-Bucket" 10 to 80), and we see that these models perform well only on the test bucket of the same sentence length. The performance overall degrades for higher sentence lengths as there are fewer and fewer such sentences in the training data.

The lower part of Section 4.5 explains the reason: models trained on short sentences always produce short output, models trained on long sentences always produce long output.

We can also express this observation in different words:

[3] As pointed out by one of our reviewers, newer positional encodings such as Press et al. (2021) or Chen (2021) generalize better in this respect.

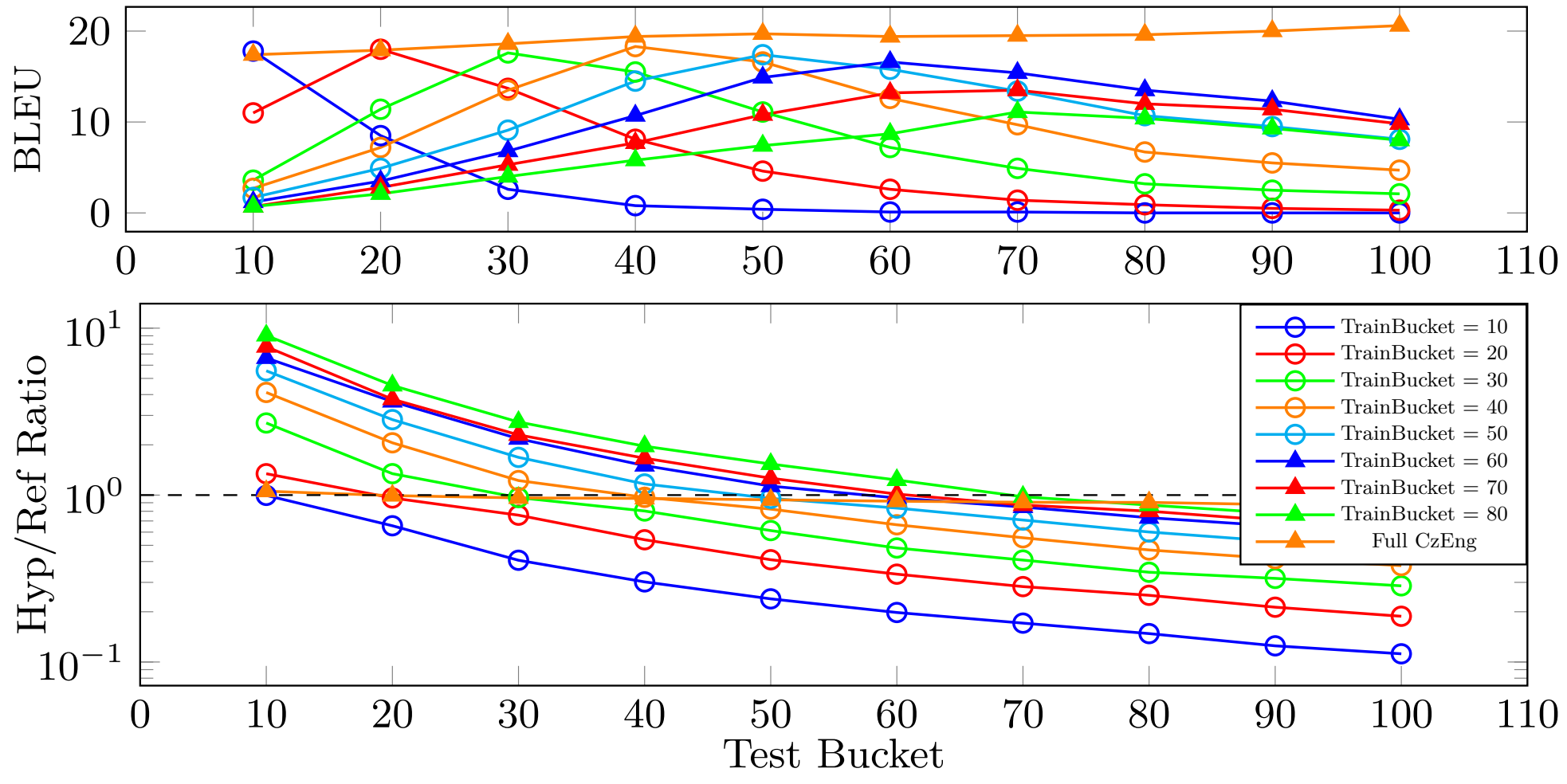

**Figure 4.6:** Transformer fails to generalize over sentence length. **Top**: Varying performance of models trained only on sentences restricted to a certain range of lengths when tested on similarly partitioned test sets. The baseline model (top line) trained on all lengths performs well on all test buckets. Length-specific models perform well on the matching test bucket and score worse on both shorter and longer test sentences. Note that BLEU scores are not comparable across different test sets (i.e. horizontally). **Bottom**: Average ratio between a hypothesis and reference. Dashed line indicates a ratio of 1.0. Systems trained on short sentences produce short outputs (see e.g. Train Bucket 10 evaluated on any other test bucket), systems trained on long sentences produce up to 10x longer outputs (Train Bucket 80 evaluated on test bucket 10). Reproduced from Variš and Bojar (2021).

**Observation 3:** *Transformer with its default positional encoding is unable to produce sentences of a different length (shorter or longer) than the training data illustrated.*

## 4.6 Catastrophic Forgetting

The adaptability of neural networks goes hand in hand with their tendency to adapt *too quickly* and lose performance on the distribution of inputs presented in past batches. This phenomenon is known under the term **catastrophic forgetting**. The network is trained in batches of inputs and expected outputs that exemplify the overall training data. It is very important that *each* batch illustrates ideally the *full range* of phenomena that will be needed at inference.

Figure 4.7 reproduced from Kocmi and Bojar (2017a) illustrates the behaviour of the attentional sequence-to-sequence model (Bahdanau et al., 2014). The model is trained for English→Czech translation on the 48.6M CzEng 1.6 corpus (Bojar et al., 2016) for exactly one epoch, varying only the order of the training examples.

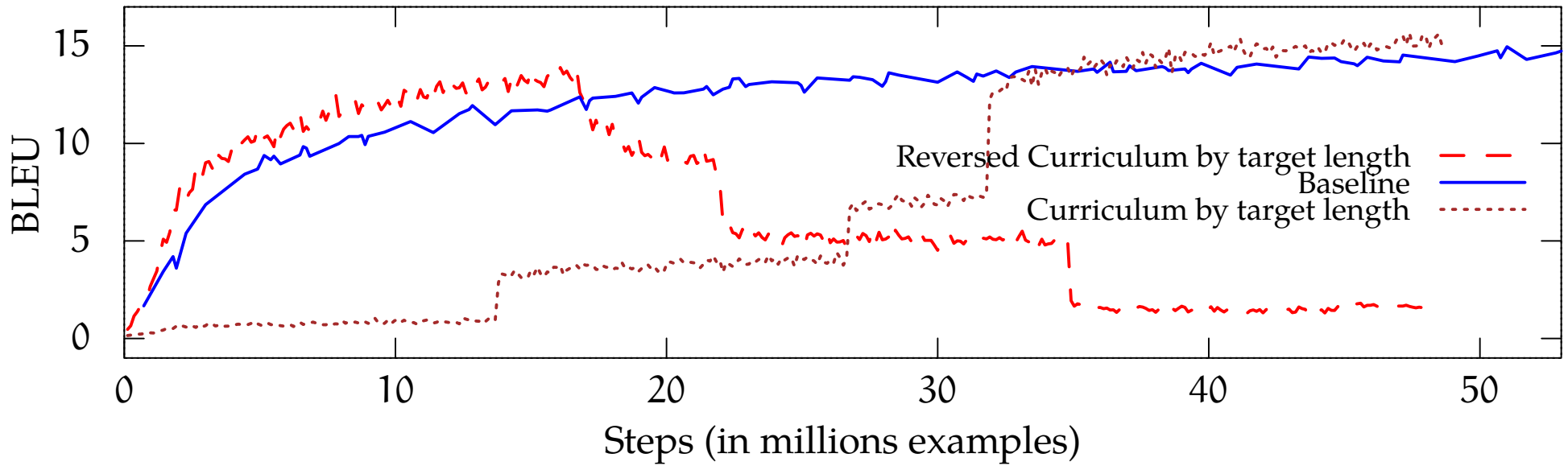

**Figure 4.7:** Catastrophic forgetting with respect to sentence length.

The baseline run uses random ordering of sentences. The strategy of Kocmi and Bojar (2017a) aimed at learning from simple examples first and increase the complexity of the learning step by step. Kocmi and Bojar (2017a) approximate the complexity of training examples simply by the target sentence length and organize sentences in buckets. While the baseline proceeds over the training data in shuffled order, "Curriculum by target length" gradually learns on buckets of longer and longer sentences.[4] The resulting performance slightly surpasses the baseline but the baseline can easily grow higher in further epochs.

Catastrophic forgetting is documented by the third curve where Kocmi and Bojar (2017a) simply reverse the sequence of training examples. The early parts of the curve seem to outperform the random baseline but this is actually an artifact of measuring progress in sentences and not words. In the same number of sentences, the "reversed curriculum" learns on more words than the baseline because longer sentences are oversampled in the early batches and thus its weights are trained better. As the data move towards buckets of shorter and shorter sentences, the network *unlearns* to produce long sentences, confirming again what we saw in the previous section, but now on a network which was able to produce long sentences in its early training stages.

**Observation 4:** *NMT outputs closely follow the patterns observed in* recent *training batches.*

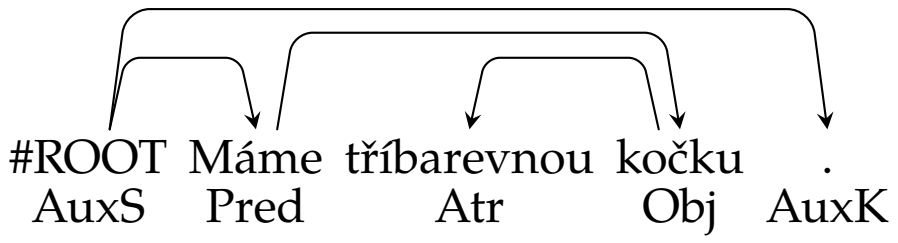

| | | |
|---|---|---|
| MT source (cs) | | Máme tříbarevnou kočku . #Translate |
| MT target (en) | | We have a three-colored cat . |
| DepHeads | src | Máme tříbarevnou kočku . #DepHeads |
| | tgt | #ROOT kočku Máme #ROOT |
| DepLabels | src | Máme tříbarevnou kočku . #DepLabels |
| | tgt | #Pred #Atr #Obj #AuxK |
| DepHeads | src | Máme tříbarevnou kočku . #DHeadsLab |
| +DepLabels | tgt | #ROOT #Pred kočku #Atr Máme #Obj #ROOT #AuxK |

**Figure 4.8:** Sample dependency tree, inputs and expected outputs of linguistic secondary tasks. Reprinted from Pham et al. (2019).

## 4.7 The Cost of Multi-Tasking

Section 3.5 presents the benefits of training the network to perform more task. In Pham et al. (2019), we looked at the idea more critically and examined also the “cost” of having multiple tasks to perform, not just its benefits.

The following graphs are based on several runs of a Transformer model trained for German→Czech translation using a corpus of 8.8M sentence pairs. The model is expected to translate, and in some variants to also predict source-side syntactic information: the governing word (“DepHeads”), or its syntactic function in the sentence, i.e. the dependency edge label (“DepLabels”), or interleaving these two (“DepHeads+DepLabels”) as illustrated in Figure 4.8.

The baseline run (just translation) also differs from the multi-task runs in that the multitask runs have a special token at the end of the sentence to indicate *which* of the possible tasks should be carried out for this input sentence. These task identification tokens are visible in Figure 4.8: “#Translate”, “#DepHeads”, etc.

The first observation is that simply adding one final token to the sentence, denoting the desired task, can change the performance of the model. We see in Figure 4.9 that the model with the explicit requirement to produce the translation (‘ MT TaskID”) flattens sooner and performs worse. If such a strange effect happens in your experiments, you have already incurred a loss before doing anything related to the examined idea.

The two plots in Figure 4.10 then illustrate one important caveat when discussing learning curves. Both plots contain the exact same runs. The top one has the training steps on the x-axis. The multi-tasking runs spend half of the training steps on the secondary linguistic tasks, and the other half on MT. The bottom plot shows the MT epochs on the x-axis, it is number of iteration through MT training data.

[4] Shorter sentences are kept in these batches to some extent, too.

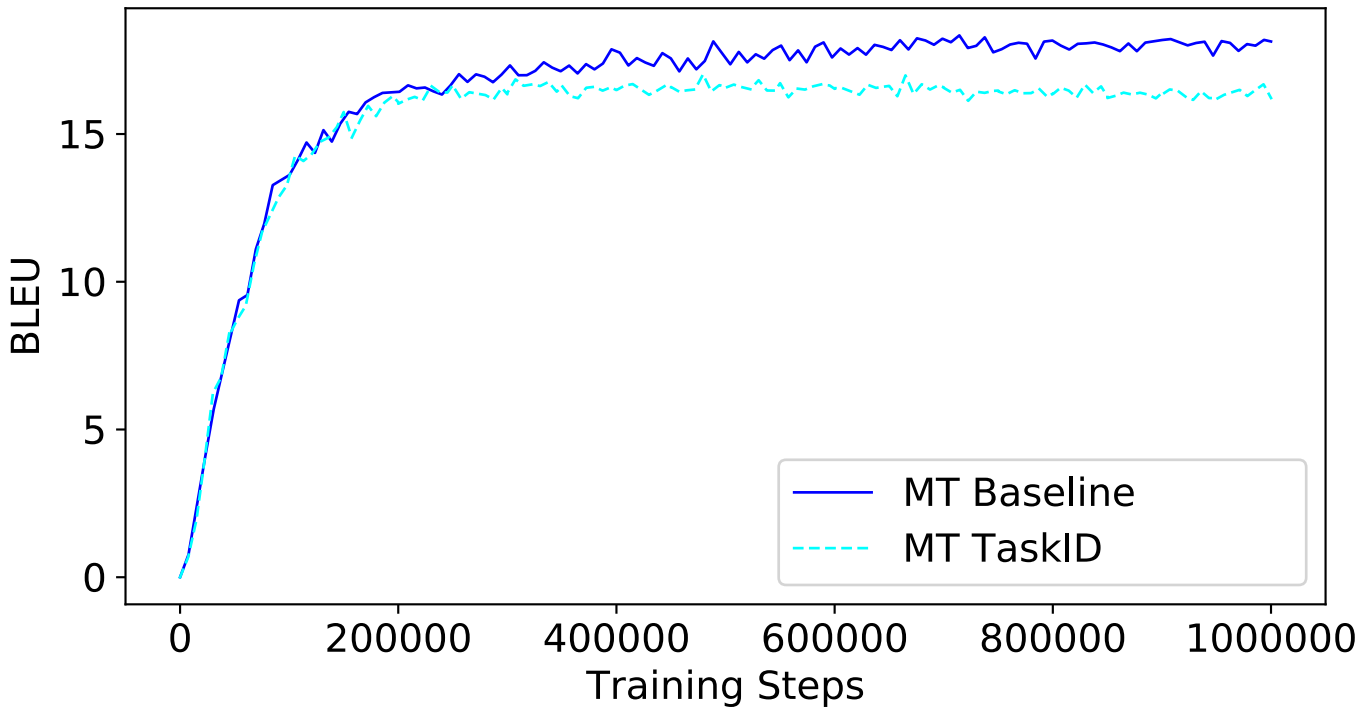

**Figure 4.9:** Training progress of a baseline MT and a setup, where each source sequence is appended by a task identification token "#Translate", although there is no other task to alternate. Reprinted from Pham et al. (2019).

We see that the particular secondary tasks examined by Pham et al. (2019) seem helpful when there the training data is limited. (Note however that the y axis shows the progress of BLEU score during training, not the BLEU score that could be achieved if the model was trained up to the convergence given a particular amount of training data.) All the multi-task curves grow faster than the baseline in the bottom plot. If we however consider the training time (which is generally proportional to the number of training steps), the baseline model is clearly better than all the multi-task ones in all stages of the training.

The next results of Pham et al. (2019) is an analysis of cost and benefits of multi-tasking with dependency parsing as a secondary task that is supposed to improve the MT quality. In reality, parsing as the secondary task degrades the MT quality. With a contrastive non-linguistic simple secondary task that is unrelated to MT, and therefore supposedly has zero benefit to MT and the same multi-tasking cost as linguistic task, the MT performance was even lower. Therefore, we conclude that the linguistic secondary task has a positive benefit to MT, but the cost of multi-tasking (e.g. swapping the targeted task by a task identification token, formatting the parse tree into the selected text format, linguistically unimportant standards of the parse tree, etc.) is higher than the benefits. More on the comparison with dummy secondary task is also in Macháček (2018), section 4.1.4.

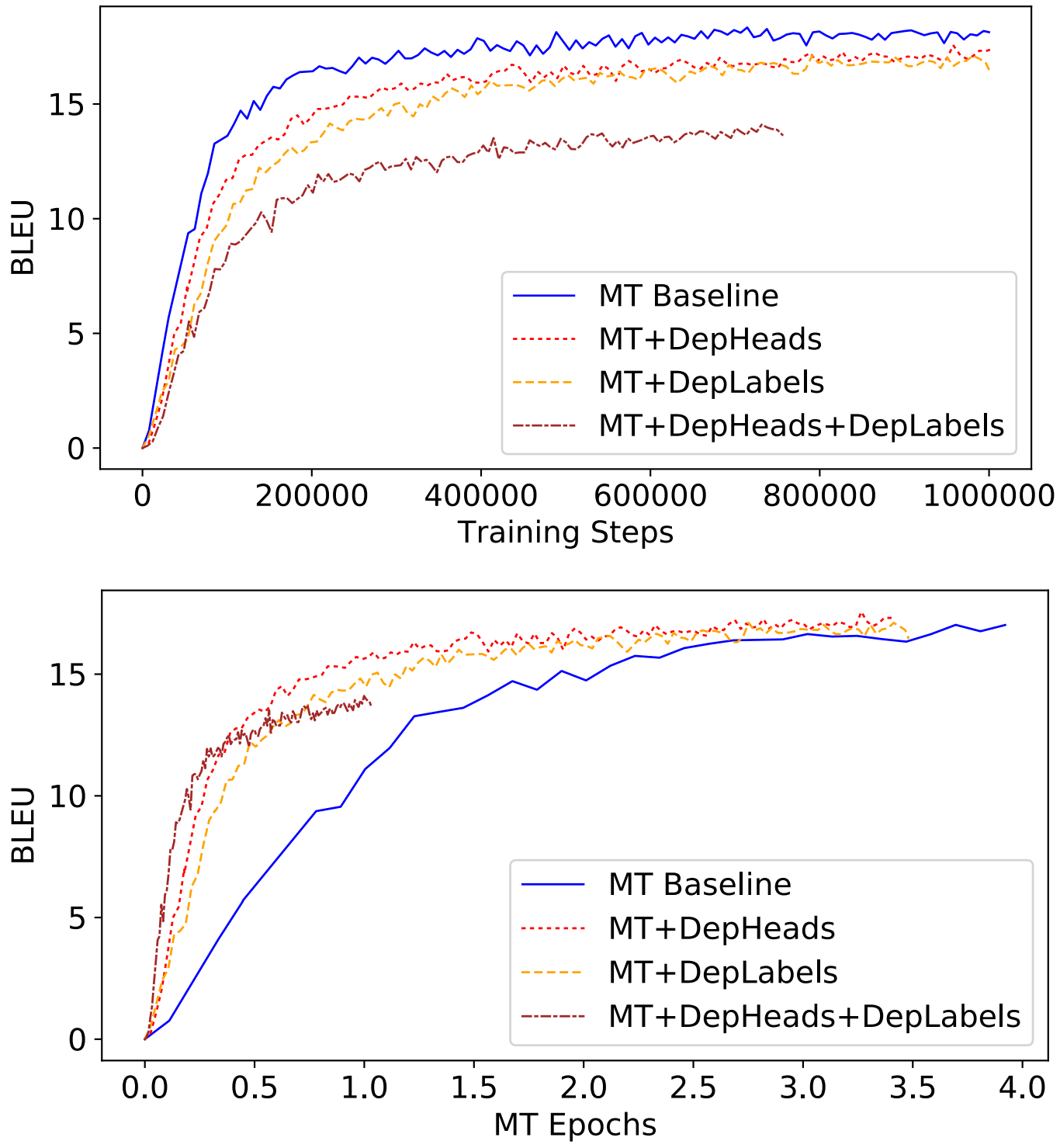

**Figure 4.10:** Learning MT BLEU curves of the MT baseline and multi-tasking runs with MT and linguistic secondary tasks over training steps (top) and MT epochs (bottom).

# 5
# The Battle Against Wishful Thinking

Wishful thinking is a formation of beliefs based on assumptions and wishes, rather than on evidence and data. The whole process of science is geared against wishful thinking: any idea has to be clearly described, implemented and tested according to the current best practices of the respective field in order to pass through the peer review barrier.

Every author on their own proceeds carefully and tries to not fall in the trap of wishful thinking, and yet the approach of deep learning increases our overall susceptibility to this.

In this chapter, we start with the basic technique: objective evaluation in machine translation. We review the current state of the art (Section 5.1) and then move its open problems: the fact that many good answers are possible and that even humans struggle to produce perfect outputs.

Even when we adopt all the best practices (including the often forgotten significance testing, Section 5.2), and we have the best references possible, the opaque nature of neural networks puts us at a high risk of wishful thinking. Without knowing the inner effective structure of the process, we are likely to see *incorrect reasons behind the good results*. It is often tempting to generalize good results from low-resource setups to high-resource (Section 5.3).

Then, we show two examples (Sections 5.4.1 and 5.4.2) of neural network analysis that seemed to be fantastic and amazing at first, but only before some further contrastive experiments were performed. Without such contrastive experiments – which are not yet commonly carried out in the field on machine translation – it is rather easy to assume the success of our novel approach while the network is actually exploiting some unintended "numerical advantage" the realization of that approach has brought. A wrong explanation in mind can easily lead us to drawing unjustified interpretations of the results and overestimating the expectations on the networks' behaviour in a new instance.

## 5.1 Machine Translation Evaluation

In order to evaluate how successful the machine is in translating, we need to define what is considered a good translation. It inherently leads to defining when two texts, in a different language, constitute the equivalent meaning. It is a difficult task, and we would need to delve into complex theoretical questions, which is out of the scope of this work. Thus, MT researchers usually evaluate MT translations by comparing it to the expert human translations.

MT evaluation is usually focusing on two main aspects called **fluency** and **adequacy**. Whenever the system produces syntactically well-formed sentences (i.e. high fluency) and does not change the semantics, the meaning of the source sentence (i.e. high adequacy), it is considered as a good translation (Hovy et al., 2002). Various ways of measuring the fluency have been proposed, and new metrics are annually evaluated in the WMT shared task (Mathur et al., 2020). As for the adequacy, it is more complicated since multiple correct translations are possible, and therefore, it is mostly evaluated by conducting the manual evaluation by humans, which is time-consuming and costly.

Bojar et al. (2013) created a method for the generation of millions of possible references that could solve the problem with having only a limited number of references (usually only one). However, their testset is restricted only to 50 prototype sentences.

### 5.1.1 Manual Evaluation

Manual evaluation campaigns are run each year at WMT to assess translation quality of both academic and commercial systems. This evaluation is considered a benchmark for identifying state-of-the-art systems of a given year.

Manual evaluation utilizes human ability to judge what is a good translation without a rigorous definition. Throughout the years, the WMT manual evaluation has changed several times based on findings from previous years. For example, comparing multiple systems together, binary yes/no decision about the translation, by directly rating one translation at a time as is the case of last several years (Graham et al., 2017), or asking annotators to mark individual errors in translations (Freitag et al., 2021). The main idea across the approaches remains similar: human judges are asked to rank presented outputs from various systems based on their intuition.

However, due to the high demand for cost and time, researchers usually do not use the manual score. Instead, we rely on automatic metrics that try to replicate human behavior as closely as possible.

### 5.1.2 Automatic Metrics

Automatic metrics for MT evaluations are often based on the estimation of similarity between the system output, and a human-produced reference translation.

Two categories of automatic machine translation metrics can be distinguished: (1) string-based metrics and (2) metrics using pretrained models. The former compares the coverage of various substrings between the human reference and MT output texts, common metrics are BLEU (Papineni et al., 2002), ChrF (Popović, 2015), or Meteor (Banerjee and Lavie, 2005). The latter category of pretrained methods consists of metrics that use pretrained neural models to evaluate the quality of MT output texts given the source sentence, the human reference, or both. Common metrics in this group are COMET (Rei et al., 2020), Prism (Thompson and Post, 2020), or BLEURT (Sellam et al., 2020). They are not strictly dependent on the translation quality of the human reference (for example, they can better evaluate synonyms or paraphrases).

The most often used metric is the **BLEU** score (Bi-Lingual Estimation Understudy, Papineni et al., 2002).[1] It is based on comparing $n$-grams of sentence units, typically words, between the system output and one or more reference sentences. It is computed as the geometric mean of $n$-gram precisions for $n = 1...4$ with penalization for short translations by **brevity penalty**, according to the following formula:

$$\mathrm{BP} = \begin{cases} 1 & \text{if } L_{sys} > L_{ref} \\ e^{(1-L_{ref}/L_{sys})} & \text{if } L_{sys} \leq L_{ref} \end{cases} \tag{5.1}$$

$$\mathrm{BLEU} = \mathrm{BP} \cdot \exp\left(\sum_{n=1}^{N} w_n \cdot \log p_n\right) \tag{5.2}$$

where $w_n$ is a positive weight summing to one, usually $\frac{1}{N}$. $L_{ref}$ denotes the length of the reference text that is closest in length to the system output, $L_{sys}$ is the length of the system output. The $n$-gram precision $p_n$ is computed by dividing the number of matching $n$-grams in the system output by the number of considered $n$-grams. The number of $n$-gram matches are clipped to the frequency in the reference when $n$-grams occur multiple times. BLEU is a document-level metric. Thus, the counts of confirmed $n$-grams are collected for all sentences in the document (or testset) and then the geometric mean of n-gram precision is computed from the accumulated counts.

It is more informative to compare system output against several references, but it is expensive to obtain multiple references. Thus, usually only one reference is used. BLEU score is usually multiplied by 100 instead of values on the interval 0 to 1 as originally described by Papineni et al. (2002).

There is a numerous criticism (Callison-Burch et al., 2006; Bojar et al., 2010) that has been observed of BLEU. For instance:

- use of a geometric mean, which makes the score 0, when there is no match at any of the $n$-gram levels (usually a problem of short testsets);
- gives no credit for synonyms or different inflected forms of the same word;
- does not consider the importance of various $n$-grams;
- it is very sensitive to tokenization.

Freitag et al. (2020) investigate how quality of references affects performance of automatic metrics. They show that typical references exhibit poor diversity and contain mainly translationese language. They suggest to generate paraphrases of references to increase quality of automatic metrics. Tamchyna and Barančíková (2016) show that automatic paraphrases generation for MT evaluation helps the most when we take the worst paraphrase sentences (highest perplexity according to language model).

[1] Marie et al. (2021) showed that 99% of papers use BLEU.

Additionally, there are several implementations of BLEU that differ in tokenization, case-sensitivity, and other details that lead to different resulting scores. Post (2018) made a call for clarity in reporting BLEU score and implemented SacreBLEU[2] tool for more comparable results. The evaluation script automatically downloads reference testset and computes performance using various metrics.

Some of the drawbacks are avoided by using pretrained methods that use deep neural model, usually language model, to evaluate translated sentences in a multidimensional embedding space. The main advantage is the possibility to evaluate paraphrases or inflected forms.

Kocmi et al. (2021) performed extensive evaluation of automatic metrics on a large corpus of human evaluations and showed that pretrained methods are superior in terms of correlations with human judgement to automatic metrics. The same conclusion has been confirmed by previous studies on a smaller corpora of human judgements (Ma et al., 2019; Mathur et al., 2020).

In conclusion, string-based methods largely depend on the quality of reference translations. However, their advantage is that their performance is predictable as it can be easily diagnosed which substrings affect the score the most. On the other hand, performance of pretrained methods is influenced by the data on which they have been trained. Moreover, the pretrained models introduce a black-box problem where it is difficult to diagnose potential unexpected behavior of the metric, such as various biases learned from training data.

## 5.2 Statistical Significance

If two translation systems differ in BLEU performance, it does not necessarily mean one is significantly better than the other for any small effect size. In order to indicate the actual quality, Koehn (2004) proposed to use the paired bootstrap resampling method to compute the statistical significance and validate the superiority of one of the systems. The method repetitively creates testsets by drawing sentences from the original testset randomly with repetition and evaluating the automatic score. Then it computes the confidence interval and statistical significance for comparing various MT systems.

Approximate randomization (Riezler and Maxwell III, 2005) can be used as an alternative test, and for metrics based on the average of sentence-level scores, such as COMET, we can also use standardized tests such as the Student t-test.

A concerning trend is presented by Marie et al. (2021), who studied 769 machine translation papers from the last 10 year. They showed that usage of statistical significance testing has been dropping and in recent years, less than 30% papers evaluate results by statistical significance testing. Additionally, Marie et al. (2021) showed that less than 10% of papers confirm automatic evaluation gains by human judgement.

[2] `https://github.com/mjpost/sacreBLEU`

## 5.3 It Works in Low-Resource

In some works, experiments with small training data and a small model show promising results. Examples of this behaviour were observed e.g. when enriching NMT with linguistic knowledge by multi-tasking of MT, POS tagging, NE tagging or parsing. The tasks may benefit from joint learning from multi-tasking as discussed in Section 3.5, but on the other hand, they may negatively interfere.

For example, Niehues and Cho (2017) report experiment with training data of 4M tokens for German-English MT. The multi-task models achieved higher translation quality than baseline single-task MT.

However, 4M tokens is a small training set. For German-English, there exist much larger corpora, for example a parallel webcrawl from 2015 with 187M German and 205M English tokens.[3] It is unclear whether the achievements of a small model on small data can be generalized to a big model and big data. If there is need for as good German-English MT as possible, there is more convenient option to train on big data, rather than small data and multi-tasking with POS and NE tagging. Macháček (2018) provided similar experiments with multi-tasking for enriching NMT. The auxiliary task improved the MT only in low resource scenario (5M training tokens, 500k sentences), but not in high resource (89M tokens, 8.8M sentences). See the plot in Figure 5.1.

One possible explanation of the success in low-resource setting is that the additional view on the data is not so useful for the generalization it illustrates (that words fall into certain classes, that sentences are formed according to certain syntactic patterns) but useful only to help the model in getting towards the area of vaguely plausible outputs as we discussed in Section 3.1.2.

We can conclude that multi-tasking with linguistic knowledge can be useful for low-resource languages, for which not many parallel data are available. However, for such languages probably do not exist linguistic annotations, so these results might be unusable in practice. Put even shorter:

**Observation 5:** *Gains observed in low-resource settings relying on additional annotation can be useless in practice because low-resource languages are not so likely to have annotated data anyway.*

Diminishing returns with growing training data are observed also in multi-lingual situations, as illustrated e.g. in Section 7.6.2 below.

[3] `https://linguatools.org/tools/corpora/webcrawl-parallel-corpus-german-english-2015/`

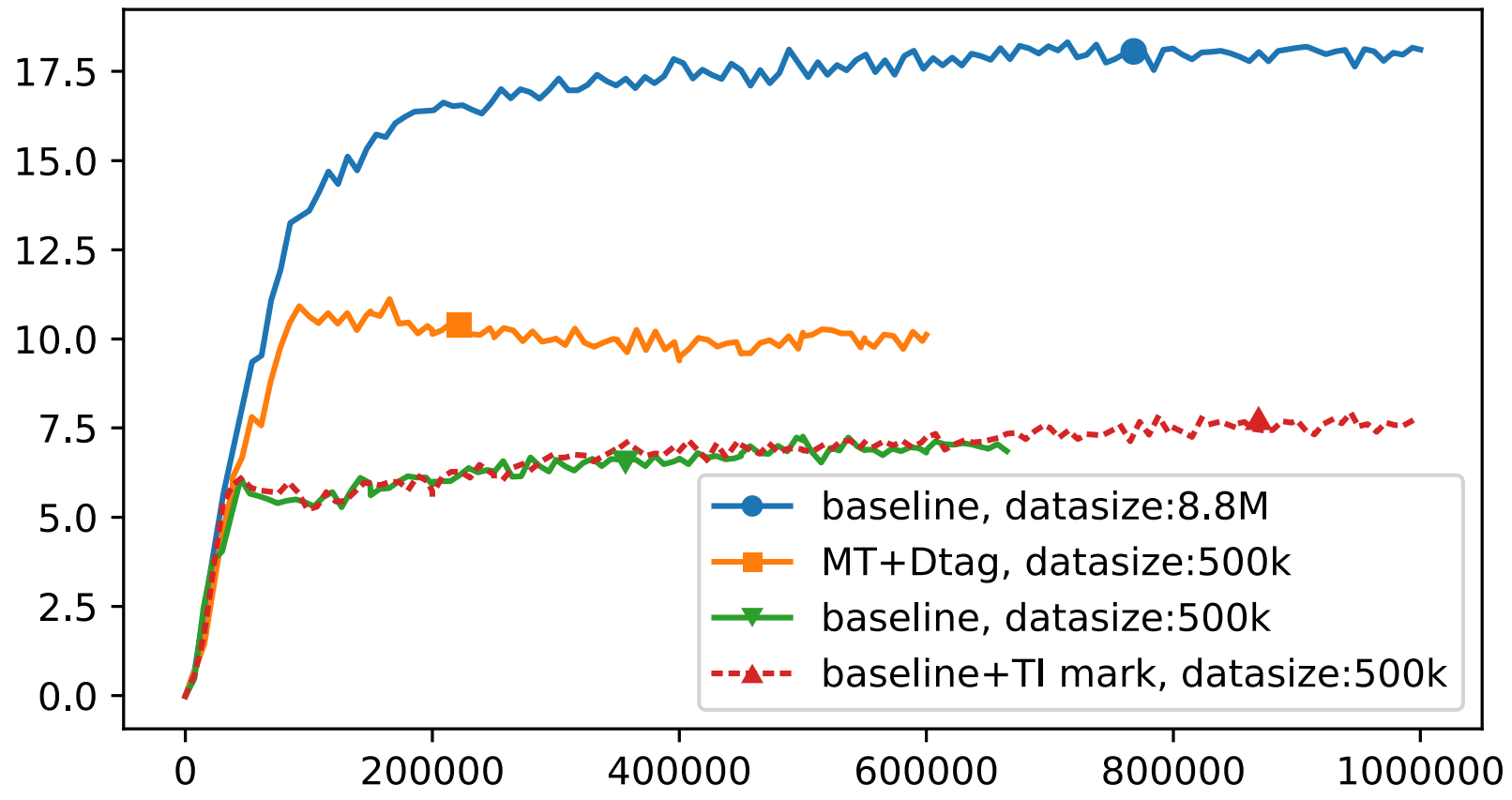

**Figure 5.1:** Learning curves for low-resource setting (500k sentences/5M tokens) with multi-tasking vs. high-resource (8.8M sentences/80M tokens). Multi-tasking MTand dependency parsing (marked with square) is useful in the low-resource setting, outperforming its baseline (downward triangle). The sanity check ("baseline+TI mark", upward triangle) confirms that task identification is not decreasing the performance as we saw in Section 4.7. However, the high-resource baseline (bullet) is much better than multi-tasking. Reproduced from Macháček (2018).

## 5.4 Simple Contrastive Task

In this section, we describe two examples how a simple contrastive task to linguistically-aware auxiliary task should be used to double-check the underlying reasons of NMT results. As the title of one of the discussed papers suggests "linguists should be replaced with dummies", i.e. the same style of annotation should be used but with dummy, uninformative content. This dummy variation should, of course, fail to deliver the improvement, but it is not always the case.

### 5.4.1 Dummy Diagonal Parse

It is well known that dependency structure is more consistent across languages than the adjacency of words (Fox, 2002). Pham et al. (2019) thus try to improve translation quality by promoting the knowledge of source syntax in Transformer NMT. One of the examined ways is joint training of MT and dependency parsing of the source from one of the attention heads.

There are two news. The good one is that when jointly training Transformer for parsing and MT, the MT quality increases over baseline, and the parsing quality is very high, approaching the state of the art.

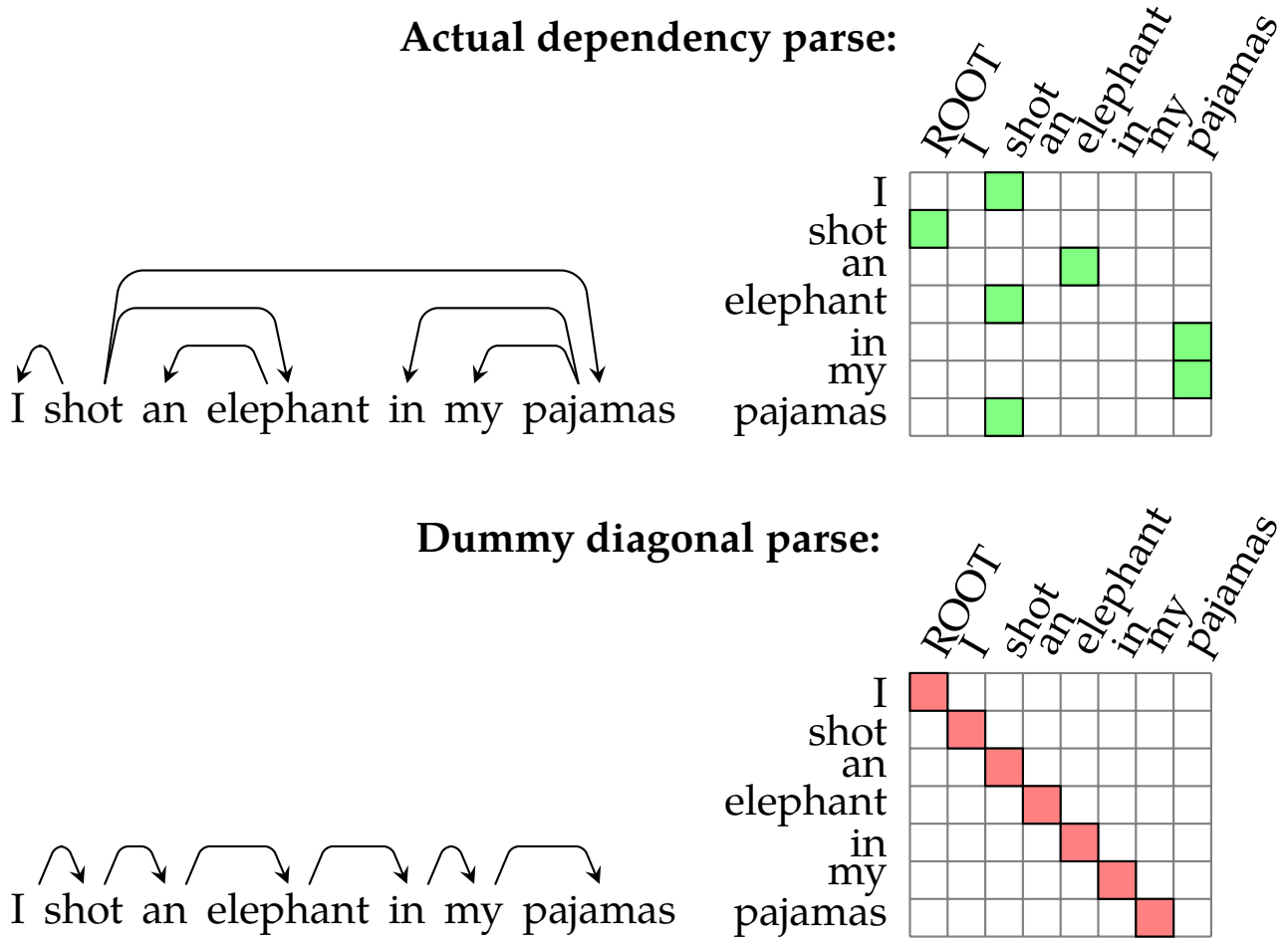

**Figure 5.2:** An example sentence with an actual syntactic dependency parse tree that supposedly benefits NMT, and a contrastive uninformative "dummy diagonal parse". The grids on the right hand side are representations of the parse tree in the form of a matrix that is retrieved from one of the Transformer attention head. The columns represent the heads, and the rows are dependents. Reproduced from Pham et al. (2019).

The bad news is that when exchanging the actual dependency tree for a dummy tree where each word is attached to the previous in the sentence, the MT quality remains increased over the baseline. See the illustration of the actual and dummy parse in Figure 5.2.

Without the contrastive task, we would conclude that multi-task training in Transformer so that it not only translates but also produces the dependency parse of the source in one of its attention heads improves translation a lot. The good performance of the dummy variant however documents that the quality gains can not be explained by the knowledge of syntax, but by something else, which has to be investigated.

### 5.4.2 Dummy Supertags

Supertags (Bangalore and Joshi, 1999) are a technique that considerably simplifies syntactic analysis of sentences by tagging words with highly informative tags. Supertags are thus one of standard techniques of encoding (part of) the syntactic structure of the sentence in a linearized form.

Motivated similarly as we discussed in the previous section, Kondratyuk et al. (2019) experiment with 12 multi-tasking setups: RNN and Transformer, interleaved and double-encoder multi-tasking, and 3 tag schemes: linguistically-aware CCG supertags (Combinatory Categorial Grammar; Steedman, 2000), and two dummy tagsets:

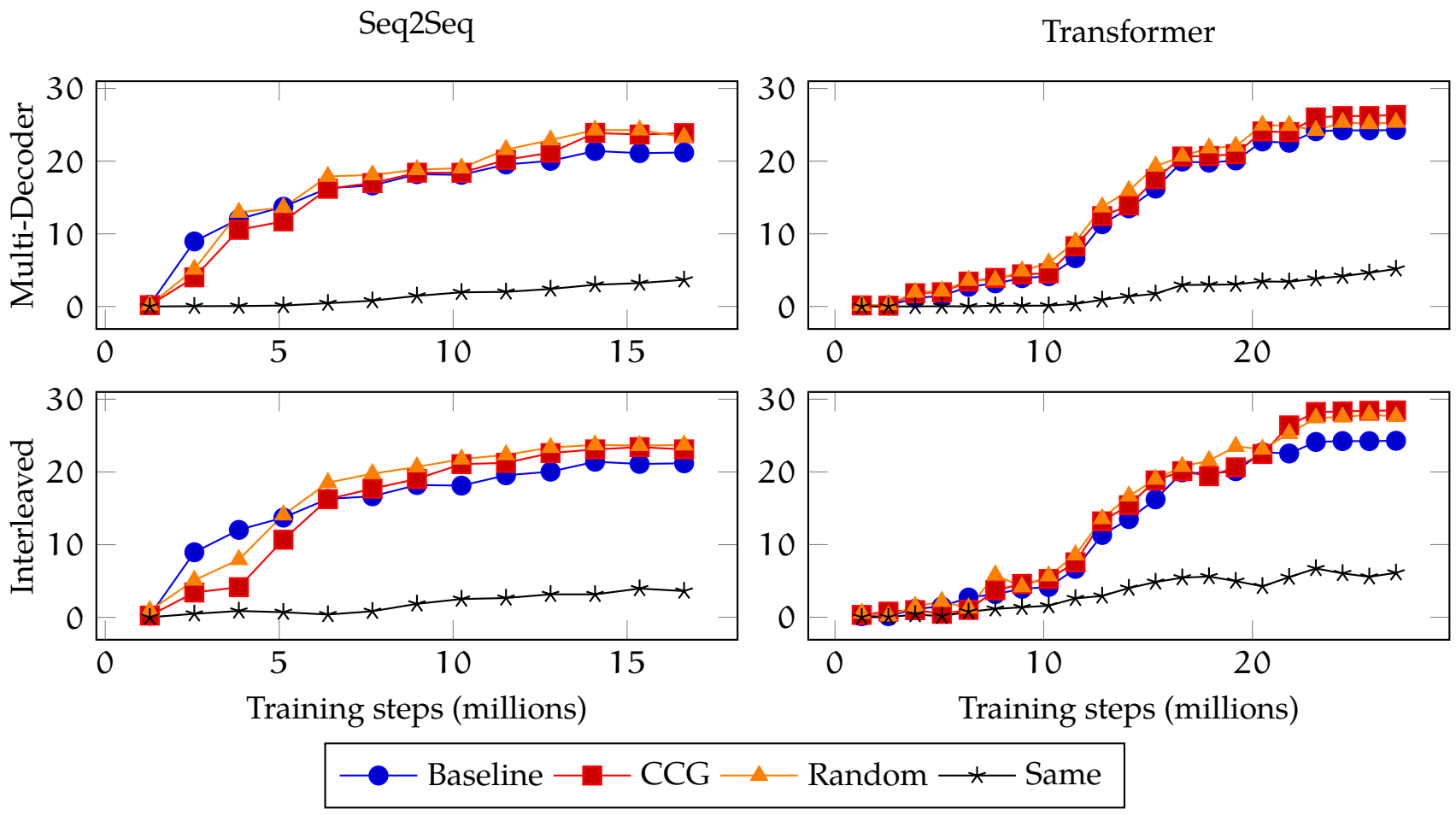

**Figure 5.3:** Performance of German-to-English translation on validation set according to BLEU score. Setups are organized by architecture (Seq2Seq, Transformer) and multi-task configuration (multi-decoder, interleaved), each one showing results for all tag schemes (CCG, random, and same tags). Baseline plots are repeated in both multi-task configurations for ease of comparison. Reproduced from Kondratyuk et al. (2019).

random tags, and one single tag. The random tags are assigned to every word in each sentence individually. This task is therefore unlearnable, or learnable only by memorizing all the training data which will never generalize to the test conditions.

Figure 5.3 observe that the setups with one single tag lead to degradation of the MT task. Roughly speaking, the model was highly rewarded for managing the simple task of generating one tag, so it had not enough motivation in learning MT.

Both CCG and random tags lead to improved MT quality over baselines. The result therefore shows that the MT improvements can not be attributed to using the linguistic knowledge but also to other, unintended factors. The decoder in the interleaved setup can e.g. benefit from the fact that it produces twice longer sequences than baseline. There is no chance to learn anything useful at the random tags positions, so the model uses this extra processing depth to refine its inner state and improve in translation itself.

**Observation 6:** *A gain observed when introducing some clever modelling technique or data manipulation, e.g. adding linguistically-aware task to multi-tasking, can actually stem from an unrelated modification of the training conditions. Dummy contrastive setups are one possible way of identifying this problem.*

# II Multilingual Models

# 6
# Overview

The second part of our book focuses on our main topic: benefiting from more than two languages in machine translation.

In the classical statistical machine translation, it was hard to share information across different language pairs and utilize knowledge learned from one language pair into another, although already Och and Ney (2001) experimented with multi-source SMT. The main issue in SMT was the increased search space leading to many search errors. With the rise of neural machine translation, many approaches to using multiple language pairs emerged.

An excellent survey of multilingual approaches to machine translation has been recently published by Dabre et al. (2020): "A Survey of Multilingual Neural Machine Translation" and we strongly recommend it to the reader. Instead of duplicating that work, we build upon it and provide our insights and reflections on work that follows that survey.

For completeness, we provide a short classification of multi-lingual models in this chapter. We then dwell upon the area of multilinguality where we have carried out considerable work ourselves, namely on transfer learning (Chapter 7). Chapter 8 is devoted to observations and recent advances in multilingual models. In Chapter 9, we move to practical aspects of deploying and training these models at smaller or medium-sized research labs. Our final remarks are collected in the conclusion, Chapter 10.

## 6.1 Classification of Multilingual Models

Dabre et al. (2020) classify multilingual models along different criteria:

**Use case.** As we discussed in Section 2.2, multilinguality can be beneficial for several reasons: improved efficiency, flexibility or quality. These possible benefits correspond more or less to the use cases highlighted by Dabre et al. (2020): multiway modelling, multi-source translation, and low resource translation.

**Level of sharing between model components.** The question here is which parts (and what proportion) of the model is common to multiple languages, i.e. trained and used in inference for sentences in that language. We reproduce the summary of this classification in Figure 6.1, including the key features of the class.

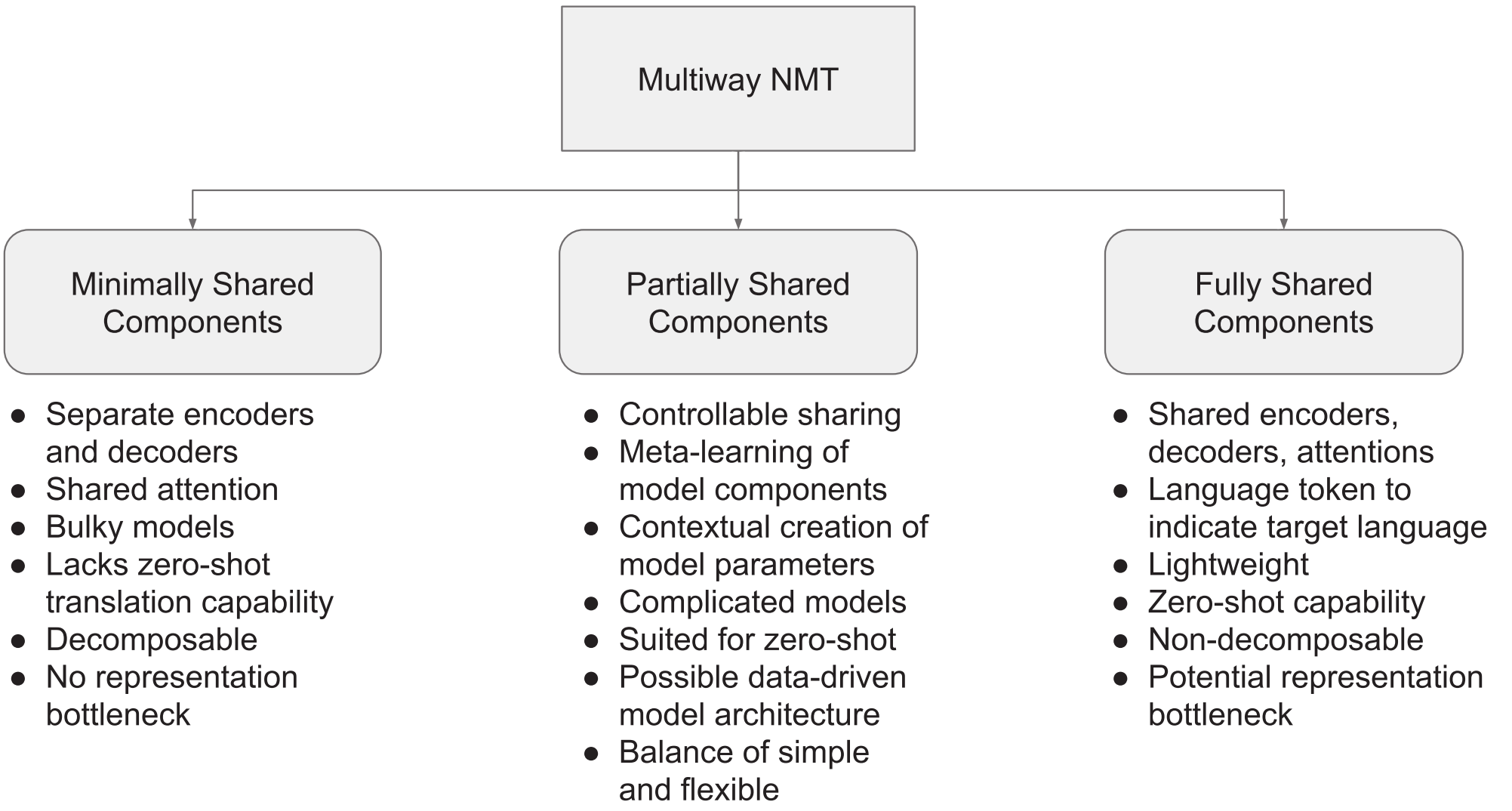

**Figure 6.1:** Classification of multi-way multilingual NMT models depending on the level of sharing of model components across languages. Reproduced from Dabre et al. (2020).

Two sub-classifications are then presented for the specific use cases of low-resource and no-resource ("unseen language pairs" in Dabre et al.'s terms) translation based on the techniques adopted.

For our purposes, it is sufficient to clarify some key terms:

**Multi-way** models include multiple "translation ways", i.e. are capable of translating among more languages. The term multi-way no longer specifies if the multiple languages are supported on the source side, on the target side or both. If we want to highlight that a multi-way model supports more languages only on one of the sides, we can talk about **multi-lingual source** or **multi-lingual target** models. Also, the sets of source and target languages can overlap. The key property of multi-way translation is that the (trained) model is capable of switching among the covered language pairs without any further training or fine-tuning.

We use the term **multi-source** to refer to models and setups that accept more inputs *at once*. Implicitly, one can assume that a multi-source model is very likely to have multi-lingual source side.[1] Similarly, we use **multi-target** for models which produce the output in multiple languages in parallel, although technically, the multi-target solution can be as simple as asking a multi-way model to process a batch with the same input repeated several times and translated into a different language in each item of the batch. We note that we avoid using the term "multi-source" for situations where the system processes one source at a time; in our terminology, these models are "(multi-way with) multi-lingual source".

[1] It is straightforward to come up with situations where a multi-source model would be uni-lingual on the source side. For instance, Zhou et al. (2019) report large gains by translating from a set of paraphrases, although that work is limited to the Bible corpus.

An important distinction between various approaches can arise from the fact which parallel data are needed during training and which inputs are needed during runtime. On the one hand, there can be multi-way models which need the full multi-parallel corpus where each sentence is available in all the source and target languages; Freitag and Firat (2020) show that this is beneficial. On the other hand, one could hope for multi-source models which would be trained on parallel data only and learn to identify the most reliable source once more inputs are provided without any multi-parallel supervision.

# 7
# Transfer Learning

Humans have the ability to utilize knowledge from previous experience when learning a new task. It helps us to learn new skills in a shorter time and with less effort. In fact, the more related a new task is, the faster we learn. In contrast, machine learning algorithms usually learn the task from random initialization on isolated data without any prior knowledge. **Transfer learning** is an approach that aims to use (in other words, to transfer) the knowledge learned for the base task to improve the performance of a new task (Bahadori et al., 2014; Farajidavar et al., 2015; Moon and Carbonell, 2017).

The base and new tasks are usually called the parent and child. They can be e.g. MT into two language pairs, or two domains, usually a large generic one as a parent, and specific narrow domain as a child (e.g. news, IT, medicine, etc.). Transfer learning can be thus used e.g. for multi-lingual NMT, and domain adaptation (Koehn and Knowles, 2017; Luong and Manning, 2015; Servan et al., 2016; Chu and Wang, 2018).

Contrary to multi-tasking (Section 3.5), where the NN is expected to be able to solve more than one task at once, and the information flow between the tasks is reciprocal, in transfer learning, the loss of ability to solve the base task is acceptable. The information flow between the parent and child task is one-directional. See illustration in Figure 7.1.

Torrey and Shavlik (2010) describe three ways of how transfer learning can improve performance. Specifically:

- improving the initial performance at the beginning of training compared to a randomly initialized model when the tasks are similar;
- shortening the time needed to reach the maximal performance;
- improving the final performance level compared to training the model without the transfer.

These three potential improvements are illustrated on learning curves in Figure 7.2.

## Structure of this Chapter

This chapter is organized as follows. First, we specify the terminology used across this book (Section 7.1). Then, we put sections about transfer learning methods: cold-start, vocabulary transformations, and warm-start (Sections 7.2 to 7.4).

The success of transfer learning is not always guaranteed. For example, when transferring from a weakly related task, it may hinder the final performance in the target task. A phenomenon known as the negative transfer has been well recognized by the transfer learning community (Pan and Yang, 2010; Wang et al., 2019b). Negative transfer is analyzed from several angles in Section 7.6.

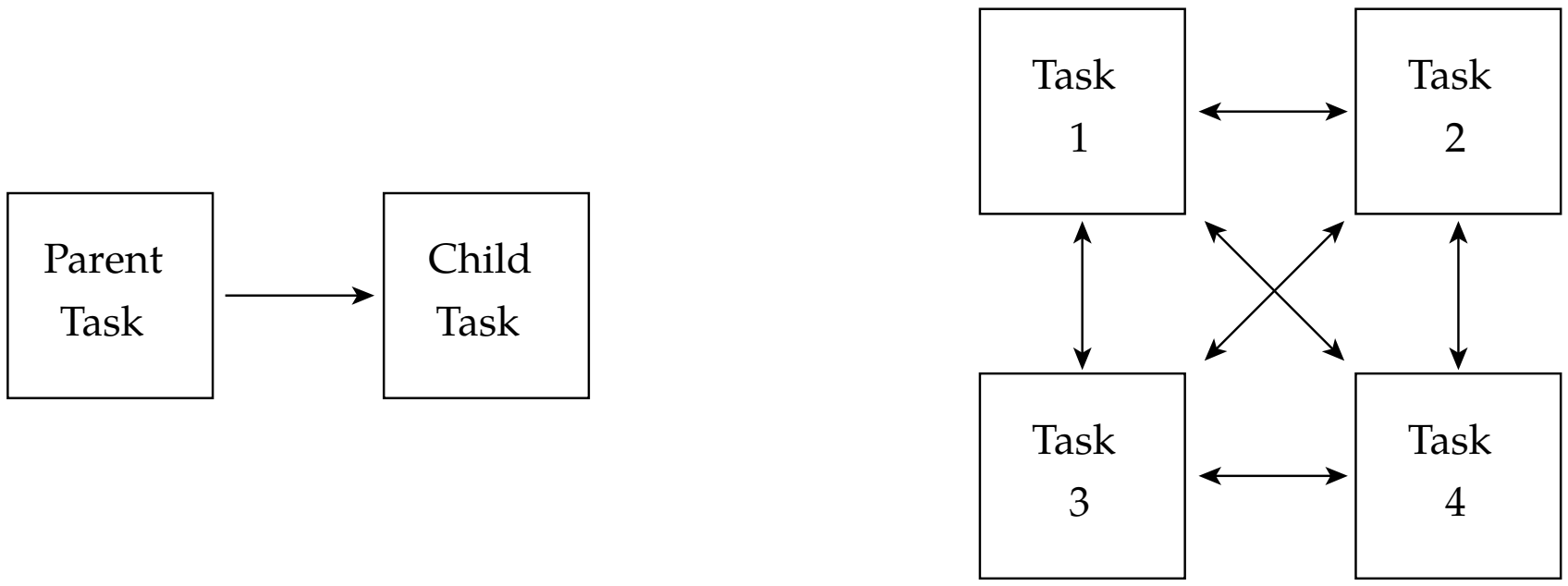

**Figure 7.1:** The information flows only in one direction from parent to the child task in transfer learning (left) compared to the multi-task learning (right), where the information flows freely among all tasks improving them altogether.

Next, three more sections with an analysis of transfer learning follow: they deal with the position of the shared language (Section 7.7), data size versus language relatedness (Section 7.8), and linguistic features versus better initialization (Section 7.9). Finally, we provide a case study of transfer learning with back-translation (Section 7.10).

In this chapter, we reuse text from Tom Kocmi's dissertation thesis (Kocmi, 2019), parts of which have been also published in two papers. The cold-start scenario is described in Kocmi and Bojar (2020), and warm-start scenario is investigated in the paper Kocmi and Bojar (2018). This chapter includes results as well as short textual parts from the two papers. For simplicity, we disregard the technical notes for reproducing the results, such as the specific implementation and model size used in particular experiments. They can be found in the dissertation thesis and in the papers.

What this chapter does not include is an up-to-date comprehensive survey of related works about transfer learning for NMT. For that, we refer to Dabre et al. (2020) and to the thesis (Kocmi, 2019).

## 7.1 Terminology

**Parent-Child** The main idea behind transfer learning is to pass on learned knowledge from one model to another. We denote the first model from which parameters are transferred as a **parent** and the designated model as a **child**.

In literature, we can often find a naming convention of *teacher* and *student*, however, it is more related to the knowledge distillation (Hinton et al., 2014), where the parent model (the teacher) is used to generate examples instead of directly sharing parameters. Another terminology is *source* and *target* tasks (Torrey and Shavlik, 2010), which is unsuitable for us as we use these terms when referring to the source and target language of the language pair.

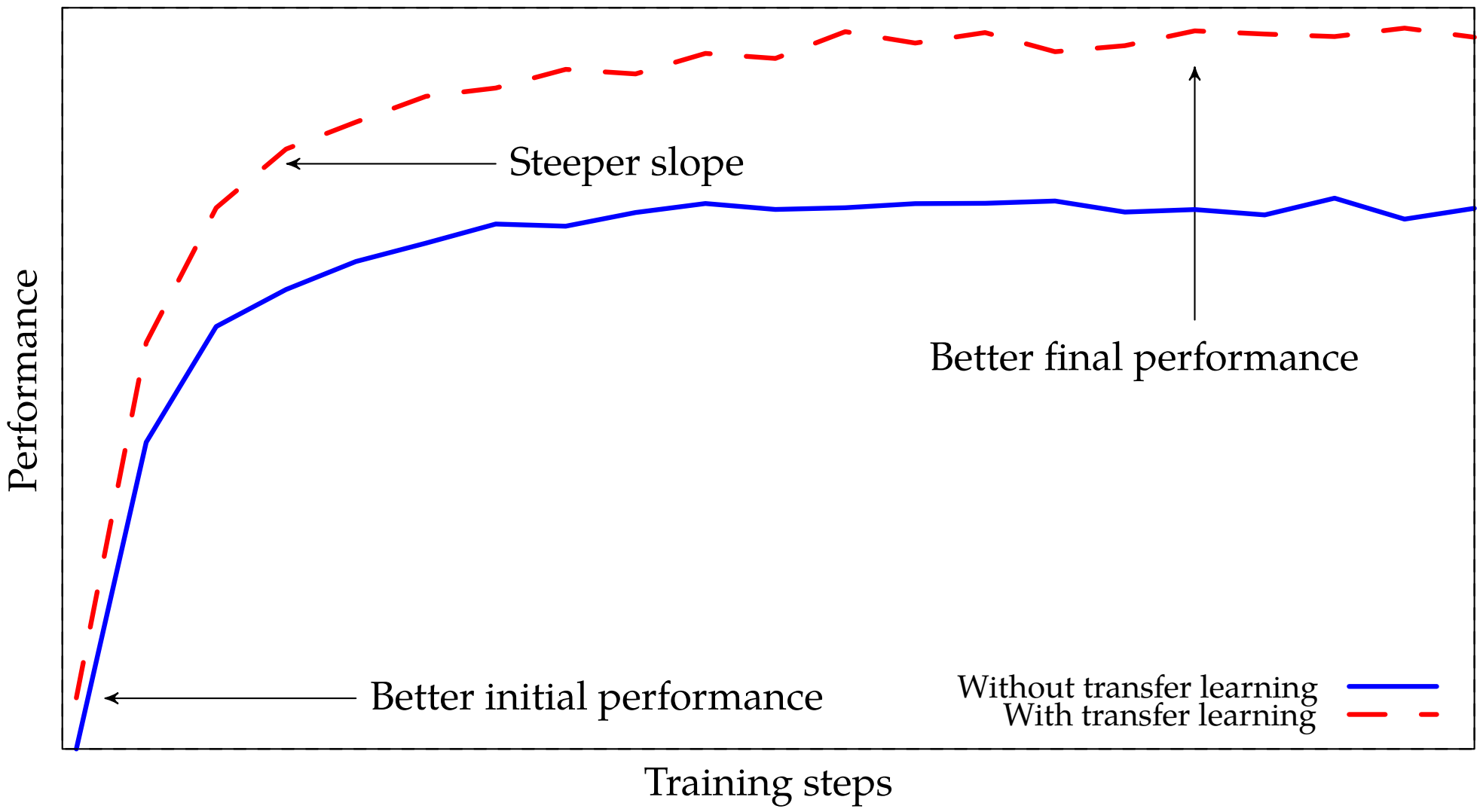

**Figure 7.2:** Three impacts where transfer learning improves the training process. These are real learning curves for warm-start transfer learning on English→Estonian (see Section 7.4).

Notably, in the NMT transfer learning it is customary to use the term "parent-child" (Zoph et al., 2016; Nguyen and Chiang, 2017), thus we are going to use it throughout this book.

Nonetheless, we use the naming convention of "parent-child" in this work for all scenarios, where there is a precedent model needed for training the second model regardless of the transferring technique. For example, during the back-translation (see Section 3.6), the parent model generates training data for the child model, but no learned parameters are shared between models. Thus, the child always specifies our designated task or the language pair for which we are trying to get the best performance. In contrast, we are not interested in the performance of the parent.

**Sharing the Languages** Both parent and child can use different language pair. For the parent model that translates from language XX to language YY (XX→YY), we recognize three scenarios:

- Shared-source language – a scenario where the source language is equal for parent and child. In other words, the child model translates XX→AA.

- Shared-target language – a scenario where the target language is equal, therefore the child model translates AA→YY.
- No-shared language – a scenario with no language shared, i.e. the child model translates AA→BB.

**Cold and Warm-Start** When we want to train the child with a different language pair than the parent, we need to tackle the problem with vocabulary mismatch between parent and child languages. There are two groups of approaches to solving the vocabulary problem based on their application either before training the parent model, or after it.

Neubig and Hu (2018) call the approaches **warm-start** and **cold-start**. In the warm-start approach, we have the child training data available at the time of training the parent, and we can take steps to prepare the parent model for transfer learning. In contrast, the cold-start approaches use a general parent that has not been modified in advance in any way for the child language pair and apply different modifications before training the child model.

Intuitively, we expect the warm-start to perform better than cold-start because it can better handle the child language pair to some extent. However, the additional training time of the child's specific parent model can be costly. For example, whenever researchers compare various language pairs or hyperparameter setting, they would need to train an individual parent for each such scenario and the time and hardware cost would considerably increase. Therefore, they can consider a scenario to train a strong general NMT system only once and use it repetitively in the cold-start transfer learning. Furthermore, there are situations where the ability to quickly learn the MT model given only a small amount of training data can be crucial. For example, when a crisis occurs in a region where people speak an under-resourced language, quick deployment of the MT system translating from or to that language can make a massive difference in the impact of the provided support (Lewis et al., 2011). In such a case, having a strong general NMT system available for a quick transfer learning of any low-resource language pair is a great advantage.

## 7.2 Cold-Start

The main problem of transfer learning is the mismatch of parent and child vocabulary. The cold-start transfer learning tackles the problem after the training of the parent by modifying the parent model right before transferring parameters to the child model.

Whenever the parent model uses vocabulary with a high overlap with the child's vocabulary, we can ignore the differences and train the child with the parent's vocabulary. We call this approach **direct transfer** and discuss it in Section 7.2.1. The second option is to transform the parent vocabulary right before the child's training in various ways to accommodate the needs of the child language pair. Approaches from the second group are discussed in Section 7.3.

All cold-start approaches rely on the ability of neural networks to quickly adapt parent parameters to new conditions, i.e. segmenting words usually to more tokens than in parent model or remapping parent subwords embeddings to unrelated child subwords. We show that NMT can quickly adapt and obtain better performance on a given child language pair than by training from random initialization.

### 7.2.1 Cold-Start Direct Transfer

Subword-based vocabulary can represent any text from any language by breaking the words down to characters or bytes.[1] In Kocmi and Bojar (2020), we exploit this behavior that a parent vocabulary can encode any child's words and use the parent model as it is. We call it direct transfer.

In the direct transfer approach, we ignore the specifics of the child vocabulary and train the child model using the same vocabulary as the parent model. We take an already trained parent model and use it to initialize parameters for a child language pair. In other words, we continue the training process of the parent model on the child training set without any change to the vocabulary or hyper-parameters. This applies even to the training meta-parameters, such as the learning rate or moments.

This method of continued training on different data while preserving hyper-parameters is essentially a domain adaptation technique, where the domain is a different language pair.

The intuition behind the direct transfer is that NN is robust enough that the vocabulary mismatch can be disregarded altogether as long as there is some overlap between the child and parent vocabulary. This is mainly due to the usage of subword tokens, which segment any text into a sequence of allowed subwords (see Section 3.2.2). However, direct transfer suffers from a deficiency in the segmentation of child words, which can lead to splitting words to individual characters or even bytes that are difficult for NMT to correctly translate.

Thanks to the simplicity of the direct transfer, it can be easily applied to existing training procedures as it does not need any modification to the NN frameworks.

In the following sections, we discuss automatically assessed translation quality. We show the results of direct transfer in Section 7.2.2, followed by an analysis of the usage of parent vocabulary in Section 7.2.3 and introduction of a problem with vocabulary mismatch in Section 7.2.5.

### 7.2.2 Direct Transfer Results

We start with the results of direct transfer method (from Kocmi and Bojar, 2020), which uses parent vocabulary without any change. The results of our evaluation are in Table 7.1. In comparison to the baseline, the performance of direct transfer is significantly better in both translation directions in all cases except for Odia and Russian, which use a different writing script than Latin.

[1] The standard implementation of BPE segmentation by Sennrich et al. (2016b) can not represent unknown characters by breaking them to bytes. However, the implementation could be extended to support encoding of bytes by escaping the byte representation in the same way as in the wordpieces.

| Translation direction | Baseline | Direct Transfer | Difference |
|---|---|---|---|
| EN→Odia | **3.54** ‡ | 0.26 | -3.28 |
| EN→Estonian | 16.03 | **20.75** ‡ | 4.72 |
| EN→Finnish | 14.42 | **16.12** ‡ | 1.70 |
| EN→German | 36.72 | **38.58** ‡ | 1.86 |
| EN→Russian | **27.81** ‡ | 27.19 | -0.62 |
| EN→French | 33.72 | **34.41** ‡ | 0.69 |
| French→Spanish | 31.10 | **31.55** ‡ | 0.45 |
| Estonian→EN | 21.07 | **24.36** ‡ | 3.29 |
| Russian→EN | **30.31** | 30.49 ‡ | 0.18 |

**Table 7.1:** "Baseline" is a model trained from random initialization with own specific vocabulary. "Direct Transfer" is using the parent vocabulary. Models in the top part of the table use the English→Czech parent model, models in the bottom part use Czech→English. The scores and difference are in BLEU. The ‡ represents significantly better results based on significance test described in Section 5.2. For experiments with Russian and Odia, we increased the threshold for maximum number of subwords from 100 to 500 in order to avoid dropping large amount of training examples due to high segmentation rate. Reproduced from Kocmi and Bojar (2020).

Importantly, our baseline, trained only on child data, has an advantage over cold-start transfer learning as it uses child-specific vocabulary. Closer baseline to our transfer learning setup would be to use the parent vocabulary even for baseline, which would lead to an even larger difference in the performance. However, we decided to use the stronger baseline.

The Estonian–English pair confirms that sharing the target language improves performance as previously shown on similar approaches (Zoph et al., 2016; Nguyen and Chiang, 2017). Moreover, we show that the improvements are significant for the translation direction from English, an area of transfer learning neglected in previous studies.

The largest improvement of 4.72 BLEU is for the low-resource language pair English →Estonian. Furthermore, the improvements are statistically significant even for a high-resource language such as 0.69 BLEU increase for a high-resource English→ French. To the best of our knowledge, we are the first to show that transfer learning in NLP can be beneficial also for high-resource languages.

**Observation 7:** *Direct transfer (simple fine-tuning on the child data) can significantly improve the performance of the child model in both translation directions for both low-resource and high-resource language pairs.*

The basic intuition behind the improvements in translation direction into English is that the models reuse the English language model in the decoder and therefore the improvements are due to better ability to generate grammatically correct sentences without the context of the source language. Although better decoder's language model could be one of the reasons behind the improvements, it can not be the only explanation since we see the improvements also for translation direction where English is the source side, and therefore the decoder has to learn a language model for the second language.

Furthermore, we get improvements even for child language pairs in the no-shared language scenario. In our study, we evaluated French→Spanish, which got a 0.45 BLEU improvement. Although, in this particular case, we need to take into account that it could be partly due to these languages being linguistically closer to the parent's source language English.

In Section 7.7, we discuss that the shared-target language is easier for NMT than the shared-source language. It is also the main reason why we compare more systems in the direction from English rather than to English.

The results of such performance boost are even more surprising when we take into account the fact that the model uses the parent vocabulary and thus splits words into considerably more subwords, which we carefully analyze in Section 7.2.5. It suggests that the Transformer architecture generalizes very well to short subwords and is robust enough to generate longer sentences.

In conclusion, the direct transfer learning improves the performance in all cases except English→Odia, English→Russian and Russian→English. In order to shed light on the failure of these languages, we need to analyze the parent vocabulary.

### 7.2.3 Parent Vocabulary Effect

The problem of OOV words is solved by using subwords segmentation at the cost of splitting less common words into separate subword units, characters, or even individual bytes, as discussed in Section 3.2.2. The segmentation applies deterministic rules on the training corpus to generate the subword segmentation that minimizes the number of splits for the observed word frequencies to fill up the vocabulary of a predefined size.

However, when using a subword segmentation created for a different language pair, the condition of the optimal number of splits is not guaranteed. Especially more linguistically distant languages that contain only a small number of common character n-grams need more splits per word.

The example in Figure 7.3 shows that using a vocabulary for a different language leads to segmenting words to substantially more tokens, especially in the case when the language contains characters not available in the vocabulary. This is most crucial as the unknown character is first transformed into byte representation ("\237;" in Figure 7.3) that is later handled as a standard text.

| | |
|---|---|
| Czech vocabulary: | {bude, doma_, end, me_, ví, vík} |
| English vocabulary: | {bud, dom, end, ho, me, week_, will} |
| Czech Sentence: | `O víkendu budeme doma.` |
| Segmented by Czech vocab.: | `O_ vík end u_ bude me_ doma_ ._` |
| Segmented by English vocab.: | `O_ v \ 2 3 7 ; k end u_ bud e me_ dom a_ ._` |

**Figure 7.3:** A toy example of using English wordpiece segmentation onto Czech sentence. For simplicity, we suppose the vocabularies contain all ASCII characters in addition to the tokens specifically mentioned.

| | Source language | | Target language | |
|---|---|---|---|---|
| Language pair | Specific | EN-CS | Specific | EN-CS |
| EN–Odia | 1.2 | 1.4 | 3.7 | 19.1 |
| EN–Estonian | 1.2 | 1.2 | 1.1 | 2.3 |
| EN–Finnish | 1.2 | 1.2 | 1.1 | 2.6 |
| EN–German | 1.2 | 1.2 | 1.3 | 2.5 |
| EN–Russian | 1.3 | 1.4 | 1.3 | 5.3 |
| EN–French | 1.3 | 1.4 | 1.6 | 2.5 |
| French–Spanish | 1.3 | 2.1 | 1.2 | 2.1 |

**Table 7.2:** Average number of tokens per word (tokenized on whitespace) when applied to the training corpora. "Specific" represents the vocabulary created specifically for the examined language pair. "EN-CS" corresponds to the use of Czech–English vocabulary.

In Figure 7.3, English vocabulary doubles the number of tokens in the investigated sentence relative to the Czech vocabulary.

Direct transfer approach uses the parent vocabulary, which can lead to segmenting the child training set into many individual tokens that could harm the MT performance. In order to study this effect, we examine the influence of using different vocabulary on the training dataset of the child.

We consider the parent Czech–English vocabulary (Popel, 2018) used in our experiments and apply it for segmentation of language pairs and compare the average number of subwords per word. We examine the language pairs and their training sets that are used in experiments regarding direct transfer.

Table 7.2 documents the splitting effect of various vocabularies. When using the language-pair-specific vocabulary (column "Specific"), the average number of subword tokens per word (denoted as **segmentation rate**) is relatively constant for English, around 1.2 subwords per word, as well as other languages except for Odia with 3.7 tokens per word, which we explain in Section 7.2.4. This regularity possibly emerges from the size of vocabulary and the number of words in both languages.

If we use the Czech–English vocabulary on the child dataset (column "EN-CS"), there is an apparent increase in the average number of subword tokens per word. For example, German has twice as many tokens per word compared to the language-specific vocabulary that has 1.3 tokens per word on average. Russian has four times more tokens per word due to Cyrillic, similarly for the Odia script.

Russian Cyrillic alphabet happens to be contained in the parent vocabulary together with 59 Cyrillic bigrams and 3 trigrams, which leads to 5.3 tokens per word. The Odia script is not contained in the Czech–English vocabulary at all, leading to the splitting of each word into individual bytes, which explains the 19.1 tokens per word (see Figure 7.3).

The first language is not affected by the parent vocabulary much (only slightly for the French-Spanish language pair) because English is shared between both the child and the parent vocabulary. The second language that differs between parent and child approximately doubles the number of splits when using parent vocabulary (see the difference between both columns "EN-CS").

**Observation 8:** *Empirically, wordpiece vocabulary roughly doubles the segmentation rate for different child languages that use the same script as the parent language, compared to the parent segmentation rate.*

It is necessary to mention that the datasets have various domains and various sizes and therefore the average number of tokens could be different on various domains even for the same language pair. The size of the vocabulary[2] is also crucial as it defines the number of available subwords. Moreover, the length relation between the source and target sentences influence the final vocabulary.

The use of a different segmentation roughly doubles the number of tokens per sentence for languages using the same writing script. Therefore the NMT models that use Direct transfer have to adapt to different sentence length in comparison to the parent. However, as we showed in Section 7.2.2, the direct transfer significantly improves the performance over the baseline showing the robustness of NMT.

### 7.2.4 Odia Subword Irregularity

We try to shed some light, why Odia has, on average, more tokens per word after subword segmentation even when using a language-specific vocabulary. The Odia script (also called Oriya script) has 52 characters, which lead to more character combinations than in English, which we believe is linked to a higher number of subwords per word.

[2] We use vocabulary with 32k subwords in all experiments.

| Child language pair | 1st language | 2nd language | Both |
|---|---|---|---|
| EN–Czech | 71.1% | 98.0% | 98.8% |
| EN–Odia | 23.3% | 0.8% | 23.9% |
| EN–Estonian | 54.8% | 39.7% | 57.0% |
| EN–Finnish | 59.2% | 54.5% | 60.6% |
| EN–German | 58.4% | 52.9% | 59.9% |
| EN–Russian | 68.6% | 67.0% | 71.4% |
| EN–French | 64.6% | 64.5% | 65.5% |
| French–Spanish | 55.1% | 54.0% | 57.1% |

**Table 7.3:** The percentage of the parent vocabulary tokens reused in the child's training set. The vocabulary is shared for both languages. The column "Both" represents the number of vocabulary tokens used by both languages.

To confirm our intuition, we investigate the Odia–English vocabulary. The average length of Odia tokens in the vocabulary is 4.2 characters compared to the 6.9 characters for English subwords in the same vocabulary. The average length of a non-segmented word in the Odia–English training set is 6.4 characters for Odia and 5.2 characters for English. With that in mind, Odia has on average longer words but uses shorter subwords than English, which leads to the substantially higher average number of tokens per word as reported in Table 7.2 in comparison to other languages.

This is mainly due to the size of vocabulary, which is not enough for the Odia–English language pair. Larger vocabulary would contain longer Odia subwords, thus would make the segmentation less frequent. This fact could be one of the reasons why the performance of direct transfer is worse than baseline as reported in Section 7.2.2. On the other hand, Odia is a low-resource language, and having large vocabulary would result in fewer examples per individual subwords in training data.

### 7.2.5 Vocabulary Overlap

Direct transfer uses parent vocabulary, and we showed how it increases the segmentation of the child's training corpus in Section 7.2.3. Now, we examine what percentage of the parent vocabulary is used by the child language pair and investigate how large is the part of parent vocabulary that is left unused with the child language pair.

In order to find tokens from the parent vocabulary that are used by the child model, we segment the training corpus of the investigated languages with the Czech–English vocabulary and count how many unique tokens from the vocabulary appear in the segmented child's training corpus.

The percentage of used tokens are in Table 7.3. Before examining the results, we need to mention that the training sets are usually noisy, and some sentences from other languages can easily appear in them. For example, it is often a case that part or whole sentence is left untranslated in the parallel corpus. For the simplicity of our argument, we do not remove any foreign sentences nor ignore the least used tokens in any way. Therefore, the actual used part of parent vocabulary by a given language is smaller than presented in Table 7.3.

With that in mind, we can notice that English always uses more tokens from the vocabulary than the second language. This is caused by English being the shared language and already present in the vocabulary. Although the reverse is true for the original dataset of Czech–English where English occupies a smaller part of the vocabulary (71.1%) than Czech (98.0%). The reason why the total is not 100% but only 98.8% (see column "Both") is possibly due to a slightly different training set as the vocabulary was prepared by Popel (2018).

We can see that the Odia does not use many of available subwords, as concluded in the previous section. Interestingly and contrary to our previous findings, Russian utilizes a substantial part of the parent vocabulary. After a closer examination of the training corpus, we noticed that the Russian part contains many Czech and English sentences. When we counted only subword tokens that contain at least one Cyrillic character, the used part of vocabulary dropped to 0.3% for Russian confirming the previous findings with extremely high segmentation rate than other languages.

The most important result is that most language pairs use around 60% of parent vocabulary. This means that remaining tokens are left unused and only slow down the training and inference because the model has to calculate the softmax over the size of the vocabulary.

## 7.3 Vocabulary Transformation

Direct transfer relies on the fact that we use subword units that can be used to encode any textual string. The pre-processing splits unseen words into several subwords, characters or possibly down to individual bytes. This feature ensures that the parent vocabulary can, in principle, serve for any child language pair, but it can be highly suboptimal, segmenting child words into too many subwords, where individual subword units do not contain relevant meaning. As expected, this is most noticeable for languages using different scripts (see the statistics in Table 7.2). Also, the child does not use the whole vocabulary leaving around 40% unused, which only slows down the training and inference process. In order to avoid both problems, the unused token positions could be reused to represent subwords more suitable for the need of the child language pair.

In this section, we describe a vocabulary transformation that we proposed in Kocmi and Bojar (2020). It changes unused subwords in parent's vocabulary with child-specific ones. We show that a straightforward replacement of subwords leads to significant improvements in performance.

NMT models associate each vocabulary item with its embedding. When transferring from the parent to the child, we can remap subwords and their assigned em-

**Input:** Parent vocabulary (an ordered list of parent subwords) and the training corpus for the child language pair.
Generate custom child vocabulary with the maximum number of subwords equal to the parent vocabulary size **for** *subword S in parent vocabulary* **do**
  **if** *S in child vocabulary* **then**
    continue
  **else**
    Replace position of S in the parent vocabulary with the first unused child subword not contained in the parent
  **end**
**end**
**Result:** Transformed parent vocabulary

**Figure 7.4:** Algorithm for transforming parent vocabulary to contain child subwords and match positions for subwords common for both of language pairs. Reproduced from Kocmi and Bojar (2020).

bedding as trained in the parent model without any modification to the architecture. The remapped subwords become associated with embeddings that initially behave as trained for the original subwords and the NMT has first to retrain them.

The algorithm for vocabulary transformation is provided in Figure 7.4.

### 7.3.1 Results with Transformed Vocabulary

Direct transfer significantly outperforms the baseline, trained only on the child data, whenever the parent vocabulary does not segment the child training sentences into many tokens. In this section, we evaluate our transformed vocabulary approaches, which remaps unused parent subwords to useful child-specific ones.

The first two columns of Table 7.4 are the same as in Table 7.1. We added a column with results of the transformed vocabulary. We see "Transformed Vocabulary" delivering the best performance for all language pairs except English→Estonian, significantly improving over "Direct Transfer" in most cases. The only exceptions are Estonian→English, English→French and French→Spanish, where neither of the systems is significantly better than the other, however, both of them are significantly better than the baseline.

Furthermore, it confirms that our transformed vocabulary successfully tackles the problem of direct transfer when the child language uses a different writing script such as English→Odia, English→Russian, and Russian→English.

| Model | Baseline | Direct Transfer | Transformed Vocab |
|---|---|---|---|
| EN→Odia | 3.54 | 0.26 | **6.38** ‡ |
| EN→Estonian | 16.03 | **20.75** | 20.27 |
| EN→Finnish | 14.42 | 16.12 | **16.73** ‡ |
| EN→German | 36.72 | 38.58 | **39.28** ‡ |
| EN→Russian | 27.81 | 27.19 | **28.65** ‡ |
| EN→French | 33.72 | 34.41 | **34.46** |
| French→Spanish | 31.10 | 31.55 | **31.67** |
| Estonian→EN | 21.07 | 24.36 | **24.64** |
| Russian→EN | 30.31 | 30.49 | **31.38** ‡ |

**Table 7.4:** "Transformed Vocab" has the same setting as Direct Transfer but merges the parent and child vocabulary as described in Section 7.3. The structure is the same as in Table 7.1. The baseline uses child-specific vocabulary. The statistical significance ‡ is measured between direct transfer and Transformed Vocabulary. Reproduced from Kocmi and Bojar (2020).

We see that cold-start transfer learning is not restricted to the low-resource scenario as it also improves high-resource language pairs: Finnish–English, German–English, Russian–English, and French–English.

**Observation 9:** *Cold-start transfer learning (fine-tuning with vocabulary adaptation) improves the performance of both low-resource and high-resource language pairs.*

Interestingly, the cold-start transfer learning technique improves even the scenario with no-shared language, in this case French→Spanish.

Furthermore, the positive results imply that the neural network is robust enough to correct the randomly assigned child-specific embeddings and therefore reuse even more capacity of the parent model.

### 7.3.2 Various Vocabulary Transformations

Our Transformed Vocabulary technique assigns unmatched subwords mostly at random. However, there are many other variants. We propose several of them and evaluate them in this section.

We noticed that the vocabulary is structured in stages roughly based on the frequency of subwords in the corpora. This is due to the vocabulary creation that adds less frequent subwords in stages until reaching the requested size of the vocabulary. Therefore, we call the technique from the previous section "Frequency-based".

In contrast to frequency, we can assign tokens at random. Either all of them or only unmatching tokens. We call the former approach "Everything random". It is when all subword tokens are first shuffled and then assigned at random. This approach does not match any tokens. Therefore the NMT needs to learn even tokens that have been used by the parent model. The latter approach is called "Unmatched random". It first assigns subwords that are in parent and child vocabulary. Then it assigns the remaining child tokens at random.

| Language pair | EN→Estonian | Estonian→EN |
|---|---|---|
| Frequency-based | 20.27 | 23.32 |
| Everything random | 16.41 | 19.84 |
| Unmatched random | **20.28** | 22.45 |
| Levenshtein distance | 20.04 | **23.66** |

**Table 7.5:** Comparison of various approaches to replacing tokens in parent vocabulary.

The last option is the assignment of subwords based on some distance. We select the Levenshtein distance, which measures the number of edits between two strings. The vocabulary created by this technique assigns subwords iteratively by increasing the allowed distance for assignment. In other words, it starts by assigning all matching subwords (distance 0), then subwords that have a distance of one edit, then two edits and so on.

The results of comparing various approaches to replacing tokens in parent vocabulary are presented in Table 7.5. All approaches reach comparable performance except the "Everything random" assignment. Thus, we found no significant differences in the performance as long as subwords common to both the parent and child keep their embeddings, i.e. are mapped to the same index in the vocabulary. The subwords unique to the child vocabulary can be assigned randomly to the unused parent embeddings. A similar result was observed by Zoph et al. (2016). They show that the random assignment of words in their approach works as well as an assigning based on lexical similarity.

**Observation 10:** *In vocabulary transformation for transfer learning, it is necessary to preserve tokens shared between the parent and child. The assignment for the remaining tokens does not play an important role.*

### 7.3.3 Training Time

Lastly, in Kocmi (2019) we also evaluate the total training time needed for cold-start transferred models to reach the best performance. The times in Table 7.6 represent the number of steps needed to reach the best performing models from Table 7.4. The steps are comparable due to the equal batch size. However, due to the training fluctuations in performance, it is not possible to define the exact step when the model converged. Therefore, the results should be seen only as a rough estimate of the training time. We use the stopping criterion as defined in Section 4.3.1.

| Language pair | Baseline | Direct Transfer | Transformed vocab |
|---|---|---|---|
| EN→Odia | 45k | 47k | **38k** |
| EN→Estonian | 95k | **75k** | **75k** |
| EN→Finnish | 420k | **255k** | 270k |
| EN→German | 270k | 190k | **110k** |
| EN→Russian | 1090k | 630k | **450k** |
| EN→French | 820k | **660k** | 720k |
| French→Spanish | 390k | 435k | **375k** |
| Estonian→EN | 70k | **30k** | 60k |
| Russian→EN | 980k | **420k** | 700k |

**Table 7.6:** The number of steps needed for a model to converge. We present the step where the model has the highest performance on the development step based on the stopping criterion described in Section 4.3.1.

Both transfer learning approaches converged in a comparable or lower number of steps than the baseline as we see in Table 7.6. The reduction in the number of steps is most visible in English→German and English→Russian, where we got to less than half of the total number of steps.

**Observation 11:** *Cold-start transfer learning converges faster than training from random initialization.*

As mentioned earlier, the results are only approximate. However, based on the broad range of our experiments where we compared nine language pairs using various scripts and various training corpora sizes we conclude that both approaches of transfer learning are faster than training the model from random initialization.

## 7.4 Warm-Start

In Section 7.2, we discussed the cold-start scenario where the parent model is trained in advance without prior knowledge about the child's language pair. The main disadvantage of that method is the need to reuse the parent vocabulary, which is associated with trained embeddings. The direct transfer, therefore, has a problem with a high segmentation rate and leaving roughly 40% of vocabulary unused. The vocabulary transformation overcame these problems by randomly assigning child's subwords to unused embeddings, but this method restricted the maximal size of child vocabulary to be equal to the parent's vocabulary. Furthermore, the randomly assigned child subwords embeddings first had to be retrained. The warm-start scenario allows us to overcome such problems by preparing the parent model in advance for the upcoming transfer learning to the child's language pair.

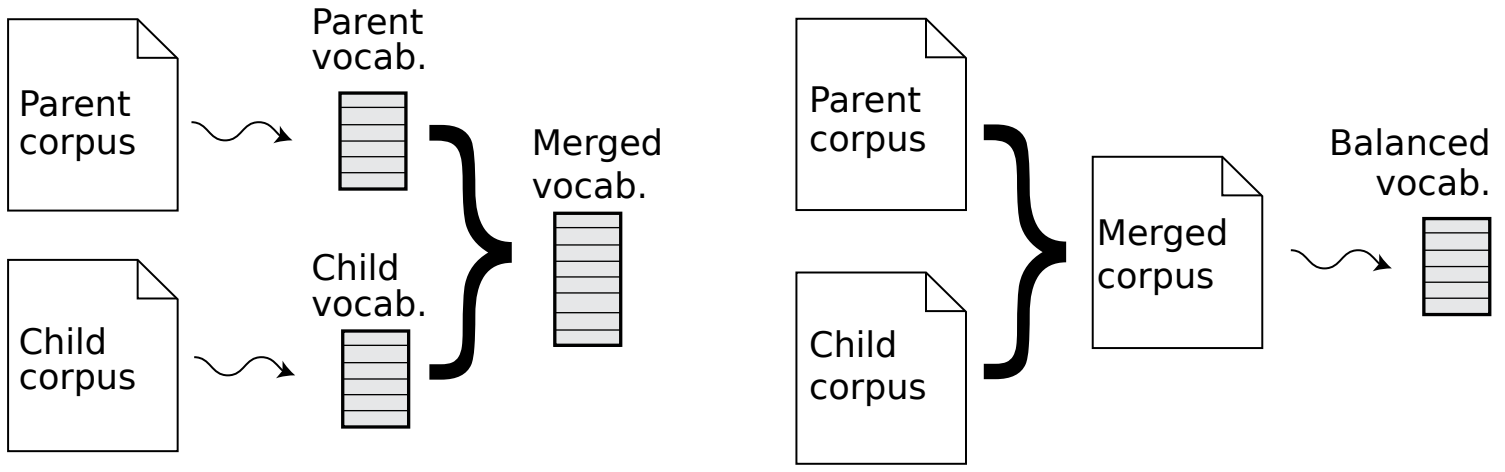

**Figure 7.5:** The process of generation of shared vocabularies: Merged (left) and Balanced (right) Vocabulary.

The basic idea of how to avoid problems with randomly assigned embeddings is to directly use child-specific vocabulary at the time of parent model training. The use of a child's vocabulary in the parent model would inevitably undermine the performance of the parent model because we restrict the parent vocabulary. We studied this effect on child models in Section 7.2.3, where we showed that inappropriate vocabulary segments text to an increased number of tokens. However, in the transfer learning task, we do not pay attention to the final performance of the parent language pair. Thus, we can ignore this performance drop.

The experiments with warm-start are published in Kocmi and Bojar (2018).

### 7.4.1 Vocabulary Shared between Parent and Child

In addition to using the child-specific vocabulary, we propose other variants of shared vocabulary in order to find the best approach for transfer learning. So far, we described two options: either to use parent-specific vocabulary (direct transfer) or child-specific vocabulary. Each of them has similar problems, but either for the parent or the child model. A solution is to create a vocabulary that is shared between both parent and child language pairs.

We propose two approaches of creating a shared vocabulary. One option of a shared vocabulary containing both the parent as well as the child subwords can be created by merging their specific vocabularies. We call this approach "Merged Vocabulary". The second option is to utilize the algorithm for subword generation (wordpiece or BPE) by applying it to the concatenated corpus of both parent's as well as the child's vocabulary. This way, it prepares a balanced vocabulary based on the subword frequency in both corpora. We call the second approach "Balanced Vocabulary". Both approaches are illustrated in Figure 7.5.

**Merged Vocabulary** Merged Vocabulary is related to the cold-start Transformed Vocabulary as it tries to combine two separate vocabularies into one. The process of vocabulary generation in Merged Vocabulary is as follows: first, we generate vocabulary for both parent and the child language pair separately, and then we merge both vocabularies and remove duplicated subwords. Because the vocabularies are text files where each subword is on an individual line, the merge operation is a simple concatenation of both vocabularies together. Therefore this approach applies only to wordpiece-type vocabulary, which is not dependent on the ordering of the vocabulary such as the BPE algorithm.

**Balanced Vocabulary** The second approach, Balanced Vocabulary, is obtained by concatenating both parent's and child's training corpora into one corpus, which is then used to generate the vocabulary by the wordpiece (possibly also BPE) algorithm. Note that this corpus is used only for the vocabulary generation purposes and we are not using it for training NMT models.

The Balanced Vocabulary method is sensitive to the total size of the corpus. Whenever one of the corpora is bigger, more subwords of that language pair get into the final vocabulary. Mainly due to our focus on low-resource language pairs, we try to balance the amount of language-specific subwords in the vocabulary by randomly selecting an equal number of sentences from both corpora instead of using all sentences. Thus, the mixed corpus contains 25% sentences from each of the four languages. We need to note that we do not take into account the number of words per sentence as it differs across corpora and also languages. Thus, we suppose that corpora are usually already segmented on a sentence level and we disregard the average length of the sentences.

**Training Procedure** Whenever we obtain the vocabulary by either of the methods, we follow the transfer learning pipeline that differs from cold-start by the fact that we need to train the parent model for each child. The training is the following: we train the parent model with a given vocabulary until convergence, followed by continued training on the child dataset. We remind the reader that in the case of warm-start transfer learning, we do not change the vocabulary nor any other (hyper)parameters when we exchange the training corpora. Therefore the NN is not forced to completely retrain embeddings when we change the training corpora from parent to child. It avoids the problem with the cold-start Transformed Vocabulary approach. Although some subwords have a different meaning in different languages, hence they need to be retrained even in the warm-start scenario.

### 7.4.2 Comparison of Shared Vocabularies

We compare the two warm-start techniques proposed earlier with parent-specific and child-specific vocabularies to figure out which one leads to the best performance.

We compare the proposed methods using both pairs with both directions (i.e. two low-resource language pairs, namely Estonian–English and Basque–English). In this comparison, we use the parent model that has English on the same side. Thus, English →Czech parent is used for English→Estonian and Czech→English parent is used for Basque→English.

| | Parent score | | Child score | |
|---|---|---|---|---|
| Method | EN→CS | CS→EN | EN→Estonian | Basque→EN |
| Direct Transfer | 22.78 | 27.81 | 15.55 | 23.29 |
| Child-specific voc. | 21.05 | 24.93 | 15.73 | 22.92 |
| Balanced Vocabulary | 22.58 | **27.93** | **16.41** | **23.63** |
| Merged Vocabulary | **22.68** | ✖ | 16.05 | ✖ |
| Baseline | – | – | 8.70 | 19.09 |

**Table 7.7:** Results of various warm-start transfer learning approaches. The first two columns show the performance of the parent model, the third and forth column is the child model based on the corresponding parent in the same row. The baseline does not use transfer learning and uses language-pair-specific vocabulary. Scores are in BLEU and are comparable only within columns.

We compare four different approaches: parent-specific (direct transfer), child-specific, Merged Vocabulary, and Balanced Vocabulary. All setups use the same layout of transfer learning and training conditions. We trained four parent models for English →Czech and four parent models for Czech→English, each with a different vocabulary. The parent training took one million steps before the transfer learning of the child.

The setups for Basque→English are from Kocmi et al. (2018). We extended the evaluation with English→Estonian translation direction, for which we downsampled the original Estonian–English (see Kocmi, 2019, Table 2.4) corpus to only 100k sentence pairs. Hence, we artificially created a less resourceful language pair.

Merged Vocabulary approach generates a larger vocabulary because merging two vocabularies of equal size leads to a bigger size of the merged vocabulary. Moreover, larger vocabulary leads to less comparable results as improvements could be attributed to different vocabulary size. We want to evaluate all methods in the closest setting as possible. Thus, for Merged Vocabulary, we generate smaller parent and child-specific vocabularies in the first place in a way that the final size after merging and removing duplicates is approximately the same as vocabulary size of other methods, in our experiments 32k subword vocabulary. Consequently, we define the size of initial two vocabularies experimentally by iteratively decreasing their size and measuring the size of merged vocabulary until we obtained the size of merged vocabulary within a 1% tolerance of the vocabulary size (the 32k subwords).

We start by investigation of the child model performance (column Child score) in Table 7.7. All four transfer learning techniques outperform the baseline trained only on the child language pair, the gains of nearly 8 BLEU points for English→Estonian and 4 BLEU points improvement for Basque→English are significant.

**Observation 12:** *Both Balanced and Merged Vocabulary warm-start techniques significantly outperform the baseline.*

The missing result of Merged Vocabulary in Basque→English is because we have not conducted a comparison of this technique in Kocmi et al. (2018).

Both direct transfer, or child-specific vocabulary setups, performed worse than the Merged and Balanced Vocabulary. We assumed that the parent-specific vocabulary would perform the worst since it introduces segmentation problems into the child language pair, as discussed in Section 7.2.3. Interestingly, the child-specific vocabulary is not the best approach even though the vocabulary is specifically tailored to the child model. Furthermore, the parent score is significantly lower (1.53–1.73 BLEU) for the child-specific vocabulary when compared to other variants, which is due to a sub-optimal vocabulary. This suggests that the translation quality of the parent model plays an important role in the transfer model. This is a rather interesting result, as we usually disregard the performance of the parent in the transfer learning setup. We study the parent performance and behavior in Section 7.7.3.

**Observation 13:** *Training the parent using child-specific vocabulary is worse for the final performance of the child than using a mixed child-parent vocabulary.*

Lastly, we want to briefly discuss the ratio of parent vs. child subwords in the variants of the vocabulary. The parent-specific vocabulary (direct transfer) in general contains no child language pair subwords. On the other hand, child-specific vocabulary contains only child-specific subwords. In between are both Merged and Balanced Vocabulary that contain roughly half of the parent's subwords and half of the child's subwords. However, it could be the case that the best approach would be different, for example, including only 30% of parent vocabulary and 70% of child vocabulary. If that would be a case, it is most likely going to be specific for the parent and child language pairs and not being general for transfer learning.

In conclusion, both Merged and Balanced Vocabulary are better than baseline and the cold-start direct transfer. Furthermore, Balanced Vocabulary leads to better quality than Merged Vocabulary. However, we are not claiming that Balanced Vocabulary is strictly better than Merged Vocabulary. Lastly, the Merged Vocabulary technique is less general as it works only for the wordpiece segmentation (see Section 7.4.1).

### 7.4.3 Warm-Starts with Ten Languages

In order to prove the general applicability of the warm-start method, Kocmi (2019) evaluates it across various parents and children, comparing on 10 genetically related and unrelated languages as well as having shared English on the source or target side. In Kocmi (2019), we use languages and resources described in Tables 7.8 to 7.10. Following results are also published in Kocmi and Bojar (2018).

| Language | Lang. family | Lang. branch | Speakers | Script |
|---|---|---|---|---|
| English | Indo-European | Germanic | 1132M | Latin |
| German | Indo-European | Germanic | 132M | Latin |
| French | Indo-European | Romance | 280M | Latin |
| Spanish | Indo-European | Romance | 534M | Latin |
| Czech | Indo-European | Slavic | 11M | Latin |
| Russian | Indo-European | Slavic | 258M | Cyrillic |
| Slovak | Indo-European | Slavic | 5M | Latin |
| Gujarati | Indo-European | Indic | 61M | Brahmic |
| Odia | Indo-European | Indic | 34M | Brahmic |
| Arabic | Afro-Asiatic | Semitic | 274M | Arabic |
| Estonian | Uralic | Finnic | 1M | Latin |
| Finnish | Uralic | Finnic | 5M | Latin |
| Basque | Basque | Basque | 1M | Latin |

**Table 7.8:** Language family, branch, number of speakers, and the writing script according to Simons (2018). Reproduced from Kocmi (2019), Table 2.5.

As mentioned earlier, we are going to discuss only the Balanced Vocabulary approach ignoring the Merged Vocabulary and the child-specific vocabulary. Thus, whenever we talk about warm-start technique, we consequently mean the Balanced Vocabulary technique.

Table 7.11 summarizes our results for various combinations of a high-resource parent and a low-resource child language pairs. All comparisons are with the English on the same translation side for both parent and child. The baseline models are trained exclusively on the child or parent parallel corpus. We do not report parent score on parent testset.

The column with the child baseline is essential as it shows the impact of transfer learning. We see that for all language pairs, the transfer learning significantly outperforms the baseline. However, as we evaluate some genetically related languages, for example, Czech and Slovak, we also evaluate the performance of parent model only on the child's testset to show that without transfer learning, the performance is strictly worse. For the Czech and Slovak, the parent alone can roughly translate given sentences and obtain 6.51 BLEU score for direction into Slovak and 11.62 BLEU for Slovak→English. In contrast, Estonian and Finnish are not as related as Czech and Slovak. Thus, the parent model does not perform well on the child testset, obtaining only 2.44 BLEU for Estonian→English translation. This highlights that the improvement brought by our method can not be solely attributed to the relatedness of languages.

| Language pair | Sentence pairs | Words | |
|---|---|---|---|
| | | First language | Second language |
| Odia–EN | 27k | 604k | 706k |
| Gujarati–EN | 173k | 1.4M | 1.4M |
| Estonian–EN | 0.8M | 14M | 20M |
| Basque–EN | 0.9M | 5M | 7M |
| Finnish–EN | 2.8M | 44M | 64M |
| German–EN | 3.5M | 73M | 77M |
| Slovak–EN | 4.3M | 82M | 95M |
| Russian–EN | 12.6M | 297M | 321M |
| French–EN | 34.3M | 1044M | 912M |
| Czech–EN | 40.1M | 491M | 563M |
| Arabic–Russian | 10.2M | 243M | 252M |
| French–Russian | 10.0M | 295M | 238M |
| Spanish–French | 10.0M | 297M | 288M |
| Spanish–Russian | 10.0M | 300M | 235M |

**Table 7.9:** Datasets sizes overview. Reproduced from Kocmi (2019), Table 2.4.

Earlier works on NMT transfer learning (Dabre et al., 2017; Nguyen and Chiang, 2017) supposed linguistically related languages. We confirm their results also with our warm-start transfer learning on linguistically similar Finnish/Estonian and Czech-/Slovak languages, moreover, it was also confirmed by Kvapilíková et al. (2020) on German/Upper Sorbian. Furthermore, the improvements are not limited only to related languages as Estonian and Finnish. Unrelated language pairs like Czech/Estonian or Russian/Estonian work comparably well.

**Observation 14:** *Warm-start transfer learning improves performance even for unrelated languages.*

The most surprising is the comparison of English→Estonian performance across various parents. We see that Finnish, the genetically related language, improves the performance the least compared to other parents. We reach an improvement of 3.38 BLEU for English→Estonian when the parent model was English→Czech, compared to an improvement of 2.71 from English→Finnish parent. This two improvements are statistically significant and differ from the conclusion of Zoph et al. (2016), who concluded that the more related the languages are, the better transfer learning works. We see it as an indication that the size of the parent training set is more important than relatedness of the languages as the Czech has 40.1M sentences and the Finnish only 2.8M sentences.

The results with Russian parent for Estonian child (both directions) show that transliteration is not necessary for our approach as used in previous works (Nguyen and Chiang, 2017). Therefore, there is no vocabulary sharing between Russian Cyrillic and Estonian Latin (except numbers and punctuation, see Section 7.4.5 for further details), the improvement could be attributed to better coverage of English; an effect similar to domain adaptation.

| Language pair | Trainset | Devset | Testset |
|---|---|---|---|
| English–Basque | IWSLT 2018 | IWSLT dev 2018 | IWSLT 2018 |
| English–Estonian | Europarl, Rapid | WMT dev 2018 | WMT 2018 |
| English–Finnish | Europarl, Paracrawl, Rapid | WMT 2015 | WMT 2018 |
| English–German | Europarl, News commentary, Rapid | WMT 2017 | WMT 2018 |
| English–Odia | Parida et al. (2018) | Parida et al. (2018) | Parida et al. (2018) |
| English–Russian | News Commentary, Yandex, and UN Corpus | WMT 2012 | WMT 2018 |
| English–Slovak | Galušcáková and Bojar (2012) | WMT 2011 | WMT 2011 |
| English–French | Commoncrawl, Europarl, Giga FREN, News commentary, UN corpus | WMT 2013 | WMT dis. 2015 |
| English–Gujarati | Bible, Dictionary, Govincrawl, Software, Wiki texts, and Wiki titles | WMT dev 2019 | WMT 2019 |

**Table 7.10:** Corpora used for each language pair in training set, development set, and the test set. The names specify the corpora from WMT News Task data except of languages from various papers. Reproduced from Kocmi (2019), Table 2.6.

We show that our method also works with the shared English on the source side. Although it is an intuitive result, we have to point out that earlier works focused only on the scenario with shared-target language (Zoph et al., 2016; Nguyen and Chiang, 2017). Similarly to cold-start, it supports the idea that improvements are not due to the better decoder's language model.

### 7.4.4 Mixing Parent and Child

In the warm-start transfer learning scenario, we could train the parent model on a mixture of parent and child training data and then fine-tuning solely on child training data. We experiment with this setting to find out if the child performance is influenced by it.

In multilingual NMT, Johnson et al. (2017) proposed a special target language identification token “<2lang>” (see Section 3.5 and also Section 8.3.2). By adding this tag to each source sentence, they could train NMT model on a mix of training corpora. Another option is to mix the training corpora without any tag.

In the following experiment, we compare Balanced Vocabulary approach with three parent models trained on various training sets: a standard no-mixing parent-only, a mix of parent and child with added tag, a mix of parent and child *without* a tag. We use the Czech–English language pair as a parent and the Estonian–English as a child.

In Table 7.12, we see that mixing corpora is slightly better for shared-target scenario. However, neither of approaches is significantly better than the others. Hence we see no difference between approaches. Although, this experiment could be influenced by the size of training corpora of the parent and the child.

| Parent | Child | Balanced Vocabulary | Baselines: Only Child | Parent |
|---|---|---|---|---|
| EN→Finnish | EN→Estonian | **19.74** ‡ | 17.03 | 2.32 |
| EN→Russian | EN→Estonian | **20.09** ‡ | 17.03 | 0.57 |
| EN→Czech | EN→Estonian | **20.41** ‡ | 17.03 | 1.42 |
| Finnish→EN | Estonian→EN | **24.18** ‡ | 21.74 | 2.44 |
| Russian→EN | Estonian→EN | **23.54** ‡ | 21.74 | 0.80 |
| EN→Czech | EN→Slovak | **17.75** ‡ | 16.13 | 6.51 |
| Czech→EN | Slovak→EN | **22.42** ‡ | 19.19 | 11.62 |

**Table 7.11:** Transfer learning with English reused either in source (encoder) or target (decoder). The baselines correspond to training on one corpus only. BLEU scores are always reported for the child language pair. The scores are comparable within lines or whenever the child language pair is the same. The ‡ represents significantly better results.

| Child model | No-Mixing | Mix with tag | Mix without tag |
|---|---|---|---|
| English→Estonian | **20.1** | **20.1** | 19.9 |
| Estonian→English | 23.4 | **23.7** | 23.6 |

**Table 7.12:** Comparison of various approaches for incorporating the child data into the parent training set. All scores are in BLEU, and neither model is significantly better than any other.

The previous experiment showed that the performance is not largely influenced whether we mix the child data into the parent training data or not. However, in that setting, the parent can perform translation of the child language pair, because it was trained together with the child training set. Thus, we evaluate four parent models, trained on the mix of corpora from the previous experiment, on the child testset.

The performance of individual parent models is reported in Table 7.13. All of them perform significantly worse than after transfer learning of the child model. However, for the scenario, where the target language is shared between Czech and Estonian pair, or when the tag marks the target desired language, it is interesting that the models' performances are only slightly lower than for transfer learning than we would expect.

We note that mixing the Estonian→English child data without language tag into the parent Czech→English model is exactly the "bi-pair" setup discussed in multilingual source below in Section 8.2 except Estonian and Czech are not related.

| Parent model | Mix with tag | Mix without tag |
|---|---|---|
| English→(Czech+Estonian) | **17.7 (20.1)** | 2.4 (19.9) |
| (Czech+Estonian)→English | **22.0 (23.7)** | 21.8 (23.6) |

**Table 7.13:** Performance in BLEU of the parent model evaluated on the child Estonian–English test set. In brackets are the results evaluated on the respective child models from Table 7.12 for comparison.

| Estonian | English | Russian | % Subwords |
|---|---|---|---|
| ✓ | - | - | 29.93 |
| - | ✓ | - | 20.69 |
| - | - | ✓ | 29.03 |
| ✓ | ✓ | - | 10.06 |
| - | ✓ | ✓ | 1.39 |
| ✓ | - | ✓ | 0.00 |
| ✓ | ✓ | ✓ | 8.89 |
| Reused parent | | | 41.03 |

**Table 7.14:** Breakdown of subword vocabulary of experiments involving Russian–English parent and Estonian–English child.

### 7.4.5 Analysis of Balanced Vocabulary

Our (Kocmi, 2019) Balanced Vocabulary method relies on the vocabulary estimated jointly from the child and parent model. In the Transformer, the vocabulary is typically shared by both the encoder and the decoder. We analyzed the vocabulary in our cold-start scenario in Section 7.2.5 where we found out that around 40% of parent vocabulary is unused after the transfer.

Balanced Vocabulary is prepared in the following way. We take an equal part of parent and child corpus and generate a wordpiece vocabulary that is used both by the parent as well as the child model. With a large overlap, we could expect a lot of "information reuse" between the parent and the child.

We take the vocabulary of subword units as created for Russian–English parent and Estonian–English child experiments. This vocabulary contains 28.2k subwords in total. We then process the training corpus for each of the languages with this shared vocabulary, ignore all subwords that appear less than 10 times in each of the languages (these subwords will have little to no impact on the result of training) and break down the total of 28.2k subwords into overlapping classes based on the languages where the particular subword was observed, see Table 7.14.

| Languages | In All | Reused Parent | Unused by Child |
|---|---|---|---|
| Finnish–EN | 19.5 % | 49.4 % | 26.2 % |
| Russian–EN | 8.9 % | 41.0 % | 29.0 % |
| Czech–EN | 20.3 % | 49.2 % | 21.2 % |
| Arabic–Russian | 4.6 % | 6.2 % | 56.3 % |
| Spanish–French | 18.4 % | 34.1 % | 28.7 % |
| Spanish–Russian | 6.0 % | 21.4 % | 45.8 % |
| French–Russian | 6.3 % | 23.1 % | 44.3 % |

**Table 7.15:** Summary of vocabulary overlaps for the various language sets. The first column specifies what is the parent language pair. The child language pair is Estonian–English for all rows. All figures represent percentage of the vocabulary.

We see that the vocabulary is reasonably balanced, with each language having 20–30% of subwords unique (see the first three rows of Table 7.14). English and Estonian share 10% subwords not seen in Russian while Russian shares only 1.39% and 0% of subwords with each of the other languages, possibly due to the Cyrillic script. Overall, 8.89% of subwords are used in all three languages.

A particularly interesting subset is the one where parent subwords are used by the child model. In other words, subwords appearing anywhere in English and also tokens common to Estonian and Russian. For this set of languages, this amounts to 20.69+10.06+1.39+0.0+8.89 = 41.03%. We list this number on a separate row in Table 7.14, as "Reused parent". These subwords get their embeddings trained better thanks to the parent model. However, the vocabulary also contains 29.04% subwords used only by the parent and unused by the child.

Table 7.15 summarizes this analysis for several language sets used in the warm-start experiments, listing what portion is shared by all the languages (column "In All"), what portion of subwords benefits from the parent training (column "Reused from Parent") and what portion of vocabulary is unused by the child (column "Unused by Child).

We see a similar picture across the board; roughly 30% of subwords are unused by the child model. And roughly 50% of subwords are unused whenever both parent languages are distinct from the child. We remind that in the direct transfer the number was around 40%. We already discussed this problem as these subwords only slow down the training and inference as they are useless to the child.

**Observation 15:** *When parent and child share one language and the vocabulary is created jointly for both of them (Balanced Vocabulary), about 30% of the vocabulary won't be used by the child.*

| Child language pair | Direct Transfer | Transformed Voc. | Balanced Voc. |
|---|---|---|---|
| EN→Estonian | **20.75** | 20.27 | 20.62 |
| EN→Russian | 27.19 | 28.65 | **29.03** ‡ |
| Estonian→EN | 24.36 | 24.64 | **26.00** ‡ |
| Russian→EN | 30.49 | **31.38** | 31.15 |

**Table 7.16:** The scores (BLEU) for cold-start methods (direct transfer and Transformed Vocabulary) with the warm-start method of Balanced Vocabulary.

We list vocabularies used in our main results in Section 7.4.3 as well as language pairs not containing any shared language between parent and child (English in our case) with the child as we report in Section 7.6.4.

The Arabic-Russian-Estonian-English stands out with the very low number of subwords (6.2%) available already in the parent, mainly due to the scripts of parent language not using Latin. The parent Arabic-Russian thus offered very little word knowledge to the child, and yet it leads to a performance gain (21.74 vs. 22.23 BLEU, see Section 7.6.4).

Our observations indicate that the key factor is the size of the parent corpus rather than vocabulary overlaps. However, the reasons for the gains are yet to be explained in detail.

## 7.5 Warm vs Cold-Start

In Kocmi and Bojar (2020), we study two approaches for transfer learning that differ in how the parent model is treated, specifically by allowing modification to the parent vocabulary before or after the training. In this section, we compare both cold-start approaches: the cold-start Direct Transform and Transformed Vocabulary with the warm-start approach of Balanced Vocabulary.

We train four parent models for warm-start approach differing in the vocabulary: two English→Czech and two Czech→English on the same parent training data. For the cold-start approaches, we used the models trained by Popel (2018) similarly as in cold-start experiments (see Section 7.2).

The results are tabulated in Table 7.16. We see that the warm-start method reaches in most cases significantly better performance in both directions. The only exception is the Russian→English and English→Estonian, where neither warm-start Balanced Vocabulary nor the Transformed Vocabulary is significantly better than the other.

The cold-start transfer learning improves the performance of high-resource language pairs (see Observation 9), and the warm-start improve the performance over the cold-start transfer learning. Hence, warm-start should work for high-resource

| Child language pair | Direct Transfer | Transformed Voc. | Balanced Voc. |
|---|---|---|---|
| EN→Estonian | 75k | 75k | 735k |
| EN→Russian | 630k | 450k | 1510k |
| Estonian→EN | 30k | 60k | 700k |
| Russian→EN | 420k | 700k | 1465k |

**Table 7.17:** Comparison of the number of steps needed for cold-start and warm-start methods to converge.

language pairs. This is confirmed by the Russian–English language pair in Table 7.16 and Slovak–English in Table 7.11 and we formulate it as another observation here:

**Observation 16:** *Warm-start transfer learning improves performance of both low-resource and high-resource language pairs.*

The better performance of the warm-start approach is understandable since the parent model is already trained with vocabulary prepared to accommodate a given child language pair. It does not have the high segmentation problem as the direct transfer has, and it does not have to retrain randomly assigned embeddings as in Transformed Vocabulary.

However, the warm-start approach has one disadvantage over the cold-start: we can not reuse *any* parent model, and we have to train the parent model for each child language pair separately. With that in mind, we investigate the number of steps needed to reach the performance as presented in Table 7.16.

Table 7.17 shows the total number of training steps. The cold-start scenarios show only steps needed for the child model convergence since we did not train the parent model by ourselves as we are using the model by Popel (2018). In contrast, the steps for Balanced Vocabulary show the total number of steps the parent and the child were training altogether.

By observing the results, we see that due to the parent training, the warm-start scenario takes more steps to train. However, the total training time of the parent can vary broadly. Furthermore, in Section 7.9.6 we show that even a parent model that did not fully converge is a good parent for transfer learning.

In conclusion, the cold-start method has the advantage of not requiring to train the parent model, it does not need any modification to the training workflow (in direct transfer) or only a trivial modification of vocabulary (in Transformed Vocabulary) and reaches a better performance than training a baseline model. This makes it the candidate to most of the current training workflows, where researchers could start using any model as a parameter initialization instead of random initialization as it is the current standard.

On the other hand, whenever the time is not a relevant criterion, but we are focused on the performance of the model, the warm-start scenario is the most recommended approach.

## 7.6 When Transfer is Negative?

In generic machine learning, transfer learning is also known for its downsides (Pan and Yang, 2010; Weiss et al., 2016). When transferring knowledge from a less related task, it may hurt the final performance on the child task in comparison to the performance obtained without transfer learning only with the use of a child model. This harmful effect is called **negative transfer**.

The main reason behind the negative transfer is often the domain mismatch between the parent and child tasks or even an unrelated parent domain (Pan et al., 2010; Ge et al., 2014), which prevents the model from the utilization of the parent model during transfer learning.

Wang et al. (2019b) proposed a formal definition of the negative transfer and evaluated it on several transfer learning approaches. They evaluated the following three critical factors influencing the negative transfer:

1. Divergence between the joint distributions of both tasks is hurting transfer learning.
2. Effectiveness of transfer learning depends on the size of child data.
3. Transfer learning should be evaluated with the same setting of the neural network to avoid adding a risk of different setups.

As for the first factor, the ideal transfer learning should figure out and take advantage of only the similar parts of tasks, however, in the real-life scenario it often takes into account also the misleading information learned from the parent task. The second factor elaborates that the less training data is available in the child domain, the more fragments are preserved from the parent task, which decreases performance on the child task. On the other hand, when we have plentiful of child data, a better baseline can be trained, which reaches a better performance than transfer learning. Thus, negative transfer is relatively more likely to occur. The last factor is to avoid misjudgment by comparing the performance of transfer learning with a baseline using different parameters. For example, fine-tuning hyper-parameters separately for the transferred model and baseline will likely lead to different results due to the hyper-parameter setting.

Transfer learning in the field of NMT emerged recently (Zoph et al., 2016). Thus, there is a lack of research on the negative transfer in this field. Zoph et al. (2016) have not discovered any problematic behavior of transfer learning. It could be due to the design of the experiment that avoids the negative transfer altogether. For example, initial works on transfer learning in NMT examine only genetically related language pairs (Nguyen and Chiang, 2017; Neubig and Hu, 2018). Another possible explanation is that the neural networks are robust enough that they can re-train any transferred parameters hence avoiding the negative transfer at all.

In this section, we try to shed some light on the negative transfer in NMT by evaluating several experiments and identifying the possible downsides of transfer learning in NMT. In Section 7.6.1, we investigate if the parent target language leaks to the outputs of the child. Then, in Section 7.6.2, we study the condition of having an extremely low-resource child and whether it hurts transfer learning. In Section 7.6.3, we study the reverse effect of having a parent with less parallel sentences than the child. Wang et al. (2019b) mentioned that another factor negatively influencing transfer learning is a divergence between parent and child distributions. Thus, we investigate the scenario with no-shared language in Section 7.6.4.

### 7.6.1 Traces of Parent

Wang et al. (2019b) mention as an effect of negative transfer that parent fragments appear in the child outputs whenever the child task has a low amount of data. This is mainly because the child has not entirely forgotten the parent task. In this section, we investigate traces of the parent language pair in child translation, such as text fragments.

During transfer learning, the neural network is not notified about the change in the language pair. This means that the NN has to forget the parent task during the training on child parallel corpus. This can result in fragments of parent target language appearing in the child model output.

In order to test if there are any traces of the parent language pair in the output of the child model, LanideNN (Kocmi and Bojar, 2017b) automatically identifies the language of transfer learning outputs, and we measure how often does the parent language pair appear in the child output.

We used LanideNN because it is able to recognize language switching within one sentence instead of labeling the whole sentence by one label; we can get distribution over each character. We calculate the score by labeling each character in the testset with the language label and then calculating the ratio of labels for each language in the testset.

We evaluate the model every 250 steps, which is roughly every three minutes of training. As the parent model, we use English→Czech and the child model is English →Estonian. These language pairs have different target language from different language family, which should help when automatically recognizing the language. For this evaluation, we use the English→Estonian development set.

In Figure 7.6, we see that the NMT model quickly stops generating Czech sentences, after just 3k steps the model generates less than 1% of Czech data. The training time for 3k steps took only 43 minutes. We remind the reader that the standard time of training even extremely low-resource language pairs is at least 50k training

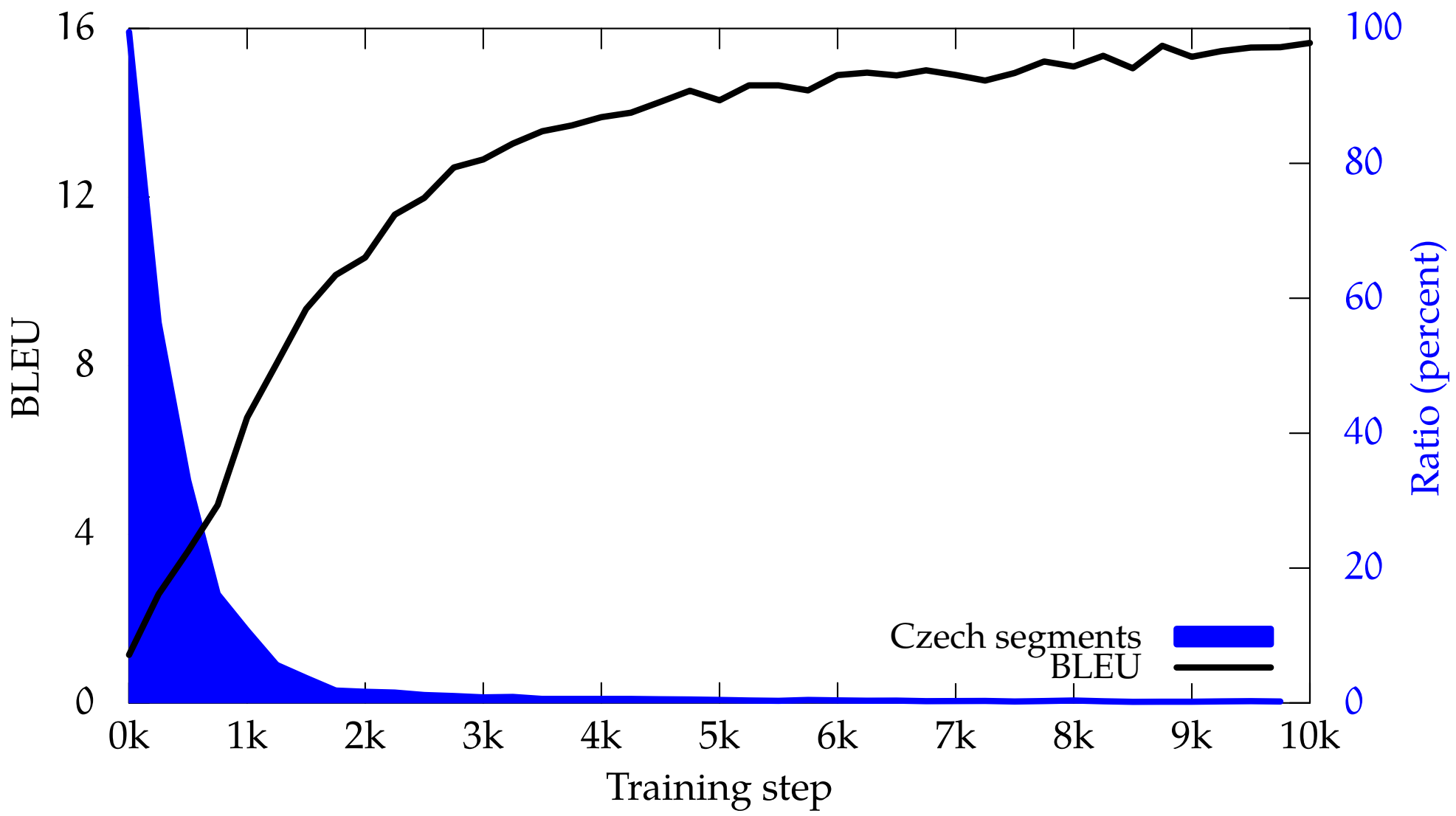

**Figure 7.6:** Traces of parent-language segments during child training. The graph represent behavior of child model during first 10k training steps. The blue area represents the ratio of Czech segments in the output of child model immediately after the transfer learning start. The black curve illustrates the BLEU score on the English→Estonian development set.

steps. Therefore it takes only a fraction of time for the NN to forget the parent target language.

**Observation 17:** *During the training of a child model, the NN almost instantly stops producing sentences in the parent target language and moves to the child target language.*

We need to mention that the results are based on an automatic measure and that LanideNN's error rate needs to be taken into account (in a multilingual setting, the error rate is less than 4%, see Kocmi and Bojar, 2017b). We note that this automatic language detection can not be reliably used for fine-grained evaluation to investigate if the child occasionally generates the parent target language.

In order to evaluate how often the child model produces parent (Czech) words, we used the final child model to translate 100k English sentences randomly selected from the parent training corpus. We chose the parent training corpus as these sentences could be memorized by NN from the parent model training and it is thus most likely that the training input sentences could trigger the parent behavior in the child model. The final child model was trained for 75k steps.

| Appearances | Czech | English Gloss |
|---|---|---|
| 49 | kámo | buddy |
| 31 | Článek | Article |
| 20 | Podívej | Look |
| 17 | jasný | clear/ok |
| 11 | Odpověď | Answer |
| 8 | strýc | uncle |
| 7 | Poznámky | Notes |

**Table 7.18:** Relic words from parent language in the child output. The English gloss is our (manual) translation of the given Czech word.

We evaluate the translated sentences both automatically as well as manually. The automatic evaluation is the same as in the previous analysis and identified 54 Czech sentences. We manually checked these sentences and found out that a few of them are Czech postal addresses or other named entities, the rest are Estonian sentences with Czech words inside them. The longest Czech sequence without named entities is "u všech čert" (an incomplete idiom "all the hell"). However, we noticed that the Czech words are often already present in the source sentence.[3]

In order to analyze the size of Czech words produced by the child model, we listed all sentences that use Czech characters not contained in Estonian, e.g. characters with diacritics. Then we removed sentences where the source also contained Czech-only characters. This way, we got 624 out of 100k sentences, which often contained only one word that we manually identified as Czech. Altogether these sentences contained 645 words with Czech-only characters (the evaluated dataset contained 1.4M words). We list few of the most frequently appearing in Table 7.18.

Despite that we evaluated the child model with parent training data, i.e. the corpus that the model could have memorized, we found only a few parent relics. Therefore, we conclude that transfer learning is not negatively affected by the parent model.

**Observation 18:** *Relic words (i.e. words from the parent target language) are very rarely produced by the child model.*

The rapid change in behavior, when only 3000 steps are enough to forget the parent target language, is one of the results of "catastrophic forgetting", a nature of a network to quickly forget or re-train previously learned features. This phenomenon has been widely studied (Kirkpatrick et al., 2017; Kemker et al., 2018) as the researchers develop methods to overcome this issue. Furthermore, we have tackled problems connected with catastrophic forgetting in Kocmi and Bojar (2017a) when experimenting with curriculum learning (Bengio et al., 2009).

[3] The Czech–English training corpus CzEng (Bojar et al., 2016) is noisy, as discussed in Section 4.2.3.

Despite catastrophic forgetting being a problem in machine learning in general, in the scenario of transfer learning, we believe it helps to avoid negative transfer from the parent model by forgetting it. However, we need to keep in mind that in future, when algorithms become more robust in terms of catastrophic forgetting, the negative transfer could emerge as a problem for transfer learning because we saw some words attributed to the parent language pair.

### 7.6.2 Extremely Low-Resourced

We showed that NMT systems quickly forget the parent language in the low-resource scenario. Now, we evaluate an *extremely* low-resource language pair to find out whether our approach helps in the extremely low-resource scenario (see Section 4.2.2) or whether insufficient data lead to negative transfer as Wang et al. (2019b) described. The results in this section are published in our paper Kocmi and Bojar (2018).

We simulate extremely low-resource settings by downscaling the data for the child model but maintaining the same parent model. It is a common knowledge that gains from transfer learning are more pronounced for child models with smaller training data. We use the English→Finnish as a parent model for English→Estonian. We mention that shared-source is more difficult transfer scenario as the model can not benefit from the parent English language model because the target language changes from Finnish to Estonian (more analysis in Section 7.7).

The results of downscaling the child training corpus are shown in Figure 7.7. We see that our approach applies even to extremely low-resource language pairs with as few as 10k sentence pairs. We see this behavior on the 10k training corpus, where the baseline reaches 1.95 BLEU. This behavior is in accordance with observations done by Koehn and Knowles (2017). For such a small amount of training data, the NMT baseline can not be properly trained. With transfer learning, NMT suddenly becomes able to train the model and reaches 12.46 BLEU.

**Observation 19:** *Transfer learning helps NMT to train models for extremely low-resource language pairs that can not be properly trained on their own.*

Sennrich and Zhang (2019) revisited the problem of extremely low-resource language pairs and showed that it could be tackled with various tricks. Furthermore, transfer learning could be used as another way of improving the extremely low-resource language pairs hand in hand with other techniques mentioned by Sennrich and Zhang (2019).

As Wang et al. (2019b) summarized, transfer learning can lower the performance of the child task whenever the amount of child training data is too low. We showed that this is not the case in NMT because transfer learning can help training the model even when the baseline can not be trained in the first place.

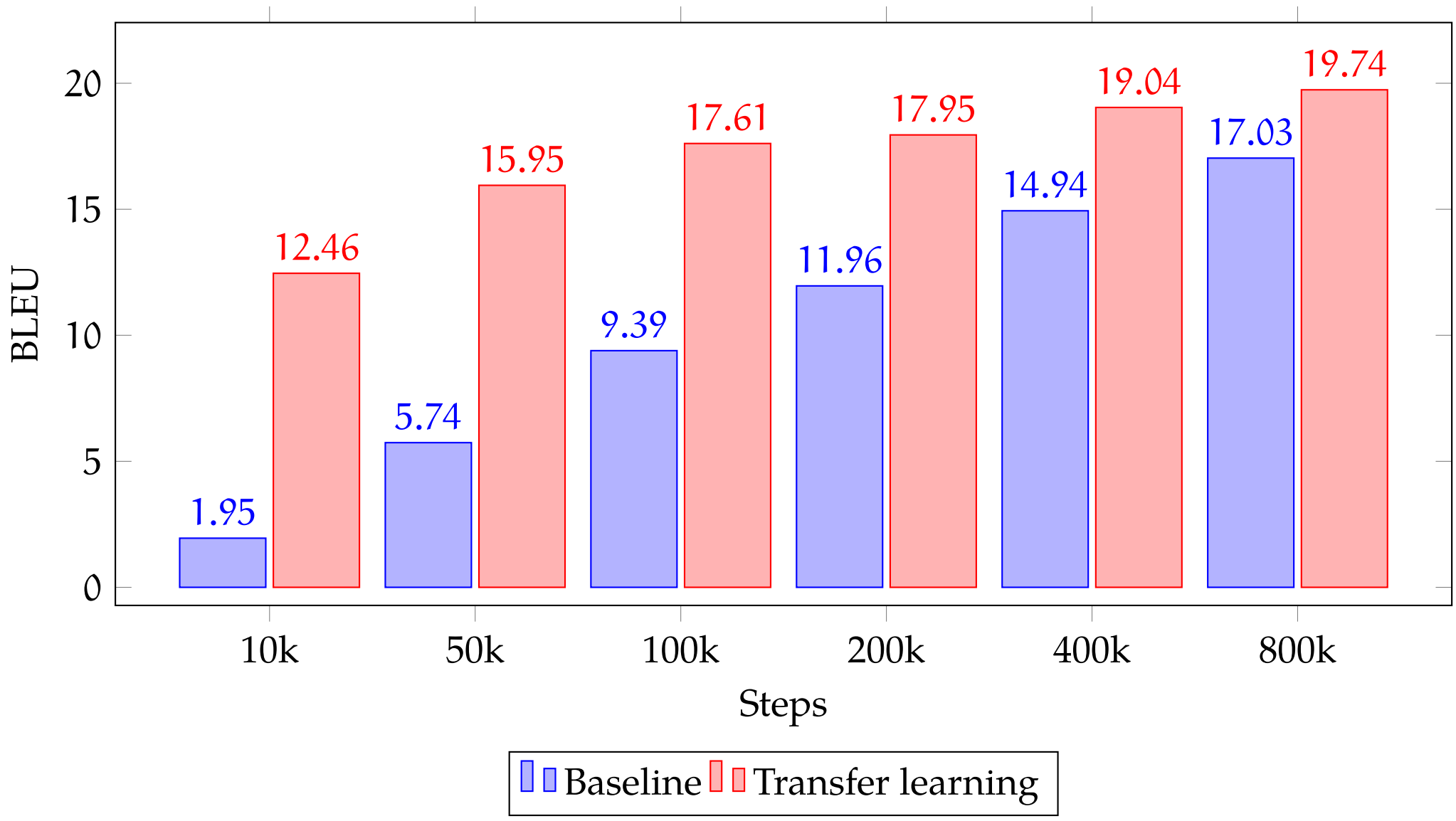

**Figure 7.7:** Diminishing returns of transfer learning. We plot the maximal score reached by English→Estonian child for decreasing sizes of child training data, trained off an English→ Finnish parent. The baselines use only the reduced Estonian–English data.

### 7.6.3 Low-Resource Parent

We showed that transfer learning is not restricted to the low-resource scenarios and improves the performance even when the child is a high-resource language pair (see Observation 9 and Observation 16). However, it is in the scenario, where both the parent and child model are high-resource.

Wang et al. (2019b) summarized that transfer learning is negatively influenced by the parent model whenever the parent has a low number of training examples. In this section, we examine this condition in the area of NMT, and we investigate the scenario where the parent has a lower amount of parallel sentences than the child model.

Zoph et al. (2016) conclude that the relatedness of languages is the main factor influencing the success of transfer learning, which we already show it is not a necessary condition in Table 7.11. However, we use genetically related languages in this experiment, because the secondary goal is to test what plays a bigger role in transfer learning – the relatedness of languages, or the size of parent data?

Results from Kocmi and Bojar (2018) are presented in Table 7.19. We see that low-resource parents do not generally improve the performance of sufficiently resourced children. The only exception is the child English→Finnish, where the child has only 3.5 times more parallel sentences than the English→Estonian parent.

| Parent | Size | Child | Size | Transfer | Baseline |
|---|---|---|---|---|---|
| EN→Estonian | 0.8M | EN→Finnish | 2.8M | **20.07** ‡ | 19.50 |
| Estonian→EN | 0.8M | Finnish→EN | 2.8M | 23.95 | **24.40** |
| EN→Slovak | 4.3M | EN→Czech | 40.1M | 22.99 | **23.48** ‡ |
| Slovak→EN | 4.3M | Czech→EN | 40.1M | 28.20 | **29.61** ‡ |

**Table 7.19:** Experiments with low-resource parent. The column "Transfer" is our warm-start method, baselines correspond to training on child corpus only. We show the sizes of corpora in millions sentences. The ‡ represents significantly better results.

**Observation 20:** *Transfer learning harms the child performance whenever the parent has substantially less training data than the child.*

Whenever the child has notably more training data, e.g. ten times more for Czech–English it even (significantly) decreases the child's performance compared to the baseline. Therefore we conclude that transfer learning is negatively influenced in scenarios where the parent has substantially less training data. We suppose it could be due to the initial warm-up steps when the network changes rapidly, thus low-resource language can skew it. However, more analysis is needed to study this behavior properly.

Furthermore, we evaluated genetically related languages where the relatedness could help improve the model. For example the Czech and Slovak are related to such extent that evaluating English→Czech system on English→Slovak testset output leads to 6.51 BLEU (see Section 7.4.3). However, the relatedness did seem not to play any role in our experiments, and transfer learning led to worse performance than training on child parallel corpus only.

**Observation 21:** *For a high-resource child, the linguistic relatedness of parent and child language pairs is less important than the size of the parent training corpus.*

### 7.6.4 No Shared Language

One of the main factors of negative transfer is the divergence in distributions between parent and child training data (Wang et al., 2019b). In NMT, one would assume that the languages in question are the key element affecting task similarity.

In Section 7.4.3, we showed that the relatedness of the languages is not the most critical for transfer learning. Moreover, the related English→Finnish parent performed worse even when compared to a parent that uses a different writing script, in our case, English→Russian with Cyrillic. Thus, we have not detected any negative transfer when evaluated on less genetically related languages.

| Parent | BLEU | nPER | nTER | nCDER | nWER | chrF3 | nCharacTER |
|---|---|---|---|---|---|---|---|
| No transfer learning | 21.74 | 54.33 | 35.66 | 41.28 | 32.70 | 49.87 | 37.70 |
| Arabic→Russian | 22.23 | 55.05 | 36.66 | 42.23 | 33.59 | 50.86 | **40.11** |
| Spanish→French | 22.24 ‡ | **55.32** | 36.58 | 42.05 | 33.69 | 50.88 | 39.59 |
| Spanish→Russian | **22.52** ‡ | 55.26 | **36.85** | **42.53** | **33.79** | **51.28** | 39.92 |
| French→Russian | 22.40 ‡ | 54.99 | 36.50 | 42.06 | 33.39 | 50.93 | 39.60 |

**Table 7.20:** No shared language scenario of transfer learning when varying the parent language pair. The child model is Estonian→English. Each row represents various metrics for measuring MT performance, where higher number is better for all metrics. The significance ‡ is computed pairwise relative to the baseline "No transfer learning".

However, our experiments always contained a language shared between the parent and child, e.g. English, which could work as a connecting bridge during transfer learning, thus preventing the negative effects. In order to test the negative transfer in NMT, we experiment with a no-shared language scenario.

We examine the performance of Estonian→English child trained on top of parents using unrelated languages, specifically Arabic→Russian, Spanish→French, Spanish →Russian, and French→Russian. The parents are trained with the UN corpus (Ziemski et al., 2016), which has 10M multi-parallel sentences across six languages.

The results from Kocmi and Bojar (2018) are shown in Table 7.20. We see mostly significant gains from transfer learning in all cases. The only non-significant gain is from Arabic→Russian, which does not share the script with the child's Latin at all, only sharing of punctuation and numbers is possible across all the tested scripts.

**Observation 22:** *Transfer learning improves the performance even in the case when there is no shared language between the parent and child pairs.*

There is no loss of performance in comparison to the baseline. This can be seen either as the evidence that transfer learning is not negatively affected by the difference in data distributions between parent and child, or that the mere distributional properties of all (tested) languages are sufficiently similar to be useful for transfer learning in NMT.

Surprisingly, the Spanish→Russian (with a target languages that uses the Cyrillic script) reached a better performance than the Spanish→French, a target language that is genetically closest to the Estonian→English from all four investigated language pairs. However, neither of these two systems is significantly better than the other. Furthermore, the gains are quite similar (+0.49 up to +0.78 BLEU), which supports the assumption that the major factor influencing transfer learning is the size of the parent (here, all parents have 10M sentence pairs). We are going to discuss this aspect in Section 7.8.1. This result can also be explained with a similar domain of parent training set. In comparison, the Czech→English parent, which has 40.1M sentences from a broader range of domains and has a shared language (English), improved the performance of Estonian→English by 3.38 BLEU.

**Observation 23:** *With a fixed parent data size, the exact parent language pair does not seem to affect the performance of the child much.*

In the no-shared language scenario, the gains can not be attributed to the language model or model parts such as shared English word embeddings. The subword vocabulary overlap is mostly due to short subwords or numbers and punctuation.

In Section 5.1.2, we discuss issues with the BLEU metrics, e.g. ignoring the importance of various $n$-grams or a high influence of tokenization. For this reason, we computed the scores for several other automatic methods. We see that in all metrics, transferred models perform better than the baseline.

The experiments presented in this section indicate that the parent simply works as a better model weight initialization in comparison to the random initialization. We investigate it more in Section 7.9.

## 7.7 Position of Shared Language Matters?

We noticed that transfer learning has different behavior for the shared-source and shared-target language scenarios. For example, English→Russian parent improved the Estonian child more than English→Finnish (20.41 vs. 19.74 BLEU), however in the opposite direction Russian→English worked as a worse parent than Finnish→English (23.54 vs. 24.18 BLEU), as we showed in Section 7.4.3.

In this section, we investigate the influence of the position of the shared language on transfer learning. Moreover, we discuss which of those tasks is harder for the NMT transfer learning.

### 7.7.1 Shared Language Position Affects Convergence Speed

We start with an investigation of the training time needed for each direction to converge. We recall the results from Section 7.5 in Table 7.21. We also added the Gujarati–English pair from our paper Kocmi and Bojar (2019). We subtracted the number of parent steps needed for the convergence and showed the results of the child model training in Table 7.21.[4]

In Table 7.21 we can see that the shared-target language scenario converges faster for both the low-resource Estonian and the high-resource Russian. In the case of Gujarati–English, the convergence is only slightly faster for the shared-target.

[4] Training steps in Section 7.5 are the sum of parent plus child training steps, but we show only the child number of steps in Table 7.21.

| Child language | Size | Shared-source | Shared-target |
|---|---|---|---|
| Estonian | 0.8M | 75k | 25k |
| Gujarati | 0.2M | 195k | 180k |
| Russian | 12.6M | 935k | 790k |

**Table 7.21:** Number of steps needed for a model to converge. The size shows the number of sentences in the corpora of each language. For both child the second language is English. The results are from Section 7.5 with subtracted time of parent model training.

**Observation 24:** *Transfer learning with a shared target language converges in fewer steps than with a shared source language.*

We need to mention that the total number of training steps does not always reflect the convergence speed because the performance usually fluctuates and thus the model can converge after a different number of steps, which depends on randomness in training. Moreover, the training time does not explicitly mean if the task is easier; there are many other factors like shared language, the noisiness of training data, and other factors.

Therefore we investigate the learning curves and look for a distinctive behavior between these tasks.

### 7.7.2 Shared Language Position Affects Slope of Learning Curve

In Figure 7.2, we described three impacts of transfer learning on the learning curve, namely higher start, higher performance, and a steeper slope. The effect of steeper slope suggests that the model trains faster, therefore we compare learning curves of shared source and shared target scenarios in order to find out which one learns faster.

We report the learning curve's Y-axis in BLEU, but any other metric could be used. The BLEU score has a disadvantage that it can not be compared across various languages or even testsets. Therefore, in order to study the slope of the learning curve, we need to scale it. We scale each learning curves by multiplicating the performance (BLEU) by a fixed constant. The constant is selected in order to align the best-reached performance of both translation directions.

Figure 7.8 presents the learning curves of three language pairs evaluated in the previous section on their respective development sets. We investigate only the first 30k training steps, where the difference in slopes is the most visible.

When comparing the learning curves of shared-source and shared-target scenarios, we see that for all three language pairs, the shared target has a higher slope than the shared source.

**Observation 25:** *Transfer learning with a shared target language has a higher slope of the learning curve.*

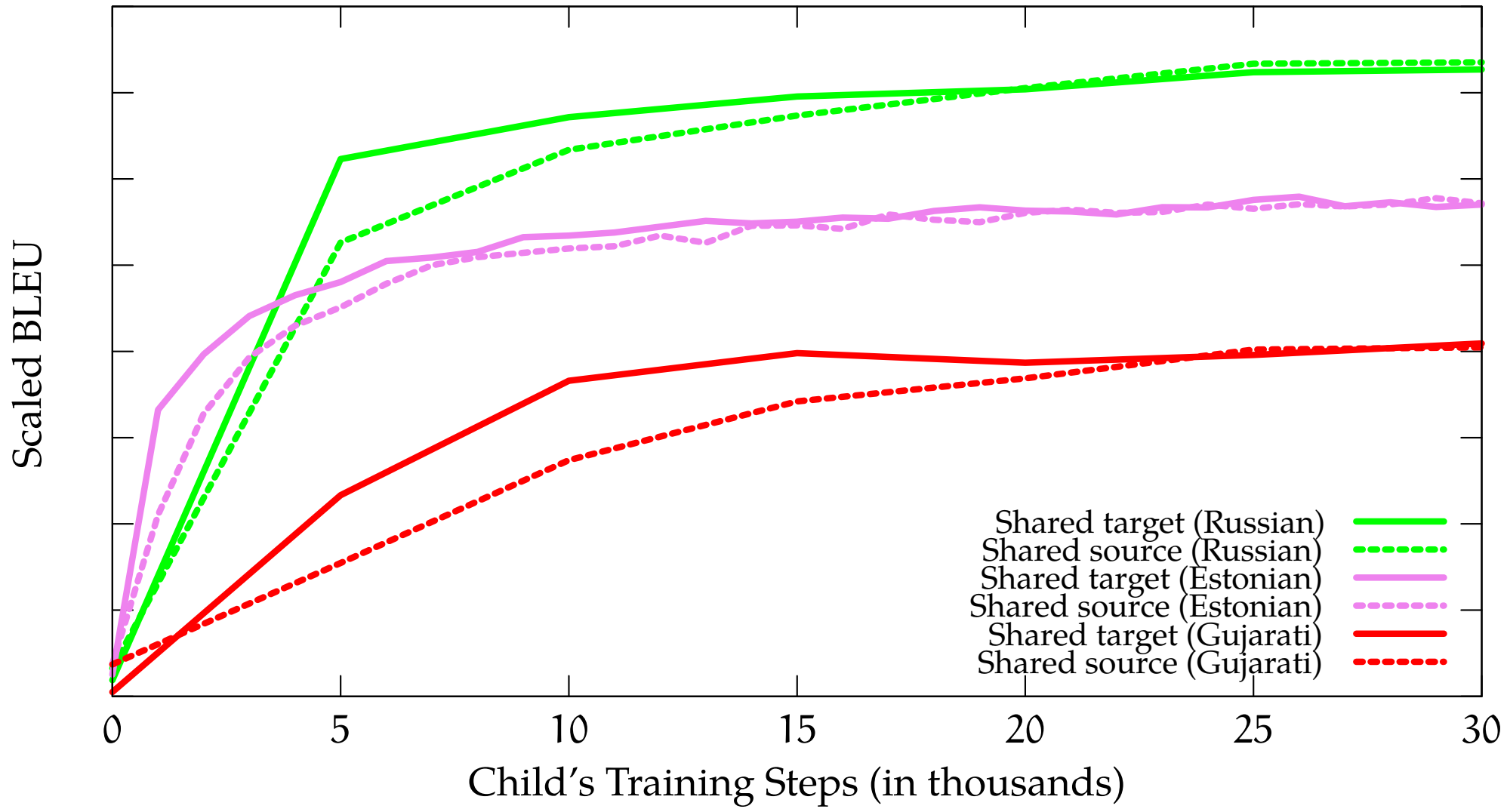

**Figure 7.8:** Speed of child training with shared source vs. shared target. Learning curves for various language pairs in both directions. The Y-axis has been scaled for each learning curve by a constant in order to match their final performance. The bracket specifies the child's second language that is paired with English. The convergence is seen only on Gujarati pair as other languages converged later than within first 30k steps.

This observation suggests that shared-target, i.e. having shared language (e.g. English) on the target side, is easier for transferring knowledge from parent to child. In this scenario, NN reaches higher performance in a shorter time compared to the shared-source scenario. This behavior is not surprising. From the neural network's point of view, it is learning to predict the shared language through the whole training process. Therefore it can utilize the language model from a parent with only learning different encoder's part of the model. We further analyze the behavior by freezing various parts of a model in Section 7.9.

### 7.7.3 Parent Performance Drop

In transfer learning, we do not pay attention to the final parent's performance as is customary in multi-task learning. In Section 7.6.1, we showed that for translation direction with the shared source, the child model quickly forgets the parent target language.

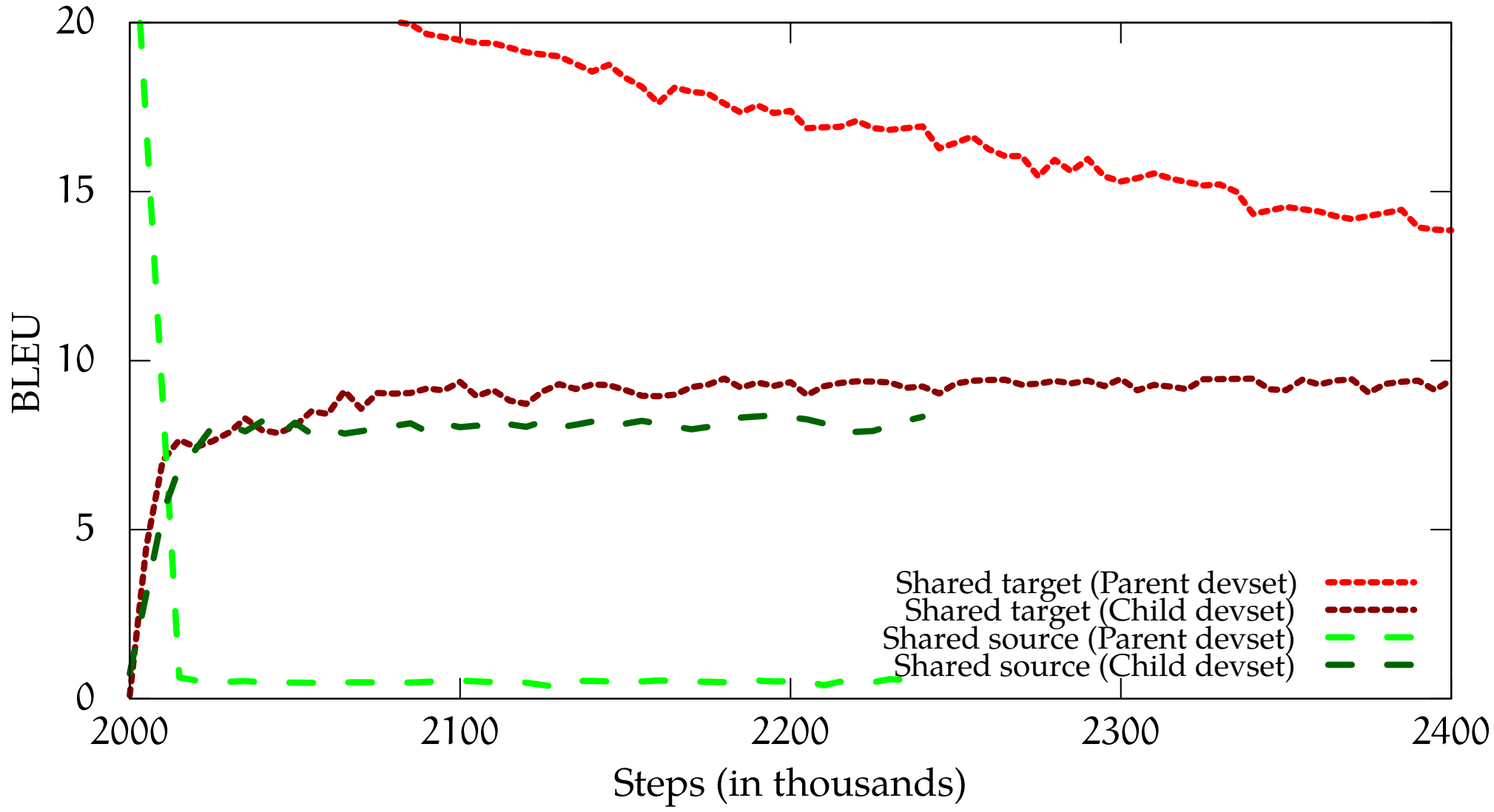

**Figure 7.9:** Performance of child model on the parent development set. Both childs are Gujarati–English. Learning curves are evaluated on a development set of analogous language pairs.

In this section, we investigate how the parent translation deteriorates during the child's training phase in both directions, and if the shared-source and shared-target scenarios behave differently.

We evaluate two scenarios, the shared-source and shared-target under our English →Gujarati and Gujarati→English child models transferred from Czech–English parent (Kocmi and Bojar, 2019).

We evaluate both models each by the corresponding child's and parent's development set. The scores between language pairs are not comparable as they are for different languages.

Figure 7.9 presents the results of our experiment. The figure starts at a global step of 2M, the end of the parent models. At this point, all parent models have the final performance higher than 20 BLEU (not visible in the figure). The learning curves with the same dashing correspond to the same model evaluated either on the child or parent development set.

We can see that from the start of the child model's training, the performance of the parent largely deteriorates on the parent's development set and that the behavior for shared-source and the shared-target is distinct. In the scenario with the shared source language, the model does not know to which language it should translate. Therefore it learns to always translate to the child target language.

Interestingly, in the shared-source language scenario, the performance drops nearly immediately. In contrast, whenever we investigate the shared-target scenario, the performance is deteriorating slower, and even after finishing the child's training, it is still able to translate Czech→English parent language pair with 15 BLEU.

**Observation 26:** *Parent performance deteriorates during the child training at different speeds depending on shared language position. The shared-source language scenario declines almost instantly. The shared-target language scenario deteriorates slowly.*

Figure 7.9 shows that the neural network forgets the parent source language slower. Thus, the phenomenon of critical forgetting is mostly concerned with the decoder part. If it forgot both of them in the same way, the drop in parent performance would be similar in both directions, i.e. English→XX and XX→English.

**Observation 27:** *A converged child model in the shared-target scenario can still translate the parent language pair to some extent. This is not possible in the shared-source scenario.*

This behavior could be a result of an error backpropagation, where the gradient vanishes as it travels back through the network, thus updates the encoder layers less than the decoder layers. This is especially true in transfer learning with shared-target because the network already knows how to generate the target language, e.g. English. Thus, it does not produce sufficient errors that would modify the encoder. Thus, the encoder does not forget the parent task as quickly. This suggests that the most important part of the model is the decoder and therefore, transfer learning with the shared target language is easier to learn as it already knows how target language should look like.

## 7.8 Rather Related Language, or More Data?

Whenever humans learn a new language, it is much easier for them if they know a related language. We suppose that it works similarly for NN with transfer learning. Zoph et al. (2016) in their transfer learning approach concluded that the "choice of parent model can have a strong impact on transfer models, and choosing better [related] parents for our low-resource languages could improve the final results". Furthermore, the use of related language pair as a way to improve the performance of a model has been widely studied, and researchers showed that related language pairs can be used as a source of improvements (Nakov and Ng, 2009; Nguyen and Chiang, 2017).

However, in Section 7.4.3, we saw that English→Russian is a better parent to English →Estonian than English→Finnish despite Estonian and Finnish are genetically re-lated, in contrary to Estonian and Russian.[5] And moreover, Estonian and Russian share the same script.[6] Thus, the main difference could be in the number of parent parallel training sentences where Russian–English has 12.6M sentences, and Finnish–English has only 2.8M sentences.

| | | | | | | | | |
|---|---|---|---|---|---|---|---|---|
| Original: | my | cat | likes | playing | with | my | other | cats |
| Option 1: | likes | playing | other | cats | my | cat | with | my |
| Option 2: | has | juggling | rather | study | research | has | those | set |
| Option 3: | tf | jha | sprlz | wshfpun | dpao | tf | vaoly | jhaz |

**Figure 7.10:** Various ways of damaging original sentences. Each column corresponds to the original word. All occurrences of a word type are replaced with the same string anywhere in the corpus, except for option 1 with shuffled words.

Besides, we showed in Section 7.6.4 that entirely unrelated languages still yield improvements in the child model. On the other side, having less resourceful parent can harm the performance of the child as we showed in Section 7.6.3.

These are indications that the relatedness of languages is not the main factor in transfer learning, and the size of the parent model has a bigger influence on child performance. In this section, we investigate the phenomenon of language relatedness in contrast to the training size of the parent model.

### 7.8.1 Artificially Related Language Pair

Many factors influence language relatedness: linguistic family, writing script, grammar phenomena, etc. Therefore, it is arguable how to measure the relatedness of various languages, especially comparing training data sizes with various degree of relatedness. In this section, we evaluate the effect of various degree of relatedness in contrast to various training data sizes. We present artificially related language where we can influence the degree of the language relatedness and measure the performance of the child. The artificial language is prepared by harmful modifications of the original training set.

There are several options on how to prepare artificial related language pair from the original training data. We can either shuffle words in the sentence, which creates a language pair with the same vocabulary, but different word order. The second option is to shuffle words within the language, e.g. "cat" would always be replaced by "juggling". The third option is to shuffle individual characters within each word type based on an exact replacement rule. The variants are visualized in Figure 7.10.

[5] Recall Table 7.8 on page 88: Estonian and Finnish are both from Uralic language family, Balto-Finnic branch, while Russian is from a different language family: Indo-European, Slavic branch.

[6] We do not transliterate the Cyrillic as done by other works (Nguyen and Chiang, 2017).

| | |
|---|---|
| Original: | **Pardon? Have you seen this cat?** |
| 70% related: | **Pardon?** Crnk **you seen this** tre? |
| 50% related: | Irfjwh? **Have you** ykkh ecsy **cat**? |
| 30% related: | Irfjwh? Crnk bwm **seen** ecsy **cat**? |
| 0% related: | Irfjwh? Crnk bwm ykkh ecsy tre? |

**Figure 7.11:** An example sentence illustrating levels of artificial relatedness via character substitution at various ratios.

The first option is prone to the word order, if we would like to have artificial language with consistent word order, we need to rely on some linguistic analysis, which would add another layer of uncertainty to this experiment. However, we investigate this option in Section 7.9.3 as a way to study word order). The second option is problematic as we want the related language to have a property that similar words behave similarly, for example, a word "cats" should be replaced with something similar to "juggling", however this would need an in-depth analysis of clusters of words and generation of rules which words are mapped to which. It is very hard, especially for an inflected language such as Czech. The third option generates a language with the same word order and language where visually similar words appear in a similar context. However, the language is unintelligible from the original. Moreover, due to the subword segmentation, the actual lengths of sentences as seen by NMT vary and thus the NN can not learn easy mapping across the sentences.

We decided to create the artificial language pair by the last option of substituting characters. We mix only the alphabet characters to match real-life conditions where most languages use the same punctuation and numbers. Furthermore, we preserved the capitalization. We modify both source and target language. Therefore there is no shared unmodified language between parent and child.

In order to scale the relatedness of languages, we randomly select X% of words from source and target language and substitute characters only in the remaining words; thus we obtain corpus with X% words unchanged. This way we get a pseudo-related language pair with a varying degree of relatedness where 0% is almost unrelated, and 100% is identical language. An example is in Figure 7.11.

In this experiment, we use modified English→Czech as the parent model with various degree of relatedness and various amount of training data. The corpus is created from CzEng 1.7.[7] We experiment with 80%, 50%, and 0% related corpus and each in 2M, 5M, 10M, and 20M parallel sentences. We use a warm-start technique where all models use the same vocabulary that is created from 50% related corpus.

[7] `https://ufal.mff.cuni.cz/czeng/czeng17`

| | 2M | 5M | 10M | 20M |
|---|---|---|---|---|
| 80% related | 18.11 | *20.46 | *21.81 | *22.10 |
| 50% related | 16.76 | 19.63 | 19.13 | 19.61 |
| 0% related | 15.09 | 16.83 | 17.91 | 18.25 |

**Table 7.22:** The results of various parent models. Each column specifies the size of parent training data, which is randomly downsampled from the original. Each row specifies parent model relatedness. The scores are in BLEU and specify performance of child model trained from the parent. For models with the star the performance dropped quickly during child training.

As the child language pair, we use 100k unmodified random sentences from the same corpus that have not appeared in our parent corpus. The performance of a child when trained solely on its training data is 7.17 BLEU.

The training process is as follows: train the parent model for 1M training steps, take the last model, and continue with the training of the child model for additional 200k steps. The best child model is selected based on the development data and evaluated against English→Czech testset.

The results in Table 7.22 present an interesting pattern. We can see that having more data can be more useful than a related language with fewer data. For example, 50% parent with 2M parallel sentences reached 16.76 BLEU, in contrast, having ten times more data but 0% related parent yields 18.25 BLEU. On the opposite, whenever the difference is only double, then the relatedness helps more with fewer data (19.13 BLEU vs. 18.25 BLEU). With 5M training data, the 50% parent already performs better than 20M with 0% (19.63 BLEU vs. 18.25).

We obtained similar results in real-life experiments as noted in the previous chapter, where more resourceful language (Czech or Russian) performed better than related language with less data (Finnish) with the Estonian child.

**Observation 28:** *Language relatedness plays a role in transfer learning. However, the amount of parent training data can improve performance even more than language relatedness.*

This finding can help when deciding which parent language pair to choose for a particular child. For example, whenever related language pair does not have enough training data, we could choose any training corpus with a high number of parallel sentences (for example Czech–English). However, our experiment is performed on artificially related languages, which could be considered as a noisy parent model. Moreover, we evaluated only 12 settings of relatedness and training size.

We noticed that with 80% relatedness the child model (labeled by a star) performs best without the training on its training data as the performance quickly deteriorated during the child training.

The language relatedness is not the only criterion or the most important one. Even unrelated parent can improve the performance of the child. In the next section, we examine the effect using parent language pair with an artificially huge number of parallel sentences.

## 7.9 Linguistic Features, or Better Initialization?

Neural networks have the reputation of being a black box. In this section, we try to understand if the gains can be attributed to some linguistic features or if the main contribution of transfer learning is simply a better initialization of weights than random initialization. There are several possible explanations of what are the reasons for the improvements:

- Knowledge of the shared language, e.g. English.
- General linguistic knowledge transferred from the parent model (e.g. word order patterns or typical lengths of sentences).
- Better hyper-parameters that are changed after the start (especially learning rate).
- Better initialization of weights (weights transformed from the parent could have more suitable distributions than the random initialization).

The main contribution could be due to the shared language between parent and child, e.g. English. Although it is possibly one of the main aspects, it is not the only one. In Section 7.6.4, we showed that transfer learning also helps languages that have both source and target language different from the parent model, i.e. Spanish→ Russian helping Estonian→English.

Zoph et al. (2016) showed that transfer learning does not utilize only the shared English, but also other parameters from the second language. Thus, other options could lead to the improvements of the child model.

Alternatively, the improvements could yield from linguistic features. For example, the child could transfer some knowledge about the word order or the ratio in source and target lengths. On the other hand, the main benefits could be attributed only to NN layout. If we compare the training model from a random initialization or a parent model, there are two main differences. The first difference is in the initial weights, where transfer learning has weights initialized already in some informed part of the parameter space compared to the random initialization. The second difference is in the learning rate because we are using non-constant learning rate depending on the global number of steps it changes through the training process, which can simply mean that different learning rate could lead to the same improvements.

In this section, we start by discussing the effect of shared language on the NMT by analyzing various layers of the network in Section 7.9.1 as well as investigating the behavior of the model when the shared language changes position from parent to child (Section 7.9.2). Then we explore some linguistic features of parent training data and analyze the generated output in Section 7.9.3, Section 7.9.4 and Section 7.9.5.

We conclude the section by analysis of learning rate influence in Section 7.9.6 and comparing transfer learning to random initialization in Section 7.9.7.

### 7.9.1 Freezing Parameters

Thompson et al. (2018) investigated, which parts of a model are responsible for the gains during domain adaptation. They used the technique of freezing model sub-networks to gain an insight into NMT system behavior during the continued training.

They segmented the RNN model (Bahdanau et al., 2014) into five sub-networks: source embeddings, target embeddings, encoder, decoder with attention mechanism and the softmax layer responsible for the generation of the output. Then they follow standard scenario of domain adaptation.

In this section, we use their technique to evaluate which parts of the neural network are crucial for transfer learning. In order to analyze transferred parameters that are the most helpful for the child model and which need to be updated the most, we follow the strategy by Thompson et al. (2018). We carry out the analysis on Estonian–English pair with Czech–English parent.

Based on the internal layout of Transformer model parameters in the T2T, we divided the model into four parts:

1. Word embeddings map each subword unit to a dense vector representation. The same embeddings are shared between the encoder and decoder.
2. The encoder part includes all the six feed-forward layers converting input sequence to the deeper representation.
3. The decoder part is again six feed-forward layers preparing the choice of the next output subword unit.
4. The multi-head attention is used throughout encoding as well as decoding, as self-attention layers interleaved with the feed-forward layers (see Section 3.3.2). We do not separate the self-attention layers used in the encoder or decoder, therefore when freezing encoder (resp. decoder) we also freeze some of the attention layer matrices.

We run two sets of experiments: either freeze only one out of the four parts and leave updating the rest of the model, or freeze everything except for the examined part.

The results are in Table 7.23. Based on the results, the most important part of NN that has to be changed is the decoder for EN→Estonian (resp. encoder for Estonian→EN) that handles the language that changes from the parent to child. With this part fixed, the performance drops the most.

The same observation is confirmed in Table 7.24: all the model parts (including the multi-head attention) can be reused precisely from the parent model as long as the decoder for EN→Estonian (resp. encoder for Estonian→EN) can learn the new language.

| Frozen part | EN→Estonian | Estonian→EN |
|---|---|---|
| All | 1.99 | 1.39 |
| Embeddings | 19.79 | 22.89 |
| Encoder | 19.65 | 20.61 |
| Decoder | 18.76 | **23.95** ‡ |
| Attention | 19.73 | 23.00 |
| None | **20.07** | 23.35 |

**Table 7.23:** Child BLEU scores when trained with some parameters frozen. Each row represents a parameter set that was fixed at the pre-trained values of the Czech→English parent. Best result for frozen parts in each column in bold.

| Non-frozen part | EN→Estonian | Estonian→EN |
|---|---|---|
| All | **20.07** ‡ | **23.35** ‡ |
| Embeddings | * | * |
| Encoder | * | 13.21 |
| Decoder | 7.87 | 5.76 |
| Attention | 6.19 | 10.69 |
| None | 1.99 | 1.39 |

**Table 7.24:** Child BLEU scores when trained with most parameters frozen. Each row represents a parameter set that was free to train; all other parameters were fixed at their pre-trained values. Best result for non-frozen parts in each column in bold. The results marked with * diverged as the model could not train anything.

We got a significantly ‡ better score when the decoder was frozen compared to when all the network were free to train in Estonian→English (23.95 vs. 23.25 BLEU). This shows that at least for examined language pair, the Transformer model lends itself very well to decoder reuse. However, we do not see the same in the opposite direction, which confirms that the position of the shared language makes the task different as we discussed in Section 7.7.

**Observation 29:** *Freezing the decoder when the target language is shared during child training can significantly improve the final performance.*

Zoph and Le (2016) needed freezing embeddings for their transfer learning to successfully work. On the other hand, freezing embeddings is harmful to our transfer learning.

Other results in Table 7.23 reveal that the architecture can compensate for some of the training deficiencies. Freezing the encoder (resp. decoder for Estonian→EN) or attention is not that critical as a frozen decoder (resp. encoder).

**Observation 30:** *Transformer model is robust enough to compensate for some frozen parts and reach a comparable performance.*

Whenever we freeze everything except a particular layer, we get a completely new picture. Results in Table 7.24 show that the network struggles to change the behavior from the parent when most of the network is frozen. Especially the parent embeddings are the least useful for the child because keeping only them leads to diverged training. The diverging results show that NN is not capable of providing all the needed capacity for the child, unlike the self-attention.

All in all, these experiments illustrate the robustness of the Transformer model in that it is able to train and reasonably well utilize parent weights even when the training is severely damaged. Interestingly, the attention is a crucial part of the network as it can compensate for the harmful effect to some extent. We use this knowledge in Section 7.9.3, where we evaluate the parent model with damaged word order.

### 7.9.2 Swapping Direction in Parent and Child

In the previous section, we showed that the side of the network with shared language is modified the least, and when frozen, it leads to the best performance.

We experimented with scenarios of shared-source, shared-target, and no-shared language. In this section, we investigate the scenario, where the shared language is switched to the other side between parent and child in order to investigate if the technique can transfer other features across various parts of the network.

In other words, we now allow a mismatch in the translation direction of the parent and child. The parent XX→English is thus followed by an English→YY child or vice versa. We use Estonian–English language pair as the child with various parents. The results are from our paper Kocmi and Bojar (2018).

This way, the child can not use the parent target language model as languages on both sides changed. It is important to note that Transformer shares word embeddings for the source and target side. However, we showed in the previous section that the word embeddings are not crucial for the training, although some improvements could be due to better English embeddings.

The results in Table 7.25 document that an improvement can be reached even when none of the involved languages is reused on the same side. This suggests that the model can transfer further knowledge across various parts or layers of the network. Although there are too many factors that could influence it (language relatedness, parent training size, etc.), thus we can not make any definite conclusion.

More importantly, the improvements are better when the shared language is aligned (column "Aligned EN"), which concludes that the shared language does play a sig-nificant role in transfer learning. This finding could be used whenever we want to train only one parent, for example, due to the time or resource restrictions and use it for more children even those with the shared language on the other side.

| Parent | Child | Transfer | Baseline | Aligned EN |
|---|---|---|---|---|
| Finnish→EN | EN→Estonian | **18.19** ‡ | 17.03 | 19.74 |
| Russian→EN | EN→Estonian | **18.16** ‡ | 17.03 | 20.09 |
| EN→Finnish | Estonian→EN | **22.75** ‡ | 21.74 | 24.18 |
| EN→Russian | Estonian→EN | **23.12** ‡ | 21.74 | 23.54 |

**Table 7.25:** Results of child following a parent with the common English appearing on the other side. "Baseline" is trained on child data only. "Aligned EN" is the more natural setup with English appearing on the "correct" side of the parent (source stays source, target stays target), the numbers in this column thus correspond to those in Table 7.11. The ‡ represents significantly better model when comparing "Transfer" and "Baseline".

However, it can not be the only source of improvements as we showed in Section 7.6.4 that the no-shared language scenario improves the performance of the child.

**Observation 31:** *The improvements of transfer learning are partly but not fully attributed to the shared language between parent and child.*

### 7.9.3 Broken Word Order in Parent Model

We showed that the shared language plays a vital role in transfer learning, now we attempt to find some particular linguistic features explaining the gains.

Intuitively, the child model could transfer linguistic knowledge from the parent such as word order, length of the target language sentences, etc. In this section, we experiment with the somehow modified parent to study the effect on the transfer of knowledge.

We use the English→Czech as a parent model and the child is English→Estonian. We used the shared-source language scenario as it is more difficult for transfer learning (see Section 7.7), and the improvements can not be attributed to the shared English target side as a better language model.

In the first experiment, we change the word order of a parent language pair in order to find out if the language word order plays an important role in transfer learning. Both of examined languages have mostly SVO (subject-verb-object) word order.

The word order is important for the attention mechanism that learns which parts of the source it should consider most.

We use sorting or shuffling of words (tokenized on whitespace) as a way to create a broken parent language. We modify only the word order of the parent model, therefore "shuffle source" means shuffling the source sentences (English in this experiment) and leaving the target language unmodified.

| Parent | Child Performance | Parent Performance |
|---|---|---|
| Unmodified parent | 20.07 | 23.48 |
| Shuffle source | 19.18 | 12.63 |
| Shuffle target | 19.16 | 2.78 |
| Shuffle both | 18.43 | 2.23 |
| Sort target | 19.45 | 2.29 |
| Shuffled sentences | 0.03 | 0.00 |
| Baseline | 15.80 | – |

**Table 7.26:** Results of transfer learning with modified parent word order. The performance is measured in BLEU.

Additionally, we created an experiment, where we shuffled all sentences which breaks the sentence pairs (row "Shuffled sentences"). This way, the parent model could learn to generate random sentences that are not related to the source sentence. Thus, it can mainly learn the decoder's language model.

Results are in Table 7.26. The sorting of target side is less damaging than shuffling. Parent with both source and target side shuffled has the worst performance. However, it is still a good parent model for transfer learning and improves the performance of the child be more than 2 BLEU points over the baseline.

Interestingly, the neural network has equal performance whenever the source and the target are shuffled despite the different performance of the parents (12.63 vs. 2.78 BLEU). We checked the performance of child model by various automatic metrics, and in all of them, the shuffled source and target reached similar scores.

This suggests that the word order of the target language is not the main feature in transfer learning. A better explanation is that the shuffling of either source or target breaks the attention mechanism in the same way, and then it needs to retrain on the child data. However, further analysis is needed as there can be other factors that can cancel each other out.

**Observation 32:** *There is little difference for the final child performance between shuffled word order of parent's source or target languages.*

The poor performance of parent on English→Czech testset is understandable because the BLEU is computed based on $n$-grams of words that are heavily disrupted by shuffling. When we studied the outputs of the parent models, we noticed that they actually learned to translate and only the shuffle the output on top of that. The "sorted target" model translates and even alphabetically sorts the outputs. This documents the high flexibility of Transformer's skills in capturing word relations: between source and target, it happily resorts to lexical relations, and within the target, it easily captures (memorizes) alphabetical sorting.

| Parent | BLEU | nPER | nTER | nCDER | chrF3 | nCharacTER |
|---|---|---|---|---|---|---|
| Baseline | 17.03 | 47.13 | 32.45 | 36.41 | 48.38 | 33.23 |
| English→Russian | 20.09 | 50.87 | 36.10 | 39.77 | 52.12 | 39.39 |
| English→Czech | 20.41 | 51.51 | 36.84 | 40.42 | 52.71 | 40.81 |

**Table 7.27:** Various automatic scores on English→Estonian testset. Scores prefixed "n" reported as (1 − score) to make higher numbers indicate better translation.

We have noticed interesting behavior of "sort target". The model generates correct words in alphabetical order. We compute the unigram BLEU score and get 51.1 (compared to an unmodified model that has 53.8 unigram BLEU). If we evaluate the BLEU on the sorted target, we obtain 14.77 BLEU (result is not in the table). This result shows that the model learns to translate without access to the word order.

**Observation 33:** *NMT models, in general, can learn to some extent even with source sentences having a broken word order. The word order is nevertheless crucial for a good performance.*

We showed that damaging the word order in the parent model leads to a slight drop in performance of the child, still staying high above the baseline. In the next section, we analyze the outputs of the child model and look for potential over-estimations of translation quality that could emerge from the usage of BLEU metrics.

Lastly, "Shuffled sentences" can not learn anything and obtain BLEU of 0.03. The parent model learned to generate one sentence for each input, and the child only changed the sentence into Estonian and generated "See on meie jaoks väga tähtis." (MT gloss: "This is very important to us.").

### 7.9.4 Output Analysis

We rely on automatic evaluation. Thus, we need to prevent some potential over-estimations of translation quality due to BLEU. For this, we took a closer look at the baseline English→Estonian model (BLEU of 17.03 in Table 7.11) and two English→Estonian children derived from English→Czech (BLEU of 20.41) and English→Russian parent (BLEU 20.09).

Table 7.27 confirms that the improvements are not an artifact of uncased BLEU. The gains are apparent with several (now cased) automatic scores.

As documented in Table 7.28, the outputs of transferred models are slightly longer in terms of words produced. In the table, we also show individual n-gram precisions and BP of BLEU. The longer output helps to reduce the incurred BP, but the improve-ments are also apparent in n-gram precisions. In other words, the observed gain can not be attributed solely to producing longer outputs.

| Parent | Length | BLEU Components | BP |
|---|---|---|---|
| Baseline | 35326 | 48.1/21.3/11.3/6.4 | 0.979 |
| English→Russian | 35979 | 51.0/24.2/13.5/8.0 | 0.998 |
| English→Czech | 35921 | 51.7/24.6/13.7/8.1 | 0.996 |

**Table 7.28:** Candidate total length, BLEU n-gram precisions and brevity penalty (BP). The reference length in the matching tokenization was 36062.

| Appeared in | English→Russian | English→Czech |
|---|---|---|
| Baseline+Reference | 15902 (44.2 %) | 15924 (44.3 %) |
| Baseline only | 7209 (20.0 %) | 7034 (19.6 %) |
| Reference only | 3233 (9.0 %) | 3478 (9.7 %) |
| Neither | 9635 (26.8 %) | 9485 (26.4 %) |
| Total | 35979 (100.0 %) | 35921 (100.0 %) |

**Table 7.29:** Comparison of child outputs vs. the baseline and reference. Each column shows child trained from different parent either English→Russian or English→Czech.

**Observation 34:** *Transfer learning helps the child model to generate slightly longer sentences, and there are also clear improvements in produced n-grams.*

Table 7.29 explains the gains in unigram precisions by checking words in the child outputs that were present also in the baseline and/or confirmed by the reference. We see that about 44+20% of words of child outputs can be seen as “unchanged” compared to the baseline because they appear already in the baseline output. (The reference confirms the 44% words.)

The differing words are more interesting: “Neither” denotes the cases when the child model produced something different from the baseline and also from the reference. Gains in BLEU are due to “Reference only” words, i.e. words only in the child output and the reference but not in the baseline. For both parent setups, there are about 9–9.7 % of such words. We looked at these 3.2k and 3.5k words, and we have to conclude that these are regular Estonian words; no Czech or Russian leaks to the output and the gains are not due to simple word types common to all the languages (punctuation, numbers or named entities). We see nearly identical BLEU gains even if we remove all such simple words from the child output and references.

| | Parent | | Child | |
|---|---|---|---|---|
| Sentence lengths | BLEU | Avg. words | BLEU | Avg. words |
| 1-10 words | 8.57 | 10.9 | 16.57 | 15.3 |
| 10-20 words | 16.21 | 15.4 | 17.48 | 15.3 |
| 20-40 words | 12.59 | 21.9 | 17.99 | 15.3 |
| 40-60 words | 5.76 | 35.5 | 16.80 | 15.5 |
| 1-60 words | 22.30 | 15.3 | 19.15 | 15.4 |

**Table 7.30:** Performance and average number of words generated over the testset. The references have average number of words 15.6 for the Czech testset 15.4 for the Estonian testset.

### 7.9.5 Various Lengths of Parent Sentences

In Section 7.9.4, we showed that the child model generates slightly longer sentences than when trained without transfer learning. In Section 4.5, we discussed how strongly the Transformer overfits towards the sentence lengths seen in the training data. In this section, we put these observations together and investigate if the effect is visible also in the child when the parents are trained on corpora with sentences limited to certain length ranges.

In this experiment, we take the Czech–English corpus and randomly select sentences of various length creating training corpus for different parents. Each of the parents is trained with the corpus of sentences with lengths in the predefined range. The length of the source and target sentences are different. Therefore, we use the sum of both lengths as the criterion.

We use four parent models. The first has sentence pairs with the length of either source or target of at most 10 words (tokenized by spaces). The second parent uses sentences of length 10 to 20 words. The third parent uses sentences of length 20 to 40, and the last parent uses sentences of length 40 to 60 words. We use English→Estonian as the child model.

We want to make the experiment comparable. Therefore, each training corpus has exactly 300M words. Thus, training set with the shortest sentences has the most sentences altogether.

Table 7.30 shows the results from our experiment. We start by discussing the parent performance. The parent model is unable to generalize from training data of different length for the training set. Although it generates sentences with slightly better length ratios than the training set, however, it can not generalize the sentence length well. For example, when training on “1-10 words”, it produces testset sentences with an average length of 10.9 words, which is more than it saw during the training. Similarly, “40-60 words” generates sentences of an average length of 35.5, which is shorter than training sentences. This closely matches Observation 4.5 on page 49.

When we discuss the performance of the child, we see that the child model generates sentences of valid lengths; the actual distribution over the testset is 15.4 words per test sentence. The performance is reaching 16.57 to 17.99 BLEU where the best parent model looks like the parent trained on "20-40 words" corpus. However, training the model on sentences of all lengths (row 1-60 words) leads to the best performance.

**Observation 35:** *Child model output lengths are not significantly influenced by the lengths of parent training sentences. The child can learn the length distribution by itself from child training data.*

Therefore, the finding that child model generates slightly longer sentences than baseline from Section 7.9.4 is not due to the lengths of sentences in the parent training set.

In conclusion, it seems that the length of outputs of the parent model is not that important factor behind the performance improvements of the child model.

### 7.9.6 Parent's Performance Influence

So far, we discussed features that could be considered linguistic. In the rest of the section, we examine various attributes that are associated with the neural architecture.

Transfer learning is, in fact, a way how to initialize weights for the child model in contrast to random initialization in the baseline. One of the explanations of the success could be attributed solely to the initialization or learning rate as these are only two attributes that change with the parent model.

In this section, we study the effect of various learning rates, and final parent performances on various models trained for a different amount of time in order to find out if there is a correlation between them and the child performance. Furthermore, we answer a question if it is necessary to train the parent model fully or is only a fragment of the training time enough?

In transfer learning, we train the parent model to its maximal performance, e.g. until convergence on the development set. In the following experiment, we compare the parent model in various stages of training and investigate the influence on the performance of the child model.

In Kocmi and Bojar (2018), we took English→Finnish as the parent model and used the warm-start transfer learning of child's English→Estonian model after 50k, 100k, 200k, 400k, and 800k of parent's steps. In order to simplify notation in this section, we label a child model that started after X parent's steps as a "X-child", e.g. 400k-child is a model trained after 400k parent's steps.

We conclude based on the learning curves in Figure 7.12 that the child's performance improvements correlate with the parent's performance. Furthermore, we see that only 50k steps are enough to outperform the baseline.

**Observation 36:** *A longer-trained parent model (or model trained until convergence) leads to a better performing child model.*

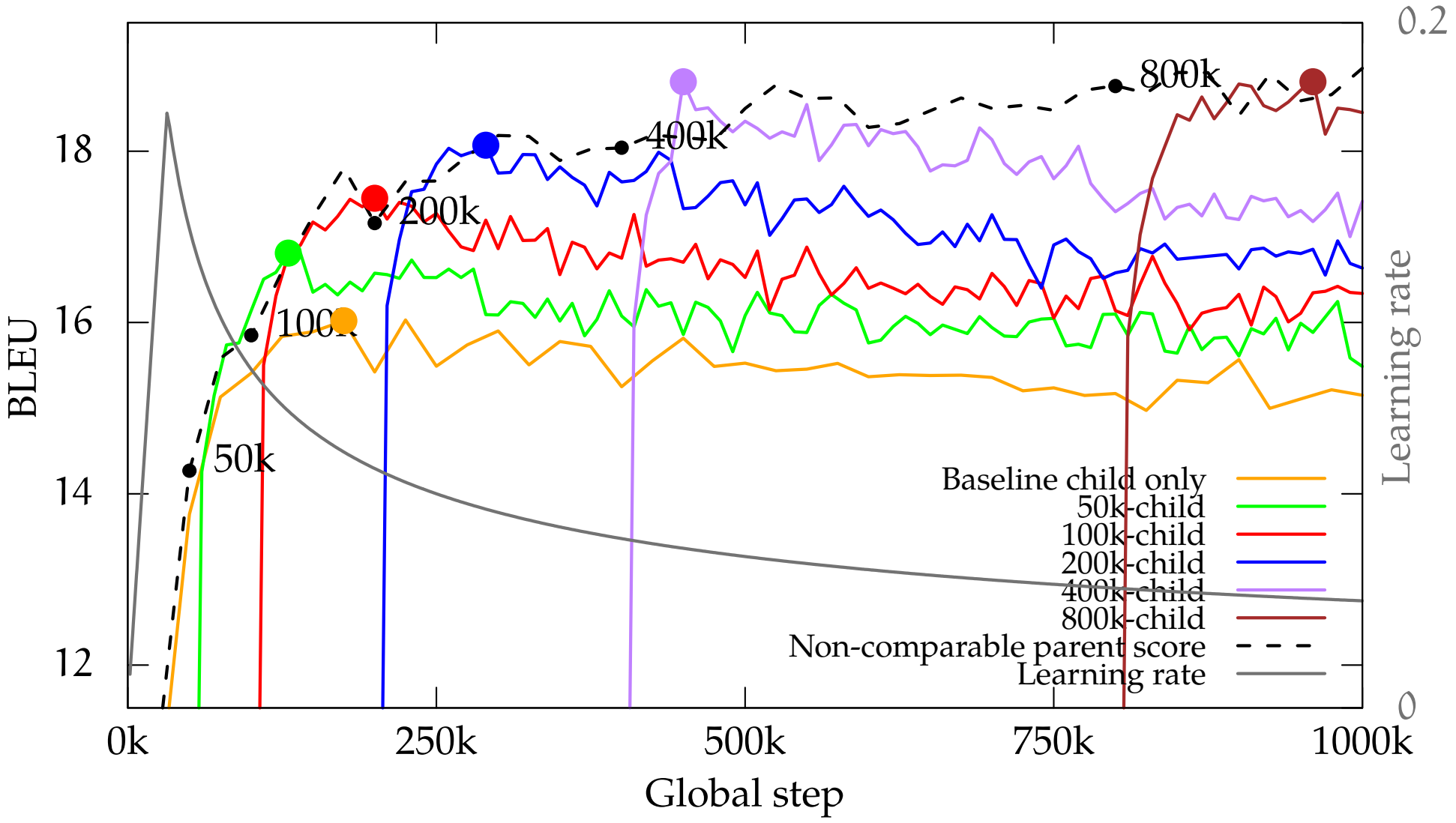

**Figure 7.12:** Learning curves on development set for English→Finnish parent and English→Estonian child. The child starts training after various number of parent's training steps. Black dots specify a performance of the parent at the moment when the child training started. The colored dots show the best performance of the child model. The grey curve shows how the learning rate depends on the global step number.

However, notice that the performance of the 400k-child is equal to the 800k-child, both having 18.8 BLEU despite the parent's performance improving from 18.0 to 18.8 BLEU.[8] This could be only an anomaly, or it could suggest that the child's performance also depends on different factors, which we try to examine further.

One such factor is the learning rate, which is the only parameter that depends on the global number step and decreases during the training. As the learning rate, we use a function of the inverse square root of steps with the warm-up stage of 16k steps, known as the "Noam" scheme. The steepness of the learning rate throughout the training is visualized in grey color in Figure 7.12.

In order examine the factor of the learning rate, we prepare experiment where we fine-tune the child model on various stages of parent training as in Figure 7.12 however we investigate various learning rates for each parent step.

[8] We remind that the BLEU scores of parent and child are not comparable as they are calculated on different testsets, and also that the scores fluctuate a lot between iterations.

| LR set as if at: | 0k | 50k | 100k | 200k | 400k | 800k | 1600k | Parent |
|---|---|---|---|---|---|---|---|---|
| 0k-child | 8.16 | 7.55 | 7.24 | 6.43 | 6.09 | 5.45 | 4.61 | 0.0 |
| 25k-child | 12.02 | 12.47 | 12.76 | 12.95 | 12.97 | 12.91 | 12.98 | 17.7 |
| 50k-child | 12.72 | 13.48 | 13.71 | 13.77 | 13.72 | 13.58 | 13.70 | 19.9 |
| 100k-child | 13.31 | 14.01 | 14.06 | 14.26 | 14.64 | 14.59 | 14.56 | 21.6 |
| 200k-child | 13.70 | 14.97 | 14.92 | 15.22 | 15.15 | 15.15 | 15.67 | 23.0 |
| 400k-child | 14.16 | 15.20 | 15.89 | 15.89 | 16.08 | 15.83 | 15.84 | 23.9 |
| 800k-child | 13.59 | 15.64 | 15.82 | 16.19 | 16.23 | 16.39 | 16.58 | 24.8 |
| 1600k-child | 14.21 | 15.68 | 15.86 | 16.29 | 16.86 | 16.61 | 16.72 | 25.2 |

**Table 7.31:** Experiment comparing various stages of parent model and learning rate (LR) stage. Each row corresponds to the same parent model as saved at given step (parent). Columns correspond to different learning rate shifts named by the global step at which it starts. The "0k-child" row is therefore baseline trained only on the child training set. The column "Parent" represent a performance of the parent at the step when the child was spawned.

Initially, we wanted to fix the learning rate to a constant value. However, it would add a new factor to the question since even during the parent model training, the learning rate slowly decreases. Instead, we change the global step value before starting the child training to pretend that the parent was trained to a different stage than it actually was. The child learning rate follows the Noam learning curve. Thus, we refer to the learning rate value using the global step index, e.g. 400k-child with 800k learning rate represent a model that fine-tunes on the parent model trained for 400k steps, and its learning rate behaves as if the parent was trained for 800k steps.

We now provide more details in the new experiment with slightly different settings. We decided to use the English→Czech as the parent model and English→Estonian as the child. We artificially downsampled the child training corpus Estonian–English to 100k sentences. We used Czech as a parent language pair in contrast to Finnish in Figure 7.12. We have used only 100k compared to 800k of child's training data in the original experiment. It was motivated to examine the effect of the less resourceful language. The initial learning rate is 0.2.

Table 7.31 presents the results of the experiment; all results are evaluated on the same testset. Thus, they are comparable, and we present them in the form of a heatmap for better visualization. However, the values within columns should be compared with care as they differ in the parent model that has been trained for a various number of steps. Note that the learning rate scheme of the parent model never changed.

From Table 7.31, we conclude that the learning rate schedule is important for the training. The 0k-child baseline trained without transfer learning (see the first row in Table 7.31) has the best performance with learning rate starting at 0k. Therefore the initial warm-up steps (16k in total) are needed for proper training. In contrast, resetting learning rate to 0k across all the transfer setups (see the column "0k") harms the final child performance.

**Observation 37:** *Warm-up steps and the peak in Noam learning rate schedule are crucial for good performance of the baseline. However, repeating this peak in child training damages the performance of transfer learning.*

Beyond that, there is not a clear pattern of the best learning rate. For all transfer learning results it fluctuates between 200k and 1600k learning rate where the differences in BLEU are mostly not significant, we suppose that it is due to the shape of learning rate that is close to being constant. Therefore, we conclude that the learning rate is not the primary source of improvements behind transfer learning.

**Observation 38:** *The improvements by transfer learning can not be attributed to a better chosen learning rate stage in its warmup-delay scheme.*

We see that with a better performing parent, the best performance of the child grows, in contrary to Figure 7.12 where the child performance has not changed between 400k-child and 800k-child. Therefore, the parent performance plays an important role in transfer learning. The best child performance of 16.86 is obtained with the parent trained for 1600k steps, which is around three weeks of training on one GPU. In comparison, the average training of the child in this experiment was 50k steps. We use stopping criterion from Section 4.3.1 and saw all child models started to overfitting.

The improvements for the child model diminish in relation to the parent's step, for example, it takes only 400k steps to reach the performance of 16.08, but additional 1200k steps to improve by 0.72 BLEU. Interestingly, the 25k-child already outperforms the baseline. The 25k steps were trained for eight hours, thus proving that the improvements in transfer learning are usable after a short period of training the parent model.

To answer our initial questions, we conclude that child performance depends on the performance of the parent model. We showed that the learning rate is not the main factor of the transfer learning gains, but it can limit the maximum gains if changed. Moreover, we showed that transfer learning significantly improves child performance over the baseline, even when the parent has not been adequately trained.

### 7.9.7 Same Language Pair in Reverse Direction

We showed that transfer learning can extract knowledge from a parent model whenever the shared language is on the different translation side as well as improve the performance whenever both languages are different. In both cases, it is the additional data from the parent that probably affect the performance of the model. However, there is yet another explanation of the improvements: transfer learning improvements could be attributed to the better initialization of weights.

| Parent | Child | Transfer | Child-only | Difference |
|---|---|---|---|---|
| EN→Estonian | Estonian→EN | **22.04** ‡ | 21.74 | +0.30 |
| Estonian→EN | EN→Estonian | **17.46** | 17.03 | +0.43 |
| EN→Finnish | Finnish→EN | **20.23** ‡ | 19.90 | +0.33 |
| Finnish→EN | EN→Finnish | **14.51** | 14.25 | +0.16 |
| EN→Odia | Odia→EN | **7.95** ‡ | 6.97 | +0.98 |
| Odia→EN | EN→Odia | **3.22** | 3.19 | +0.03 |
| Spanish→French | French→Spanish | **28.54** ‡ | 27.89 | +0.65 |
| French→Spanish | Spanish→French | **27.69** ‡ | 27.21 | +0.48 |

**Table 7.32:** Results of child following a parent with swapped direction. The ‡ represents significantly better results.

When we train the neural network, we have to decide on the initialization of the whole network. As Glorot and Bengio (2010); Mishkin and Matas (2016), NNs are sensitive to the variance of random initialization, and bad initialization can have a huge effect on the final model performance.

We can perceive transfer learning as a way of finding a better initialization of weights for the training of the child. Thus, it could improve the performance mainly due to the effect of having weights initialized to better values.

In order to investigate this explanation, we prepare an experiment where the parent model does not have any additional training data. We train the parent on reversed training data than the child, in other words, the parent is XX→YY model, and the child is YY→XX. Thus, the child model does not have access to any new training data. The experiments for Estonian–English are from our paper Kocmi and Bojar (2018). We selected languages to cover low-resource languages (Estonian–English) as well as high-resource languages (Spanish-French).

Table 7.32 shows a particularly exciting result: the parent does not use any other parallel sentences, but the very same corpus as a child with source and target side swapped and obtained a performance improvement. We see gains in both directions, although not always statistically significant.

**Observation 39:** *Transfer learning improves performance even in the situation where no new data are available and the child is trained on the parent training corpus in reverse direction.*

One explanation of the improvements could be that the training corpus is noisy and often contains English sentences on the wrong side, for example, in the Estonian part of the corpus. In order to verify it, we ran an automatic language identification and found out that Estonian part of the corpus contains only 0.1% English sentences, Finnish contains 3.4%, and Odia contains 0.0% of English sentences. The Estonian and Odia can not be attributed to the noisiness as there is nearly zero English sentences and the 3.4% for Finnish is low that we do not think it is the main reason behind the improvements.

The low-resource language as Odia–English reached low performance due to insufficient data for model training (see Section 7.2.4). In contrast, the high-resource language pair of French-Spanish reached significant improvements in both directions.

It is an exciting result. The model did not have access to any new data, yet it could extract new features from the reverse language pair, which it would not learn only from the original direction. Similar behavior has been shown in Niu et al. (2018), where they mixed both directions and added an artificial token indicating the target language.

The results from this experiment support our alternative explanation that the main improvements are simply from a better initialization of the model. Although, we showed that other features further improve the final performance of the model, such as shared language between parent and child, language relatedness, or size of parent training data.

## 7.10 Back-Translation with Transfer Learning

Back-translation is helpful mainly in scenarios where the parent model, which is used to translate monolingual data, has a reasonable performance (Hoang et al., 2018; Bawden et al., 2019). However, for the low-resource language it is difficult to train any suitable initial model.

In our paper Kocmi and Bojar (2019), we propose to use transfer learning on the low-resource language pair for training the initial model with a reasonable score. We train two models in parallel, one for each translation direction. The models iteratively generate back-translated data for the other one. We show this approach on two low-resource languages Gujarati–English and Kazakh–English.

As a parent model, we use Czech–English for Gujarati–English, and Russian–English for Kazakh–English. The training procedure is as follows.

First, we train two high-resource parent models for each studied language until convergence: English→Czech, Czech→English, English→Russian and Russian→English.

Then, we apply transfer learning with the use of an authentic dataset of the corresponding low-resource language pair. We preserve the English side: Czech→English serves as the parent to Gujarati→English, and English→Czech to English→Gujarati. The same strategy is used for transfer learning from Russian to Kazakh.

After transfer learning, we select one of the translation directions to translate monolingual data (model ①). As the starting system for the back-translation process, we have selected the English→Gujarati and Kazakh→English. The decision for Kazakh––English is motivated by choosing the better performing model, see Table 7.33. This is however only a rough estimate because higher BLEU scores across various language pairs do not always need to indicate better performance; the properties of the target language such as its morphological richness affect the absolute value of the score. For the Gujarati–English, we decided to start with the model with English source side in contrast to Kazakh→English.

| Training dataset | EN→GU | GU→EN | EN→KK | KK→EN |
|---|---|---|---|---|
| Authentic (baseline) | 2.0 | 1.8 | 0.5 | 4.2 |
| Parent dataset | 0.7 | 0.1 | 0.7 | 0.6 |
| Authentic (transfer learning) | ① 9.1 | 9.2 | 6.2 | ① 14.4 |
| Synth generated by model ① | – | ② 14.2 | ② 8.3 | – |
| Synth generated by model ② | ③ 13.4 | – | – | 17.3 |
| Synth generated by model ③ | – | ④ 16.2 | – | – |
| Synth generated by model ④ | 13.7 | – | – | – |
| Averaging + beam 8 | 14.3 | 17.4 | 8.7 | 18.5 |

**Table 7.33:** Testset BLEU scores of our setup. Except for the baseline, each column shows improvements obtained after fine-tuning a single model on different datasets beginning with the score on a trained parent model. The circled names points to the systems in the right side of the table.

After the back-translation, we mix the synthetic data with the authentic parallel corpus and train the first back-translated model ②. We repeat this process: Use the improved system (③ and then ④) to back-translate the monolingual data, and use this data in order to train the improved system in the reverse direction. We make two rounds of back-translation for both directions on Gujarati–English and only one round of back-translation on Kazakh–English.

The baseline models in Table 7.33 are trained on the authentic data only, and it seems that the number of parallel sentences is not sufficient to train the NMT model for the investigated language pairs (we obtained performance of 0.5 to 4.2 BLEU). The remaining rows show incremental improvements as we perform the various training stages. The last stage of model averaging takes the best performing model and averages it with the previous seven checkpoints that are one and a half hours of training time from each other.

Figure 7.13 above shows the progress of training of Gujarati–English models in both directions. We can notice that after each change of parallel corpus, there is a substantial improvement in the performance. The learning curve is computed on the development data. The corresponding scores for the testsets are in Table 7.33.

We also run a standard approach of back-translation without transfer learning, where we first trained baseline model to translate monolingual data and then trained a model (in reverse direction) on those synthetic data with concatenation with authentic data. We visualize the training with orange in Figure 7.13. When comparing with a transferred model trained with the first round of back-translation data (light blue), we can see the clear improvement in performance due to transfer learning.[9]

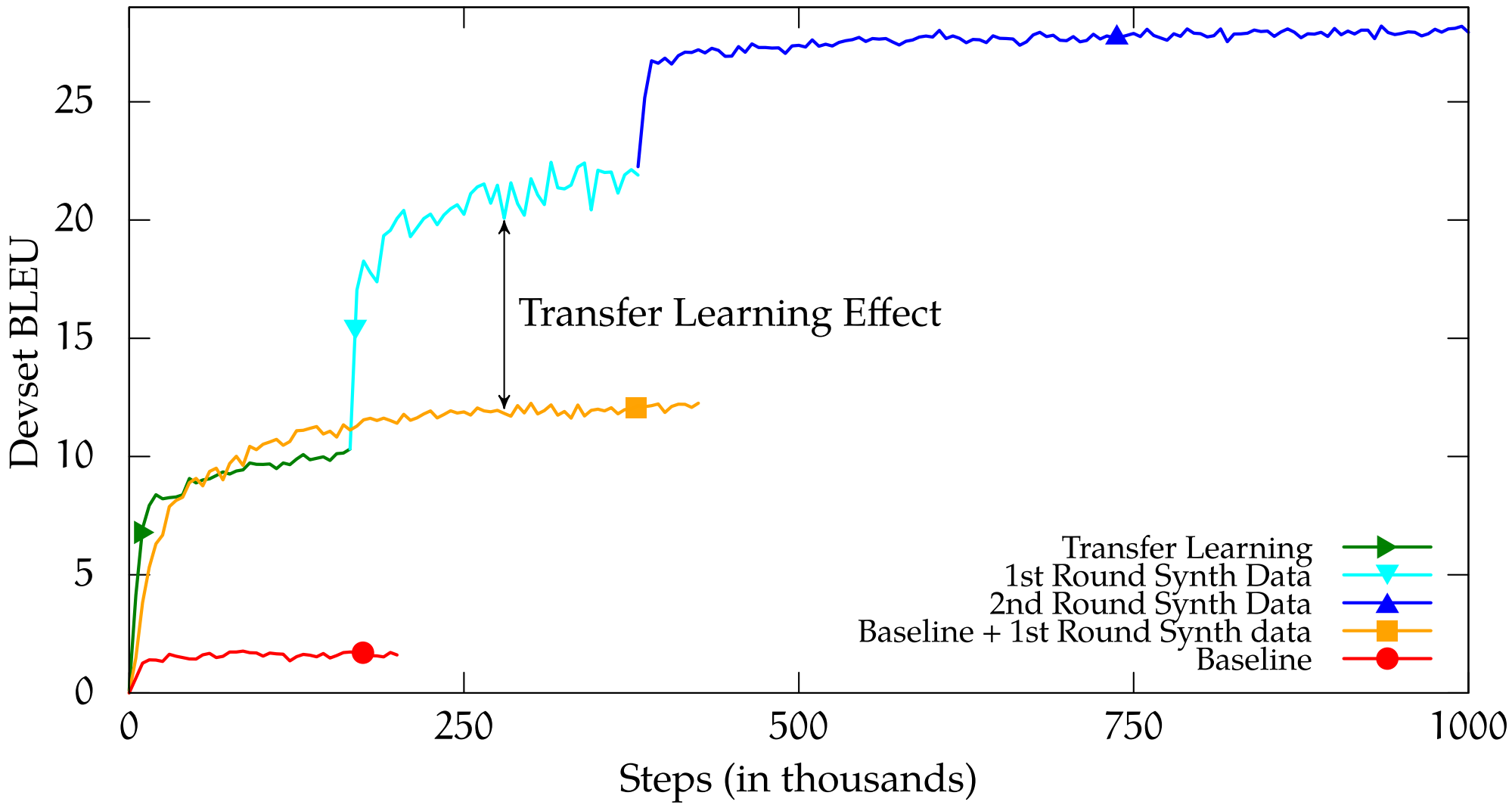

**Figure 7.13:** Learning curves of Gujarati→English models. Our approach is combination of four steps. The first is to train the parent model for 2000k steps (not in the figure), then transfer learning (marked with forward triangle). Then we continue with two rounds of back-translated data (downward and upward triangle). Baseline without transfer learning and back-translation is marked with a dot. Standard approach of back-translation without transfer learning is marked with a rectangle.

**Observation 40:** *Transfer learning can be used as an initial step for low-resource training in combination with back-translation technique to largely improve the performance.*

In conclusion, we showed that transfer learning can be used in combination with other techniques. Furthermore, it can be used to generate a reasonably good initial model for back-translation technique in the low-resource scenario, where it is difficult to train the model from random initialization.

[9] Each scenario (marked with rectangle and downward triangle) use synthetic data generated by different model from an equal monolingual corpus. The model generating the synthetic data either used transfer learning (downward triangle) or did not (rectangle).

# 8
# Observations and Advances in Multilingual MT

As we suggested, there is no way a book publication could sensibly cover an area as exploding in activity as we observe in multilingual translation in the last few years.

In addition to being in the focus of researchers, the area also serves as a battlefield of major technological companies. The language barrier is a tangible problem in the globalized world. Matching human performance in text or speech translation is an attractive and easy-to-explain goal. Offering high quality online translation service is resource intensive and any saving is highly desirable for the companies.

One of the widely recognized and visible activities in NLP are shared tasks. For instance WMT,[1] the flagship collection of shared tasks on MT, has been seeing and also specifically inviting multilingual MT systems. We would like to note that somewhat different from sports, the point of scientific shared tasks is not to claim victory and beat other participants but rather to evaluate competing methods in comparable conditions and learn from fellows. One of the reasons to focus on comparison of methods rather than participants' ranking is that only experienced researchers in the area are able to interpret the results considering all the fineprint. Simplified summaries or excerpts are risky; they can create false impressions. For instance, our English→Czech system scored better than human translation in 2018, see Table 9 in Bojar et al. (2018) because the evaluation was performed on isolated sentences. Translators produce sentences in the context of the document. When evaluated in isolation, human translations sound less natural than translations from MT which are also created in isolation. We can formulate this as an observation:

**Observation 41:** *Shared tasks are a great means for empirical comparison of methods but the results have to be interpreted always in the full context, including deficiencies of the evaluation.*

Coming back to multilingual MT, we remind the reader of the survey by Dabre et al. (2020) and cover key advances beyond 2020 in one critical area: massive models. As in the previous chapters, we primarily aim to help cut down on hype and highlight the risks the hype might bring for realistic and sustainable progress for the society.

[1] `https://www.statmt.org/wmt06` till `wmt21`

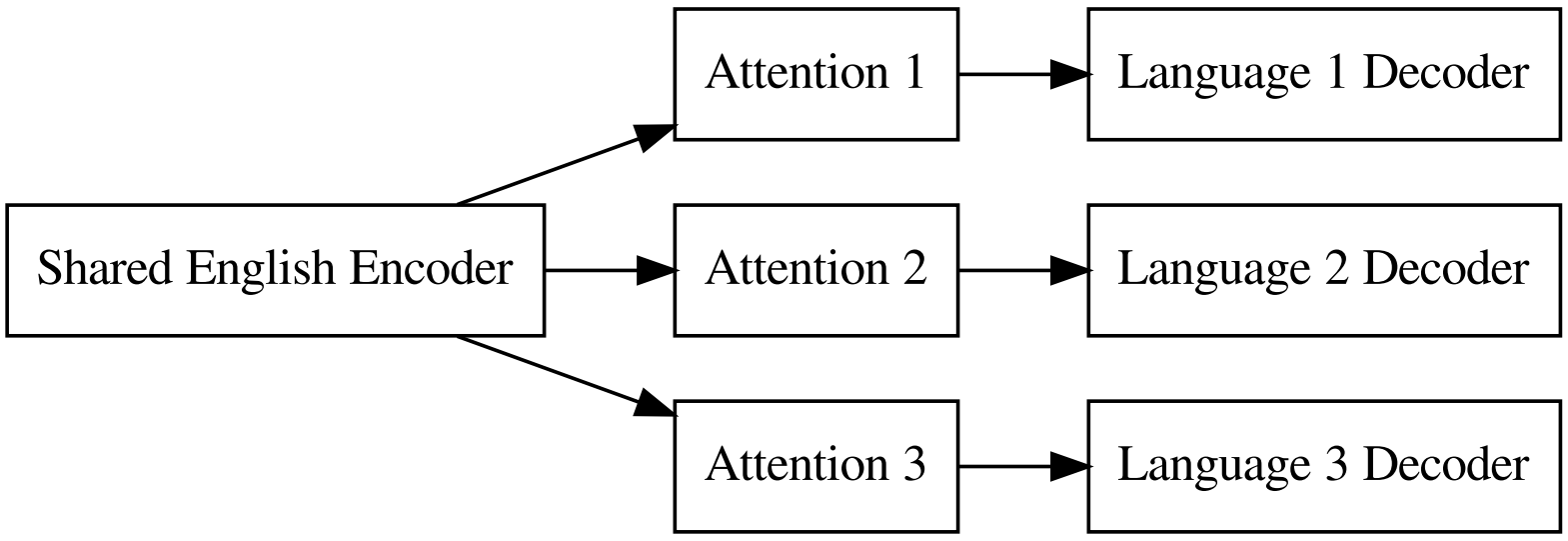

**Figure 8.1:** A sketch of the architecture by Dong et al. (2015). The model contains a single English encoder and separate attention and decoder for each of target languages.

## 8.1 Single Source, Multilingual Target

One of the first works of approaching machine translation in a one-to-many way was done by Dong et al. (2015). They used RNN NMT systems (Bahdanau et al., 2014) and shared same encoder across all language pairs. The attention mechanism and decoder are different for each target language pair. The architecture is visualized in Figure 8.1. Thanks to the shared encoder, their approach has fewer training parameters than several bilingual NMT models would have.

Dong et al. (2015) use homogeneous batches, i.e. there are sentences from one language pair in each batch. They resolve the scheduling problem as discussed in Section 3.5.2 by switching the language pair used in training batches every 1000 iterations.

Dong et al. (2015) evaluated the model on four language pairs English→French, English→Dutch, English→Spanish, and English→Portuguese, with all training data originating in Europarl corpus (Koehn, 2005). Two settings are evaluated: medium-size (2M sentence pairs per language pair), and subsampled to low-resource (300k sentence pairs). It is not clear from the paper but rather likely that the English sides of the training data are *not* shared. In other words, training on more language pairs implies larger training data of the source English. We reproduce the results in Table 8.1.

It is important to note that this early experiment was still running NMT without subword units and given the medium-to-small data size it performed comparably to PBMT. The observed gains of 1 BLEU point on average are promising but they might also come simply from the larger source-language data.

Dong et al. (2015) further report that the training of multi-lingual models not only reaches higher scores but it also trains faster (BLEU grows faster with the same number of training steps). This is, of course, useful, but again a targeted validation should be carried out so that we know how much of the gains can be attributed to the larger source-language data, how much to reuse of linguistic knowledge (e.g. ordering patterns in the languages or "predictive structure", as Dong et al., 2015, put it) and how much to "plain model regularization" which almost random additional data could bring.

| Language pair | Single NMT | Multi NMT | Single NMT subsampled | Multi NMT subsampled |
|---|---|---|---|---|
| English→Spanish | 26.65 | **28.03** | 26.65 | **28.29** |
| English→French | 21.22 | **22.47** | 21.22 | **21.89** |
| English→Dutch | 28.75 | **29.88** | 26.59 | **27.85** |
| English→Portuguese | 20.27 | **20.75** | 18.26 | **19.32** |

**Table 8.1:** Multi-target neural translation in comparison to a single language pair model given large-scale corpus in all language pairs. Columns subsampled show results where Dutch and Portuguese had only 15% of corpora size. Results are from Dong et al. (2015).

Dong et al. (2015) argue that the reason for better convergence is that the encoder parameters are shared across different language pairs and therefore improve the source language representation. This is especially useful for low-resource language pairs, where there are not enough data available to learn good source language representation. Moreover, it helps to overcome overfitting and data scarcity problem by having data in different language pairs. They support their reasoning by analysing nearest neighbours of various words and showing that that multilingual approach has more relevant near neighbour words than single language pair models, e.g. "provide" is close to "deliver", "providing" or "give" in the multi-lingually trained model whereas the low-resource single-pair model sees "though", "extending" and "parliamentarians" as the nearest neighbours, instead. The better modelling of the source lexical space is indisputable but the same effect might happen thanks to larger source data only.

## 8.2 Single Target, Multilingual Source

As an illustration of a multi-way model with multilingual source, we choose the work by Neubig and Hu (2018).

Neubig and Hu (2018) combined the multilingual MT (Johnson et al., 2017) with transfer learning by Zoph et al. (2016). Their approach has two stages. First, they train a universal multilingual system that can translate from all investigated languages into English. Second, they fine-tune the universal model for one specific source language ("Sing." for single-pair, e.g. English→Slovak). They evaluate the multilingual parent on 58 source languages, the largest reported multilingual system at that time. Detailed results are reported for selected very low-resource source language like Azerbaijani (AZE), Belarusian (BEL), Galician (GLG) or Slovak (SLK).

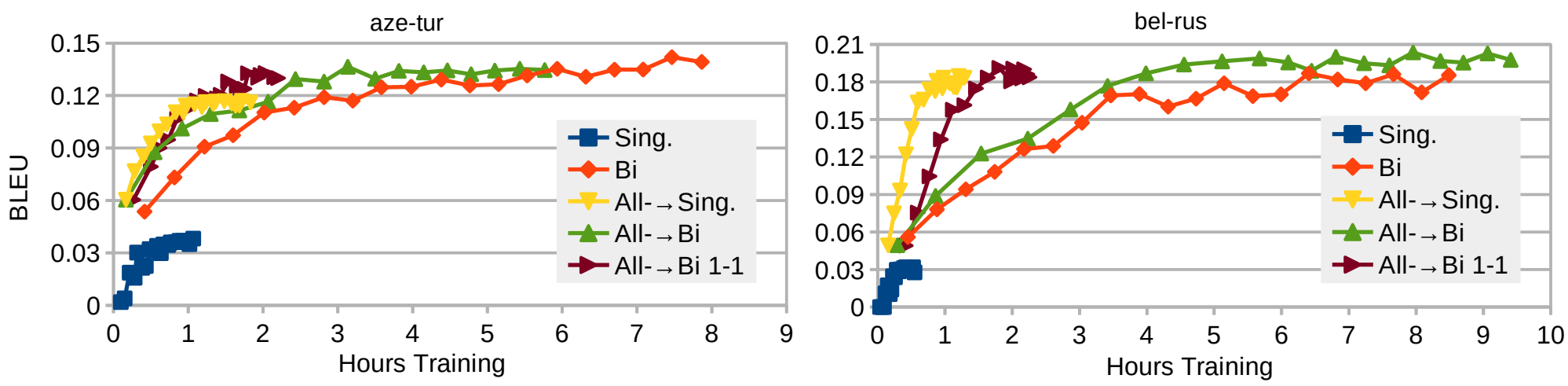

**Figure 8.2:** Learning curves of baseline MT (1-to-1, "Sing.") for low-resource Azerbaijani and Belarusian, complemented with mid-resource Turkish and Russian (2-to-1, "Bi") and transfer learning from a multilingual source model. Reproduced from Neubig and Hu (2018).

Moreover, they introduced a technique of using a helper related language pair to improve low-resource language performance, by mixing related language data into the fine-tuning training set. For example, when creating a model for Slovak→English, they include Czech→English parallel sentences in the fine-tuning phase, too, to boost the performance of Slovak via high-resource Czech. These 2-to-1 models are denoted as "bi-pairs" but they are evaluated only for the low-resource source language.

Figure 8.2 reproduces the learning curves for Azerbaijani (supported by Turkish, TUR) and Belarusian (supported by Russian, RUS). Clearly, fine-tuning a model converges faster and obtains better results than training a bi-pair model. This confirms our results from Observation 11 and suggests that adapting models from a universal one is a good strategy for the rapid construction of MT systems in new languages.

**Observation 42:** *Fine-tuning from a multi-lingual model speeds up training both for 1-to-1 child models (Observation 11) as well as for 2-to-1 models.* (Neubig and Hu, 2018)

We reproduce a subset of the final results by Neubig and Hu (2018) in Table 8.2. We see that for these low-resource languages, PBMT outperforms the off-the-shelf setup of NMT ("Single pair"). Training on two source languages ("Bi-pairs" in Table 8.2) is sufficient to surpass these baselines while training on all 58 languages is generally a little less effective.

Transfer learning is successful both when transferring for a bi-pair parent as well as when transferring from the highly multilingual model ("All → Single pair"). The best results are achieved when transferring from the highly multilingual to the bi-pair data.

A big confounder here, however, is the common target English. We suspect that already in bi-pairs, the gains can be more related to the increased size of the English data. Furthermore, this study neglects the very effective option to use back-translation discussed in Section 3.6. Back-translation is very likely going to be a rather strong baseline e.g. for the bi-pair models. Consider the Slovak and Czech case. For a fair comparison, the baseline should take all the English sentences from both Slovak-English and Czech-English data and translate them to Slovak with an initial English→Slovak model. The desired Slovak→English model should be then trained on both the authentic and synthetic Slovak-English data. A massive multilingual model, which we discuss below, by Fan et al. (2021) was analyzed in this respect for test languages reaching lower quality scores (BLEU between 2 and 10) by the multi-lingual model alone. Aside from a few exceptions, all these languages further benefited from back-translation.

| English from: Strategy | AZE (+ TUR) | BEL (+ RUS) | GLG (+ POR) | SLK (+ CES) |
|---|---|---|---|---|
| PBMT | 5.9 | 10.5 | 22.3 | 23.0 |
| Single pair | 2.7 | 2.8 | 16.2 | 24.0 |
| Bi-pairs | 10.9 | 15.8 | 27.3 | 26.5 |
| All | 9.7 | 16.7 | 26.5 | 25.0 |
| Bi-pair → Single pair | 11.4 | 16.3 | 27.5 | 27.1 |
| All → Single pair | 10.1 | 17.5 | 28.2 | 27.4 |
| All → Bi-pair | **11.7** | **18.3** | **28.8** | **28.2** |

**Table 8.2:** Results for translation from four low-resource languages into English when using the multilingual system by Neubig and Hu (2018). "Single pair" is trained solely on one language pair. "Bi-pair" represents model trained on a corpus with an auxiliary related source language (in brackets) included in the data. "All" represent model trained on 58 source languages. The last three rows represent results for fine-tuned models (warm-start, i.e. the main target language data was available when the parent model was prepared). The results are by Neubig and Hu (2018).

Fan et al. (2021) also ran a controlled experiment on 11 Slavic languages. As Figure 8.3 documents, back-translation alone reaches almost the same performance as the multilingual system.

**Observation 43:** *Simple back-translation and multilingual many-to-one setups reach comparable gains compared to the pairwise baseline. Combining both is helpful for low-resource pairs.* (Fan et al., 2021)

## 8.3 Multilingual Many-to-Many

As mentioned, multi-tasking can be achieved with larger or smaller component sharing of the model. We discuss examples of both situations in this section.

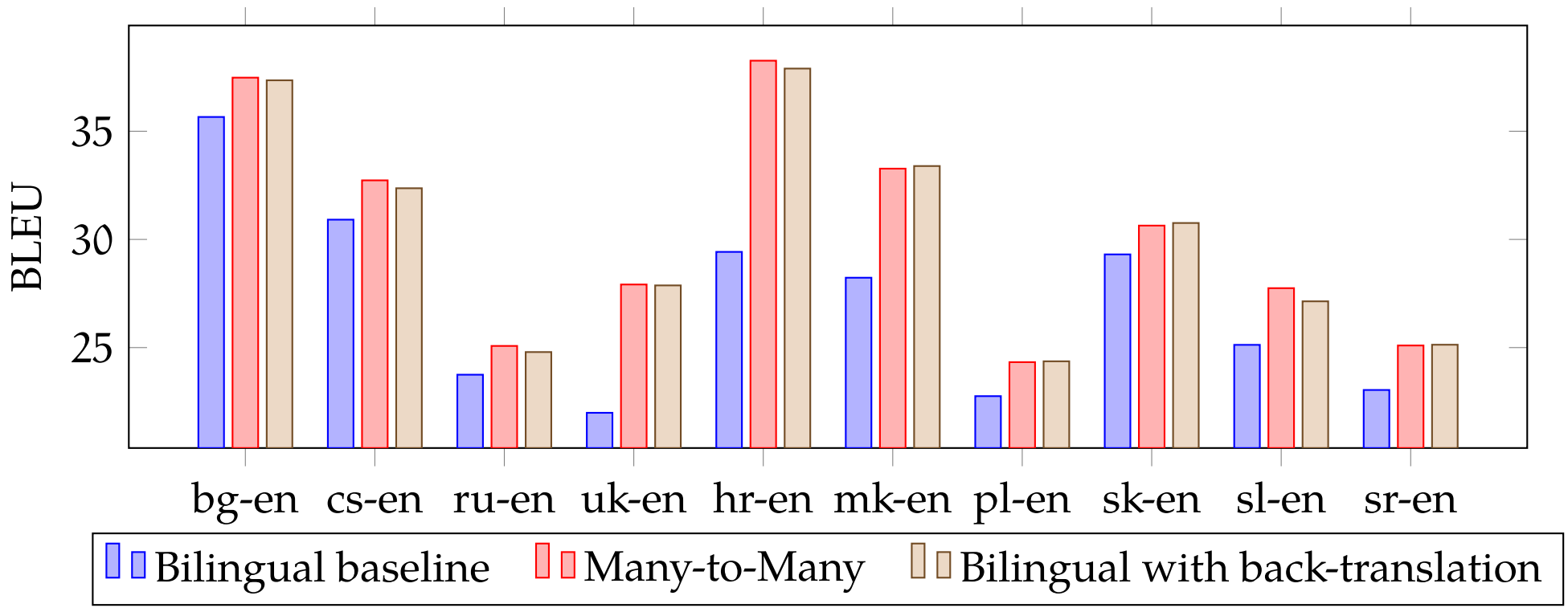

**Figure 8.3:** Performance of systems from 11 Slavic languages into English, comparing baseline pairwise systems, a multilingual system and baseline systems with back-translation. Data from Fan et al. (2021).

### 8.3.1 Limited Sharing in Multilingual Many-to-Many Models

One of the first approaches to multilingual many-to-many NMT was proposed by Firat et al. (2016). They approached multi-way multilingual systems as a way of a generalisation of work by Dong et al. (2015) (see Section 8.1), with the primary goal of reducing the total number of parameters needed to cater multiple source and target languages. The system uses individual encoders and decoders for each language, with only the attention mechanism shared. They allow to use different types of encoders to handle various source languages. In order to attend to any of them, their outputs have to be scaled to a common dimensionality, so a linear projection layer is added. With this setup, the number of model parameters grows linearly with the number of languages compared to quadratic in case of having an individual parallel model for each language pair. The scheme of the architecture is in Figure 8.4.

The model is trained on one language pair at a time, i.e. with homogeneous batches. To keep all the language pairs active in the model and avoid catastrophic forgetting (see Section 4.6), a special training schedule is needed. The suggested approach is to change the language pair after each network update (one batch evaluation).

Firat et al. (2016) experimented with six languages in an English-centric way: The training data were between English and one of the remaining five languages, namely English–French, English–Czech, English–German, English–Russian and English–Finnish. They showed that the multilingual setup with shared attention outperforms the single-pair baseline on low-resource languages in all translation directions. The low-resource scenario was prepared by downsampling training corpora, a setting we discussed in Observation 5 on page 59.

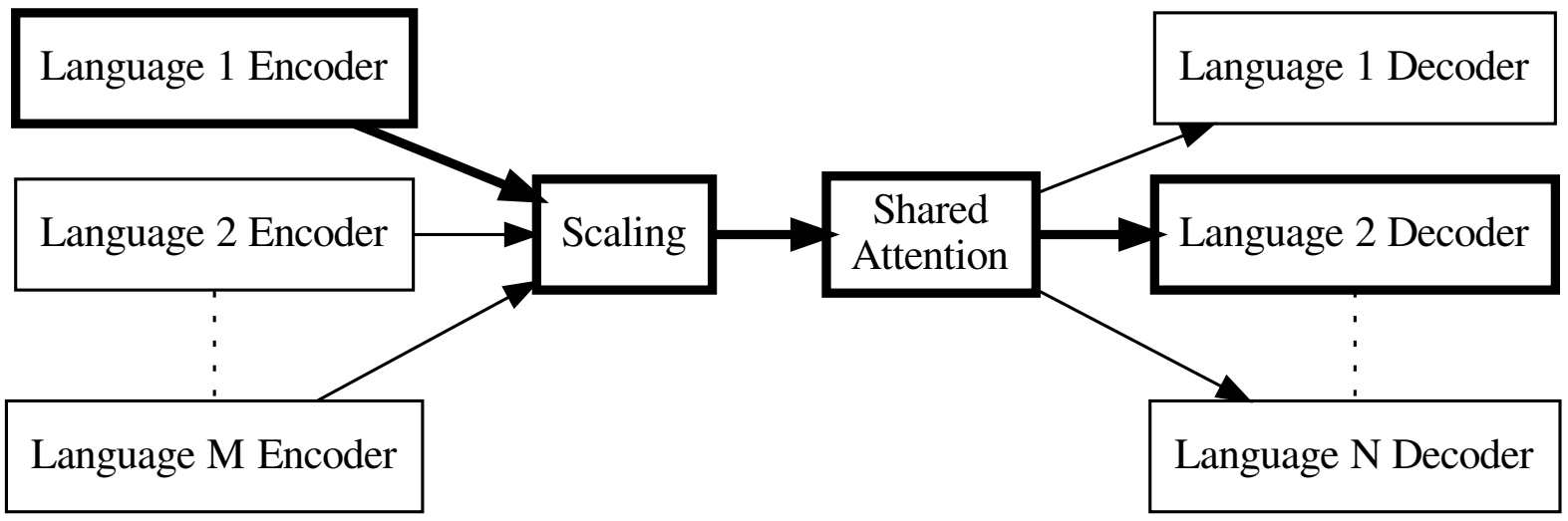

**Figure 8.4:** The multi-way architecture proposed by Firat et al. (2016). Each language has a separate encoder and decoder. The model contains one shared attention mechanism with linear transformation layer. During the training process, only one encoder and one decoder is used for each batch, as highlighted.

| Language pair | French | | Czech | | German | | Russian | | Finnish | |
|---|---|---|---|---|---|---|---|---|---|---|
| | Into EN | From EN | Into EN | From EN | Into EN | From EN | Into EN | From EN | Into EN | From EN |
| Single-lang NMT | 20.94 | **29.70** | 20.32 | **13.84** | 24.00 | **21.75** | 22.44 | **19.54** | 12.24 | **9.23** |
| Multi-lang NMT | **28.06** | 27.88 | **20.57** | 13.29 | **24.20** | 20.59 | **23.44** | 19.39 | **12.61** | 8.98 |

**Table 8.3:** Medium to high-resource BLEU scores of a English-centric multi-way multilingual system. Results are from Firat et al. (2016).

**Observation 44:** *A shared attention multilingual model outperforms single-pair translation models when the amount of available parallel data is small.* (Firat et al., 2016)

When using high-resource languages with more than two million training sentences, the multi-way system outperformed the single-pair baseline only when English was on the target side. With English on the source side, the baseline performs on par or better than the multi-way model. The results are in Table 8.3.

As Firat et al. (2016) mentioned, the fact that gains are observed only when translating into English confirms that larger target-side data are more important than multilinguality. This does not diminish the useful property that the multi-lingual model needs significantly fewer parameters than the single-pair NMT models together.

**Observation 45:** *For high-resource language pairs, multilingual multi-way models with shared attention outperform standard pairwise models only in the direction into the shared target language (English).* (Firat et al., 2016)

| Model | Single | Multi | | | |
|---|---|---|---|---|---|
| Dimensionality | 1024 | 1024 | 1238 | 1536 | 1782 |
| Number of parameters | 3B | 255M | 367M | 499M | 650M |
| English→Japanese | 23.66 | 21.10 | 21.17 | 21.72 | 21.70 |
| Korean→English | 25.42 | 22.87 | 23.46 | 24.00 | 24.67 |
| English→Portuguese | 38.40 | 37.35 | 37.42 | 37.80 | 37.92 |
| German→English | 31.77 | 31.17 | 31.65 | 32.24 | 32.32 |
| average diff | - | -1.72 | -1.43 | -0.95 | -0.76 |

**Table 8.4:** Comparison of a single model baseline and increasing size of the multilingual model. The multilingual model can translate between 12 language pairs, but we show results only for a subset of them. Results are from Johnson et al. (2017).

Firat et al. (2016) suggest that transferring knowledge with the language shared on the source side between the parent and child is a more difficult scenario than transferring with the shared language on target side. We analyzed this situation in Section 7.7.

### 8.3.2 A Universal Model for Multilingual Many-to-Many

Johnson et al. (2017) introduce another multilingual approach. They prepend a special word (or token) to each sentence which indicates the desired target language, "<2lang>". Then they mix all parallel sentences from all language pairs together. While the model can recognize the target language by the token, the architecture effectively shares all parameters between all languages, not only the attention as in Firat et al. (2016). The model implicitly learns translation between all languages. However, the performance of the multilingual system is worse than the single-pair baselines. This is mainly due to the same number of parameters as in the baseline that is used between several language pairs. Effectively there are thus fewer parameters available for each language pair. This loss in model capacity is not sufficiently compensated by the hoped-for generalization across languages.

**Observation 46:** *Universal multilingual NMT models have lower performance when compared to single language pair models of the same size.* (Johnson et al., 2017)

Johnson et al. (2017) support the hypothesis by increasing the size of the multilingual model, which reduces the performance gap to the baseline. The results from their experiment are in Table 8.4. As they increase the model size for the multilingual model, the average difference (loss) in BLEU is reducing. Most importantly, the twelve single language pair models have in total of 3 billion parameters. In contrast, the largest multilingual model has only 650 million parameters and has performed, on average, only 0.76 BLEU lower.

**Observation 47:** *Increasing the number of parameters of multilingual NMT reduces the loss behind simple pairwise models.* (Johnson et al., 2017)

### 8.3.3 Sentence Representations in Multilingual Models

The fact that a multilingual model can reach performances comparable to a set of standalone pairwise models despite having fewer parameters than the sum of parameters of these models suggests that there are indeed shared elements in the learned representation. To learn more about this sharing, Kudugunta et al. (2019) analysed a model trained on 102 languages to and from English (204 direct language pairs).

They used SVCCA designed by Raghu et al. (2017) to investigate the behaviour. They computed SVCCA scores for layer-wise activations of a fully trained model. Their visualization via Spectral Embeddings (Belkin and Niyogi, 2003) is in Figure 8.5. In the figure, we can observe that the space of encoder representations, when evaluating all sentences from the test set, is clustered in line with language families. Kudugunta et al. (2019) note that the similarity goes even further than just language families (e.g. Slavic), to branches within families (e.g. South Slavic), branches within those branches (e.g. Western Subgroup), down to dialects (e.g. Serbo-Croatian).

**Observation 48:** *Encoder representations in universal multi-lingual MT models of sentences in different languages are clustered based on the linguistic similarity of language families.* (Kudugunta et al., 2019)

Following this analysis, they discuss, if the writing system (like Cyrillic, Roman, or Ge'ez) is also an important factor in the representation of languages. They visualised a subset of languages using different writing systems, which we reproduce here as Figure 8.6. We can observe languages from the same language family but using different writing scripts to appear closer to each other than to languages from a different language family sharing a common script. As Kudugunta et al. (2019) noted, a point worth special attention is the closeness between Serbian (sr) and Croatian (hr). Although these two languages are widely considered dialects of the same language, Serbian is written in Cyrillic, whereas Croatian is written in the Roman script. However, we see that they are mutually closest neighbours. Since they have no overlap in subword vocabulary, Kudugunta et al. (2019) conclude that the closeness stems purely from their distributional similarity.

Furthermore, when observing the representation on the final encoder layer, "enclast", we can see that the similarity grows further and Serbian and Croatian even become superimposed. We can generalize this and say that:

**Observation 49:** *Later encoder layers in multilingual systems reflect the writing script of languages less and related languages use similar representations, regardless the writing script.* (Kudugunta et al., 2019)

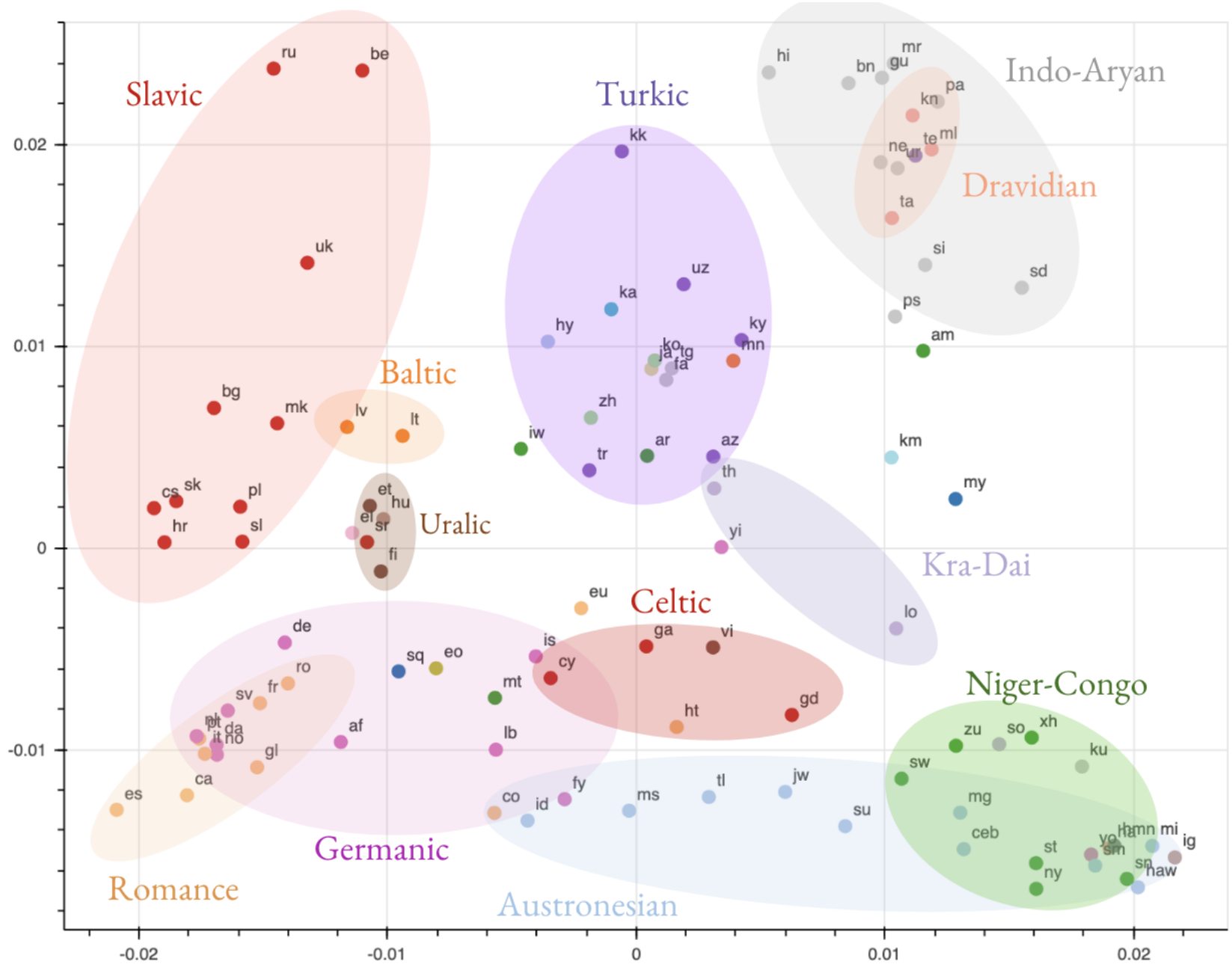

**Figure 8.5:** Visualization of encoder representations of 102 languages based on their SVCCA similarity. Cluster ellipses added manually to highlight language families. Figure reproduced from Kudugunta et al. (2019).

This observation brings a new question: how the representation evolves through the layers and whether the representation depends on the target language. Kudugunta et al. (2019) studied each layer of the multilingual model and computed pairwise SVCCA scores between language pairs across layers of the network. Figure 8.7, reproduced from Kudugunta et al. (2019), illustrates the behaviour. While the representations of sentences across languages differ (SVCCA scores around 0.5 to 0.6 in the left part of Figure 8.7 based on X→English models), there is a tendency to grow the representations more similar throughout the layers of the encoder. In the decoder part, English as the common language is represented very similarly across the translation pairs in the early layers (SVCCA scores close to 1.0) and diverges in later layers. Kudugunta et al. (2019) put this in relation to findings on translationese (Koppel and Ordan, 2011) where the translated text is predictive of the source language.

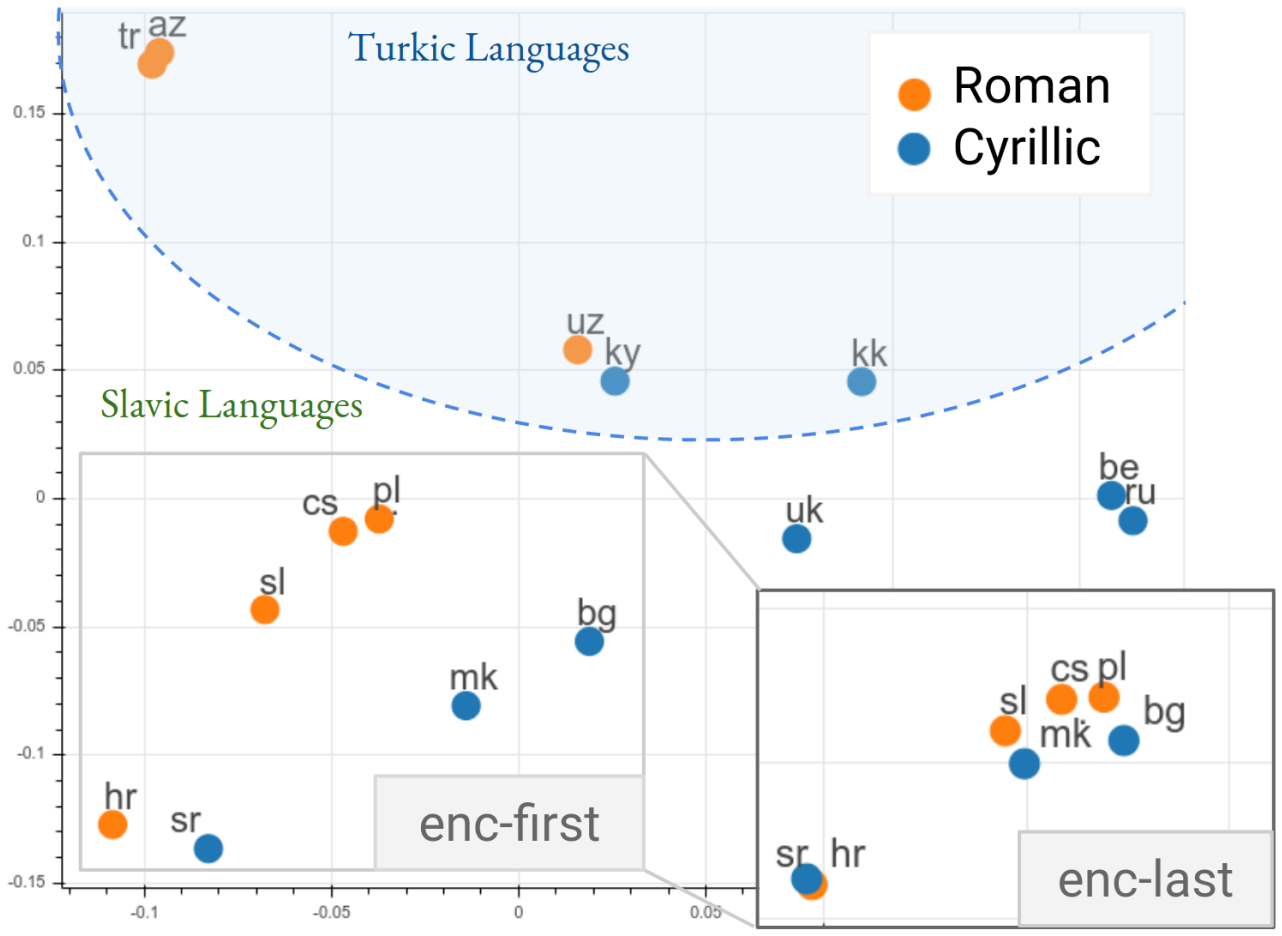

**Figure 8.6:** A detail of representations of the Turkic and Slavic languages written in Latin ("Roman") vs. Cyrillic script at the first and last layers of the encoder. Reproduced from Kudugunta et al. (2019).

For English→X, the trend is similar: the common English at the beginning starts with very similar representations. These diverge throughout the encoder layers and the dissimilarity gets fully pronounced in the multilingual decoder. This behaviour is also confirmed by Xu et al. (2020), who discovered that translation into target language happens already in the encoder.

**Observation 50:** *Representations of a source language learned by the encoder are dependent on the target language, and vice-versa* (Kudugunta et al., 2019)

## 8.4 Massively Multilingual Models

As the number of supported languages in a single multi-way model grows beyond a dozen or a couple of dozen languages, researchers start talking about **massively multilingual model**.

Aharoni et al. (2019) build on top of the work Johnson et al. (2017) and develop a massively multilingual system with 102 language pairs in English-centric setting, i.e. translation to and from English. They prepend target language identification token to each source sentence. They use the Transformer model (Vaswani et al., 2017) with a variation in hyperparameters.

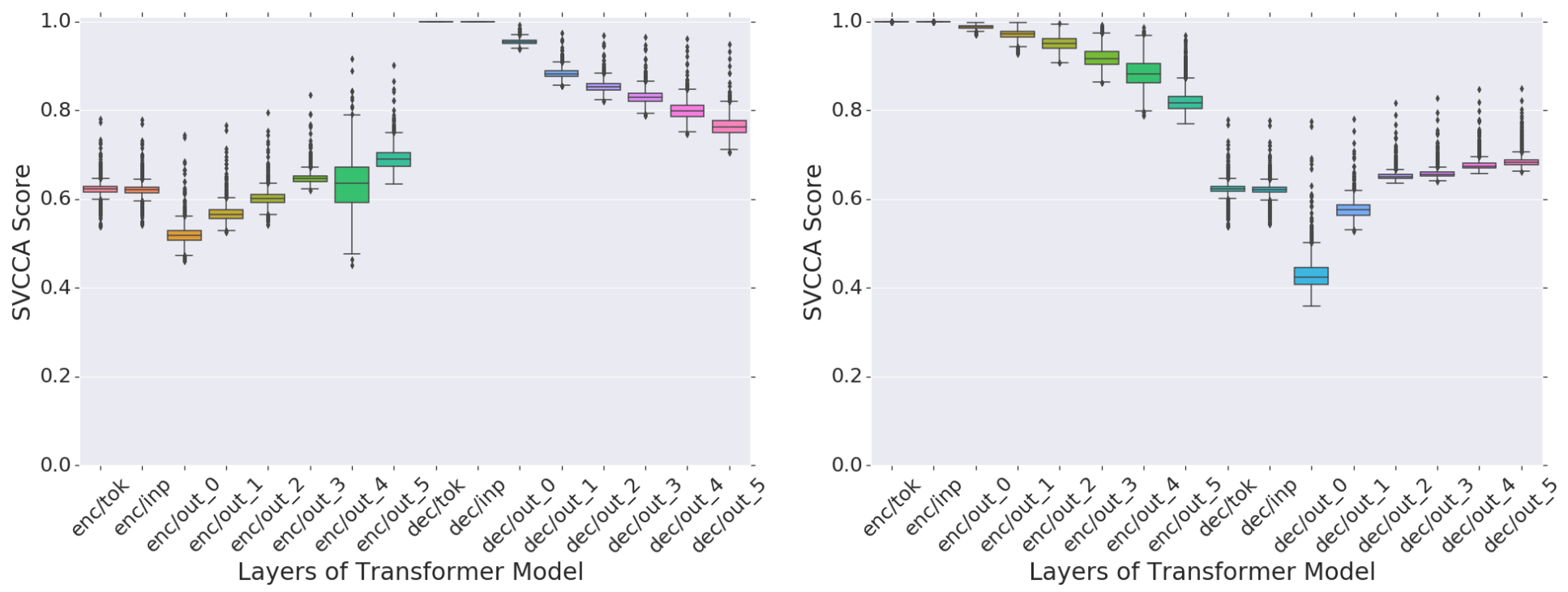

**Figure 8.7:** Development of pairwise SVCCA scores between language pairs across layers of X→English (left) and English→X multilingual models. We see that while the encoder on the left and the decoder on the right have dissimilar representations across languages, the English representations (the decoder on left and the encoder on the right) start very similar and diverge throughout the layers depending on the language X. Reproduced from Kudugunta et al. (2019).

For the first set of experiments, they use the base model with about 93M trainable parameters and stay in the low-resource constraints. They train the model on TED talks with 59 languages (58 languages to and from English). Their many-to-many model (X→English plus English→X, i.e. $2 \times 58$ language pairs) outperforms the many-to-one (X→English) model when translating into English. This is a surprising result as both the many-to-one and many-to-many models use the same data and an equal number of parameters and the evaluation is run for the many-to-one case. The many-to-many model is expected to perform worse because it needs to use some parameters to learn translation into the other languages.

We see Aharoni et al. (2019) as one of the rare exceptions where the authors carry out a detailed analysis of such surprising results. Indeed, they find a reason. Apparently, the win of the many-to-many model is not so much caused by its better performance but rather by a decreased performance of the many-to-one setup. As Aharoni et al. (2019) document and we reproduce here in Figure 8.8, the many-to-one setup overfits on the English side of the corpus because, in this corpus, the English sentences overlap between languages. As we see, the many-to-one training curve grows to BLEU scores of over 0.6 but the corresponding performance on the development set does not surpass BLEU of 0.3 and actually decreases from about 50k training steps. On the other hand, the many-to-many model does not suffer this overfitting and both training and development set performances keep growing. The conclusion that we can draw from this is that the additional target languages in the many-to-many serve as a regularizer and prevent the model from overfitting. It is this overfitting in many-to-one model that makes many-to-many setup win.

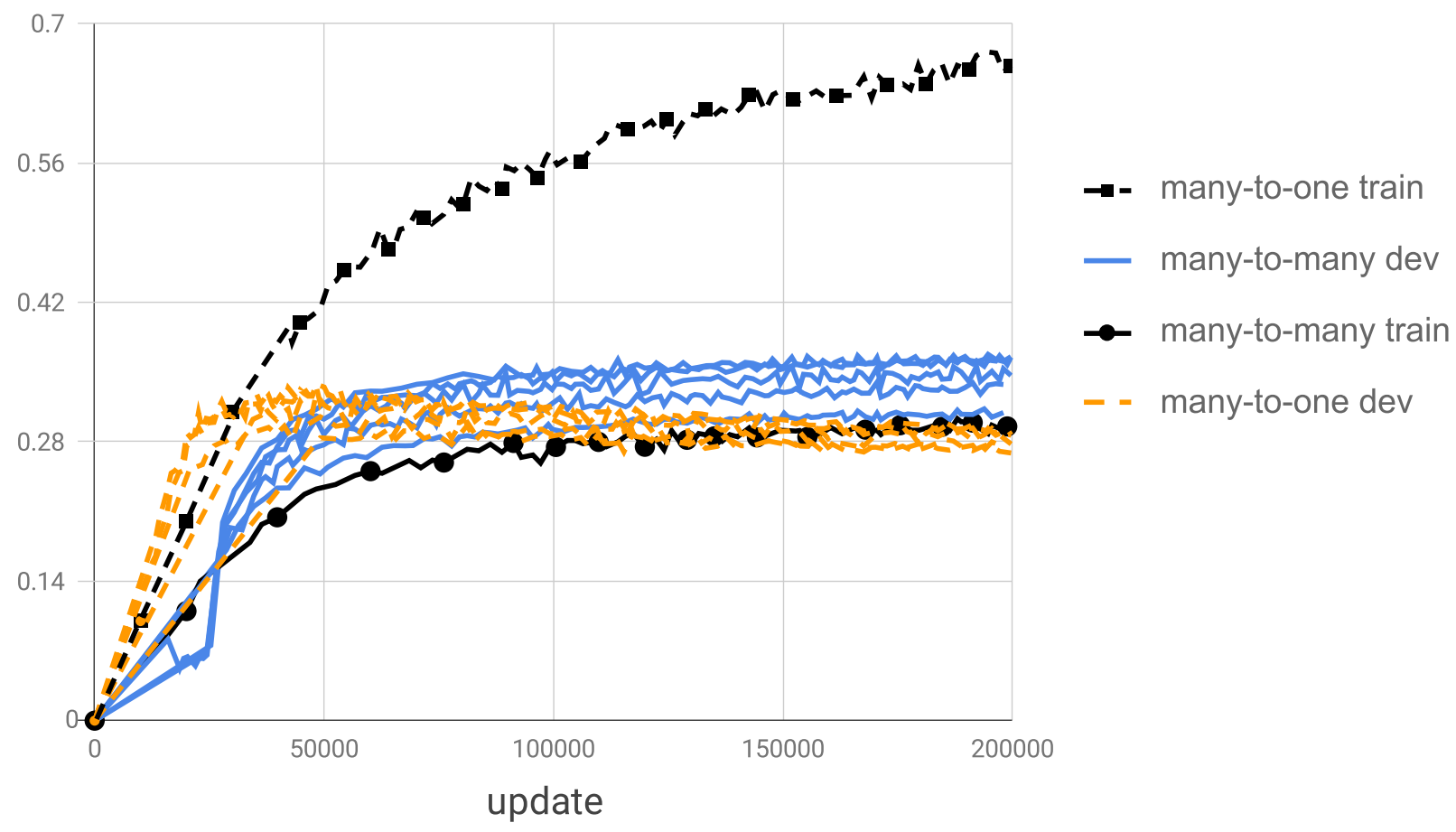

**Figure 8.8:** Development and training BLEU on {It,Ro,Nl,De,Ar}→En for the many-to-one and many-to-many models. The plot is adapted from Aharoni et al. (2019).

**Observation 51:** *Many-to-one multilingual models in small data setting can overfit on the target side due to too much repetition of the target language. A many-to-many model, regularized by the additional target languages, may thus seem to perform better on a many-to-one test set.* (Aharoni et al., 2019)

In the reverse direction, comparing many-to-many and one-to-many model, Aharoni et al. (2019) observe the opposite and expected behaviour. The many-to-many has a significantly worse performance than the one-to-many model.

For the second set of experiments, Aharoni et al. (2019) use an in-house dataset with 103 language pairs. The model is enlarged to about 474M parameters. One of the reasons of this increase is the need to double the vocabulary size to 64k subword unit types. With the standard 32k entries, most of the subwords would be actually individual letters, which would lead to an undesired length of input and output sequences.

The results are in Table 8.5. On average across 8 tested language pairs, the single-pair baselines are outperformed by the multilingual models but one-to-many and many-to-one models perform better than the many-to-many model.

| | Ar | Be | De | He | It | Nl | Ro | Tr | Avg. |
|---|---|---|---|---|---|---|---|---|---|
| baselines | 23.34 | 21.93 | 30.18 | 31.83 | **36.47** | 36.12 | 34.59 | 27.13 | 28.33 |
| many-to-one | **26.04** | **25.36** | 35.05 | **33.61** | 35.69 | **36.28** | 36.33 | **29.75** | **31.01** |
| many-to-many | 22.17 | 23.03 | **37.06** | 30.71 | 35.00 | 36.18 | **36.57** | 27.64 | 29.97 |
| baselines | 10.57 | 15.30 | 23.24 | 19.47 | 31.42 | 28.68 | 27.92 | 15.54 | 19.13 |
| one-to-many | **12.08** | **15.60** | **31.39** | **20.01** | **33.00** | **31.06** | **28.43** | **17.68** | **21.68** |
| many-to-many | 10.57 | 14.30 | 28.48 | 17.91 | 30.39 | 29.67 | 26.23 | 15.58 | 20.11 |

**Table 8.5:** Performance of English-centric multilingual systems on a dataset of 103 languages, tested on 8 languages translating into English (top) and out of English (bottom). Reproduced from Aharoni et al. (2019).

**Observation 52:** *Massively multilingual many-to-many models can outperform single pair models. However, they have a worse performance than many-to-one or one-to-many models.* (Aharoni et al., 2019)

We note that in papers on multilingual and especially massive models, the baselines are hardly ever subject to careful hyperparameter optimization, probably simply due to the overall training costs. Aharoni et al. (2019) are one of a few papers that explicitly mentions the limitation. As e.g. Sennrich and Zhang (2019) observed for low-resource languages single-pair models, hyperparameters can have a critical effect on the final performance. The comparison between single-pair and multilingual models can thus be somewhat unfair also for this reason.

Because massively multilingual systems have a worse performance than many-to-one or one-to-many, Aharoni et al. (2019) analysed the behaviour of multilingual NMT when decreasing the total number of languages. Their results are reproduced in Table 8.6. All these models are similarly-sized, i.e. about 474M parameters. The scores clearly show that a lower number of languages performs better. Aharoni et al. (2019) discuss that the behaviour is due to bottleneck in the number of available parameters per language. With more languages in the training data, only fewer parameters are available for each language pair.

**Observation 53:** *Adding many language pairs into a multilingual many-to-many model results in the model capacity bottleneck. The number of parameters is not sufficient to learn all the languages (despite knowledge sharing) and the fewer languages in the model, the better is the performance.* (Aharoni et al., 2019)

## 8.5 Massive Massively Multilingual Models

Scaling up the number of parameters in models could solve the problem where many-to-many models perform worse than many-to-one or one-to-many. However, scaling up model capacity has limits in the form of GPU memory, amount of computation needed and also available data size. Specialized model composition and parallelization techniques are used to train **massive models**. We discuss their application in NMT here.

| | Ar–En | En–Ar | Fr–En | En–Fr | Ru–En | En–Ru | Uk–En | En–Uk | Avg. |
|---|---|---|---|---|---|---|---|---|---|
| 5-to-5 | **23.87** | **12.42** | **38.99** | **37.30** | 29.07 | **24.86** | **26.17** | 16.48 | 26.14 |
| 25-to-25 | 23.43 | 11.77 | 38.87 | 36.79 | **29.36** | 23.24 | 25.81 | **17.17** | 25.80 |
| 50-to-50 | 23.70 | 11.65 | 37.81 | 35.83 | 29.22 | 21.95 | 26.02 | 15.32 | 25.18 |
| 75-to-75 | 22.23 | 10.69 | 37.97 | 34.35 | 28.55 | 20.70 | 25.89 | 14.59 | 24.37 |
| 103-to-103 | 21.16 | 10.25 | 35.91 | 34.42 | 27.25 | 19.90 | 24.53 | 13.89 | 23.41 |

**Table 8.6:** Performance while varying the number of languages involved. Results are from Aharoni et al. (2019).

Huang et al. (2019) address the problem with task-independent model parallelism. They develop a batch-splitting pipelining algorithm, which results in an almost linear speedup with a growing number of GPUs. Their algorithm divides the input mini-batch into smaller micro-batches, enabling different GPUs to work on different micro-batches simultaneously. Gradients are applied synchronously at the end.

Huang et al. (2019) demonstrated their algorithm on massively multilingual NMT. They trained five different models with increasing capacity. The notation T(L, H, A) refers to a Transformer-big model (Vaswani et al., 2017) with L layers in encoder and decoder, a dimension H of hidden feed-forward layer, and A attention heads. Their results across 103 languages in visualised in Figure 8.9. Later, the model was further expanded to the size of 50 billion parameters, as mentioned in the corresponding Google blog post.[2]

Observing Figure 8.9, we can notice various interesting patterns. When comparing high-resource languages (left side of the figure) with low-resource languages, we see that multilingual setting helps more the low-resource languages but only for huge models. For models below a billion parameters, high-resource languages are actually losing.

**Observation 54:** *Huge quality improvements (5-15 BLEU) are observed for low-resource languages in massive models, when compared against standard bilingual baselines.* (Huang et al., 2019)

Huang et al. (2019) also compared trade-off between the depth (number of layers L) and width (number of heads H) with the same 1.3B parameters in total. As we see, the deeper model T(24, 8192, 16) outperforms wider model T(12, 8192, 32) for almost all language pairs. This effect is most visible for low-resource languages.

[2] https://ai.googleblog.com/2019/10/exploring-massively-multilingual.html

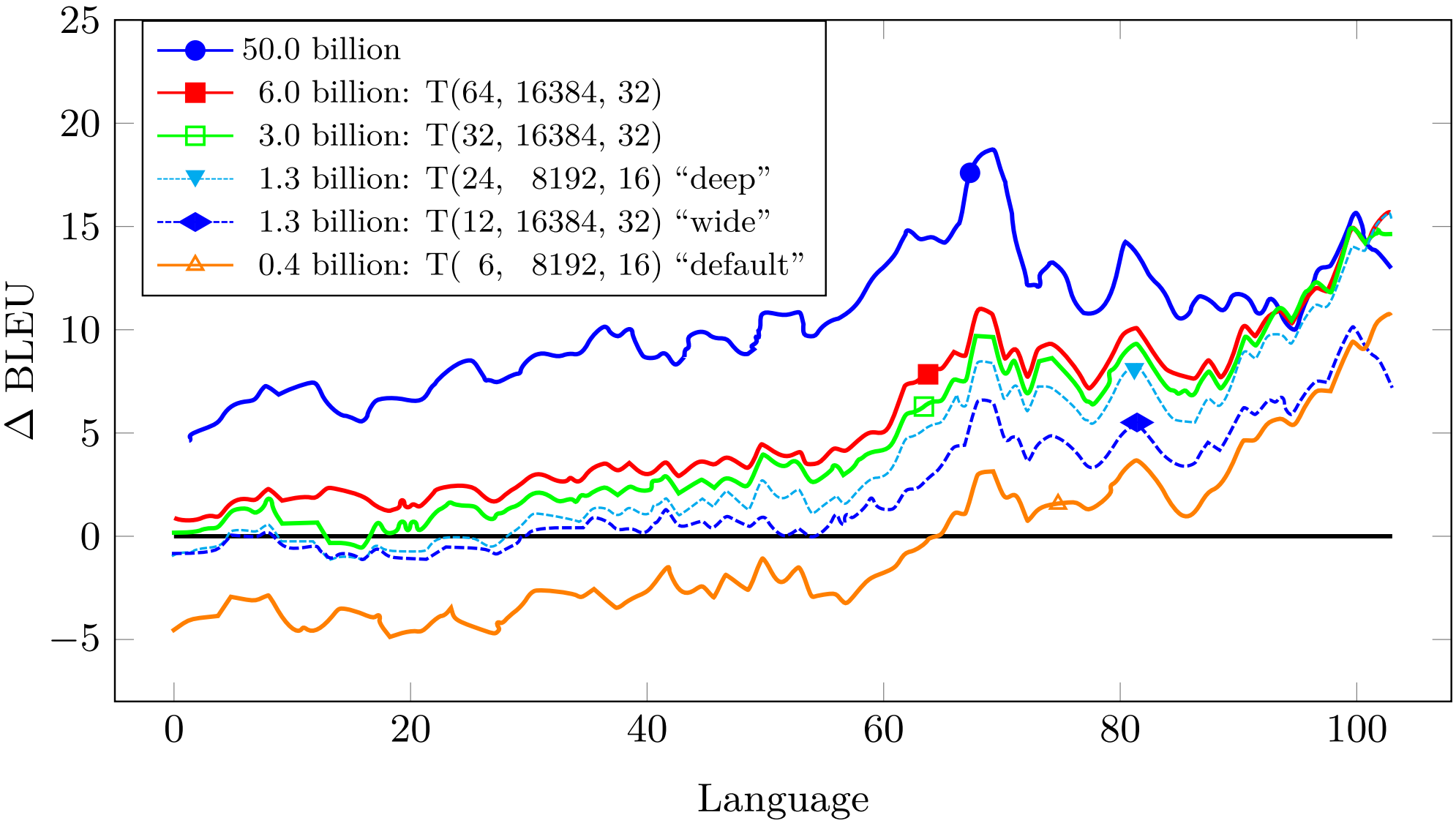

**Figure 8.9:** Translation quality of X→English across 102 source languages with increasing model capacity evaluated on 102 languages (on X-axis) from high (left) to low-resource (right) ones. The Y-axis represents the difference in BLEU compared to the pairwise model of 375M parameters for that language. The plot combines data from Aharoni et al. (2019) and the Google blog post referred here in Footnote 2.

**Observation 55:** *Esp. in low-resource setting, a massive deeper model (more layers) outperforms by huge margins a wider model (more attention heads) of a comparable size, suggesting that increasing model depth might be better for generalisation* (Huang et al., 2019)

Substantially increasing model size starts to help even high-resource languages to outperform single-pair (bilingual) baselines.

**Observation 56:** *Increasing the capacity of a many-to-many multilingual model from 400M to 1.3B parameters improves performance across all languages. Scaling up the model from 1.3B parameters to 6B parameters shows further improvement, especially for high-resource languages.* (Huang et al., 2019)

As discussed by Arivazhagan et al. (2019), the gains of these massively multilingual models differ based on the translation direction. Table 8.7 demonstrates this on the model with 103 languages at three model sizes.

| | Any→English | | | English→Any | | |
|---|---|---|---|---|---|---|
| | High 25 | Med 52 | Low 25 | High 25 | Med 52 | Low 25 |
| Bilingual | **37.61** | 31.41 | 21.63 | 29.34 | 17.50 | 11.72 |
| 400M | 33.85 | 30.25 | 26.96 | 28.03 | 16.91 | **12.75** |
| 1.3B Wide | 37.13 | 33.21 | 27.75 | 28.36 | 16.66 | 11.14 |
| 1.3B Deep | 37.47 | **34.63** | **31.21** | **29.46** | **17.67** | 12.52 |

**Table 8.7:** Average BLEU of many-to-one and one-to-many massively multilingual models across groups of languages based on their training data size. BLEU scores are comparable only within each column. Reproduced from Arivazhagan et al. (2019)

Each column of Table 8.7 reports the average BLEU across 25 high-resource, 52 mid-resource and 25 low-resource languages, into and from English, respectively. Because the scores are based on different languages and averaged across different sets of languages, they are comparable only within each column. Actually, even the differences observed between lines of a column are not comparable to differences in other columns, but the order of the approaches is informative. While in Any→English low-resource, the 1.3 billion parameter deep model clearly wins by a very large margin (BLEU increased from 21.63 for the baseline to 31.21), no such gain is observed in the other direction.

We agree with Johnson et al. (2017) and reported also here as Observation 53 on page 142 that a universal model of a fixed size has insufficient capacity for higher-resource languages as the number of languages is growing. On the other hand, we would like to know the performance of single-pair systems trained with 1.3B, 3.0B and 6.0B parameters.

As Melvin Johnson mentions in his talk[3] on Google's multilingual system, it can be considered unfair to attempt to fit more languages in the same size of a model. While we admire the subsequent technological achievement reaching massive models (and feel a little envy over the available resources), we find the presented comparison unfair in the other direction: high-resource language pairs are likely offer more information in their data than what did fit into their pairwise baseline model. Once the model is massively scaled up ("to accommodate all the small languages without eating up space of the big ones"), it is very likely that the small languages *do not make use* of all the capacity provided to them. The massive model thus has *free* capacity for the big languages.

We fear that the gains in high-resource languages in the massive models are primarily thanks to this extra model capacity and not due to any knowledge transfer from the low-resource ones.

[3] `https://www.youtube.com/watch?v=nR74lBO5M3s`

Indeed as one can find, the supplementary material of Huang et al. (2019)[4] mentions: "For most bilingual experiments, we used a larger version of Transformer Big model containing 375M parameters, and a shared source-target SPM [SentencePiece] vocabulary with 32k tokens."

A fair baseline, which we would be asking for, is to train single-pair models esp. for high-resource languages also at this massive scale.[5] We anticipate even better performance than that of the massive multilingual model. More as a warning than a completed observation, we formulate:

**Observation 57:** *Performance of models can not be assessed without considering their number of parameters. If a technique requires an increase in model size, increasing model size should be tested with the baseline approach, too, to ensure that the gain can be attributed to the proposed technique and not* solely *to the increased model size.*

Finally, the Google team are not alone in pushing their models to hardware limits. The same approach is taken by Facebook/Meta (Fan et al., 2021) and Microsoft (Kim et al., 2021).

[4] `https://papers.nips.cc/paper/2019/hash/093f65e080a295f8076b1c5722a46aa2-Abstract.html`

[5] We are fully aware of the cost of this massive training. We are not suggesting that *all* the 2×102 language pairs should be trained pairwise. A handful of examples or just one or two of the high-resource language pairs would be very likely sufficient to support or reject our hypothesis. We also acknowledge that the tested pairwise baselines with 375M parameters each add up to about 80 billion parameters in total. A 6B replacement thus undeniably *is* a huge saving. Our point is different: the *source* of the gains may the bigger model size, not cross-lingual knowledge sharing.

# 9
# Practical Aspects

In this chapter, we describe practical aspects of multilingual MT systems.

We start with the deployment issues and lessons learned from two research projects that integrate multi-target NMT in a practical application: KIT Lecture Translator and ELITR.

Next, we move to briefly discuss our experience with the necessary hardware equipment to train and run these models.

Being researchers ourselves, we find it rather inspiring and important to also consider and discuss these practicalities. On the one hand, by deploying a system, we can directly learn from its users and prioritize problematic elements of the automatic processing, on the other hand, being aware of the hardware conditions and infrastructure helps to understand scaling, cooperation, integration and other issues that are inevitable in any collaborative research.

We realize that this experience and knowledge is far from unique: many companies are regularly handling these processes at much larger scales than what we experienced ourselves. Yet, some companies (rightfully) treat this knowledge as part of their assets and are unlikely to share it openly. For that reason, we hope our observations will be useful.

## 9.1 KIT Lecture Translator

At the Karlsruhe Institute of Technology in Germany, there are many foreign students attending lectures in German although their knowledge of German is limited. Therefore, the institute offers an automatic simultaneous translation of lectures into the students' native languages since 2012 (Cho et al., 2013). The lecturer speaks into the microphone and to the audience in German. The speech is simultaneously transcribed by hybrid HMM-DNN Janus ASR model (and newer architectures in later years), then translated by either bilingual, or multi-target MT (Pham et al., 2018) with 24 target languages (see Figure 9.1). Finally, the live translation of what is being said at the lecture is presented to the students on a web page on their personal devices.

Müller et al. (2016) describe an interesting user study. They ran the system for all lectures during the semester and collected feedback from students by a questionnaire. It is therefore valuable evidence from real-life sessions from real users, and not approximate, laboratory test. The authors conclude that the system indeed helps students to understand lectures in German better.

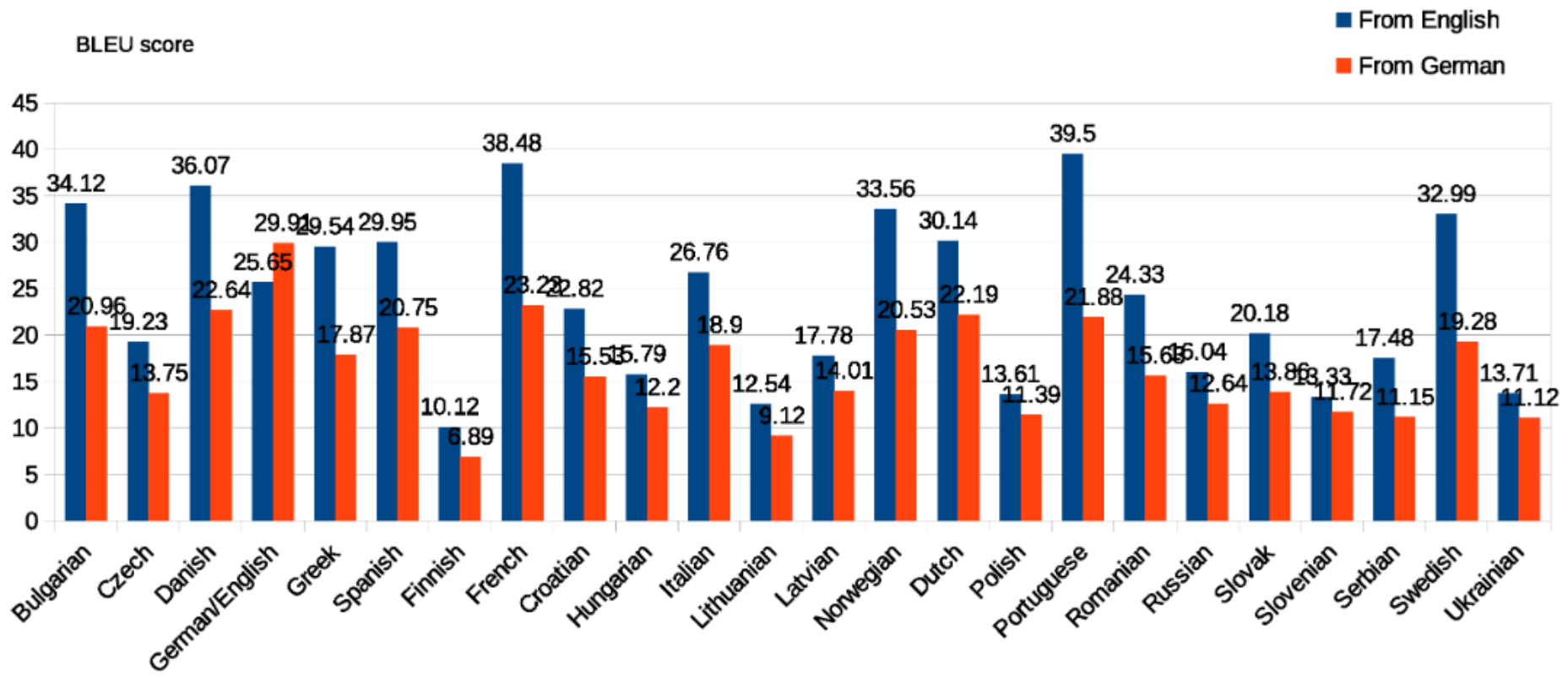

**Figure 9.1:** MT Performance when translating from English or German to 24 European languages. Figure reprinted from Dessloch et al. (2018)

.

## 9.2 ELITR: European Live Translator

**ELITR** (European Live Translator, Bojar et al., 2020, 2021a)[1] is an EU research project that has been running from January 2019 to March 2022. A consortium of research partners Charles University, Karlsruhe Institute of Technology, University of Edinburgh, industrial partners Alfaview and PerVoice, and the affiliated user partner Supreme Audit Office of the Czech Republic approach the goal (beside others) to provide simultaneous speech translation to a congress of supreme audit institutions EUROSAI. They focused on 6 source spoken languages and 42 target text languages. See Bojar et al. (2020) for a brief project description and Bojar et al. (2021b) for details of the deployment of the system at the congress.

In the following sections, we describe several interesting lessons about multi-lingual MT, that the investigators learned during the project.

### 9.2.1 Many Languages, Many Problems

In the second part of this book, we were presenting many research works that train and evaluate multi-lingual systems with up to a hundred target languages. ELITR's goal meets that scale, producing live translations in 42 target languages.

The reported evaluations are, however, only automatic. A rare exception of a multilingual research which does include manual evaluation is that of Fan et al. (2021).

[1] `http://elitr.eu/`

The lack of manual evaluation of highly multilingual systems is natural because the research team does not have the skills to verify the output quality across the many languages. The evaluation can be outsourced to a language service provider company, or crowd-sourced as often done in shared tasks like WMT, but this is costly and time consuming for the regular research. Even if the funding was available, manual evaluation of translation quality is not an easy task if the results should be reproducible and comparable. In multiple situations, we faced the issue of unreliable informants: they were not careful enough to spot the errors or simply hesitant to report them. Essentially, only research groups with long-term localization and translation experience are in the position to be able to manually evaluate the wide range of languages.

**Observation 58:** *Multilingual systems strengthen our reliance on automatic evaluation methods. Manual evaluation is much more difficult due to the wide language expertise needed.*

Relying only on researchers' own language skills and automatic metrics, the following problems can be easily neglected and ran into only at deployment:

**Small, Language-Specific Problems.** Many small problems that can appear anywhere in MT development are easy to solve in bilingual MT, especially if the MT engineer who develops the MT system speaks the source and target languages at least at a basic level. The easiest option is when the engineer can spot the problems right away. In multi-lingual MT, it is likely that a single engineer can not speak dozens of target languages. Therefore, he or she has to rely on the feedback from experts or speakers of the language. With more languages, he or she has to ask many persons, and that may be complicated and time-consuming.

The small problems that an engineer who is not a speaker of the language can not notice may be missing or wrongly formatted diacritics (e.g. in Czech, Slovak, Polish), effects of noisy training data (e.g. mixing lowercase "l" and uppercase "I"; in some fonts, they are identical, so it may not have mattered in the original data, but it is unexpected in MT), mixed or wrong languages (e.g. swapping Russian and Ukrainian might make a serious political problem), wrong script (e.g. Kazakh has changed the official script three times in the past century, from Arabic through Cyrillic to Latin, so the Kazakh MT training data may be noisy, and it is important to check the result), wrong casing (e.g. in German, all nouns have to be capitalized), postprocessing of subword pairs (e.g. too many short words ending with "@@" may be a bug, a relic of subword processing), the writing direction, especially for right-to-left languages (e.g. Hebrew, Arabic), readable typesetting of non-Latin scripts (e.g. Chinese, all scripts of Japanese, Korean, Georgian, Hindi, etc.), not forgetting any language (e.g., Montenegrin language is identical to Serbian, except two subtle differences that were introduced by government and not adopted by the population. So, there is little technical reason to handle it separately, but the Montenegrins prefer a labeled option with "Montenegrin", even if it would be a copy of Serbian.), the variant of Norwegian (there are two official variants of Norwegian. They differ by spelling and some grammar rules: Bokmål and Nynorsk) and many others.

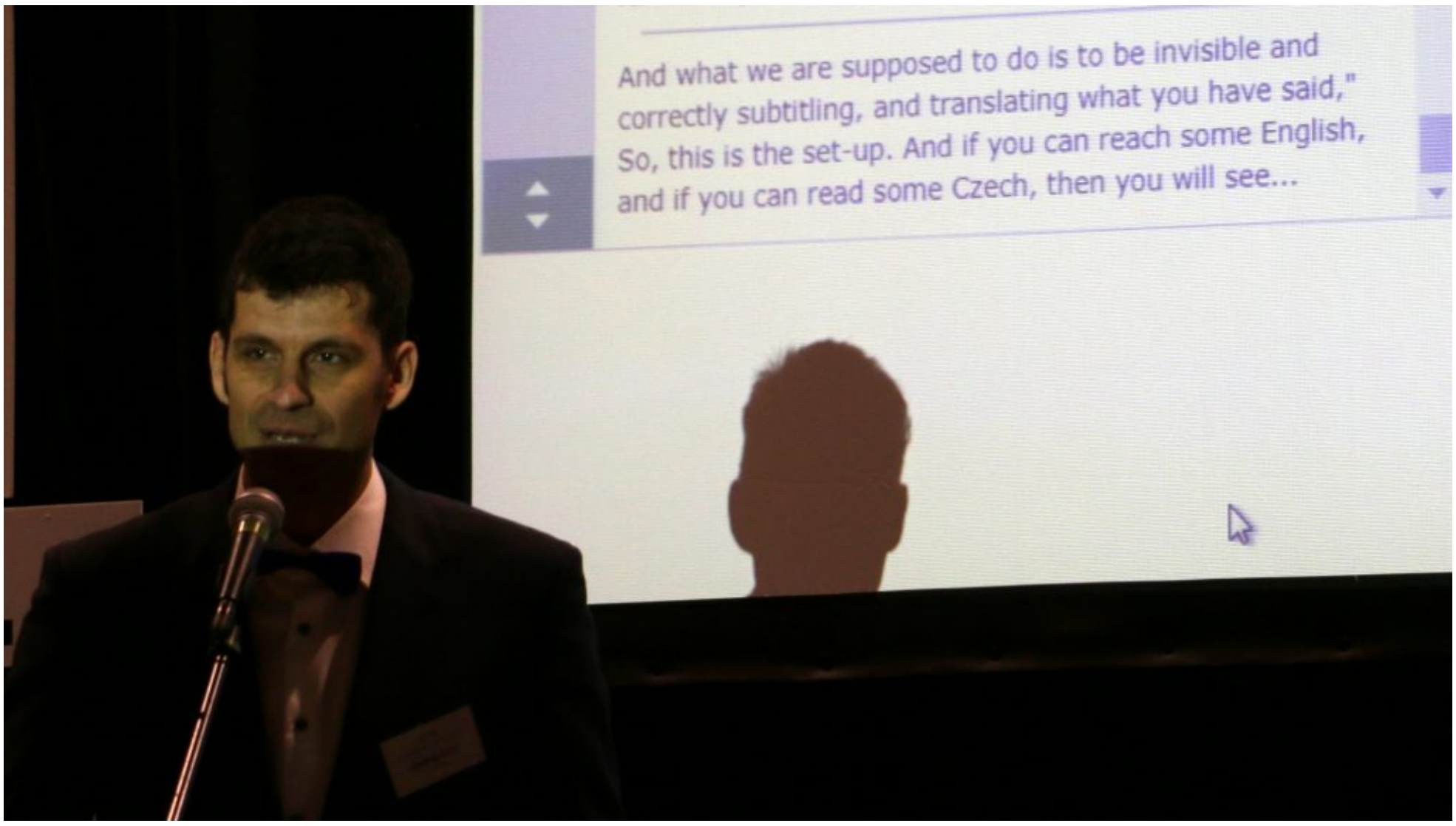

**Figure 9.2:** A picture from the first testing event of ELITR multilingual live subtitling. The speaker is speaking slowly and clearly in English. There is a live ASR transcript projected behind him, and translations into 4 languages, out of picture scope.

Every separate problem might be simple but many of them at once can be serious. Moreover, fixing one problem can influence the others. Many drops make a shower.

**Social Implications of Errors** The very first ELITR dry-run test with real audience was at an international event for high school students.[2] There was a main stage, on which a group of Romanian students was presenting their mock company. The automatic translation in 5 languages was projected on the screen behind their backs, the same way as illustrated in Figure 9.2.

At one point, the students in the audience started laughing, pointing at the screen, taking photos and sharing them on social networks. See the photo in Figure 9.3. The presenting students were confused, they were not aware of what is happening.

At that point, the output quality was very poor, due to various reasons starting with bad sound quality, specific accent and erroneous speech recognition. The MT neural network produced random nonsense words. At one point, an inappropriate word appeared in the outputs, which fellow students took a picture of and circulated it on a social network. Luckily, an apology from the ELITR team was sufficient for the Romanian students, but more serious mishaps are possible.

[2] See the blog post about the event at `https://elitr.eu/clearest-voice-at-trade-fair/`.

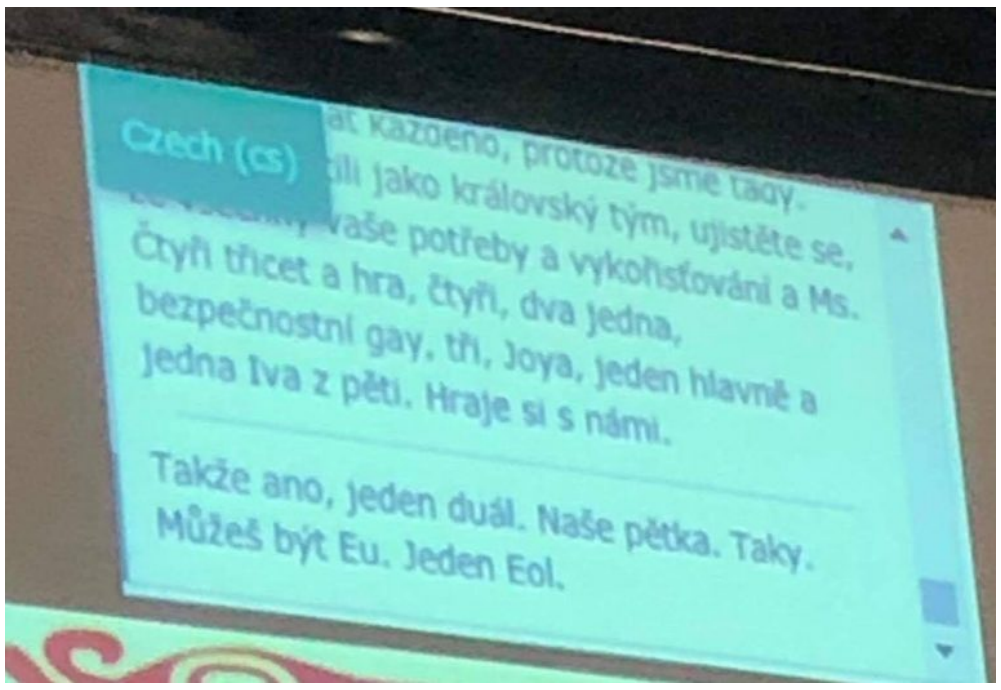

**Figure 9.3:** A detail of the Czech translation screen of Romanians speaking English that the teenage Czech students in audience found amusing. The "translation" is totally wrong and unrelated to the source, looking like a nonsense text randomly generated by a Czech bigram language model. The keyword on which the teenagers reacted is probably the 3-letter g-word at the 5th row from top.

The lesson learned at the technical level is that neural networks are very creative. On the one hand side, this can be useful. NMT can for instance create existing words that could not be found in a fixed vocabulary derived from some fixed training data. On the other hand, it can create inappropriate words that can be hardly anticipated and detected by rules.

At another test occasion, this time without audience, ELITR saw a personal name totally mis-recognized. Instead of "Mister, …", the speech recognition output was "As for my girl, she is the floor of yours." This sentence does not contain any bad words and yet it is rather offensive. Listening to the original, the recognized sentence somewhat resembles the original utterance with the German name. However, translated to any target language, no user can guess where this offence is coming from.

The challenge of **profanity filtering** has to be faced for important meetings.

One solution is to employ a human operator who filters out the translations with inappropriate words or sentences. Ideally, one would however need one operator for each target language because the inappropriate word can be created by NMT only in some of the output languages. In multi-target MT, the coordination of human operators may be complicated.

The second option to avoid negative effects of inappropriate or offensive words in translation, may be a disclaimer. There can be a notification for the users that the automatic translation can be wrong, that they are neither reviewed nor authorised by any human, and that inappropriate or offensive words can appear as translation errors. However, the more target languages, the more complicated the communication with users who speak only their native language is. Messages from the operators have to be prepared in advance, and translated by many professional translators. Changes to the set of supported messages in the last minutes are impossible, if the translators are not paid to be available.

### 9.2.2 Distributed Development and Deployment

The ELITR speech translation pipeline (Franceschini et al., 2020) is distributed between the consortium partners. There is a central server called Mediator. The research partners in consortium connect their workers offering a service, e.g. ASR, MT, or an intermediate punctuation and segmentation tool.

A client requests Mediator to connect a cascade of workers. For example, Czech audio → Czech ASR → Czech punctuation → Czech-English MT → multi-target MT English to 42 languages → 42-times presentation on the web, under different language labels.

During ELITR events, the Mediator was operated by PerVoice, Italy, and hosted on their rented servers in West Germany. Charles University (CUNI), Czech Republic, provides Czech ASR, punctuation, and Czech-English MT. CUNI runs them on its own computer cluster, which is located in two places in two quarters of Prague. The 42-target MT was run by University of Edinburgh, United Kingdom, and located on their cluster in Edinburgh. Karlsruhe Institute of Technology, Germany, offers English and German ASR and punctuators, and other languages.

The data in the processing pipeline are shipped across Europe there and back as numerous data streams. This distributed development has both advantages and disadvantages.

**Advantages:**

- Easy access. The workers provided by different labs are accessible any time by a client, if they are connected to the mediator. The members of different teams do not need to have access to the other clusters just for using the service.
- Easy deployment. The workers can depend on their specific software and hardware environment, and they do not have to be unified across workers. Different teams have different habits and rules for using their clusters. They do not need to unify.
- Distributed responsibility. There are so many different workers with so many specifics that one person would not easily be expert on all of them. The responsibility is distributed to multiple developers.
- Flexibility. It is easy to replace one version of model with another. This is beneficial for immediate deployment of better versions of systems as well as for various bug fixes, e.g. inserting a trivial script between workers to normalize them.

- Protection of intellectual property. If some of the partner's services can not be released under any license, this distributed setup still allows to include them in complex pipelines.

**Disadvantages:**

- Network delay. Some delay in processing due to distances and other network traffic is inevitable. Luckily, this delay is low compared to ASR and MT processing time.
- Complicated debugging. The client operator does not have access to worker logs on other clusters. When debugging, the worker operator must be contacted. If it is unclear which worker is problematic then all the operators may be needed. Synchronizing more people is always more complicated than handling the issues oneself.
- Coordination. Multiple developers have to be asked to run the workers at a specific time and in specific versions, to allow for a correct composition of the pipeline. If the workers crash then the client operator has to ask again and wait.
- Complicated pipeline. It is not easy to understand the specifics of all the workers and all the wrapping scripts. Some development issues may be caused by misconfiguration.
- Complicated updates. The important changes in pipeline codes or in workers have to be explained and reported to others.
- Security. The worker logs with the inputs and outputs may stay on multiple clusters, and it might be more complicated to keep them confidential than in one controlled environment. This is important for strictly confidential meetings that may be automatically translated by ELITR.
- The more locations, the more unexpected problems. We mentioned that ELITR distributed infrastructure relies on 4 partners running their services on clusters at 5 locations. At each of these locations, unexpected problems such as power, disk or network outage may happen at random times independently of others. The more places, the lower chance that everything works well.

In our opinion, the most unbelievable and most unexpected event that happened and complicated an important demonstration of the ELITR project was a suspected bomb from World War II (1939-1945) that was found close to KIT premises with the servers room.[3] It caused evacuation that affected 7 000 people in the area (see Figure 9.4) and prevented KIT IT administrators from entering the building where their cluster was running. Because of that, they were unable to make a simple manual operation on their computers for several days, and therefore the KIT ASR service was not running. Although it was not during the demonstration event but shortly before it, the ASRs were necessary for testing and debugging the other components in the pipeline.

[3] https://bnn.de/karlsruhe/von-moeglicher-bombenentschaerfung-koennten-am-donnerstag-7-000-karlsruher-betroffen-sein

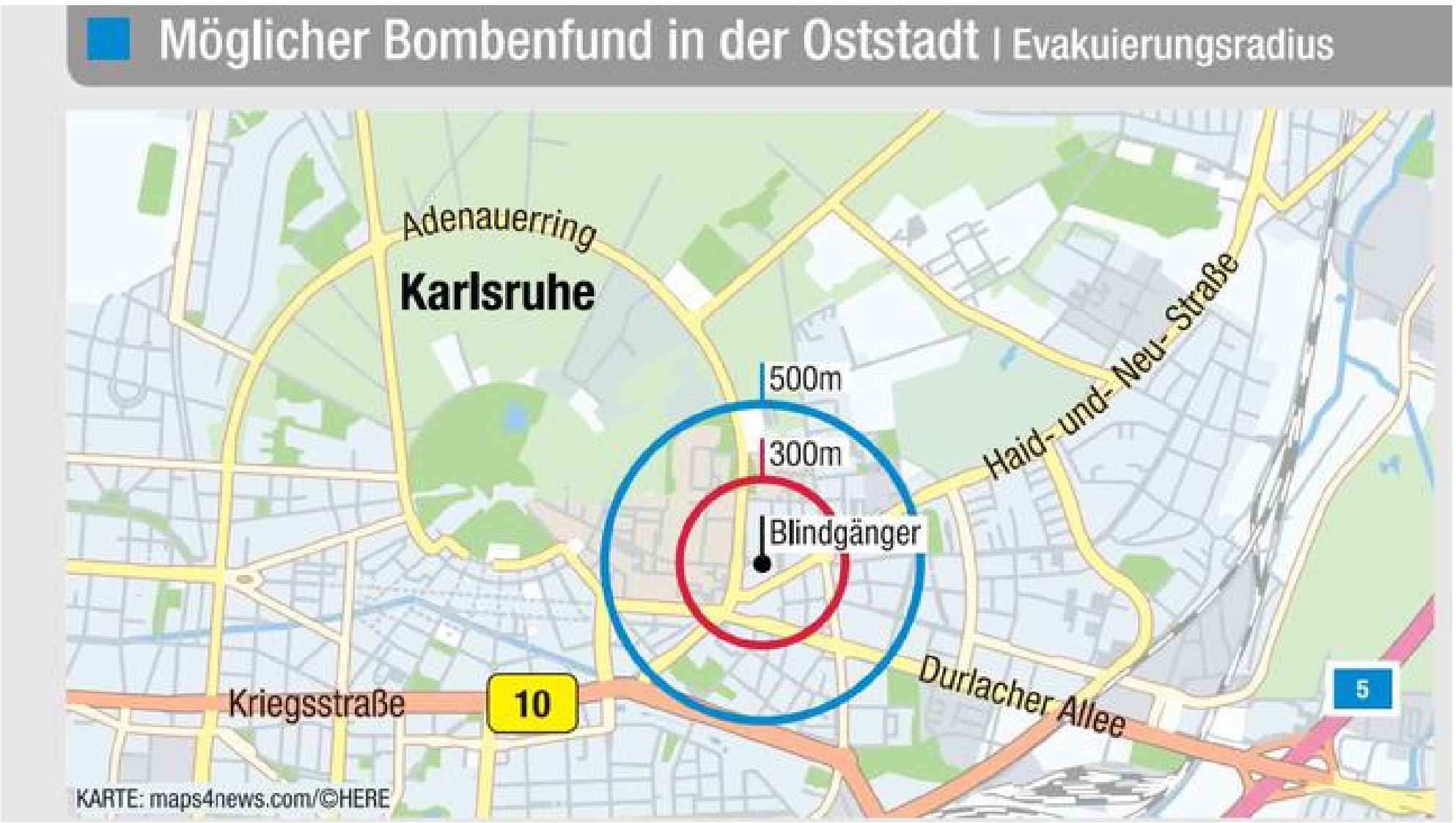

**Figure 9.4:** A map of evacuation radius because of a suspected bomb from World War II (1939-1945) in Karlsruhe, Germany. Unfortunately, the KIT computer cluster is within this area, which eventually complicated an important ELITR demonstration event. The map is reproduced from German local news server `bnn.de`.

In the end, it was found out that the object was not a bomb but an old pipe. Despite of that, this event was a large complication but it also gave us an important lesson.

## 9.3 Computer Clusters

Standard NMT today requires training and running the NMT models on one or more GPUs. Administrating GPUs and the underlying hardware is a special technical skill rather distant from what NMT researchers may primarily like to do. More so if there are many GPUs and many machines connected in a computer cluster. And much more so, if many researchers use the same computer cluster simultaneously, persistently and for very specific use cases. What the use cases have in common is that every now and then somebody needs to run many experiments at once, or at least in a short time, and the number of required GPUs is often lower than available. Then, cluster management becomes a very valuable craft.

In this section, we describe experience with the computer cluster at ÚFAL,[4] the home institute of us authors, and contrast it with stories and experience of our colleagues from their internships and stays at other institutes and companies.

In our opinion, a strong technical background and good cluster management are necessary conditions for high-quality NMT research. We feel that this prerequisite is worth mentioning, despite being neglected as obvious in most papers. We have a good experience with the situation when the technical administration and cluster management is delegated to an experienced technician (or a team of technicians) who have it as a long-term full-time job.

There are many nation-wide cluster solutions, e.g. the MetaCentrum[5] for academic institutions of the Czech Republic or Finnish Computing Competence Infrastructure[6] for Finland and also higher-tier clusters, e.g. PRACE[7] as one of the European research infrastructures. ÚFAL cluster is unique in that it provides similar services to these large infrastructures but is essentially run in-house. Aside from somewhat better availability of free nodes and little or no restrictions e.g. on computing job duration, the added benefit is the flexibility if a specific solution for distributed computing is needed.

As an alternative, paid cloud services such as Amazon Elastic Compute Cloud (EC2)[8] are possible. While they are interesting for short and somewhat unplanned experiment opportunities, they are generally more expensive and possibly difficult to scale beyond individual compute nodes. Furthermore, with the increasing overall uptake of AI technologies, it does happen that even huge provides like Amazon run out of their GPU capacity.

### 9.3.1 Computer Clusters from the Users' Perspective

One of the most salient impressions of our fellow NLP researchers who returned from their stays and internships at the institutes and companies abroad is the practical experience with the techniques used for high-performance computing, the layout of the computer cluster at the visited institution. Those who return back report that the cluster operated by ÚFAL is the best one they have ever seen. From their subjective stories,[9] we set the following analogy of clusters to historical epochs.

- Prehistory. Every researcher has one desktop computer under the table. It has one GPU and there is no system for sharing data or GPU resources between the computers in a lab. Everyone takes care of his or her computer. Nobody can break anything to others but everybody spends time on administrating and everybody can use at most one GPU.

[4] Institute of Formal and Applied Linguistics, Faculty of Mathematics and Physics, Charles University, Prague, Czech Republic. `https://ufal.mff.cuni.cz/`

[5] `https://metavo.metacentrum.cz/`

[6] `https://www2.helsinki.fi/en/infrastructures/fcci`

[7] `https://prace-ri.eu/`

[8] `https://aws.amazon.com/ec2`

[9] We have not heard back from those who have not returned.

- Classical Era. There is no in-house computer cluster. The cluster is rented from a large cloud provider. Everybody is responsible for administrating his or her own sub-part of the rented cluster. There is no global system for sharing the data and GPU resources, every researcher thus has to spend time on it. After some initialization steps, everybody has an access to a larger number of GPUs at once, and after some termination steps, the GPUs can be released. Data management is typically the major issue, researchers have to juggle with their training data and trained models in order to cut data storage costs.
- The Middle Ages. The lab has an in-house cluster of several GPU machines, but there is very limited IT support, very limited up-to-date global installations, and no centralized scheduling. Allocating a GPU is very simple, but fragile: log in to a computer, inspect the GPU load, pick one that seems empty, and start your process. It may later show up that the GPU is occupied by a process of someone else and the process was actually initializing when you were inspecting the node. So a conflict arises and one of you must leave for another GPU. If it is on a different machine and there is no shared file system, he or she may need to synchronize the data first.
  The technicians moreover terminate any machine every now and then without prior warning. In a later stage, there is a shared spreadsheet where the researchers notify others which GPUs they allocate but they can simply forget and their processes can terminate at night and no other process replaces them automatically during the same night.
  A particular variant is the situation when the cluster is administered by the team of researchers, e.g. on a rotation duty schedule every few months. While technically, there is always a person responsible for cluster issues and support, their experience is limited and any stability of long-term solutions is hard to achieve.
- Early Modern Era. There is a large in-house cluster with decent up-to-date software, disk sharing failing only occasionally, some global and some up-to-date installations, and decent IT support. However, there is no global task scheduler that prevents the conflicts and puts processes into operation as soon as the previous ones terminate. Or, there once was an attempt to introduce it, but it was not done properly, it caused more problems than benefits and it was not adopted. Alternatively, there is a task scheduler but not everybody uses it, or it can be used together with spontaneous allocations, leading to risks of a conflict.
- Modern Era. The ÚFAL cluster. Automatic scheduling, disk sharing, IT support are always available and very helpful, up-to-date software and global installations of frequently used libraries, flexible, transparent, large, etc. See more in the following section.

### 9.3.2 Conditions and Features of a Good Cluster

In this section, we describe the necessary conditions for running a cluster at the level of a research department or a school and features that make the cluster good for its users.

- Power supply and costs. GPU computing requires lots of electric power. In cluster planning, not only its running costs have to be taken into account but also the actual availability of sufficient electricity supply. The latter can be a problem in densely populated parts of a city.
- Cooling. Intensive computing generates heat, and that requires cooling facility in the servers room. This is another power consumption and object of maintenance.[10]
- Up-to-date hardware. New types of GPUs are constantly being developed, they must be purchased and installed every now and then, to keep track with current research trends.
- Up-to-date software. The software on the machines must be up-to-date and compatible with the hardware and recent NMT frameworks. In an ideal world, it should be installed globally on all machines for all cluster users, instead of everyone installing it individually, wasting their time and disk space. In practice, this is often complicated because different researchers need different tools in older or newer development versions, which typically need particular versions of the GPU libraries.
- Shared disks. Ideally, the disks should be shared among all cluster computers automatically. The user may then very flexibly change the computers without any need of data synchronization. The disks should also be backed-up every now and then (or run in a replicated mode where data loss is prevented), globally by the administrators, not by the users individually.
- Task scheduling. When many people use one computer cluster, they may very quickly run more tasks than available GPUs. One task should be run on one GPU at a time, running more of them at once causes out-of-memory conflicts and significantly slows down both processes. We find it critical that there is a robust, fully automatic task scheduler that:
  - Makes sure that no GPU is used for more than one task at a time.
  - Puts the tasks into a queue.
  - As soon as one task finishes, it starts another one from the queue.

[10] There was an interesting story in Czech news regarding power cost and heating by intensive computing: The management of the Czech Parliament noticed large bill for electricity in an official subsidized apartment of one newly elected Member of Parliament. The investigation showed that the MP was mining cryptocurrency on his 3-GPU computer because he was cold. He used it as a temporary heating before his regular heating device was installed. He clarified the energy costs and profit and claimed that the profit was slightly larger than the cost (in late 2017). He apologized, paid the bill and donated more than the profit to charity. Source: `https://zpravy.aktualne.cz/domaci/pirat-vymazal-v-byte-jsem-vytezil-1442-korun-asistentku-mi-d/r~1d2ff42c602a11e89297ac1f6b220ee8/`.

- Is flexible: Allows to set task priority, require some minimal GPU memory size, a larger number of GPUs, etc.
  - Is transparent so that everybody sees the tasks of the others, to estimate when they end, or to suggest to a particular colleague to be more efficient and cooperative, e.g. not flooding the whole cluster with thousands of tasks for weeks when others also have their deadlines. The tasks and processes are owned by real users, not some common account or even the root.
  - Logs the traces of failed jobs for inspection and debugging.
  - Allows direct terminal access with enabled GPU for debugging.
- CPU cluster. Although GPUs are the critical equipment for NMT training, some data processing and other auxiliary tasks can be run on CPUs. For that, it is useful to have an easy option for massive parallelization on CPUs. Therefore, besides the GPUs at ÚFAL cluster, there are also many CPU machines. Some of them are equipped with large RAM. Every single CPU thus can be allocated through a scheduler the same way as GPUs. It is extremely easy to parallelize a data processing task and achieve a speedup of e.g. 500.
- Abstraction. The users benefit from an "abstract" view of the cluster where the machine to process a task is selected and operated fully automatically in the background, without explicitly targeting the machine by name but rather by hardware features that may be met by multiple machines. The software should be identical on all the machines so the software versions will not cause any problems.
- Direct access. At the same time, it should be also possible to directly access (e.g. ssh log in) the machine where a task is running, for inspection and debugging.
- Remote access. Remote working is the standard now. The cluster must be accessible to the users from anywhere, not only from the workplace.
- Troubleshooting and support. The technicians should be available and willing to receive and solve issue reports and to give support. Researchers should also support each other by giving advice.
- Cooperativeness. The whole cluster works on the rules that everybody accepts because of monitoring and instant penalty (e.g. disk quota) as well as on trust, overall cooperativeness and good manners: *Do not treat others in ways that you would not like to be treated.* For example, it is possible to bypass the scheduler and run the tasks directly, but the memory conflict usually fails all the conflicting jobs and a complaint of a colleague prevents repeating the mistake.
- Shared costs and benefits. At ÚFAL, all the grant projects contribute parts of the funds for expanding and upgrading the cluster, and indirectly also to the power costs. The benefit is that every researcher at the institute has a free unlimited access to the whole cluster for any work-related computing.

# 10
# Conclusion: The Prospects

We have covered general properties of training complex neural networks and common argumentation errors when trying to understand the source of the observed gains in Part I. Then, in Part II, we thoroughly discussed transfer learning as the simplest example of benefiting from multilinguality, and then the more complex models supporting many more languages at once. We saw cases where even respected research teams made some errors in explanation and we hope you are not going to repeat them. We also discussed our practical experience with multilingual MT both from the end users' point of view as well as from the position of a researcher at a moderately-sized research lab.

Before we finish, we want to highlight three topics that we find most relevant for the next decade: (1) the idea of mutual understanding between us and the models: Do the models understand the content they are processing, or will they ever? Do we understand what the model is doing and why?, (2) the general interest in massive models, and the implications this shift of interest might have on research, and finally (3) the ecological trace of NLP research.

## 10.1 Mind the Gap in Understanding

We all have experienced the sense of mutual understanding with a communication partner. This understanding can happen at various levels, starting from the personal one (the perceived unity of opinions on the "life, universe and everything"), over the practical one (being confident that my student or secretary understands and will carry out a task "as agreed"), up to the "world" one (our ability to interpret the actions of our fellow or enemy, the motivations behind them, and the ability to predict the next ones).

NLP systems are growing to such qualities that we would like to see them as communication partners. With appliances and services like Amazon Alexa, Google Home, Apple's Siri or Microsoft Cortana, users are being persuaded that that era is here.

We can distinguish two main directions of the understanding: do the models understand us, and do we understand the models? In both cases, the short answer is no.

| | |
|---|---|
| Source | V Penny jej nyní prodávají tuším za 27,90Kč. |
| CUBBITT | Penny is now selling it for, I think, \$27.90. |
| Source | V Penny jej nyní prodávají tuším za 27,90 Kč. |
| CUBBITT | Penny's now selling it for, I think, £27.90. |

**Figure 10.1:** The brittleness of CUBBITT (Popel et al., 2020). The translation is flawless *except* for the currency unit. The Czech crown ("Kč") was once "translated" to U.S. dollar and once to the British pound, depending whether there was a space between the number and the unit.

### 10.1.1 Deep Models Do Not Understand Us

Anyone who has attempted to rely on even the most recent systems of machine translation that reach the qualities human parity of human translation in certain types of evaluations (Hassan et al., 2018; Popel et al., 2020) has surely and rather quickly run into a "catastrophic error", an error that severely changed the meaning of the sentence. In Figure 10.1, we demonstrate the sensitivity of CUBBIT, the system of Popel et al. (2020) for English→Czech translation,[1] to very subtle details of the input.

With just a single occurrence of such a catastrophic error, it becomes clear than the systems are not processing the *meaning* of the sentence. Instead, they benefit from the huge amounts of training data and limited novelty of test examples.[2]

Test sets focusing specifically on the ability to reason about the meaning of the sentence have been explored. The best known is the Winograd Schema Challenge (WSC), see the recent review of the challenge by Kocijan et al. (2020). Each of WSC examples appears in two very similar variants where some form of common sense knowledge or reasoning is required to correctly identify the correct antecedent of a referring expression. An example is provided in Figure 10.2. While the examples should be unambiguous for humans, so a perfect performance could be expected, in practice only the accuracy of 92% was achieved, probably due to insufficient concentration of the respondents (Bender, 2015). The most recent systems relying on huge language models (see also Section 10.2 below) get to about 60% with simple techniques (Trinh and Le, 2018; Klein and Nabi, 2019) and are able to reach the human benchmark with accuracies of up to 90%, see Kocijan et al. (2020). As Kocijan et al. (2020) note, these systems have however not demonstrated any other aspect of understanding, such as answering simple questions about the text.

Sentences in WSC are not always suitable for evaluating MT systems because the target language may allow to preserve the ambiguity of the source, see Davis (2016) for details. Stanovsky et al. (2019) and Kocmi et al. (2020) used WSC-like constructions to evaluate gender bias in MT: While the sentence is unambiguously indicating a particular gender for a noun (e.g. "teacher"), MT systems tend to ignore this information and translate the word using the stereotypical gender for that occupation (e.g. "Lehrer" vs. "Lehrerin" as two German counterparts of "teacher").

[1] `https://lindat.mff.cuni.cz/services/transformer/`

[2] This is true even if the test set was created anew, after the training data has been released, and can not thus be accidentally included in the training data. Our feeling, which has yet to be explored, is that the level of repetitiveness in natural language is much higher than generally believed. Novel test sets can be thus "excessively similar" to the training data and inflate the true generalization capacities of a system.

| | |
|---|---|
| SRC | Fred and George knocked on the door of Jane and Susan's apartment, but they did not answer. |
| MT | Fred a George zaklepali na dveře bytu Jane a Susan, ale * **neotevřeli**. |
| Gl. | Fred and George … , but * **they [F&G] did not open**. |
| SRC | Fred and George knocked on the door of Jane and Susan's apartment, but they did not get an answer. |
| MT | Fred a George zaklepali na dveře bytu Jane a Susan, ale nedostali odpověď. |
| Gl. | Fred and George … , but they [F&G] did not get an answer. |
| SRC | Jane and Susan knocked on the door of Fred and George's apartment, but they did not answer. |
| MT | Jane a Susan zaklepaly na dveře bytu Freda a George, ale * **neotevřely**. |
| Gl. | Jane and Susan … , but * **they [J&S] did not open**. |
| SRC | Jane and Susan knocked on the door of Fred and George's apartment, but they did not get an answer. |
| MT | Jane a Susan zaklepaly na dveře bytu Freda a George, ale nedostaly odpověď. |
| Gl. | Jane and Susan … , but they [J&S] did not get an answer. |

**Figure 10.2:** Failure of CUBBITT on variations of a sentence from Winograd Schema Challenge. The target Czech language indicates the gender of the plural subject in the verb's ending: "y" is for feminine, "i" is for masculine. The system guesses the gender of all the participants correctly but the meaning of "did not answer" is shifted to "did not open" and its subject is identified wrongly (marked with a star and boldfaced), as can be seen in the gloss ("Gl.").

Figure 10.2 illustrates the errors made by CUBBITT on a WSC sentence pair. We see that this rather complex sentence is translated almost perfectly, with correct declination of English names ("Freda a George"), with perfect punctuation, with correct grammatical agreement between the subject and the verb, except for the common-sense challenge itself.

It is also well known that the performance at the sentence level is still considerably higher than the translation quality if the whole document is assessed (Läubli et al., 2018). Vojtěchová et al. (2019) provided a test suite for WMT 2019 where all systems totally failed on preserving the identity of parties when translating a sublease agreement. Both parties of the agreement were denoted with the same term in the target language, rendering the translation useless for the end user.

### 10.1.2 We Do Not Understand Deep Models

As a research community, we have gained considerable knowledge about the behaviour of the Transformer model, as illustrated throughout this book for the MT use case. Significant progress is being made in the area of **explainable AI** (XAI), see e.g. the survey by Danilevsky et al. (2020) or the books by Mareček et al. (2020) and Søgaard (2021) for examples of techniques that reveal certain similarities of represen-tations of different inputs, connect predicted outputs to training items to document why a decision was made, etc.

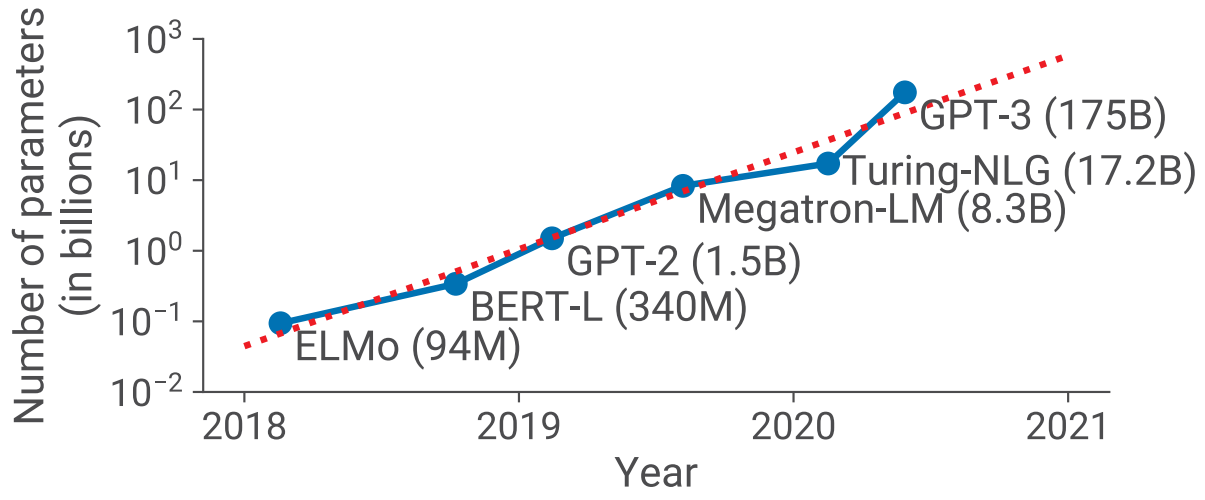

**Figure 10.3:** Exponential growth in the number of parameters in pre-trained language models. Reproduced from Narayanan et al. (2021).

In our case of multilingual MT, some evidence is available supporting the hope that the additional languages bring *linguistically relevant* information to the model and that the model abstracts competence across languages in a human-like way. At the same time, we provided numerous examples where there was real gain observed but it stemmed from somewhat hidden changes brought along with the intended modification. These changes included additional directly usable data (e.g. monolingual English texts brought with the added languages, see Observation 43 on page 133), effectively deeper models (Section 5.4.2), some unclear regularization advantage (Section 5.4.1), a larger number of model parameters or simply larger model capacity for the main task in multi-tasking (Observation 57).

We see these unnoticed changes as highly undesirable. Attributing improvements to wrong reasons is inevitably going to slow down further research. Such a mistake in explanation effectively creates a dead-end path that many of us might pursue.

These cases clearly warn us how limited our understanding is and how easily we fall into the trap antropomorphising deep neural models (Proudfoot, 2011; Watson, 2019; Marcus, 2020).

## 10.2 Massive Models in NLP

Over a couple of recent years, the whole area of NLP and AI is seeing an *exponential* growth in model sizes. For MT, we mentioned some examples in Section 8.5.

In Figure 10.3 reproduced from Narayanan et al. (2021), we see the growth for language modelling tasks. Recent language model are based on the Transformer architecture anyway, taking either the encoder or the decoder to predict a masked or damaged word in a sentence (BERT-like), or predicting the next token (GPT-like), respectively.

He et al. (2021)[3] state that "larger pre-trained models have shown better generalization results", mentioning e.g. T5 (Raffel et al., 2020) or GPT-3 (Brown et al., 2020). There is no denying of the great scores in these tasks but we strongly disagree with the claim of "better generalization". With hardly curated training data and test sets reused from other research teams without any detailed knowledge of their content and inherent biases, anyone can easily misinterpret the results. Many types of noticeable and unnoticeable data overlaps and biases can be exploited by the current over-parametrized techniques. See also our Footnote 2 above.

The more important remark we want to make here is that DeBERTa of He et al. (2021) with up to 1.5 billion parameters needs 30 days of training on 16 GDX-2 nodes, each of which contains 16 Nvidia V100 GPUs, i.e. 30 days on 256 GPU cards.

Narayanan et al. (2021) contemplate about the effectiveness of large-LM methods and report two main problems, GPU memory and number of compute operations needed: Training GPT-3 with 175 billion parameters (Brown et al., 2020) would require approximately 288 *years* with a single V100 Nvidia GPU. Narayanan et al. (2021) conclude that "this calls for parallelism" and proceed to devise such methods.

Instead, we would like to argue that this should call for task reformulation.

The size of the models is now a new factor which can prevent the reproduction of the method at an independent site. While the practice is to release the pre-trained models, they can be sometimes too big even just to load and use at runtime.

Bender et al. (2021) bring additional arguments against too much focus on these huge training sets and models, including societal implications, e.g. on diversity.

## 10.3 Ecological Trace of NMT Research

One of the early motivations of the shift from statistical machine translation to neural machine translation (Cho et al., 2014a) was the *decreased* memory demands and runtime complexity of the large-scale SMT models at that time. Today, less than eight years after those experiments, we face a very similar problem: NMT models have grown so huge that they require significant costs for computation resources, restricting the set of labs that can afford to examine them. Moreover, the intensive computing in NMT research creates environmental concerns. Intensive computing consumes electricity whose production, in many countries, has a significant carbon footprint; a highly undesirable effect in our times of global warming.

In the rest of this section, we summarize environmental concerns from the global point of view and then quantify the carbon footprint generated by the PhD dissertation of Kocmi (2019) which constitutes the experiments with transfer learning reported in this book directly.

[3] `https://github.com/microsoft/DeBERTa`

### 10.3.1 General Environmental Concerns

From our point of view, the problem with large carbon footprint of NMT can be faced from several facets. The first two, the energy resources and hardware development, are beyond the scope of this book and in a little to no control of the researchers, the targeted audience of this book, but we remind them anyway because in our opinion, some contributions should be expected also from their sides.

The latter two facets, Green Deep Learning and "green" NLP and NMT research are of course under control of the researchers.

**Renewable Energy Resources** A typical computer cluster in a research lab is connected to the public electricity network and consumes the power that is available in the local network. Depending on the country and locality, the electricity is generated more or less from a renewable ("green") resource, or from a non-renewable energy resource with a significantly large carbon footprint. The green resource should be preferred, see e.g. Google Cloud claiming carbon neutrality,[4] or a Siberian data center powered by cheap hydroelectric supply.[5]

**Power Efficient Hardware.** The hardware developers can come up with new processing units with lower energy consumption.

**Green Deep Learning.** The researchers may focus on methods that make the current massive DL models more energy efficient, e.g. smaller and faster in training thanks to knowledge distillation, quantization, NN pruning, etc. The industry may consider running the inference of large models on power efficient hardware such as small portable devices or distributed systems on remote power efficient clusters. When training NNs, researchers should consider reusing old pretrained models as much as possible, and the models should be shared.

As discussed by Bender et al. (2021), giving more prominence to methods different from just scaling up existing models can actually increase chances of new breakthroughs.

The category of methods that make NN energy efficient is called **green deep learning**. See Schwartz et al. (2020) or the survey by Xu et al. (2021).

**Green NLP and NMT Research.** Strubell et al. (2019) in her opinion paper at ACL 2019 estimated the impact of several deep learning models and experiments in NLP on the $CO_2$ emissions and on the cost. They call upon fellow NLP researchers to report training time and sensitivity to hyperparameters (thus indicating the extent to which a costly hyperparameter search is needed) and to prioritize work on computationally-efficient algorithms and use energy-efficient hardware. As discussed here in Section 10.2, Strubell et al. (2019) also warn about the dangers of non-inclusive research conditions due to available resources.

[4] https://cloud.google.com/sustainability/

[5] https://www.euronews.com/next/2021/06/28/how-this-siberian-data-centre-is-attracting-bitcoin-miners-with-cheap-green-power

Based on these considerations, ACL itself is drafting its policies to promote efficient NLP research.[6]

### 10.3.2 Carbon Footprint of Our Transfer Learning Experiments

Strubell et al. (2019) published work on the impact of deep learning on $CO_2$ emissions. They estimated that a single Transformer architecture hyper-parameter search produced 284 tonnes of $CO_2$.[7] Many researchers pointed out that it reports numbers based on the U.S. energy mix. For example, Google claims that its platform is 100% renewable.[8] Moreover, the most carbon-intensive scenario in the study costs between $1 million and $3 million (Strubell et al., 2019), which is not an everyday expense.

We try to calculate the impact of experiments performed for our own experiments on transfer learning reported in this book and originally in Kocmi (2019). We roughly calculate the wall-clock time, power consumption, and $CO_2$ emissions.

We start by calculating the total wall-clock time on GPUs. Our departmental cluster (see Section 9.3 for more details) saves information about GPU usage every 10 minutes. We recovered logs from the 20 months during which almost all of our experiments were performed and calculated how many GPU hours were spent on this.

The average power consumption is harder to calculate. Based on a discussion with our IT department, the average GeForce GPU card takes 200W per hour and Quadro P5000 160W per hour, when entirely in use. We need to take into account also the air-conditioning of the room, which can be roughly calculated by coefficient 1.3 to 1.4, we take the 1.35 as a middle value. With this in mind, we use 270W for Tesla and GeForce cards and 216W for Quadro P5000.

An important notice is that our energy consumption is *very roughly* estimated. It is based on GPU in full use, and the effect of air-conditioning is only estimated. Therefore the numbers are more likely the upper bound.

Our experiments produced a similar amount of $CO_2$ as seven round-trip-flights from Prague to New York one passenger or as an average person produces within two years. For more comparisons, see Table 10.1. For ourselves, we see this amount of consumption as higher than expected but at least in proportion to the standards before the Covid-19 pandemic: participation at around 6–7 high-profile venues including an occasional distant one like Australia is a nice achievement of a PhD student and would be definitely funded by the department.

[6] `https://www.aclweb.org/portal/sites/default/files/Efficient%20NLP%20policy%20document%20full%20document.pdf`

[7] The paper reports numbers in imperial units, thus we recomputed the numbers into SI units.

[8] Source: `https://cloud.google.com/sustainability/`

| **Source of Emissions** | | **$CO_2$ Emissions** |
|---|---|---|
| Experiments in the dissertation thesis of Kocmi (2019) | | 9.80 t |
| One person air travel: | | |
| New York, USA → San Francisco, USA | | 0.65 t |
| Prague, Czech Republic → New York, USA | | 0.70 t |
| London, UK → Sydney, Australia | | 1.79 t |
| Frankfurt, Germany → New Delhi, India | | 0.64 t |
| Average person per year | 2019 | 2020 |
| Global | 4.92 t | 4.62 t |
| EU27 | 6.61 t | 5.91 t |
| USA | 15.30 t | 13.68 t |
| India | 1.87 t | 1.74 t |
| China | 8.10 t | 8.20 t |
| Brazil | 2.25 t | 2.11 t |
| Russia | 12.36 t | 11.64 t |

**Table 10.1:** Comparison of $CO_2$ emissions. The numbers for air travel are according to `travelnav.com` and take into account the usual aircraft on the route. Average personal emissions are from `https://edgar.jrc.ec.europa.eu/report_2021`. We report the year 2019, the last one before Covid-19 pandemic, and year 2020 where the pandemic influenced personal $CO_2$ emissions.

Nonetheless, we find it very important to discuss these issues openly, pragmatically, and based on at least some approximate calculations.

Some may argue that the massive pre-trained models we criticized in Section 10.2 are one of the possible remedies to this problem. If released and reused, these models could be seen as a *saving* of $CO_2$ emissions. Our experiments with the cold-start scenario (Section 7.2) confirm that by helping to reduce the total training time and improving translation performance. The only catch is that the models need to be far more often reused than updated.

Furthermore, machine learning can help to tackle climate change in various ways: predicting the electricity demand; flexibly managing the household or vehicle demand; optimizing transportation routes; forecasting extreme climate events; modeling disease spreading; and many more as summarized by Rolnick et al. (2019).

In conclusion, deep learning is energy-intensive, and further research is needed to find the best ways to minimize the impact. However, restrictions on research are not the solution. The source problem here is the production of electricity. In our case in the Czech Republic, more than a half of the electricity still comes from fossil fuel power plants. As mentioned earlier, Google claims that its AI cluster is 100% run by renewable energy; Amazon claims 50%.[9]

We can only hope that overall, countries will keep moving towards more environmentally friendly electricity production. While waiting for this change, we can contribute by steering our focus to efficiency and clever methods instead of brute force.

## 10.4 Final Words

We have thoroughly discussed the use of more than just two languages in machine translation. The idea of transfer learning was explored in detail and we also glanced over the most recent techniques for multilingual MT models, up to the massive ones.

Many of the reported experiments undisputably show gains, from more languages or from more data in general. At the same time, we highlighted multiple cases where a comparable improvement was achieved with a totally uninformed method. In our opinion, the key virtue of a good scientist is a healthy level of scepticism – not towards the veracity of fellows[10] but towards one's own understanding of *why* it helps. The harshest reviewers of successful experiments should be the researchers themselves. The enemy is the complexity of the task *and* the great ability of neural networks to learn. The problem we all are facing is the *misalignment between our learning and the learning of the networks*. We tend to think the network does what we think we are doing in the task but this has never been the case. Under this wrong initial assumption, we easily miss a nasty trick the network has played at us and our test sets are typically too brittle to reveal that.

We find it vital to carry out research of realistic problems (avoid simulated low-resource) and with very strictly comparable setting and dummy baselines, in a vigorous attempt to avoid wishful thinking. We would be very much delighted if such a good practice was adopted in machine translation and ideally spilled over to other areas on NLP and AI as well.

[9] Source: `https://aws.amazon.com/about-aws/sustainability/`

[10] Trustworthiness in science is an absolutely fundamental component, if we lose it, we can't proceed at all.

# List of Observations

# Index

## P

## Q

## R

## S

## T

## U

## W

## X

## Z